%% file: 0_aamain.tex
\documentclass[11pt]{report}
\pdfoutput=1 
\usepackage{microtype}
\usepackage{graphicx}

\usepackage{natbib}
\usepackage{natbib}
\usepackage{mkolar_definitions}
\usepackage{multirow}
\usepackage{amsmath}
\usepackage{prettyref}
\usepackage{comment}
\usepackage{amsmath}
\usepackage{amssymb}
\usepackage{amsthm}
\usepackage{mathtools}
\usepackage{comment}
\usepackage{tikz}
\usepackage{float}
\usepackage[algo2e,ruled,linesnumbered]{algorithm2e}

\newcommand{\dist}{\mathcal{D}}
\newcommand{\DA}{\mathcal{D}_A}
\newcommand{\DB}{\mathcal{D}_B}
\newcommand{\closure}{cl(\mathcal{H})}
\newcommand{\Bern}{\text{Bernoulli}}

\newcommand{\inparen}[1]{\left ( #1 \right )}
\newcommand{\insquare}[1]{\left [ #1 \right ]}

\newcommand{\abs}[1]{\left\lvert #1 \right\rvert}

\newlength{\dhatheight}

\DeclareMathOperator*{\Ex}{\mathbb{E}}
\DeclareMathOperator*{\E}{\mathds{E}}

\DeclareMathOperator*{\Prob}{Pr}

\renewcommand{\Pr}{\mathbf{Pr}}

\newcommand{\calA}{\mathcal{A}}

\newcommand{\calF}{\mathcal{F}}
\newcommand{\calG}{\mathcal{G}}

\newcommand{\calI}{\mathcal{I}}

\newcommand{\calL}{\mathcal{L}}

\newcommand{\calO}{\mathcal{O}}

\newcommand{\calU}{\mathcal{U}}

\newcommand{\calX}{\mathcal{X}}
\newcommand{\calY}{\mathcal{Y}}

\newcommand{\ERM}{\textsf{ERM}\xspace}

\newcommand{\vc}{{\rm vc}}

\newcommand{\MAJ}{{\rm MAJ}}

\newcommand{\OPT}{\mathsf{OPT}}
\newcommand{\OPTDMax}{\mathsf{OPT}^{\mathcal{D}}_{\max}}
\newcommand{\OPTSMax}{\mathsf{OPT}^S_{\max}}

\renewcommand{\vec}[1]{\mathbf{#1}}
\newcommand{\RLoss}{\ell^{\text{rob}}}

\newcommand{\removed}[1]{}

\usepackage{color-edits}[suppress] 
\addauthor{pco}{purple}
\addauthor{kms}{orange}
\addauthor{sa}{teal}
\addauthor[suppress]{as}{cyan}
\usepackage{alg}
\usepackage{multirow}
\usepackage{subfig}
\usepackage{graphicx}
\usepackage{dsfont}

\usepackage[utf8]{inputenc} % allow utf-8 input
\usepackage{amsmath}
\usepackage{amsfonts}
\usepackage{amssymb}
\usepackage{mathrsfs}  
\usepackage{bbm}

\usepackage[T1]{fontenc}    % use 8-bit T1 fonts
\usepackage[hidelinks]{hyperref}       % hyperlinks
\usepackage{url}            % simple URL typesetting
\usepackage{booktabs}       % professional-quality tables
\usepackage{amsfonts}       % blackboard math symbols
\usepackage{nicefrac}       % compact symbols for 1/2, etc.
\usepackage{microtype}      % microtypography
\usepackage{xcolor}   
\usepackage{comment}
\usepackage{thmtools} 
\usepackage{thm-restate}
\usepackage{natbib}

\newcommand{\cX}{{\mathcal X}}
\newcounter{note}[section]

\def\ie{\mathrm{IE}}
\def\q{q}
\def\u{u}

\def\f{\mathsf{f}}
\def\t{\mathsf{t}}

\def\p{\mathsf{tpr}}
\def\n{\mathsf{fpr}}

\def\bypass{\mathsf{bypass}}

\newtheorem{thm}{Theorem}[section]
 \newtheorem{lem}[thm]{Lemma}
 \newtheorem{clm}[thm]{Claim}
 \newtheorem{obsr}[thm]{Observation}
 
 \newtheorem{cor}{Corollary}

 \newtheorem{defn}{Definition}[section]
 
 \newtheorem{rem}{Remark}

\def\ie{\mathrm{IE}}
\def\q{q}
\def\u{u}

\def\f{\mathsf{f}}
\def\t{\mathsf{t}}
\newcommand{\X}{\mathcal{X}}

\def\p{\mathsf{tpr}}
\def\n{\mathsf{fpr}}

\def\bypass{\mathsf{bypass}}
\newcommand{\juba}[1]{\textcolor{red}{[Juba: #1]}}
\newcommand{\saeed}[1]{\textcolor{red}{[Saeed: #1]}}
\newcommand{\kevin}[1]{\textcolor{red}{[Kevin: #1]}}
\newcommand{\ali}[1]{\textcolor{red}{[Ali: #1]}}
\newcommand{\lee}[1]{\textcolor{red}{[Lee: #1]}}

\renewcommand{\juba}[1]{}
\renewcommand{\saeed}[1]{}
\renewcommand{\kevin}[1]{}
\renewcommand{\ali}[1]{}
\renewcommand{\lee}[1]{}

\newtheorem{assumption}[theorem]{Assumption}

\newcommand{\manipconj}[1]{c^*_{conj} \left(#1\right)} %manip power in the conjunctive case
\newcommand{\manipseq}[1]{c^*_{seq} \left(#1\right)} %manip power in the sequential case
\newcommand{\cost}{c} %in case we wamt to change the cost notation
\newcommand{\reals}{\mathbb{R}}

\title{Fairness, Accuracy, and Unreliable Data}
\author{Kevin Stangl}

\begin{document}

\include{notations}

\maketitle
\chapter*{Abstract}
This thesis investigates three areas targeted at improving the reliability of machine learning; fairness in machine learning, strategic classification, and algorithmic robustness.   
Each of these domains has special properties or structure that can complicate learning.
%and there are substantial connections to fairness in machine learning in each paper.

A theme throughout this thesis is thinking about ways in which a `plain' empirical risk minimization algorithm will be misleading or ineffective because of a mis-match between classical learning theory assumptions and specific properties of some data distribution in the wild.  

The overarching research goal for these related topics is to provide a crisp mathematical model for each learning scenario that  exposes different failure modes and makes trade-offs between important metrics explicit in order to provide algorithmic advice or recommendations to practitioners and expose gaps for future research. 

By tuning our learning algorithms to be more distribution specific in these scenarios, the resulting learned system will exhibit higher utility and avoid catastrophic failure modes.   
This research is grounded in the theory of machine learning and is fundamentally mathematical in nature, with empirical support when appropriate.

Theory is particularly important in these sensitive domains as it is unclear which poor behavior in deployed systems is a natural or benign consequence of a learning system with the underlying distribution,contrasting with problematic but correctable behavior caused by an error in algorithm design or implementation, how to mitigate these issues, or what a successful outcome even looks like in each problem. 
Theoretical understanding in each domain can help guide  best practices and allow for the design of effective, reliable, and robust systems.

\chapter*{Acknowledgements}

The research in this thesis was seven years in the making but really was a slow burn over thirty years. 
In absolutely no order of priority, I would like to thank my mother Gerilyn Maloney for her endless encouragement of my scientific and mathematical pursuits.
She was my first mathematics teacher and remains deeply invested in my education and personal growth.
So many good things in my life started with her prompting and I am deeply appreciative of her support in managing my speech impediment and becoming \emph{mostly} comprehensible. %To my grandmother, Marilyn Pozzi Maloney, who attempted to bequeth to me the gift of gab and I deeply miss.

To Joseph Mann, Matthew Orton, Adam Warmoth, and Olivia Santiago--I look very fondly back on playing water polo with you all and laughing together; the zest for life each one of you has inspires me each day.

To UCLA friends and the Boelter Library math-study crew: 
Without your assistance in challenging courses, and especially Math 275A, I would not have been ready to take on the Phd. 
Thanks to Michael Shi, Ammar Doo, Timur Celikel, Anastasia Borovich, Prianna Ahsan, Brandon Ayers, Samuel Birns, and many others. I would like to specifically acknowledge Professors Deanna Needell and Jonathan Peterson, whose mentorship during summer research programs gave me my first real exposure to research.

I gratefully acknowledge the support of the Toyota Foundation and TTIC and the Simons Foundation and more concretely, the American taxpayer. My Phd was enabled by the priority the United States puts on scientific advancement and I hope such support persists and grows.

In a very real sense, everyone at TTIC supported me in this journey, and I am honored by the support and friendships of the students, staff, and faculty. First, I want to thank Nati Srebro for admitting me as a prospective student; I thoroughly enjoyed the lively 

To the Salonika Breakfast Crew, Nick Kolkin, Chip Schaff, Davis Yoshida, Shane Settle, Blake Woodworth, Phillip Sossenheimer, and David Yunis; chatting with you before getting down to work started my days on a light and joyful note, even when I felt rudderless and doubted my ability to walk this path. 

To Yixin and Tom Sunbear, I am endlessly inspired by your zest for life and happiness. You are a dynamic duo. 

To each of my collaborators, Omar Montasser, Saba Ahmadi, Ali Vakilian, Lee Cohen, Saeed Sharifi-Malvajerdi,Princewill Okoroafor, Juba Ziani, and Aadirupa Saha, I thank each and every one of you for being extremely patient with me and supportive of me during each and every one of our meetings. 
I am proud of the work we have done together and hope it will inspire others.

To my Phd Committee; thank you for serving and  giving generously of your time and energy. Thank you Madhur for your helpful advice early in my phd and relentlessly positive attitude.

To Ali Vakilian: Advisor and collaborator, my favorite part of working together is this little smile you get when you particularly enjoy a proof or theorem. 
To Juba Ziani: Advisor and collaborator,  your devious and playful attitude towards research made research together so fun!

I cannot say enough superlatives about my advisor Avrim Blum. 
He radiates professionalism, kindness, and curiosity wherever he goes and treats everyone he meets with respect. 
He has always been unstintingly generous with his time and I am honored by his faith in my academic potential. 
He never failed to help re-center me and keep my efforts oriented towards true north.
The computer science I have learned from Avrim pales in comparison with just learning from how he moves through the world; I often ponder to myself `What would Avrim do?'
I then try to do that, no doubt falling short, but hopefully getting closer each day.

I would also like to acknowledge the people, unknown to me, from many groups and backgrounds, who suffer adverse decisions/outcomes from AI systems or fail to benefit from such systems.
I am not a technological solutionist, but firmly believe in the power of technology to improve and elevate the human condition and allow more people to live healthy and fulfilling lives. 
Technological advancements often come with winners and losers and un-expected harms [as well as benefits]. 
My guiding light in this research program has been use to some of limited knowledge granted to me to try to mitigate some of these harms and spread the benefits of the progress in AI. 
I hope my work will contribute to this mission.

%My work is merely a small piece in the broader trustworthy AI ecosystem, a community of scholars, technologists, hackers and activists whose work I find deeply inspiring and encouraging, even when we perhaps disagree. Much work remains to be done.

Finally, I am honored to dedicate this thesis to Figaro Stark, Oleanna Stark, and Eleanor Trier Kirk, each of whom bring endless joy and laughter to my days and I cannot imagine my life without. I love you all and am excited to begin this new stage of our life together.

\tableofcontents

\chapter{Introduction}
 Machine learning is a transformative technology that leverages large amounts of data and computation to create actionable and accurate predictions and decisions
These systems offer the promise of making many accurate decisions for tasks where the `correct' answer cannot be directly programmed in the form of an imperative algorithm and instead must be learned from data.
For many tasks, a machine learned tool is the only option other than crude heuristics. 

%Roughly, machine learning `works' and off-the-shelf tools are effective when the test distribution remains connected to the training distribu
In general, the goal of a learning system is to maximize test time accuracy, by finding a hypothesis 
$h \in \mathcal{H}: \mathcal{X} \rightarrow \mathcal{Y}$ that predicts labels accurately from features.
Generally in this thesis, we will focus on $\mathcal{Y}=\{0,1 \}$.

The fundamental learning problem for supervised classification\footnote{Note, generally un-supervised classification problems reduce to a supervised problem. } over this `true' distribution is defined as follows:
\[ \min_{h \in \mathcal{H} } \quad \mathbb{E}_{(x,y) \sim \mathcal{D}} [ l^{0,1}(x,y, h(x)) ] = \min_{h \in \mathcal{H} }  P_{(x,y) \sim \mathcal{D} } [h(x) \neq y ] \]
However, instead of access to the true distribution $\mathcal{D}$, the learner must select $h$ by estimating an empirical loss on a training data set drawn according to distribution $\mathcal{D}$.

When the hypothesis class $\mathcal{H}$, the form of the training data $x \in \mathcal{X}$, and the label (and labeling method) really do come from this distribution $\mathcal{D}$,
then Vapnik's learning theory \cite{vapnik:71}  and the Fundamental Theorem of Statistical Learning \cite{blumer:89} provide \textit{the} answer for how to solve a statistical learning problem; namely, implement an Empirical Risk Minimization algorithm. 
With enough data the $l_{train} \approx_{\epsilon} l_{test}$ due to uniform convergence. 

However, this classical story becomes more complex when we zoom into the details of each step of this process.
How was the distribution created? The data was not found lying on the ground; rather the schema of the data was designed and collected by humans intentionally at a specific time and place for an instrumental objective. 
Moreover, the labelling process is similarly contingent with the predicted label often a proxy for a more open ended or hard to quantify objective (like satisfactory employment outcomes after six months). 

Perhaps the labels on some groups of people are more reliable, while on another group, the labels are more pessimistic.
How would this impact the classifier our learning algorithm will output?

Even if an algorithm designer  primarily focuses on test-time accuracy as a reasonable objective,  we will see how un-modeled complications of the learning problem likes these interacting with a plain ERM algorithm can result in substantially sub-optimal performance overall, or poor performance on one specific sub-group.

%Average accuracy across a data-set is a coarse objective that can be misleading for some tasks. 
%For instance, imagine there a model $h$ with $95\%$ classification accuracy.
%If there is a sub-group that is at most $5\%$ of the training data, then that overall accuracy of $95\%$ accuracy is consistent with the model having arbitrarily poor accuracy on that sub-group. 

Maybe for some tasks, issues like this are tolerable, but for high stakes decision making like loan decisions, recidivism prediction, medical treatment allocation, being incorrectly classified could be extremely harmful to the \textit{recipients of that decision}, an intense asymmetry that might only roughly be captured overall test accuracy.

This risk of acute harm motivates fairness in machine learning, a burgeoning field concerned with disparate error in learning systems, the social consequences of these errors, and algorithmic improvements to reduce these risks. 
This field has important social implications \emph{and} probes the fundamental limits of learning and statistical prediction in the presence of uncertainty. 

In this thesis we will first investigate core fairness questions about how fairness constraints perform subject to unreliable data, which motivated and led to questions investigating fairness in  screening problems. In a real sense, this is the beating heart of the thesis and its spiritual center.  
As we progress through the thesis, we shall introduce the fairness notions relevant to that project.

While my initial research projects centered on \emph{fairness in machine learning}, over time these interests broadened and blossomed into a research interest in trustworthy AI as a whole, encompassing directions in strategic classification and adversarial robustness [and connections between these areas].

It is hard to formally pin down the meaning of trustworthy AI, but I would suggest the core meaning of the term is AI that performs effectively in learning scenarios complicated by human action, human error, human bias, human strategic behavior or other phenomena that weaken the train-test connection that makes machine learning normally so effective. Often these obstacles coincide with the high stakes settings in which `trust' in the outputs of a learned system is critical. 

A note: the technical content of each chapter is based on published and peer reviewed research\footnote{Other than the experiments in Section \ref{sec:fairexp}.}. 
Some of the commentary and interpretation that connects the papers is novel to this thesis and is solely the opinion of this author, and may not reflect the views of my co-authors.

\section{Summary of Contributions}
Now we briefly summarize the sections of the thesis and their relationships to each other. Sections \ref{sec:fairnoise} and \ref{sec:malnoise} will focus on algorithmic fairness and unreliable data, Sections \ref{sec:multistagescreen} and \ref{sec:stratscreen} will focus on screening problems centered on fairness and strategic behavior respectively, and Section \ref{sec:robust} will focus on adversarial robustness\footnote{e.g. adversarial examples \cite{goodfellow2014explaining}}.

\subsubsection{Fairness with Unreliable Data}
The classical way of mitigating the harms of disparate performance in machine learning is by imposing group error constraints on each group during the training process or as a post-processing step, designed to equalize performance \cite{hardt16}.
Ideally, these constraints would incentivize a firm to invest in improving worst case accuracy over relevant sub-groups by collecting more accurate/reliable data or improving its model development process some other way. 

Foundational work in fairness in machine learning \cite{kleinberg2016inherent,chouldechova2017fair} shows that Equalized Odds and Calibration, two natural fairness constraints that roughly align with both sides of the COMPASS Debate \cite{angwin2016machine,flores2016false}, are mutually incompatible unless a learning rule has perfect accuracy. 

This type of `impossibility' result substantially complicates thinking about un-fairness because these notions are so natural and seem to capture critical aspects of effective and fair decision making.
Additionally, claims that requiring \cite{menon2018cost} fairness can reduce the resulting model's accuracy, which is consistent with the simple observation that more constraints to an optimization will tend to reduce the objective function value of the optimal solution. 

\emph{Fairness, Accuracy, and Biased Data:}
In Section \ref{sec:fairnoise} \cite{blum2019recovering}, we probe these ideas by making a strong fairness realizability type assumption on the true data distribution, with two groups in a population.
However, instead of the well behaved distribution, the learner has access to a corrupted data-set where the corruption concentrates on and harms one demographic group.
The goal of the learner is to use this biased training data to learn an accurate and fair model on the true data distribution. 
We study the extent to which different fairness notions might help correct for this biased data problem. 

In particular, our results show robust recovery of the correct model by using Equal Opportunity \cite{hardt16} while other fairness notions exhibit poor performance and even unfairness amplification.

We supplement these theoretical claims with synthetic and semi-synthetic experiments that intentionally corrupt training data according to our bias models, and support our theoretical results.

\emph{Fairness Constrained Learning and Malicious Noise:}
In Section \ref{sec:malnoise} \cite{blum2023vulnerability} we continue on a related line of work in response to \cite{lampert} that tests the robustness of fair-ERM in the presence of malicious noise \cite{malnoise}.
Specifically, the question is to what extent a small amount of malicious noise can increase the error rate of an ERM classifier subject to various fairness constraints.
Malicious noise is a much stronger noise model than we considered in \cite{forc2020}, which roughly corresponds to an extension of random classification noise \cite{angluin:88}.

\cite{lampert} exhibit a somewhat pessimistic perspective on fairness constrained learning.
For their first result, for a proper learner constrained to output a hypothesis satisfying Demographic or Statistical Parity \cite{dwork2012fairness}, a small amount of adversarial data can force a learner to produce a classifier with much higher error rate (even constant error rate for small amounts of malicious data) especially when group sizes are imbalanced.

%for parity, the authors show that the excess accuracy loss (on the true distribution) in the presence malicious noise will be $\Omega(\alpha)$ and the demographic parity gap will be $\Omega(\frac{\alpha}{P_0})$ where $\alpha$ is fraction of malicious examples and $P_{0}$ is the size of the smaller demographic group. 
%When $\frac{\alpha}{1-\alpha} \geq 2 P_0 P_1$, the authors show, parity gap will be $\Omega(1)$.

In contrast, for Demographic Parity, we exhibit an improper randomized classifier that has $O(\alpha)$ excess accuracy loss while satisfying parity, regardless of group size, which is optimal with the \cite{malnoise} lower bounds for unconstrained learning.
This contrasts with the pessimistic view exhibited by \cite{lampert} and more closely aligns fairness constrained learning with normal PAC learning with malicious noise.

We also exhibit upper and lower bounds for Equal Opportunity, Equalized Odds \cite{hardt16}, Calibration variants \cite{dawid1982well}, and Minimax fairness  \cite{minimaxfair}.
At a high level, our results provide a more optimistic view of the robustness of fairness constrained learning and we provide clear separations between different fairness notions in terms of this robustness. 

This work is critical since some scholarship in fairness in learning tends to argue for fairness constraints as normative requirements. 
However, if imposing these requirements results in a learning process that is unstable or highly sensitive to malicious noise or training distribution, the equity benefits might be outweighed by the degraded effectiveness of resulting models. 
Our results help shed light on this important and still very much unsettled debate.

\subsubsection{Screening Problems}
In Section \ref{sec:multistagescreen} we consider how to enforce fairness constraints on a sequential screening process. 
These processes split one decision into multiple tests/assessments, like in a hiring process, and passing each stage requires a certain minimum result.
This work was originally motivated by exploring in detail a way Under-representation Bias could enter a data-set, from Section \ref{sec:underrep}.

We (\cite{blum2022multi}) exhibit methods for enforcing fairness constraints on the pipeline while maximizing precision and recall, a linear combination of these quantities, and discuss the `Cost of Fairness' of different group constraints.

 In Section \ref{sec:stratscreen}, we continue on the screening setting but shift our focus to strategic classification \cite{hardt2016strategic}.
Consider that screening problems can often be quite high stakes, e.g. whether or not one is hired for a job. 
This prompts strategic adaption and manipulation, where individuals may mis-report or change their features.

Our initial research question, which is explored in \cite{cohen2023sequential} is how to combine the screening model with strategic classification  \cite{hardt2016strategic}. 
We introduce a novel model for that allows agents to strategically adapt in-between classification steps.
We show that the ability of an agent to manipulate in between steps substantially increases their ability to manipulate successfully and complicates defense.

\begin{comment}
\emph{Sequential Strategic Screening and Fairness--[Ongoing work]}
There is a rich field \cite{hu2019disparate,socialcost18} of thinking about the social consequences of strategic behavior in machine learning. 
At a high level, these methods argue that strategic adaptation/gaming can amplify existing disparities (e.g. wealthy families can take SAT prep classes that do not change their true qualification). 
In some cases, the disparities are increased when a Firm modifies a deployed model to take into to be more pessimistic. 
In our ongoing work we are first considering a more expansive  detailed consideration of the action space of the Firm from \cite{cohen2023sequential} (who deploys the screening process) and how they should mitigate or shape the actions of the Agents who are modifying their features. 
We want to zoom in on the fairness implications of \cite{cohen2023sequential}.

%Once we have a clearer view of the action space, we 
\end{comment}

\subsubsection{Robustness}
In Section  \ref{sec:robust} we will consider our work \cite{ahmadi2023certifiable} which focuses on the problem of using an ERM oracle to obtain a classifier that is robust against patch-attacks in the challenging non-realizable regime. 
%While, this may be somewhat of a diversion from the emphasis on fairness in the rest of the thesis, 

Additionally, in \cite{ahmadi2023certifiable} we provide algorithms for our  multi-robustness notion, in which we want one classifier that is robust for multiple (possibly) overlapping subgroups, and competes with the best classifier on each group.

This notion is similar but distinct from to multi-calibration notions \cite{multicalib}. 
%I believe fairness and robustness problems are in fact closely connected as will be argued in depth and explored in the full thesis.  
\cite{blum2023vulnerability} neatly shows the intersection of fairness and robustness since the same data corruption can be conceptualized as a data corruption that is the cause of the fairness issue or an explicit adversary exploiting the fairness constraints.

Now we will begin the technical chapters of the thesis, starting with my work on fairness constraints and biased data.

\chapter{Fairness and Biased Data} %p
\label{sec:fairnoise}

\input{eoppmain}

\section{Experimental Support}
\label{sec:fairexp}
\input{p1exp}

\chapter{On Fair Learning and Malicious Noise}
\label{sec:malnoise}
Now we we shift gears somewhat and study a related but similar problem, originally published as \cite{blum2023vulnerability}. 
In some sense, this a converse problem to Chapter \ref{sec:fairnoise}.
As we are considering a stronger threat model, we instead characterize how much accuracy is lost when an adversary uses their fairness constraints to amplify his power.

We consider the vulnerability of fairness-constrained learning to small amounts of malicious noise in the training data. \cite{lampert} initiated the study of this question and presented negative results showing there exist data distributions where for several fairness constraints, any proper learner will exhibit high vulnerability when group sizes are imbalanced. Here, we present a more optimistic view, showing that if we allow randomized classifiers, then the landscape is much more nuanced. For example, for Demographic Parity we show we can incur only a $\Theta(\alpha)$ loss in accuracy, where $\alpha$ is the malicious noise rate, matching the best possible even without fairness constraints. For Equal Opportunity, we show we can incur an $O(\sqrt{\alpha})$ loss, and give a matching $\Omega(\sqrt{\alpha})$ lower bound. In contrast, \cite{lampert} showed for proper learners the loss in accuracy for both notions is $\Omega(1)$. The key technical novelty of our work is how randomization can bypass the way an adversary uses the fairness constraints to amplify his power.

We also consider additional fairness notions including Equalized Odds and Calibration. For these fairness notions, the excess accuracy clusters into three natural regimes $O(\alpha)$,$O(\sqrt{\alpha})$, and $O(1)$. These results provide a more fine-grained view of the sensitivity of fairness-constrained learning to adversarial noise in training data.

\input{mal_intro.tex}

\input{mal_preliminaries}

\input{mal_main_results}

\section{Main Results: Calibration}
\input{mal_calibration}

\input{mal_discussion}

\input{mal_appendix}

\chapter{Fairness and Multi-Stage Screening Problems}
\label{sec:multistagescreen}
We will now shift gears somewhat to screening problems, an important area of study in machine learning because of how often these processes are used for high stakes decisions.

Consider an actor making selection decisions (e.g., hiring) using a series of classifiers, which we term a  {\em sequential screening process}.
The early stages (e.g. resume screen, coding screen, 
phone interview) filter out some of the applicants, and in the final stage an expensive but accurate test (e.g. a full interview) is applied to those individuals that make it to the final stage. 
Since the final stage is expensive, if there are multiple groups with different fractions of positives in them at the penultimate stage (even if a  slight gap), then the firm may naturally only choose to apply the final (interview) stage solely to the
highest precision group
%group of highest precision,  
which would be clearly unfair to the other groups.  
Even if the firm is required to interview all those who pass to the final round, the tests themselves could have the property that qualified individuals from some groups pass more easily than qualified individuals from others.
%Given these concerns, 

Accordingly, we consider requiring Equality of Opportunity (qualified members of each group have the same chance of reaching the final stage and being interviewed).
We then examine the goal of
maximizing quantities of interest to the decision maker subject to this constraint,
%The action space of our algorithm is modifying how we promote individuals through the process,  base based on performance at the previous stage.  d on performance at the previous stage.
via  modification of the probabilities of promotion through the screening process at each stage based on performance at the previous stage.

We exhibit algorithms for satisfying Equal Opportunity over the selection process and maximizing precision (the fraction of interviews that yield qualified candidates) as well as linear combinations of precision and recall (recall determines the number of applicants needed per hire) at the end of the final stage. 
We also present examples showing that the solution space is non-convex, which motivate our combinatorial exact and (FPTAS) approximation algorithms for maximizing the linear combination of precision and recall. 
Finally, we discuss the `price of' adding additional restrictions, such as not allowing the decision-maker to use group membership in its decision process.

\input{fairscreen_content}

\input{fairscreen_missing-proofs}

\input{fairscreen_appendix}

\chapter{Sequential Strategic Screening}
\label{sec:stratscreen}

\input{0_abstract}

\input{1_intro}

\input{1.1_related}

\input{1.2_informal}
\section{Our Model}

\input{2_model}

\section{Best Response of Agents in a Screening Process with Oblivious Defender}\label{sec:bestRes}
\input{3_manipulation_charac}

\input{3.1_best_response}

\input{3_0greedycomparison}

\section{Manipulation Resistant Defenses}\label{sec:defense}

\input{4_defense}

\section{Discussion}

\input{5_discussion}

\section{Proofs of Section \ref{sec:bestRes}}
\label{sup:bestRes}

\input{sup-3_bestRes}
\section{Proofs of Section \ref{sec:defense}}\label{sup:defense}

%\subsection{Hardness Proof}
%\input{supplementary/sup-6_hardness.tex}

\subsection{Conservative Defense Proofs}\label{app:conservative_proof}
\input{sup-4_defense}

\chapter{Agnostic Multi-Robust Learning Using ERM}
\label{sec:robust}
Finally, we will briefly consider adversarial robustness \cite{goodfellow2014explaining}.
This is an exciting an important research area that in my opinion is rapidly growing in importance. 
The development and proliferation of consumer facing generative AI systems has exposed new attack surfaces for adversarial behavior that makes this research area even more critical. 

Going back to our definition of trustworthy AI the core areas of trustworthy AI are fairness, strategic behavior, and adversarial behavior. 
For instance, adversarial attacks can be used to allow a malicious user to evade safety fine-tuning
of llms and generate hate-speech or other forms of objectionable content \cite{wei2024jailbroken,zou2023universaltransferableadversarialattacks}.
At a higher level, fairness problems, especially in our framing in Chapter \ref{sec:fairnoise} and Chapter \ref{sec:malnoise} can be thought of as train-test time mis-match, which also captures some robustness work.
In particular in this chapter, we also consider a `multi-group' notion similar to \cite{hebert2018multicalibration}.
Now we shift in detail to the technical content of this work.
%[EXPAND, talk about connections with fairness]

\subsection{Introduction}
A fundamental problem in robust learning is asymmetry: a learner needs to correctly classify every one of exponentially-many perturbations that an adversary might make to a test-time natural example. In contrast, the attacker only needs to find one successful perturbation. \cite{xiang2022patchcleanser} proposed an algorithm that in the context of patch attacks for image classification, reduces the effective number of perturbations from an exponential to a polynomial number of perturbations and learns using an ERM oracle. However, to achieve its guarantee, their algorithm requires the natural examples to be robustly realizable. 
This prompts the natural question; can we extend their approach to the non-robustly-realizable case where there is no classifier with zero robust error?

Our first contribution is to answer this question affirmatively by reducing this problem to a setting in which an algorithm proposed by \cite{DBLP:conf/colt/FeigeMS15} can be applied, and in the process extend their guarantees. Next, we extend our results to a multi-group setting and introduce a novel agnostic multi-robust learning problem where the goal is to learn a predictor that achieves low robust loss on a (potentially) rich collection of subgroups.

\input{rob_introduction}

\input{rob_single-population}

\input{rob_unified-boosting}

\input{rob_appendix_2.tex}

%%%%%%%%%%%%%%%%%%%%%%%%%%%%%%%%%%%%%%%%%%%%%%%%%

\chapter{Conclusion}
In this thesis, we have explored a collection of issues centered on fairness in machine learning, strategic classification, and adversarial robustness.

The key problem that this thesis tries to address is how should stakeholders of machine learning technology interpret and respond to observed disparities and adversarial behavior in their learning pipelines due to an underlying mis-match between training and test distributions.
In a very real sense, the last several decades of scientific and engineering advancements in machine learning have proven the success of the fundamental learning problem from a finite data-set and generalizing to an un-seen test distribution, presumably very closely related to or identical to the training distribution.

As machine learning goes further and further from the laboratory and into challenging, dynamic environments, with complex interaction with humans, the tight coupling of the train-test distributions may loosen or break, requiring mathematical and empirical understanding to maintain performant AI systems. 

To that end, the research in this thesis is a concrete and specific instantiation of that larger train-test mismatch.

We have considered how fair-ERM interacts with benign and malicious noise, how to enforce fairness and mitigate strategic behavior in screening processes, and finally explored adversarial robustness, an area which I believe is promising for further research. 

\subsection{Prospective Role of Theory in Fair-er Machine Learning}
There was once a hope for a single uniform fairness constraint that could rule them all and provide a simple technical answer to issues with disparities, somewhat analogous to the role differential privacy plays as a unifying notion in privacy.

This notion has been critiqued extensively first by the impossibility results \cite{kleinberg2016inherent,chouldechova2017fair} and later by social scientists and computer scientists who contend that these statistical criteria are not expressive enough to capture important normative aspects of fairness and equity \cite{nlpcritical, mitchell2021algorithmic}. 
% of these scholars then become very skeptical of solely technical approaches to fairness in learning.

In this thesis, I have implicitly pushed back on these critiques by instead centering the role of computer scientists and algorithmic approaches on the `right' aspects of algorithmic fairness, meaning using a technical toolkit to characterize the behavior of learned systems in the presence of biased data.

%by correctly characterizing the state-of-the-art 
Rather than solely algorithmic approaches `solving' the problem of bias in social technical systems, a well grounded and extensive theory of machine learning in adversarial, biased, and strategic regimes can help practitioners develop robust learning systems and then have confidence in the outputs of those systems. 

Machine learning based systems can provide substantial and non-replicable utility in many contexts. 
Due to issues of scale and speed in many times a learned solution is the only option other than ineffective baselines. 
Having a calibrated understanding of the risks of algorithmic bias and the effectiveness of possible fair learning approaches will allow the development and deployment of systems that strike the correct risk-benefit balance.

Ideally, these algorithmic innovations and best practices will result in robust systems that provide reliable, calibrated predictions.
Then  when faced  with an issue of bias in a system, the algorithm designer can provide a mix of solution concepts that sweep through a range of classifiers on the relevant Pareto frontier, to be selected among by stakeholders with domain specific knowledge. 
The work in this thesis is a step in this broader vision but much remains to be done.

\bibliography{ref.bib}
\end{document}

%% file: notations.tex
\newcommand{\vect}[1]{\ensuremath{\mathbf{#1}}}

%% Useful
%\newcommand{\p}[1]{\left( #1 \right)}
\newcommand{\br}[1]{\left[ #1 \right]}

\newcommand{\ev}[1]{\mathbb{E}\left[{#1}\right]}
\newcommand{\evd}[2]{\mathbb{E}_{#1}\left[{#2}\right]}

%% Algortihm notations
\newcommand{\bigO}[1]{O \left( #1 \right )}

%% Calibration
\newcommand{\I}[1]{\mathbb{I}\left[#1\right]}       % Indicator
\newcommand{\calerr}{\mathrm{calerr}}   % Calibration error
\newcommand{\game}{\textsf{Sign-Preservation}}      % Name of the game
\newcommand{\spgame}{\textsf{Sign-Preservation}}     
\newcommand{\sprgame}{\textsf{Sign-Preservation with Repetitions}}      % Name of the game
\newcommand{\spmgame}{\textsf{Sign-Preservation with Multiplicity}}      % Name of the game
\newcommand{\fspgame}{\textsf{Fractional Sign-Preservation}}      % Name of the game
\newcommand{\maxerr}{\mathrm{maxerr}}       % Maximum cumulative calibration error
\newcommand{\opt}{\mathrm{SP}}     % Optimal game value of Sign-Preservation
\newcommand{\spg}{\mathrm{SP}}     % Optimal game value of Sign-Preservation
\newcommand{\usp}{\mathrm{USP}}     % Optimal game value of Unanimous Sign-Preservation
\newcommand{\tsp}{\mathrm{TSP}}     % Optimal game value of TSP
\newcommand{\spr}{\mathrm{SPR}}     % Optimal game value of SPR
\newcommand{\spm}[1]{\mathrm{SPM}\left(#1\right)}     % Optimal game value of SPM
\newcommand{\fsp}[1]{\mathrm{FSP}\left(#1\right)}     % Optimal game value of SPR

\newcommand{\A}{\mathcal{A}}    % Algorithm
\newcommand{\Ber}{\mathrm{Ber}}
\newcommand{\Ecover}{\event^{\textrm{cover}}}   % Event that covered epochs exist
\newcommand{\Enegl}{\event^{\textrm{negl}}}     % Event that negligible epochs exist
\newcommand{\Epoch}{\mathsf{Epoch}}
\newcommand{\eps}{\epsilon}     % epsilon
\newcommand{\Etruth}{\event^{\textrm{truth}}}   % Event that all epochs are truthful
\newcommand{\event}{\mathcal{E}}    % Events
\newcommand{\Int}{\mathcal{I}}      % Interval
\newcommand{\poly}{\operatorname*{poly}}    % Polynomial
\newcommand{\pr}[1]{\Pr\left[#1\right]}     % Probability
\newcommand{\red}[1]{{\color{red} #1}}
\newcommand{\SPinner}{\mathsf{SP}^{\textrm{inner}}}
\newcommand{\SPouter}{\mathsf{SP}^{\textrm{outer}}}
\newcommand{\Tact}{T^{\mathrm{actual}}}     % Actual stopping time
\newcommand{\prodspace}{\mathcal{X}\times A \times \mathcal{Y}}

%% Fair ERM Notation
\newcommand{\error}[1]{ \left| \mathbb{E}_{(x,y) \sim \mathcal{D}} \ [\one (#1(x) \neq y)] - \ \mathbb{E}_{(x,y) \sim \mathcal{D}} \ [\one (h^*(x) \neq y)] \right|}
\newcommand{\htilde}{\tilde{h}}
\newcommand{\hhat}{\hat{h}}
\newcommand{\hstar}{h^*}
\newcommand{\hclass}{\mathcal{H}}
\newcommand{\posrate}[1]{ P_{(x,y) \sim \DA} [#1 (x)=1]}

\newcommand{\DAC}{\widetilde{\mathcal{D}}_A}
\newcommand{\DBC}{\widetilde{\mathcal{D}}_B}
\newcommand{\RA}{P_{(x,y) \sim \dist } [x \in A]}
\newcommand{\RB}{P_{(x,y) \sim \dist} [x \in B]}
\newcommand{\normalF}{F}
\newcommand{\corruptF}{\widetilde{F}}

%% file: eoppmain.tex
Machine learning (typically supervised learning) systems are automating decisions that affect individuals
in sensitive and high stakes domains such as credit scoring \cite{scoredsociety} and  bail assignment \cite{machinebias,flores}. 
This trend toward greater automation of decisions
has produced concerns that learned models may reflect and  
amplify existing social bias or disparities in the training data. 
Examples of possible bias in learning systems include the Pro-Publica investigation of COMPAS (an actuarial risk instrument) \cite{machinebias}, accuracy disparities in computer vision systems \cite{shades}, and gender bias in word vectors \cite{kalai}. 

In order to address observed disparities in learning systems, an approach that has developed into a significant body of work is to add demographic constraints 
to the learning problem that encode criteria that a fair classifier ought to satisfy. 

%A number of different such 
Multiple constraints have been proposed in the literature \cite{eodds, dwork12}, each encoding a different type of unfairness one might be concerned about, and there has been substantial work on understanding their relationships to each other, including incompatibilities between the fairness requirements \cite{costfairness, chouldechova, inherent,pleiss2017fairness}. 

In this section, corresponding to the paper \cite{forc2020}, we take a different angle on the question of fairness.  Rather than argue whether or not these demographic constraints encode intrinsically desirable properties of a classifier, %(for arguments against this view \cite{costfairness,mismeasure}),
we instead consider their ability to help a learning algorithm to recover from biased training data and to produce a {\em more accurate} classifier. 

In particular, adding a constraint (such as a fairness constraint) to an optimization problem (such as ERM) would typically 
result in a lower quality solution. However, if the objective being optimized is skewed (e.g., because training data is corrupted or not drawn from the correct distribution) then such constraints might actually help prevent the optimizer from being led astray, and yield a higher quality solution when accuracy is \textit{measured on the true distribution}. 

More specifically, we consider a binary classification setting in which data points correspond to individuals, some of whom are members of an advantaged Group A and the rest of whom are members of a disadvantaged Group B. 
We want to make a decision such as deciding whether to offer a candidate a loan or admission to college.  
We have access to labeled training data consisting of $(x,y)$ pairs where $x$ is some set of features corresponding to an individual and $y$ is a label we want to predict for new individuals. 

The concern is that the training data is potentially biased against Group $B$ 
in that \emph{the training data systematically misrepresents the true distribution over features and labels in Group $B$}, while the training data for Group $A$ is drawn
from the true distribution for Group $A$. 
We consider several natural ways this might occur.  One way is that members of the disadvantaged group might show up in the training data at a lower rate than their true prevalence in the population, and worse, {\em this rate might depend on their true label}. 

For instance, if the positive examples of Group B appear at a much lower rate in the training data than the negative examples of Group B (which might occur for cultural reasons or due to other options available to them),
then ERM might learn a rule that classifies all or most members of Group B as negative.  

A second form of bias in the training data we consider is bias in the labeling process.  
Human labelers might have inherent biases causing some positive members of Group B in the training data to be mislabeled as negative, which again could cause unconstrained ERM to be more pessimistic than it should be.  Alternatively, both processes might occur together. 
We examine the ability of fairness constraints to help an ERM learning method recover from these problems.

\section{Summary of Results}
Our main result is that ERM subject to the \textbf{Equal Opportunity} fairness constraint \cite{eodds} recovers the true Bayes optimal hypothesis under a wide range of bias models, making it an attractive choice even for decision makers whose overall concern is purely about accuracy on the true data distribution. 

In particular, we assume that under the true data distribution, the Bayes optimal classifiers $h_A^*$ and $h_B^*$ classify the same fraction $p$ of their respective populations as positive\footnote{$p=P_{\mathscr{D}_A} (h_{A}^{*}(x) = 1) = P_{\mathscr{D}_B} (h_{B}^{*} (x) = 1)$. 
We will allow the classifiers to make decisions based on group membership or alternatively assume we have sufficiently rich data to implicitly infer the group attribute.},
$h_A^*$ and $h_B^*$ have the same error rate $\eta$ on their respective populations, 
and that these errors are uniformly distributed.  

However, during the training process we do not have access to the true distribution. We only have access to a biased distribution in a way that implicates the distinct social groups and causes the classifier to be overly pessimistic on individuals from Group $B$.

We prove that, subject to the above conditions on $h_{A}^{*}$ and $h_{B}^{*}$, even with substantially corrupted training data either due to the under-representation of positive examples in Group B or a substantial fraction of positive examples in Group B mislabeled as negative, or both, the Equality of Opportunity fairness constraint will enable ERM to learn the Bayes optimal classifier $h^{*}=(h_{A}^{*},h_{B}^{*})$, subject to a pair of inequalities ensuring that the labels are not too noisy and Group $A$ has large mass.

Expressed another way, this means that 
\emph{the lowest error classifier on the biased data satisfying Equality of Opportunity is the Bayes optimal classifier on the un-corrupted data.} These results provide additional motivation for considering fairness interventions, and in particular Equality of Opportunity, even if one cares primarily about accuracy. 

 Other related fairness notions such as Equalized Odds, Demographic Parity, and Calibration %\textcolor{red}{Calibration}
 do not succeed in recovering the Bayes optimal classifier under such broad conditions.
 In fact, we show that given data subject to Under-Representation Bias, Calibration can actually {\em amplify} the effects of the bias, and so can be worse than doing nothing  and instead learning with plain ERM (see Section \ref{sec:underrep}).
 
Our results are in the infinite sample limit and we suppress issues of sample complexity
\footnote{Our notion of sample complexity is typical. 
Let $S$ be the biased training data-set and $ERM_{\mathscr{H}}(S) = \hat{h}$.
Given $\epsilon, \delta > 0$, $m(\epsilon, \delta)$ samples ensures with
probability greater than $1-\delta$ that
$L_{\mathscr{D}}(\hat{h}) \leq L_{\mathscr{D}}(h^*) + \epsilon$.}
in order to focus on the core phenomenon of the data source being unreliable. 
%%%%%%%%%%%%%%%%%%%%%%%%%%%%%%%

\subsection{Related Work}
This chapter is directly motivated by a model of implicit bias in ranking \cite{implicit}.
In that paper, the training data for a hiring process is systematically corrupted against minority candidates
and a method to correct this bias 
increases both the quality of the accepted candidate and the fraction of hired minority candidates.
However, that fairness intervention, the Rooney Rule, does not immediately translate to a general learning setting. 

Our results avoid triggering the known impossibility results between high accuracy and satisfying fairness criteria 
\cite{chouldechova, inherent} by assuming we have equal base rates across groups. 
This assumption may not be realistic in all settings, however there are settings where bias concerns arise and there is empirical evidence that base rates are equivalent across the relevant demographic groups, e.g. highly differential arrest rates for some alleged crimes that have similar occurrence rates across groups \cite{predictandserve, dirtydata}. 

Within the fairness literature there are several approaches similar to ours. 
In particular, our concern with positive examples not appearing in the training data 
is similar in effect to a selective labels problem \cite{kleinbergselective}. 
\cite{chouldechovaselectivelabels} uses data augmentation to experimentally improve generalization under selective label bias. 

\cite{impossibility, discriminative} also consider the training and test data distribution gap we experience in our model and posit differing interpretations of fairness constraints under different worldviews. 
While we do not explicitly use the terminology in these papers, we believe our view of the gap between the true distribution and the training time distribution is aligned with Friedler et al's concept of the gap between the construct space and the observed space. 

Our second bias model, Labeling Bias, is similar to \cite{labelbias}.
In that paper, the bias phenomenon is that a biased labeler makes poor decisions on the disadvantaged group and intervenes with a reweighting technique, one that is more complex than our Re-Weighting intervention.
However, that paper does not consider the interaction of biased labels with different groups appearing in the data at different rates as a function of their labels.

%%%%%%%%%%%%%%%%%%%%%%%%%%%%%%%%%%%%%%%%%%%%%%%%%%%%%%%%%%%%%%%%%%%%%%%%%%%%%%%%%%%%%%%%%%%%%%

\section{Model}
In this section we describe our learning model, how bias enters the data-set, and the fairness interventions we consider. 

We assume the data lies in some instance space
$\mathscr{X}$, such as $\mathscr{X} = \mathbb{R}^d$.
There are two demographic groups in the population, Group $A$ and Group $B$.
Their proportions in the population are given by
 $P(x \in A) = 1-r$ and $P(x \in B)= r$ for $r \in (0,1)$.
$x \in A$ can be read as individual $x$ in demographic Group $A$.
Group $B$ is the disadvantaged group that suffers from the effects of the bias model.

Assume there is a special coordinate of the feature vector $x$ that denotes group membership.
The  data distribution is given by 
$\mathscr{D}$, and is a pair distributions $(\mathscr{D}_{A}, \mathscr{D}_B)$,
with $\mathscr{D}_{A}$ determining how $x \in A$ is distributed and $\mathscr{D}_B$ determining how $x \in B$ is distributed.

\subsection{True Label Generation: } \label{truelabels}
 Now we describe how the true labels for individuals are generated. 
Assume there exists a pair of Bayes optimal classifiers $h^* = (h_{A}^*, h_{B}^*)$ with $h_{A}^{*}, h_{B}^{*} \in \mathscr{H}: \mathscr{X} \rightarrow \{0,1\}$.

We assume that the  Bayes optimal classifier $h_B^*$ for Group B may be different from the Bayes optimal classifier $h_A^*$ for Group A.  
If $h_A^*$ was also optimal for Group B,
then  we can just learn $h^*$ for both Groups $A$ and $B$ using data only from Group $A$ and biased data concerns fade away.
Thus we are learning a pair of classifiers, one for each demographic group. 

When generating samples, first we draw a data-point $x$. 
With probability $1-r$,
 $x \sim \mathscr{D}_{A}$ (and thus $x \in A$)
and with probability $r$, $x \sim \mathscr{D}_{B}$ (so $x \in B$).

Once we have drawn a data-point $x$, we model the true labels as being produced as follows; 
evaluate $h^{*}(x)$, using the classifier corresponding to the demographic group of $x$.
If $x \in A$, then $h^* (x) = h^{*}_{A}(x)$. If $x \in B$, then $h^*(x) = h^{*}_{B}(x)$.
However, we assume that $h^*$ is not perfect and independently with probability $\eta$, the true label of $x$ does
not correspond to the prediction $h^{*}(x)$.
\begin{align*}
y = y(x) = 
\begin{cases}
& \neg \; h^{*}(x) \quad \text{with probability} \quad \eta \\
& h^*(x) \quad \text{ w.p. } \quad 1-\eta \\
\end{cases}
\end{align*}
The labels $y$ after this flipping process
are the \textit{true labels} of the training data.\footnote{Note this label model is equivalent to
the Random Classification Noise model \cite{angluin}.
However the key interpretative difference is that in RCN, $h^*(x)$ is the correct label and those that get flipped are noise, but in our case 
the $y$ are the
true labels and $h^*$ is merely the Bayes optimal classifier given the %observable features.
observed features.
}
 We assume that $p=P(h^*_{A}(x) = 1 | x \in A) = P(h_{B}^{*}(x) = 1 | x \in B)$.
This combined with the assumption that $\eta$ is the same for classifiers from both groups implies that the two groups
have equal base rates (fraction of positive samples) i.e $p(1-\eta)+(1-p)\eta$ (un-normalized).
 
 We denote this label model as $(x,y) \sim P_{\mathscr{D},r}(h^*, \eta)$ for a pair of classifiers
$h^*= (h_{A}^*, h_{B}^*)$ with $h_{A}^*, h_{B}^* \in \mathscr{H}$
where $\mathscr{H}: \mathscr{X} \rightarrow \{ 0,1 \}$ is some hypothesis class with finite VC dimension.

\subsection{Biased Training Data}
\label{sec:underep}
Now we consider how bias enters the data-set. 
Consider the example of hiring where the main failure mode will be a classifier that is too negative on the disadvantaged group.
We explore several different bias models to capture potential ways the data-set could become biased. 

The first bias model we call \textbf{Under-Representation Bias}. 
In this model, the positive examples from Group $B$ are under-represented in the training data.  Specifically, the biased training data is drawn as follows:
\begin{enumerate}
\item $m$ examples are sampled from the distribution $\mathscr{D}$. 
Thus each $x \sim \mathscr{D}$. 
\item The label $y$ for each $x$ is generated according to the 
label process from Section \ref{truelabels} with hypothesis $h^*=(h_{A}^*, h_{B}^*)$ and $\eta$. 
\item For each pair $(x,y)$, if $x \in B$ and $y=1$, then the data-point $(x,y)$ is discarded from our training set independently with probability $1-\beta$. 
\end{enumerate}
Thus we see fewer positive examples from Group $B$ in our training data.
$\beta$ is the probability a positive example from Group $B$ stays in the training data and $  1 > \beta> 0 $.

If $\eta=0$, then the positive and negative regions of $h^{*}$ are strictly disjoint, so if we draw sufficiently many examples,
with high probability, we will see enough positive examples in the positive domain of $h^{*}$ 
to find a low empirical error classifier that is equivalent to $h^*$.\footnote{We would learn with ERM and Uniform Convergence, using the fact that 
$\mathscr{H}$ has finite VC-dimension.}

In contrast for non-zero $\eta$, our label model interacting with the bias model can induce a problematic phenomenon that fools the ERM classifier.
For non-zero $\eta$ there is error even for the Bayes optimal classifier $h^*$ and thus
in the %positive region of the classifier
region classified as positive by the Bayes optimal classifier $h^*$
there are positive examples mixed with negative examples. 
The fraction of negative examples is amplified by the bias process. 

\begin{figure}
  \centering
  \subfloat[Un-Corrupted Data]{{\includegraphics[width=0.5\textwidth]{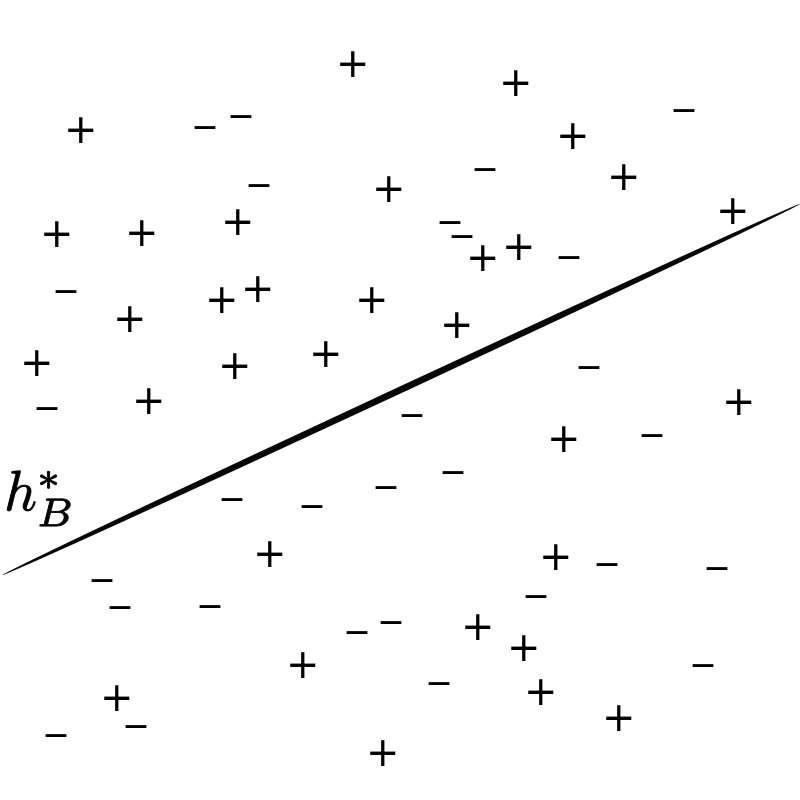}}}
\qquad
\subfloat[Corrupted Data: Under-Representation Bias]{{\includegraphics[width=0.5\textwidth]{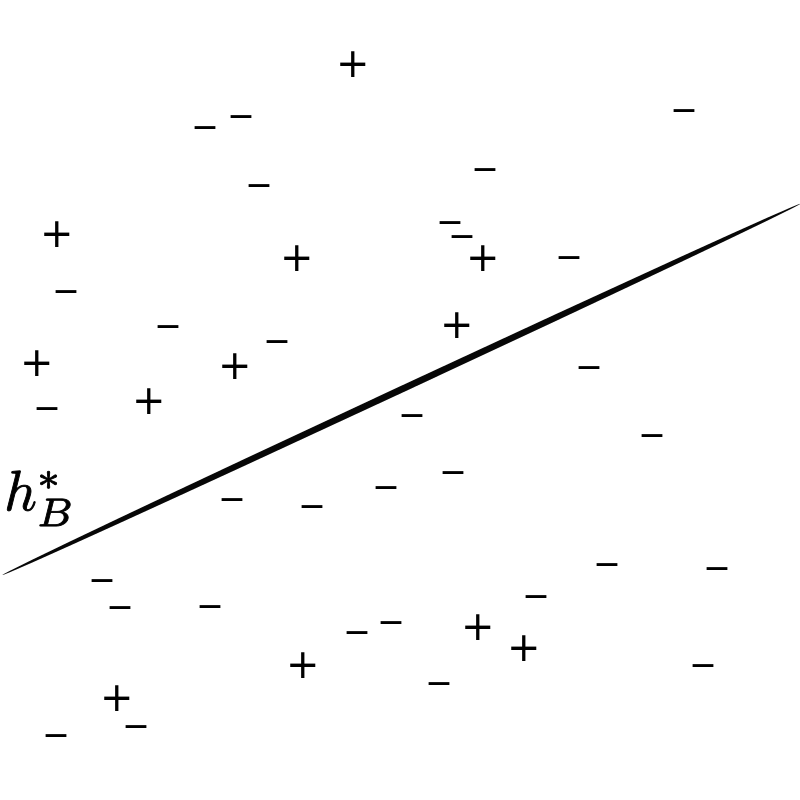}}}
\caption{The schematic on the left displays data points with $p=1/2$, $h^{*}_{B}$ as a hyperplane, and $\eta=1/3$. 
The schematic on the right displays  data drawn from the same distribution subject to the Under-Representation Bias with $\beta_{POS}=1/3$.
Now there are more negative examples than positive examples above the hyperplane so the lowest error hypothesis classifies all examples on the right as negative.}
\label{underrep}
\end{figure}

If $\beta$ is sufficiently small, there could in fact be more negative examples of Group B  than positive examples in the positive region of $h_B^*$.  If this occurs, then the bias model will snap the unconstrained ERM optimal hypothesis (optimal on the biased data) to classifying all individuals from Group $B$ as negatives. 
This can be observed in Figure \ref{underrep}.
             
Under-Representation Bias is related to selective labels in \cite{kleinbergselective} since we are learning on a filtered distribution where
the filtering process is correlated with the group label.
Our model is functionally equivalent to over-representing the negatives of the  in the training data, an empirical phenomenon observed in \cite{dirtydata}. 
In Chapter \ref{sec:multistagescreen}, we shall see another way in multiple classifiers with only small amounts of disparity will be amplified into a large amount of Under-Representation Bias.

 \subsection{Alternative Bias Model: Labeling Bias}
 \label{sec:labelbias}
We now consider a bias model that captures the notion of implicit bias, which we call \textbf{Labeling Bias.}  In particular, a possible source of bias in machine learning is the label generating process, especially in applications where the sensitive attribute can be inferred by the labeler, consciously or unconsciously.  
For example, training data for an automated resume scoring system could be based upon the historical scores of resumes created by a biased hiring manager or a committee of experts. 
This source of labels could then systematically score individuals from Group $B$ as having lower resume scores, an observation noted in\ randomized real world investigations  \cite{bertrand2004emily}.

Formally, the labeling bias model is:
\begin{enumerate}
\item $m$ examples are sampled from the distribution $\mathscr{D}$. 
Thus each $x \sim \mathscr{D}$. 
\item The labels $y$ for each $x$ are generated according to the 
label process from Section \ref{truelabels} with hypothesis $h^*=(h_{A}^*, h_{B}^*)$ and $\eta$. 
\item For each pair $(x,y)$, if $x \in B$ and $y=1$, then independently with probability $\nu$,  the label of this point is flipped to negative.
\end{enumerate}
This process is one-sided, so true positives become negatives in the biased training data, 
so apparent negatives becomes over-represented.  We are making a conceptual distinction that the \textit{true} labels (Step 2) 
are those generated by the original label model and these examples that get flipped
by the bias process (Step 3) are not really negative, instead they are just mislabeled. 

As $\nu$ increases, more and more of the individuals in the minority group
appear negative in the training data.
Once the number of positive samples is smaller than the number of negative samples above the decision surface $h_{B}^*$, then the optimal unconstrained classifier (according to the biased data) is to simply classify all those points as negative.
%%%%%%%%%%%%%%%%%%%%%%%%%%%%%%%%%%%%
\subsection{Under-Representation Bias and Labeling Bias} \label{ilfprime}
We now consider a more general model that combines Under-Representation Bias and Labeling Bias, and moreover we allow either positives or negatives of Group B (or both) to be under-represented. 
Specifically, we now have {\em three} parameters: $\beta_{POS}$, $\beta_{NEG}$, and $\nu$. Given $m$ examples drawn from $P_{\mathscr{D},r}(h^*, \eta)$, we discard each positive example of Group B with probability $1-\beta_{POS}$ and discard each negative example of Group B with probability $1-\beta_{NEG}$ to model the Under-Representation Bias. 
Next, each positive example of Group B is mislabeled as negative with probability $\nu$ to model the Labeling Bias. 
Note that the under-representation comes first: $\beta_{POS}$ and $\beta_{NEG}$ represent the probability of {\em true} positive and {\em true} negative examples from Group B staying in the data-set, respectively, regardless of whether they have been mislabeled by the agent's labelers.
\subsection{Fairness Interventions} 
Now we introduce several fairness interventions and define a notion of successful recovery from the biased training distribution. 

We consider multiple fairness constraints to examine whether the criteria have different behavior in different bias regimes.
The fairness constraints we focus on are \textbf{Equal Opportunity}, \textbf{Equalized Odds}, \textbf{Demographic Parity}, and \textbf{Calibration}.

\begin{definition}
Classifier $h$ satisfies  \textbf{Equal Opportunity} on data distribution $\mathscr{D}$ \cite {eodds} if 
\begin{align}
P_{(x,y) \sim \mathscr{D} }(h(x)=1 | y=1, x \in A) =  P_{(x,y) \sim \mathscr{D} } (h(x)=1 | y=1, x \in B) \label{Equal Opportunity} 
\end{align}
\end{definition}
This requires that the true positive rate in Group $B$ is the same as the true positive rate in Group $A$. 

\textbf{Equalized Odds} is a similar notion, also introduced in \cite{eodds}. In addition to requiring Line \ref{Equal Opportunity}, Equalized 
Odds also requires that the false positive rates are equal across both groups.
Equivalently, we can define \textbf{Equalized Odds}
as $h \perp A | Y$, meaning that $h$ is independent of the sensitive attribute, conditioned on the true label $Y$.
We also consider \textbf{Demographic Parity} := $P(h(x)=1 | x \in A) = P(h(x)=1 | x \in B)$ \cite{dwork12}.
For each of these criteria, the overall training procedure is solving a constrained ERM problem.\footnote{We do not consider methods for efficiently solving the constrained ERM problem.}

An alternative intervention we study \textbf{data Re-Weighting}, where we 
change the training data distribution to correct for the bias process and then do ERM on the new distribution. 
The overall gist of how the training data becomes biased in our models is that the positive samples from Group $B$
are under-represented in the training data so we can intervene by up-weighting the observed fraction of positives 
in the training data from Group $B$ to match the fraction of positives from the Group $A$ training data. 

In the training process we only have access to samples from the training distribution and thus when
using a fairness criterion to select among models \emph{we check the requirement on the biased training data}.

%\textcolor{red}{
The last fairness intervention we consider is \textbf{Calibration}. 
Calibration \cite{flores, dieterich2016compas,chouldechova,pleiss2017fairness} requires that when interpreted as probabilities, the same score communicates the same information for individuals from different demographic groups.
%mean what they claim to.
Specifically, in the bucket of individuals receiving score $s$, the same fraction in both demographic groups is in fact truly positive. We focus on Calibration for the case of our binary classifier where there are only two scores, e.g. the scores $0$ and $1$, so in order for classifier $h=(h_A, h_B)$ to satisfy Calibration, the following equalities must hold.
\footnote{If one of the conditioned on events never occurs, such as a classifier that never classifies anyone from Group B as positive, we treat the associated equality as satisfied.}
\begin{align*}
& P_{x \sim \mathscr{D}_A}(y=1 | h_A (x)=1 ) = P_{x \sim \mathscr{D}_B}( y = 1 | h_B (x)=1)   \\ %P_{x \sim \mathscr{D}_A}(y=1 | h_A (x)=1 ) = P_{x \sim \mathscr{D}_B}( y = 1 | h_B (x)=1)
& P_{x \sim \mathscr{D}_A}(y=1 | h_A (x)=0 ) = P_{x \sim \mathscr{D}_B}( y = 1 | h_B (x)=0) %P_{x \sim \mathscr{D}_A}(y=1 | h_A (x)=0 ) = P_{x \sim \mathscr{D}_B}( y = 1 | h_B (x)=0) 
\end{align*}

While the other fairness criteria are vigorously debated, Calibration is less contested as an important desiderata of machine learning models. 
Calibration has been used to defend the epistemic validity of risk prediction instruments 
\cite{flores,dieterich2016compas} and it is claimed that mis-calibrated classifiers may have serious harms and 
induce undesirable behavior when scores are used by a human actor \cite{pleiss2017fairness}.

Observe that in our model of label generation, the Bayes optimal classifier on the true distribution is the $h^*$ used to generate the labels initially, regardless of the values of $\eta$ and $r$.
Thus our goal for the learning process is to 
recover the original optimal classifier $h^*$, subject to training data from a range of bias models and the true label process with $(x,y) \sim P_{\mathscr{D},r}(h^*, \eta)$.
A more effective learning method would recover $h^*$ in a wider range of the model parameters (the parameters that characterize the bias process and the true label process). 
Accordingly we define \textbf{Strong-Recovery}$(r,\eta)$:
\begin{definition}
A Fairness Intervention in bias model $B$ satisfies 
Strong-Recovery$(r_0 , \eta_0)$ if for all $\eta \in [0,\eta_0)$ and all $0 < r < r_0$, when given data
corrupted by bias model $B$, the training procedure recovers
the Bayes optimal classifier $h^*$, given sufficient samples, for all $\beta_{POS}, \beta_{NEG} \in (0,1]$, $\nu \in [0,1)$, and $p \in (0,1]$.
\end{definition}

\section{Recovery Behavior Across Bias Models}
\label{sec:recoverysummary}
There are two failure modes for learning a fairness constrained classifier that we will need to be concerned with. 
First, the Bayes optimal hypothesis may not satisfy the fairness constraint evaluated on the biased data.
Second, within the set of hypotheses satisfying the fairness constraint,
another hypothesis (with higher error on the true distribution) may have lower error than the Bayes optimal classifier $h^{*}$ on the biased data.
We now describe how the multiple fairness interventions provably avoid or fail to avoid these pitfalls in increasingly complex bias models. We defer formal proofs to Section \ref{overview}.

\subsection{Under-Representation Bias} \label{sec:underrep}
Equal Opportunity and Equalized Odds both perform well in this bias model and avoid both failure modes, subject to an identical constraint on the bias and demographic parameters.

First, from the definition of the Under-Representation Bias model, observe that $h^*$ satisfies both fairness notions on the biased data, so the first failure mode does not occur.

Second, Equal Opportunity intuitively prevents the failure mode where a hypothesis is produced that appears better than $h^*$ on the biased data, such as classifying all examples from Group $B$ as negative, by forcing the two classifiers to classify the same fraction of positive examples as positive.  So, 
if we classify all the examples from Group B as negative, we have to do the same with Group A, inducing large
error on the training data from the majority Group A. 
In particular, so long as the fraction $r$ of total data from Group B is not too large and $\eta$ is not too close to $1/2$, this will not be a worthwhile trade-off for ERM (saying negative on all samples will not have lower perceived error on the biased data than $h^{*}$) and so it will not produce this outcome.

A formal proof of correctness is given in Section \ref{maintheoremproof}.
Specifically, we prove that Equal Opportunity strongly recovers from  Under-Representation Bias so long as 
\begin{align}
(1-r)(1-2\eta) +r((1-\eta)\beta-\eta) > 0 \label{Equal Opportunityineq}
\end{align}
Note that this is true for all $\eta<1/3$ and $r \in (0,1/2)$, so we have that Equal Opportunity satisfies Strong-Recovery($1/2$,$1/3)$ from 
Under-Representation Bias.  
Alternatively, we see that if $r = 1/4$ then the inequality simplifies to at least $3/4 (1-2 \eta) - \eta/4 = 3/4-(7/4) \eta$ 
so we have Strong-Recovery$(1/4,3/7)$.
Equalized Odds also recovers in this bias model with the same conditions as Equal Opportunity. 

\begin{figure}[ht]
\label{regions}
  \centering
  \includegraphics[width=0.6\textwidth]{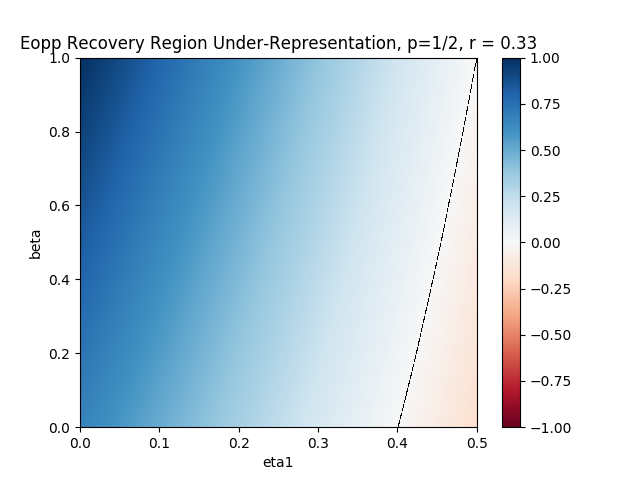}
  \caption{This figure indicates the parameter region such that Equal Opportunity Constrained ERM recovers $h^*$
  under the Under-Representation Bias Model and is a visualization of Equation \ref{Equal Opportunityineq}. $r=1/3$ and $p=1/2$. We label each pair $(\eta, \beta)$ with blue if it satisfies the inequality and red otherwise. This plot shows how smaller $\eta$ means we can recover from lower $\beta$. 
  Blue means $h^*$ is recovered.
  The dashed black line indicates the boundary between recovering $h^*$ and failing to recover $h^*$.}
\end{figure}

In contrast, Demographic Parity fails to recover $h^*$ even if $\eta=0$.
If $p=1/2$, $\eta=0$, and $\beta=1/2$ and we originally had $n$ samples, then
the Bayes optimal classifier does not satisfy Demographic Parity on the biased data
since the fraction of samples that will be labelled positive is $\frac{1}{3} \neq \frac{1}{2}$. 

Similarly, if we let 
$\eta \neq 0 , \beta < 1 $,
then in order to match the fraction of positive classifications made by $h_{A}^{*}$,
$h_{B}$ is forced to classify a larger region of the input spaces as positive than $h_{B}^*$ would in the absence of biased data and so we do not 
recover $h_{B}^{*}$.

Another way to intervene in the Under-Representation Bias model
would just be to re-weight
 the training data to account for the under-sampling of positives from Group $B$.
If we really know positives from Group $B$ are under-represented, we can change our objective function 
$ min \sum_{i=1}^{m} I(h(x) \neq y) $
 by changing each indicator function such that minimizing the sum of indicators measures the loss on the true distribution 
 and not the loss on the biased training distribution. 

Define $B^{+} = \{ x \in B \; s.t.\; y=1 \}$. 
Then let, 
\[
I'(h(x),y) = 
\begin{cases}
\frac{1}{\beta} \quad h(x) \neq 1 \quad and \quad x \in B^{+} \\
0 \quad  h(x) = 1 \quad and \quad x \in B^{+} \\
I(h(x) \neq y) \quad otherwise 
\end{cases}
\]
Then we use this new indicator in the objective function.  
This new loss function is an unbiased estimator of the true unbiased risk, so uniform convergence on this estimator will suffice to learn $h^*$.
We can infer the value of $\beta$ from the data for Group A if we know the data from Group B is corrupted by this bias model.  One concern  with re-weighting in general is that the functional form of the correction is tied to the exact bias model.

As we show 
in Section \ref{calib}, Calibration has strange results in this bias model. 
Specifically, when the bias is such that ERM fails to recover $h^*$ (i.e when $(1-\eta)\beta < \eta$), then the Calibration constraint can only be satisfied by a trivial classifier that assigns all of Group $A$ to one label and all of Group $B$ to the alternative label. 
For typical parameters, this will result in Group $B$ being given the negative label and Group $A$ will be
assigned as all positive.
This will not recover $h^*$ and is in fact substantially worse than merely using ERM. 
Un-constrained ERM would learn badly on Group $B$ but would recover $h_{A}^{*}$ for Group $A$. %\\ \\

When the bias regime is such that $(1-\eta)\beta > \eta$, plain ERM recovers $h^*$, while enforcing Calibration will lead to excess true error 
on both demographic groups over the true error of $h^{*}$.
In particular, satisfying Calibration on the biased data requires
intentionally classifying some negative input space from Group $A$ as positive and classifying some positive input space from Group $B$ as negative.
These results suggest that Calibration is an actively harmful intervention (for both groups) in our model, when compared to plain ERM, across all model parameters.

 In summary, for the Under-Representation Bias model, 
 the fairness interventions Equalized Odds, Equal Opportunity, and Re-Weighting recover $h^*$ under a range of parameters. 
 However, Demographic Parity is inadequate even for $\eta=0$ and will not recover $h^*$ for non-vacuous bias parameters. 
%%%%%%%%%%%%%%%%%%%%%%%%%%%%%%%%%%%%%%%%
\subsection{Labeling Bias} \label{labelbias}
In Section \ref{overview}, we prove that Equal Opportunity constrained ERM on 
data biased by the Labeling Bias model
 also finds the Bayes optimal classifier, under similar parameter conditions to the previous bias model.

Interestingly, in contrast to Under-Representation Bias, \textit{Labeling Bias cannot be corrected} 
by \textit{Equalized Odds}.  
The problem is the first failure mode. 
For example, consider $\eta =  0$ but where $\nu \neq 0$.
The Bayes optimal classifier $h_{A}^{*}$ for Group $A$ has false positive rate of 0 and true positive rate of $1$.
However, since $\nu>0$, there is no classifier for Group $B$ that achieves both of these rates simultaneously. In particular, the only way to classify the negative individuals in the positive region as negative is for the classifier to decrease its true positive rate from $1$. Therefore, Equalized Odds rules out usage of $h_A^*$.  This violation holds for $\eta \neq 0$ as well. 

%Thus, the Bayes Optimal Classifier $h^* = (h_{A}^*, h_{B}^*)$ does not satisfy Equalized Odds on this corrupted data set. 
In contrast, $h^*$ does satisfy Equal Opportunity on the biased data, and given the conditions in Theorem \ref{maintheorem}, it will be the lowest error such classifier on the biased data. 
%the lowest error classifier satisfying Equal Opportunity is the Bayes Optimal Classifier, $h^*$. 

When just Labeling Bias is present, observe that $h^{*}$ still satisfies 
Demographic Parity on the biased data, since in contrast to the Under-Representation Bias case, the positives that are flipped to negative still appear in the training data.
In this case, Demographic Parity will experience strong recovery when $(1-r)*(1-2\eta) + r((1-eta)(1-2\nu)-\eta) > 0$. 
This inequality  is a simple variation of the first inequality in Theorem \ref{maintheorem}, and a simplification of that proof will yield this result, if the only present bias is Labeling Bias.

The Re-Weighting intervention 
%from the first model 
is to change the weighting of observed positives in the training data for Group $B$
so that we have the same  fraction of positives in Group $B$ as in Group $A$.
Define $p_{A,1}:= $ the fraction of positive individuals in Group $A$ and $p_{B,1}:=$ the \emph{observed} fraction of positives in $B$ in the biased data. $p_{A,0}$ and $p_{B,0}$ refer to the observed fraction of negative individuals in Group $A$ and Group $B$ in the biased data.

We need a re-weighting factor $Z$ such that:
\begin{align*}
 \frac{p_{A,1}}{p_{A,0}} &= \frac{Z p_{B,1} }{p_{B,0}} \\
  \frac{p_{A,1}}{1-p_{A,1}} &= \frac{Z p_{A,1}(1-\nu)}{p_{A,0} + p_{A,1}\nu} = \frac{Z p_{A,1}(1-\nu)}{1-p_{A,1} + p_{A,1} \nu } \\
  %\frac{p_{A,+}}{1-p_{A,+}} &= \frac{Z P_{A,+}(1-\eta_2)}{1-P_{A,+} + p_{A}\eta_{2}}   =  \frac{Z P_{A,+}(1-\eta_2)}{1-P_{A,+}(1 -\eta_{2})}  \\
 Z &= \frac{1-p_{A,1}(1-\nu) }{(1-\nu)(1-p_{A,1})}  
\end{align*}
%Using the fact that the positive examples are flipped from positive to negative at the same rate in all the regions of the input space, 
We prove in Section \ref{reweightarg} that this correction factor will lead to the positive region of $h_{B}^{*}$ having a higher weight of positive examples than negative examples and simultaneously the negative region of $h_{B}^{*}$ having a higher weight of negative examples than positive examples. 
%positive to negative samples from Group $B$ in the regions classified as positive and negative by $h^*$, 
This causes ERM to learn the optimal hypothesis $h^{*}$.  We can infer the value of $\nu$ by comparing the fraction of positives in Group $A$ and Group $B$. 

In summary, Equal Opportunity, Demographic Parity, and the Re-Weighting Interventions recover well in this bias model (Labeling Bias)
while Equalized Odds is inadequate.
%More details are in the Appendix in Section \ref{reweightarg}. 

%%%%%%%%%%%%%%%%%%%%%%%%%%%%%%%%%%%%
\subsection{Under-Representation Bias and Labeling Bias} 
\label{sec:reweightinglb}
In this most general model that combines the two previous models, Re-Weighting the data is now no longer sufficient to recover the true classifier.  For example, consider the case where $\eta=0$ and $p=1/4$, $\nu=1/2$ and $\beta_{NEG}=1/3$ and $\beta_{POS} = 1 $. 
If there were $n$ points originally from group $B$, 
then in expectation $3n/4$ were negative and $n/4$ were positive.
After the bias process, in expectation there are $n/4$ negatives on the negative side of $h^*$, 
and on the positive side of $h^*$ we have $n/8$ correctly labelled positives and what appear to be $n/8$ negative samples.

The Re-Weighting intervention will not do anything in expectation because the overall fractions are still correct; we have $n/2$ total points
with one quarter of them labeled positive.
ERM is now indifferent between $h^*$ and labeling all samples from Group $B$ as negative. 
If we just slightly increase the parameter $\nu$ and reduce $\beta_{POS}$ then in expectation ERM will strictly prefer labeling
all the samples negatively. 

While the Re-Weighting  method fails, we prove that Equal Opportunity-constrained ERM recovers the Bayes optimal classifier $h^*$ as long as we satisfy a condition ensuring that Group A has sufficient mass and the signal is not too noisy.
 As with the previous models, Demographic Parity and Equalized Odds 
are not satisfied by $h^*$ on minimally biased data and so they will not recover the Bayes optimal classifier.

\section{Main Results} \label{overview}
We now present our main theorem formally.
Define the biased error of a classifier $h$ as its error rate computed on the biased distribution. 
\begin{theorem} \label{maintheorem}
Assume true labels are generated by  $P_{\mathscr{D},r}(h^*, \eta)$ corrupted by both Under-Representation bias and Labeling bias with parameters $\beta_{POS}, \beta_{NEG},\nu$, and assume that
\begin{align}
 (1-r)(1-2\eta) + & r( (1-\eta)\beta_{POS} (1-2\nu) - \eta \beta_{NEG}) > 0 \label{proofc1} \\
& \quad  \text{and} \nonumber \\
(1-r)(1-2\eta) + & r ((1-\eta) \beta_{NEG} -(1-2\nu )\beta_{POS} \eta ) > 0 \label{proofc2}
\end{align}

Then $h^*=(h_{A}^{*}, h_{B}^{*})$ is the lowest biased error 
classifier satisfying Equality of Opportunity on
the biased training distribution and thus $h^*$ is recovered by Equal Opportunity constrained ERM. 

Note $\beta_{POS}, \beta_{NEG} \in (0,1]$, $\nu \in [0,1)$, $\eta \in [0,1/2)$, $r \in (0,1)$ and $p \in (0,1]$.
Condition \ref{maintheorem} refers to Equation \ref{proofc1} and Equation \ref{proofc2}.
\end{theorem}

This case contains our other results as special cases and in the next section we prove our main theorem in this bias model.
Note that if Equation \ref{proofc1} is not satisfied then the all-negative hypothesis will have the lowest biased error among hypotheses satisfying
Equal Opportunity on the biased training distribution. 
Similarly, if Equation \ref{proofc2} is not satisfied then the all-positive hypothesis will have the lowest biased error among hypotheses satisfying
Equal Opportunity on the biased training distribution. 
Thus Theorem \ref{maintheorem} is tight. 
%needed since it is possible for the unconstrained classifier to be tricked by an over-representation of positives. 
%To see this, imagine the case when $\beta_{POS} >> \beta_{NEG}$.
To give a feel for the formula in Theorem \ref{maintheorem}, note that the case of small $r$ 
is {\em good} for our intervention, because the advantaged Group $A$ is large enough to pull the classification of the disadvantaged Group $B$ in the right direction. For example, if $r \leq \frac{1}{3}$ then the bounds are satisfied for all $\eta < \frac{1}{4}$ (and if $r \leq \frac{1}{4}$ then the bounds are satisfied for all $\eta < \frac{1}{3}$) for {\em any} under-representation biases $\beta_{POS},\beta_{NEG}>0$ and {\em any} labeling bias $\nu<1$.

Thus, Equal Opportunity Strongly Recovers with $(1/4,1/3)$ and $(1/3,1/4)$ in the Under-Representation and Labeling Bias model. 

Table \ref{summarytab} summarizes the results in the three core interventions and the three core bias models.
%Demographic Parity is omitted from the table since it cannot recover under the bias models when $\eta = 0$ and thus is inadequate. 
The contents of each square indicate if recovery is possible in a bias model with an intervention 
and what constraints need to be satisfied for recovery.
%and if recovery is possible the so what constraints on the parameters need to be satisfied. 
\begin{table}[ht]
\centering
    \begin{tabular}{ |p{2cm}|p{2cm}|p{2cm}|p{2cm}| }
    \hline
	Intervention   & Under-Representation & Labeling Bias & Both   \\ \hline
	Equal Opportunity-ERM & Yes: $(1-r)(1-2\eta) +r((1-\eta)\beta-\eta) > 0$
 & Yes:  $(1-r)(1-2\eta) + r ( (1-\eta)(1-2\nu) -  \eta) > 0 $ & Yes: Using Condition \ref{maintheorem} \\ \hline
	Equalized Odds & Yes: $(1-r)(1-2\eta) +r((1-\eta)\beta-\eta) > 0$ & No & No \\ \hline
	Re-weighting Class B: & Yes & Yes & No \\
    \hline
    Demographic Parity : & No & Yes & No \\    \hline
    \end{tabular}
    \caption{Summary of recovery behavior of multiple fairness interventions in bias models.}
    \label{summarytab}
\end{table}

\vspace*{-0.2in}
\subsection{Proof of Main Theorem} \label{maintheoremproof}
In this section we present the proof of the main result, \textbf{Theorem} \ref{maintheorem}.
We want to show that the lowest biased error classifier satisfying Equal Opportunity on the biased data is $h^*$, given Condition \ref{maintheorem}.

The first step of the proof is to show that $h^*$ satisfies Equal Opportunity on the biased training data.
Note: the lemmas and claims here are all in the Under-Representation  Bias combined with Labeling Bias Model, the most general bias model. 
\begin{lemma}\label{lemma:bayesopt}
$h^*=(h^*_{A}, h_{B}^*)$ satisfies Equal Opportunity on the biased data distribution. 
\end{lemma}

\begin{proof}
First, let's consider the easiest case with $\eta=0$, $\beta_{POS}=\beta_{NEG}=1$, and $\nu=0$.  
Recall that $h^*$ is the pair of classifiers used to generate the labels. 
When $\eta=0$, $h^*$ is a perfect classifier for both groups so Equal Opportunity is trivially satisfied.  
Now, let's consider arbitrary $0 \leq \eta < 1/2$.  
Recall that $p=Pr_{\mathscr{D}_A}(h_{A}^*(x) = 1 | x \in A) = Pr_{\mathscr{D}_B}(h_{B}^*(x) = 1 | x \in B) $.

By our assumption that Group A and Group B have equal values of $p$ and $\eta$ we have 
\begin{align*} 
&\Pr(h^*_A(x)=1|Y=1, x \in A) = \frac{p(1-\eta)}{p(1-\eta)+(1-p)\eta} \\
&= \Pr(h^*_B (x) =1|Y=1, x \in  B) 
\end{align*}

Next consider when we have both Under-Representation Bias and Labeling Bias.
Recall that $\beta_{POS}, \beta_{NEG}>0$ is the probability that a positive or negative sample from Group $B$ is \emph{not filtered} out of the training data
while $\nu<1$ is the probability a positive label is flipped and this flipping occurs after the filtering process. 
Then,
\begin{align*}
& \{ \text{True Positive Rate on Group A} \} := \Pr(h^*_A(x)=1|Y=1, x \in A) = \\
&\frac{p(1-\eta)}{p(1-\eta)+(1-p)\eta} =
\frac{p(1-\eta)\beta_{POS}(1-\nu)}{p(1-\eta) \beta_{POS}(1-\nu)+(1-p)\eta \beta_{POS}(1-\nu)}   \\
&= \Pr(h^*_B (x) =1|Y=1, x \in  B) := \{ \text{True Positive Rate on Group B} \}
\end{align*}
so Equal Opportunity is still satisfied.  

In words, the bias model removes or flips positive points from Group $B$ independent of their 
location relative to the optimal hypothesis class. 
Thus positive points throughout the input space are
are equally likely to be removed, so the overall probability of true positives being classified as positives is not changed.
\end{proof}

\begin{figure}[ht]
  \centering
  \includegraphics[trim = 0 90 0 90, clip, width=0.5\textwidth]{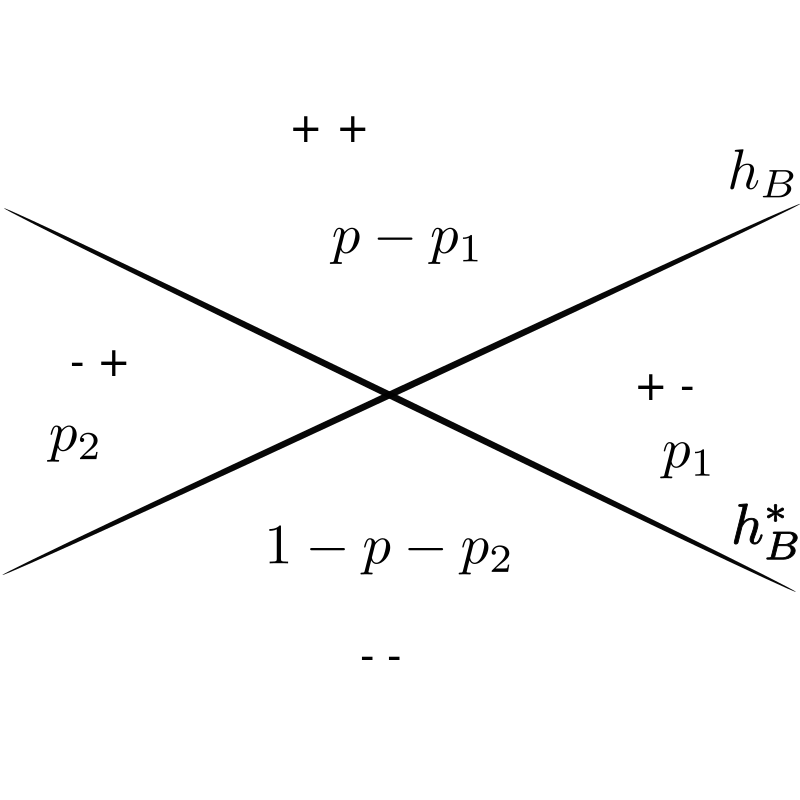}
  \caption{Differences between $h_B$ and $h_{B}^*$ \label{regionsh*}
  measured with probabilities in the true data distribution  (before the effects of the bias model).}
\end{figure}

Now we describe how a candidate classifier $h_{B}$ differs from $h^*_{B}$.
We can describe the difference between the classifiers by noting the regions in the input space that each classifier gives a specific label. 
This gives rise to four regions of interest with probability mass as follows:
\begin{align*}
    & p_{1B} = P_{1B}(h_{B}):= P_{x \in \mathscr{D}_B} (h_{B}^{*}(x)=1 \land h_{B} (x) =0) \\
    &  p_{2B} = P_{2B}(h_{B}) := P_{x \in \mathscr{D}_B}(h_{B}^{*}(x)=0 \land h_{B} (x) =1) \\
    & p-p_{1B} = P_{x \in \mathscr{D}_B}(h_{B}^{*}(x)=1 \land h_{B} (x) =1) \\
    & 1-p-p_{2B} = P_{x \in \mathscr{D}_B}(h_{B}^{*}(x)=0 \land h_{B} (x) =0) 
\end{align*}
These probabilities are made with reference to the regions in input space \emph{before} the bias process. 
$p_{1B}$ and $p_{2B}$ are functions of $h_{B}$ to make explicit that there may be multiple hypotheses with different functional forms that could allocate the same amount of probability mass to parts of the input space where $h_{B}^{*}$ and $h_{B}$ agree on labeling as  positive and negative respectively.
The partition of probability mass into these regions is easiest to visualize for hyperplanes but will hold with other hypothesis classes.
$p_{1A}$ and $p_{2A}$ are defined similarly with respect to $h_{A}^*$ and $\mathscr{D}_A$. 
A schematic with hyper-planes is given in Figure \ref{regionsh*}.
To show that $h^*$ has the lowest error on the true distribution, we first show how given any pair of classifiers $h_A$ and $h_B$, which jointly satisfy Equal Opportunity (Equal Opportunity) on the biased distribution, we
can transform $\{ h_{A}, h_{B} \}$ into a pair of classifiers still satisfying Equal Opportunity with at most one non-zero parameter from  $\{p_{1B}, p_{2B} \}$, and at most one non-zero parameter from $\{p_{1A}, p_{2A} \}$, while also not increasing biased error. 

The final step of our proof argues that out of the family of all hypotheses with (1) at most one non-zero parameter for the hypothesis on Group $A$, (2) at most one non-zero parameter for the hypothesis on Group $B$, (3) and jointly satisfying Equal Opportunity on the biased data, $h^*$ has the lowest biased error. 

These steps combined imply that $h^*$ is the lowest biased error hypothesis that satisfies Equal Opportunity. 

\begin{lemma}
\label{shrink}
Given classifiers $h_A$ and $h_B$ which satisfy Equal Opportunity on the biased data, there exist classifiers $h_{A}^{'}$ and $h_{B}^{'}$ (not necessarily in $\mathcal H$) satisfying 
\begin{enumerate}
    \item At most one of $\{P_{1A}(h_{A}^{'}),P_{2A}(h_{A}^{'})\}$ is non-zero and at most one of  $\{P_{1B}(h_{B}^{'}),P_{2B}(h_{B}^{'}) \}$ is non-zero.
    \item $(h_{A}^{'}, h_{B}^{'} )$ has error at most that of $(h_A, h_B)$ on the biased distribution. 
    \item $h_{A}^{'}$ and $h_{B}^{'}$ satisfy Equal Opportunity. 
\end{enumerate}
\end{lemma}
\begin{proof}
We want to exhibit a pair of classifiers with lower biased error that zeros out one of the parameters. 
We do this by modifying each classifier separately, while keeping the true positive rate on the biased data fixed to ensure we  satisfy Equal Opportunity.

First, consider Group $A$ and suppose that $P_{1A}(h_{A}), P_{2A}(h_{A}) > 0 $ since otherwise we do not need to modify $h_A$.
Imagine holding the true positive rate of $h_A$ constant and shrinking $p_{2A}$ towards zero.
As we shrink $p_{2A}$, we must shrink $p_{1A}$ towards zero in order hold the true positive rate fixed (and thus satisfy Equal Opportunity). 

The un-normalized\footnote{The normalization factor for these rates for Group $A$ and Group $B$ is the same so this term can be cancelled.} True Positive Rate (constrained by Equal Opportunity) is
$(p-p_{1A})(1-\eta) + p_{2A} \eta = p(1-\eta) -p_{1A}(1-\eta) + p_{2A} \eta   = (p-p_{1B})(1-\eta) + p_{2B} \eta$. 
Since the $p(1-\eta)$ term is independent of the classifier $h_A$, keeping the true positive rate constant is equivalent to keeping $C:= -p_{1A}(1-\eta) + p_{2A} \eta $ constant.  

Define $f(\Delta) = \Delta\frac{\eta}{1-\eta}$.
If $C \leq 0$ then we can shrink $p_{2A}$ to 0 and reduce $p_{1A}$ by $f(p_{2A})$, keeping $C$ constant.  If $C \geq 0$ we can instead shrink $p_{1A}$ to 0 and reduce $p_{2A}$ by $f^{-1}(p_{1A})$.

\iffalse
Let $\Delta$ be the amount we shrink $p_{2A}$ and let $f(\Delta)$ be the amount we must shrink $p_{1A}$ to keep $C$ fixed. 
Then, 
\begin{align*}
%& (p_{2A}-\Delta)\eta -(p_{1A}-f(\Delta))(1-\eta) = C  \\
%& p_{2A} \eta - \eta \Delta - p_{1A} (1-\eta) + f(\Delta) (1-\eta) = C \\
%& C - \eta \Delta + f(\Delta) (1-\eta) = C \\
& f(\Delta) = \Delta \frac{\eta}{1-\eta}.
\end{align*}
We continue shrinking these parameters until either $p_{1A}$ or $p_{2A}$ has hit zero. 
To this see can be always be done, note that $p_{1A} = -C' + p_{2A} \frac{\eta}{1-\eta}$ for $C'= \frac{C}{1-\eta}$. 
Which term hits zero will depend on the sign of $C$. 
\fi

Observe for Group $A$ this process will clearly reduce training error since we are decreasing both $p_{1A}$ and $p_{2A}$ and the error on group $A$ is monotone increasing (and linear) with respect to $p_{1A}+p_{2A}$. 

We then separately do this same shrinking process for group $B$.
Now we show the biased error decreases for Group $B$.  For a given amount $\Delta$ by which we shrink $p_{2B}$,
\begin{comment}
\begin{align*}
& \text{Change in Biased Error on }  =  \Delta[\eta \beta_{POS} (1-\nu) - (1-\eta) \beta_{NEG} - \eta \beta_{POS} \nu] \\
& \text{Change in Biased Error after Shrinking} P_{1B}  =
f(\Delta)[\eta \beta_{NEG} + (1-\eta)\beta_{POS} \nu - (1-\eta) \beta_{POS}(1-\nu)]
\end{align*}
\end{comment}
%Thus 
the overall biased error change for Group $B$ is $\Delta[\eta \beta_{POS} (1-\nu) - (1-\eta) \beta_{NEG} - \eta \beta_{POS} \nu] + f(\Delta)[\eta \beta_{NEG} + (1-\eta)\beta_{POS} \nu - (1-\eta) \beta_{POS}(1-\nu)]$, %of %the two above terms %
and simplifies to become
\begin{multline*}
    = \Delta \eta \beta_{POS} (1-\nu) - f(\Delta) (1-\eta) \beta_{POS} (1-\nu)  \\ 
 + \Delta(- (1-\eta) \beta_{NEG} - \eta \beta_{POS} \nu)  + f(\Delta)( \eta \beta_{NEG} + (1-\eta) \beta_{POS}\nu)
\end{multline*}
 The first two terms vanish because of $f(\Delta)= \Delta \frac{\eta}{1-\eta}$.
 \begin{align*}
 & = \Delta(- (1-\eta) \beta_{NEG} - \eta \beta_{POS} \nu) + 
 f(\Delta)( \eta \beta_{NEG} + (1-\eta) \beta_{POS}\nu)\\ 
 & =\Delta(- (1-\eta) \beta_{NEG} - \eta \beta_{POS} \nu) + 
 \Delta\frac{\eta^2}{1-\eta} \beta_{NEG} + \Delta \eta \beta_{POS}\nu\\ 
 &= \Delta ( \frac{\eta^2}{1-\eta} \beta_{NEG} - (1-\eta) \beta_{NEG}) < 0
 \end{align*}
Since this term is negative, we have shown that this modification process decreases error on the biased training data for both Group $A$ and Group $B$ while keeping the true positive rate fixed.  $h_{A}^{'}$  and $h_{B}^{'}$ are then any functions satisfying these $p$'s (e.g. $p_{1A}, p_{2A}$ etc).
\end{proof}

\begin{lemma} \label{delta}
If $h_A$ and $h_B$ satisfy the Equal Opportunity constraint and each classifier has at most one non-zero parameter, then $p_{1B} = p_{1A} $ and $p_{2B}=p_{2A}$.
\end{lemma}

\begin{proof}
Recall that the Equal Opportunity constraint requires that these expressions be equal.
\begin{align*}
    & (p-p_{1A})(1-\eta)  + p_{2A} \eta = (p-p_{1B})(1-\eta) + p_{2B} \\
    &  p_{2A} \eta - p_{1A} (1-\eta) =  p_{2B} \eta - p_{1A}(1-\eta)
\end{align*}
Then the theorem follows from inspecting the second equality.  
\end{proof}
This lemma makes explicit that when the classifiers each have only one non-zero parameter and satisfy Equal Opportunity, then the non-zero parameter corresponds to the same region. 

%Combining  Lemmas \ref{shrink} and  \ref{delta}, we can take a pair of classifiers $(h_A, h_B)$ and modify them to a classifier that 
%satisfies equal opportunity while error has strictly decreased and only one parameter from each classifier is non-zero (and this parameter) is the same for each classifier. 
\begin{lemma}
Of hypotheses satisfying
 ($p_{1A}=p_{1B}$ and $p_{2A}=p_{2B}=0$) or 
($p_{1A}=p_{1B}=0$ and $p_{2A}=p_{2B}$), if these inequalities hold:
\begin{align*}
 (1-r)(1-2\eta) + & r( (1-\eta)\beta_{POS} (1-2\nu) - \eta \beta_{NEG}) > 0 \\
& \quad  \text{and} \\
 (1-r)(1-2\eta) + & r ((1-\eta) \beta_{NEG} -\eta \beta_{POS} (1-2\nu ) ) > 0 
\end{align*} 
then the lowest biased error classifier satisfying Equal Opportunity on the biased data is $h^*=(h_{A}^{*}, h_{B}^{*})$.
\end{lemma}

\begin{proof}
First, we sketch the proof informally. 
Consider three cases which depend on how the bias process affects the unconstrained optimum for Group $B$ on the biased data. 
In the first case, in the biased data distribution, the region $X^{+} :=  \{ x \; s.t. \;  h_{B}^{*} (x) = 1 \}$ has more positive than negative samples in expectation and the region $X^{-} :=  \{ x \; s.t. \;  h_{B}^{*} (x) = 0 \}$ has more negative than positive samples in expectation.
 In the second case, there are more positive than negative samples throughout the entire input space  in the biased distribution.
 In the third and final case, there are more negative than positive samples throughout the input space in the biased distribution. 

In these three cases, the optimal hypothesis is exactly one of $\{ h_{B}^*, h_{B}^{1}, h_{B}^{0} \}$, respectively.  
The second two hypotheses mean labelling all inputs as positive and labelling all inputs as negative, respectively.
These three hypotheses correspond to hypotheses with at most one non-zero parameter. 

For instance,  $h_{B}^{1}$ occurs when $p_{2B}=1-p$ and $p_{1B}=0$.
Each of the three hypotheses occur when the one non-zero parameter attains a location on the boundary of its range of values.
When $p_{2B}$ is allowed to be non-zero, if instead $p_{2B}=0$ (and thus it also must be that $p_{1B}=0$), the hypothesis is equivalent to $h_{B}^{*}$. 
A similar relationship holds for $h^{0}$ and $p_{1}$. 

In order to show the theorem, we prove that if $h^*$ has lower biased error than $h^{1}=(h_{A}^{1}, h_{B}^{1})$ and $h^{0}=(h_{A}^{0}, h_{B}^{0})$ on the biased data distribution, then $h^{*}$ has the lowest error among all hypotheses with at most one non-zero parameter and satisfying Equal Opportunity.

To see this, consider $h_A$ and $h_B$ with the same non-zero parameter equal to $\Delta$.
Then the error of $h_A$ is a linear function of $\Delta$. 
Similarly, the error of $h_B$ is a linear function of $\Delta$.
The overall error of $h=(h_A, h_B)$ is a weighted combination of  the error of $h^*$ and the error of $h^{0}$ or $h^{1}$,
so the overall error of $h$ is thus linear in $\Delta$, so the optimal hypothesis parametrized by $\Delta$ must occur on the boundaries of the region of $\Delta$, so the optimal hypothesis is one of $\{h^{*} , h^{0}, h^{1} \}$.
We then show that the inequalities we assume in the theorem enforce that $h^*$ has strictly lower error than
$h^{0}$ or $h^{1}$. 
Formally, we enumerate the possible events: 
\begin{center}
    \begin{tabular}{ |l | l | l | l | p{5cm} |}
    \hline
      Type & Sign of  $h^*$ &  Label in Biased Data & Un-Normalized Probability of Event \\ \hline
      A    & +    & +     & $R_1 = (1-r)p(1-\eta)$ \\ \hline 
      A    & +    & -     & $R_2 = (1-r)p\eta$ \\ \hline 
      A    & -    & +     & $R_3 = (1-r)(1-p)\eta$ \\ \hline 
      A    & -    & -     & $R_4 = (1-r)(1-p)(1-\eta)$ \\ \hline 
      B    & +    & +     & $R_5 = rp(1-\eta)\beta_{POS}(1-\nu)$ \\ \hline 
      B    & +    & -     & $R_6 = rp[(1-\eta)\beta_{POS}\nu+\eta \beta_{NEG}]$ \\ \hline 
      B    & -    & +     & $R_7 = r(1-p)(\eta \beta_{POS})(1-\nu)$ \\ \hline 
      B    & -    & -     & $R_8 = r(1-p)[(1-\eta)\beta_{NEG} + \eta \beta_{POS} \nu ]$ \\ \hline 
    \end{tabular}
\end{center}
The probabilities on the far right hand side are not normalized. 
First we show that the $err(h^*) < err(h^{1})$.
$err(h^*)= R_{2}+R_{3}+R_{6}+R_{7}$ and $err(h^{1})=R_{2}+R_{4}+R_{6}+R_{8}$, thus
$err(h^*) < err(h^1)$ if and only if $R_{3} + R_{7} < R_{4} + R_{8}$ or thus if
\begin{multline*}
 (1-r)(1-p)\eta + r(1-p)(\eta \beta_{POS})(1-\nu)  \\ 
 < (1-r)(1-p)(1-\eta) + r(1-p)[(1-\eta)\beta_{NEG} + \eta \beta_{POS} \nu ] 
\end{multline*}
 %\begin{equation*}
 %\begin{split}
%(1-r)(1-p)\eta + r(1-p)(\eta \beta_{POS})(1-\nu) & <  \\
%(1-r)(1-p)(1-\eta) & + r(1-p)[(1-\eta)\beta_{NEG} + \eta \beta_{POS} \nu ] 
%\end{split}
%\end{equation*}
Equivalently,
\begin{align}
& 0 < (1-r)(1-2\eta) + r [(1-\eta)\beta_{NEG} - \eta \beta_{POS} (1-2\nu) ] \label{const1}
\end{align}
Now we consider $h^*$ compared to $h^{0}$.
Then  
$err(h^{0})=R_{1}+R_{3}+R_{5}+R_{7}$
Then $err(h^*) < err(h^{0})$ if and only if $R_{2} + R_{6} <R_{1} + R_{5}$. 
\begin{align*}
 (1-r)p\eta+ rp[(1-\eta)\beta_{POS}\nu+\eta \beta_{NEG} ] < (1-r)p(1-\eta) + rp(1-\eta)\beta_{POS}(1-\nu) %\nonumber
 \end{align*}
 Equivalently,
 \begin{align}
 0 < (1-r)(1-2\eta) + r((1-\eta)\beta_{POS}(1-2\nu) - \eta \beta_{NEG}) \label{const2}
\end{align}
Thus we have shown that the error of $h^*$ is less than the error of of $h^{1}$ and
$h^{0}$ if and only if both Lines \ref{const1} and \ref{const2} are true, which we
assume in our theorem.

Now we show that we error of $h=(h_A, h_B)$ is linear in $\Delta$.
There are two cases depending on what parameter of $h$ is non-zero.

Let $h$ be a hypothesis such that $P_{1A}(h_{A} ) = p_{1B} = \Delta$ and $P_{2A}(h_A ) = p_{2B} = 0$ and
$\Delta \in [0,p]$.
\begin{align*}
& err(h) = R_1 \frac{\Delta}{p} + R_{2} \frac{p-\Delta}{p}+R_{3} + R_5 \frac{\Delta}{p} + R_6 \frac{p - \Delta}{p} +R_7 \\
& = \frac{\Delta}{p} err(h^{0}) + \frac{p-\Delta}{p} err(h^*) 
\end{align*}
On the other case let $P_{1A}(h_{A} ) = p_{1B} = 0$ and $P_{2A}(h_A ) = p_{2B} = \Delta$ and $\Delta \in [0,1-p]$.
\begin{align*}
    & err(h)=R_2 +  \frac{1-p-\Delta}{1-p}R_3 +\frac{\Delta}{1-p} R_4 + R_6 + \frac{1-p-\Delta}{1-p}R_7  + \frac{\Delta}{1-p} R_8 \\
    & =\frac{\Delta}{1-p} err(h^{1}) + \frac{1-p-\Delta}{1-p} err(h^*) \\
\end{align*}
Thus the error of $h$ is linear in $\Delta$ and boundary values for $\Delta$ correspond to the hypotheses in $\{ h^{*}, h^{0}, h^{1} \}$.
These two arguments show that:
\begin{enumerate}
    \item Any single parameter $h$ is a weighted sum of ($h^*$ and $h^{0}$ ) or is a weighted sum of ($h^*$ and $h^{1}$) and so is linear in $\Delta$. 
    The boundary values of $\Delta$ correspond to $\{h^{*}, h^{0}, h^{1} \}$.
\item Since the optimal value of a linear function occurs on the boundaries of its range, the optimal Equal Opportunity classifier with at most one non-zero parameter is one of $\{h^{*}, h^{0}, h^{1} \}$.
\item  The inequalities in the theorem statement enforce that $h^*$ has lower biased error than either $h^0$ or $h^1$, so $h^*$ has the lowest biased error of any single parameter hypothesis satisfying Equal Opportunity. 
\end{enumerate}
\end{proof}
%Tightness Arg
If the conditions in the Theorem \emph{do not hold}, then $h^{*}$ will not have lower error than $h^{0}$ and $h^{1}$.  

\subsection{Verification Re-Weighting Recovers from Labeling Bias}  \label{reweightarg}

The way we intervene by Reweighting is we multiply the loss term for mis-classifying positive examples in Group $B$ by a factor $Z$  such that the weighted fraction of positive examples in biased data for Group $B$ is the same as the overall fraction of positive examples in Group $A$. 

The goal of this reweighting is to ensure that the ratio of positive to negative samples
in the positive region of $h_{B}^*$ is greater than $1$ while the ratio is less than $1$ in the negative region of $h_{B}^*$.
Thus the re-weighted probabilities need to simultaneously satisfy:
\begin{align*}
& \frac{P(y=1|h_{B}^*(x)=1)}{P(y=0 | h_{B}^*(x) =1 ) } = \frac{Z[(1-\eta )(1-\nu )] }{ (\eta + (1-\eta)\nu)}  > 1   \\
& \frac{P(y=1 |h_{B}^*(x)= 0 )}{P(y= | h_{B}^*(x) = 0  ) } = \frac{Z[\eta (1-\nu)]}{((1-\eta) + \eta \nu)} < 1\\
\end{align*}
 The two constraints are equivalent to requiring that:
\begin{align} \label{ilfweight}
& \frac{\eta + (1-\eta) \nu }{ (1-\eta)(1-\nu)} < Z < \frac{1-\eta + \eta \nu }{\eta (1-\nu)} 
\end{align} 
Recall from Section \ref{labelbias} that  $Z = \frac{1-P_{A,1}(1-\nu)}{(1-\nu)(1-P_{A,1})}$

First we show the right hand inequality.
\begin{align*}
    &\frac{1-p_{A,1}(1-\nu)}{(1-\nu)(1-p_{A,1})} < \frac{1-\eta + \eta \nu }{\eta (1-\nu)} \\
    & 0 <  \frac{1-\eta + \eta \nu }{\eta } - \frac{1-p_{A,1}(1-\nu)}{(1-p_{A,1})} \\
\end{align*}
Observe that both terms are linear in $\nu$. When $\nu = 0$, the inequality becomes $\frac{1-\eta}{\eta}- \frac{1-p_{A,1}}{1-p_{A,1}} = \frac{1-\eta}{\eta} - 1 > 0 $. 
In our bias model $\nu \in [0,1)$, but if  $\nu = 1$, the inequality becomes $\frac{1}{\eta}-\frac{1}{1-p_{A,1}} > 0 $.
Thus Equation \ref{ilfweight} holds if both $\frac{1-\eta}{\eta} -1 > 0$ and
$\frac{1}{\eta}-\frac{1}{1-p_{A,1}} > 0$. 

$\frac{1-\eta}{\eta} -1 > 0$ is clearly true because $0 < \eta < 1/2 $.

To see that $\frac{1}{\eta}-\frac{1}{1-p_{A,1}} > 0$, note that this is equivalent to $\eta<1-p_{A,1}$, where the right-hand-side is the overall fraction of negative examples in $A$.  This is clearly true because the positive region of $h_A^*$ has exactly an $\eta$ fraction of negatives, and the negative region of $h_A^*$ has a $1-\eta>\eta$ fraction of negatives.
%\comment{
%Observe if $\eta < 1-p_{A,1}$, then $\frac{1}{\eta}-\frac{1}{1-p_{A,1}} > 0 $. 
%To see that $\eta < 1 - p_{A,1}$:
%\begin{align*}
%& \eta < 1 - p_{A,1} = 1-(p(1-\eta)+(1-p)\eta) = 1- (p+\eta -2p\eta) = 1- p - \eta +2p\eta \\
%& \eta < 1-\eta + p(2\eta-1) \\
%& p(1-2\eta) < 1-2\eta
%\end{align*}
%The final inequality clearly holds since $p \in (0,1]$. 
%
%Thus we have shown that the right hand inequality in Line \ref{ilfweight}
%is satisfied for $\nu$ on its boundaries and  since the inequality is linear in 
%$\nu$, clearly the inequality holds for for $\nu \in [0,1)$. 
%Thus this constraint is satisfied for all parameters in our bias model.

Now we show the left hand inequality in Equation \ref{ilfweight}. 
\begin{align} 
& \frac{\eta + (1-\eta) \nu }{ (1-\eta)(1-\nu)} < \frac{1-P_{A,1}(1-\nu)}{(1-\nu)(1-P_{A,1})} \nonumber \\
& \frac{\eta + (1-\eta) \nu }{ (1-\eta)} < \frac{1-P_{A,1}(1-\nu)}{1-P_{A,1}} \nonumber \\
&  0 < \frac{1-P_{A,1}(1-\nu)}{(1-P_{A,1})} - \frac{\eta + (1-\eta) \nu }{ (1-\eta)} \label{linear}
\end{align} 
We follow a similar linearity argument to above.
For $\nu=1$, Equation \ref{linear} becomes $\frac{1}{1-p_{A,1}} - \frac{1}{1-\eta} > 0$.
This holds if $1-p_{A,1} < 1-\eta \iff \eta < p_{A,1}$.  This is clearly true because the negative region of $h_A^*$ has exactly an $\eta$ fraction of positives, and the positive region of $h_A^*$ has a $1-\eta>\eta$ fraction of positives.
For $\nu=0$, Equation \ref{linear} becomes $1-\frac{\eta}{1-\eta} >0 $ which holds since $0< \eta < 1/2$.
\begin{comment}
\begin{align*}
    & \eta < p_{A,1} = p(1-\eta) + (1-p) \eta = p - p\eta + \eta - p\eta \\
    & 0 < p-2p\eta
\end{align*}
The last line holds since $\eta < 1 /2$.
%\comment{
Observe that the left hand constraint
$ \frac{(1-\eta) (1-\nu) }{ \eta + (1-\eta) \nu } < 0 $ since $ Z > 0$. 
However since $\eta  \in (0,1/2)$ and $\nu \in [0,1]$ as long as the 
right hand side is well defined (i.e. $\eta  \neq 0$) and $\nu \neq 1$ then this requirement is satisfied.
Note that $Z$ from Section \ref{labelbias} i.e. $Z = \frac{1-P_{A,+}(1-\nu)}{(1-\nu)(1-P_{A,+})}$ satisfies these constraints-this is consistent with our figures.
\begin{align*}
&  \frac{1-\eta + \eta \nu }{\eta (1-\nu)}  > \frac{\eta + (1-\eta) \nu }{ (1-\eta)(1-\nu)} 
\end{align*}

\begin{align*}
  \frac{1-\eta + \eta \nu }{\eta (1-\nu)} & > \frac{\eta + (1-\eta) \nu }{ (1-\eta)(1-\nu)} \\
((1-\eta) + \eta \nu )(1-\eta) & > \eta (\eta + (1-\eta)\nu ) \\
    1-\eta + \eta \nu - \eta +\eta^{2} - \eta^2 \nu & > \eta^2 + \eta \nu - \eta^{2} \nu \\
 1- 2 \eta + \eta^2 & > \eta^2 \\
 1- 2 \eta & > 0 
\end{align*}
This always holds so we can always find the correct $Z$ satisfying both constraints. 
Thus to find the optimal $Z$ we can simply increment $Z$ from $0$ until the classifier snaps out of 
saying $0$ on all samples from the dis-advantaged group and before it snaps into saying $1$ for all
samples on the advantaged group. 
\end{comment}
\section{Calibration Results} \label{calib}
%First, we consider a less precise theorem that captures the core failings of Calibration in the Under-Representation Bias Model.
\begin{theorem} \label{calibinformal}
Assume the training data is corrupted by Under-Representation Bias with parameter $\beta < 1$.
For any such $\beta$, $h^*$ does not satisfy Calibration on the biased data and
thus Calibration constrained ERM will return a hypothesis that has strictly worse true error than the true error of $h^*$.
%then the biased error of any hypothesis $h$ Calibration constrained ERM solution is greater than the biased error $h^*$ and $h^*$ does not satisfy Calibration on the biased data and thus cannot be recovered.
This occurs even when $(1-\eta) \beta > \eta $, i.e. in the bias regime such that plain ERM on the biased data would recover $h^*$.

Moreover, if bias is such that $(1-\eta)\beta < \eta$ and thus ERM on the biased data will not recover $h^*$, then the unique ERM solution that satisfies Calibration on the biased data is a trivial classifier, meaning that all individuals from Group $A$ receive one label (the positive label) and all individuals from Group $B$ receive the opposite label.
%Clearly, the error in this case is much greater than the error of $h^*$. 
\end{theorem}
%Thus requiring Calibration when Under-Representation Bias is present performs strictly worse than plain ERM across all model parameters.
\begin{proof}
Recall that Calibration of hypothesis $h=(h_A, h_B)$ requires that both Eq. \ref{calp} and \ref{caln} hold simultaneously.
\begin{align}
& P_{x \sim \mathscr{D}_A}(y=1 | h_A (x)=1 ) = P_{x \sim \mathscr{D}_B}( y = 1 | h_B (x)=1)  \label{calp} \\
& P_{x \sim \mathscr{D}_A}(y=1 | h_A (x)=0 ) = P_{x \sim \mathscr{D}_B}( y = 1 | h_B (x)=0) \label{caln}
\end{align}
We assume that if one of the terms is vacuous in the Calibration constraints %(e.g if $h_B(x) =  1 $ for no $x$)
, then that constraint is still satisfied. 
In other words, if one bin is non-empty for one group while the corresponding bin for the other group is empty, we assume that bin satisfies Calibration. 
%\comment{
%Define $z_A$ such that $z_A := P_{x \sim \mathscr{D}_A} (h_A (x) =1 |  h_{A}^{*}(x) = 1 )$, e.g. the fraction of the  positive region of $h_{A}^{*}$ that our %classifier $h_A$ correctly classifies as positive.
%(in the previous notation $z_A$ is $\frac{p-p_{1A}}{p}$.
%Alternatively, let $v_A := P_{x \sim \mathscr{D}_B} (h_{A}(x) = 1| h_{A}^{*}(x) = 0 )$. 
%Let $z_B, v_B$, be defined similarly.
%These parameters determine whether classifiers $h_A, h_B$ satisfy calibration.}
Due to the effects of the bias model positive samples from Group $B$ appear in the training data with lowered frequency
and so the equalities in Equations \ref{calp} and \ref{caln} become: 
\begin{align}
& P_{x \sim A}(y=1 | h_{A}^{*} (x)=1 ) > P_{x \sim B}( y = 1 | h_{B}^{*} (x)=1)  \label{cal1} \\
& P_{x \sim A}(y=1 | h_{A}^{*} (x)=0 ) > P_{x \sim B}( y = 1 | h_{B}^{*} (x)=0) \label{cal2}
\end{align}
Thus $h^*  = ( h_{A}^*, h_{B}^{*})$ violates calibration for any $\beta < 1$ and any other hypothesis satisfying calibration will have strictly greater 
error on the true data distribution.
Intuitively, for $h$ to be Calibrated it will need to reduce the left-hand side of Equation \ref{cal1} because
it cannot increase the right-hand side and will have to increase the right-hand side of Equation \ref{cal2} because it cannot decrease the left-hand side.
As a result, its true error will be strictly larger than that of $h^*$.

Now, consider $(1-\eta)\beta < \eta$.
%A trivial classifier is one that assigns all of the input space to one label.
In this case, plain ERM will not recover $h^*$.
With this amount of bias, then:
\begin{align*}
   & P_{x\sim A} (y = 1 | h_{A}^{*}(x) =1 ) > P_{x\sim A} (y = 1 | h_{A}^{*}(x) =0  ) \\
    & >   P_{x\sim B} (y = 1 | h_{B}^{*}(x) =1 )  > P_{x\sim B} (y = 1 | h_{B}^{*}(x) =10)  \end{align*}
Satisfying Calibration with non-trivial classifiers requires achieving an equality with one side being a non-negative combination of the first two probabilities, and the other side being a non-negative combination of the second two probabilities.
Since these inequalities are all strict, this is clearly not possible, so the only way to satisfy calibration is to use a trivial classifier that assigns all of Group $A$ to one label, and all of Group $B$ to the other label.\footnote{Which trivial classifier is selected by ERM will depend on $p$ and $r$. If $1-r > r$ and $p > 1/2$, then Group $A$ will be all positive and Group $B$ all negative. While if $1-r > r $ and $p< 1/2$, then then Group $A$ will be all positive and Group $B$ all negative.}
\end{proof}

%% file: p1exp.tex
    In the previous sections, we introduced the formal study of fairness constraints in the presence of biased data, analyzing when they can help recover the optimal classifier on unbiased data.
    
    Our results in Sections \ref{sec:recoverysummary}-\ref{calib} show clear separations between different fairness notions,  in terms of when they will or will not recover that classifier, under a clean model of the target function and the types of bias introduced.  In this section, we conduct an empirical investigation.

    Specifically, we introduce synthetic and semi-synthetic experiments that are directly analogous to those in \cite{blum2019recovering}.
    \emph{Our results show close alignment with the theoretically expected behavior in \cite{blum2019recovering}.}

    In particular, our synthetic and semi-synthetic experiments support our the effectiveness of Equal Opportunity constrained ERM to Under-Representation Data Bias and Label Noise [and both at once].

    We also observe the effectiveness of re-weighting based methods, and it appears re-weighting is somewhat more effective in these experiments than our theoretical arguments suggest, though the Lower Bound in Section \ref{sec:reweightinglb} still holds.  
   
    In general, our empirical results emphasize and support both the benefits and limitations of theoretical analysis in the original \cite{forc2020} paper.
    
\subsection{Method}
The code to replicate these experiments is available at \url{https://github.com/kevstangl}.
We have two sets of main experiments, one using fully synthetic data, and one using semi-synthetic data. 
For the synthetic data, we generate one-dimensional data that perfectly matches the theoretical assumptions in our work, e.g. labels really come from our label generating process in \ref{truelabels}.

In both the semi-synthetic and synthetic experiments, we split the data into artificial groups, [so we know the ground truth matches our label generation assumptions in Section \ref{truelabels}].
Then we corrupt training data according to the bias model at hand, apply fairness constraints using the fairlearn \url{https://github.com/fairlearn/fairlearn} package or our own implementation [re-weighting], and then report the test accuracy on true data.
For most experiments, we use a logistic regression classifier from sklearn.
Throughout our experiments we focus on fairness-aware classifiers \cite{dwork2012fairness}, meaning that the classifiers have access to the group feature. \footnote{Or equivalently, that the feature space is sufficiently high-dimensional that group attributes are redundantly encoded.}

\section{Under-Representation Bias}
For ease of visualization, we plot recovery behavior with one type of fairness constraint at a time, i.e. one dimensional, using a discretization of the bias parameter.

When appropriate we plot the theoretical bounds from \cite{forc2020}. 
Recalling, the expected behavior, our theoretical arguments claims that Equal Opportunity and Equalized Odds should have strong recovery when $(1-r)(1-2\eta) +r((1-\eta)\beta-\eta) > 0$. Re-weighting should have strong recovery as long as there is a non-zero amount of positive samples from Group $B$.
We shall see that our experiments support these claims.

\subsection{Fully Synthetic Experiment}
In this section we report the fully synthetic experiment in more detail and show the results for data corrupted with Under-Representation Bias.

\begin{figure}[H]
\centering
\includegraphics[width=1.0\textwidth]{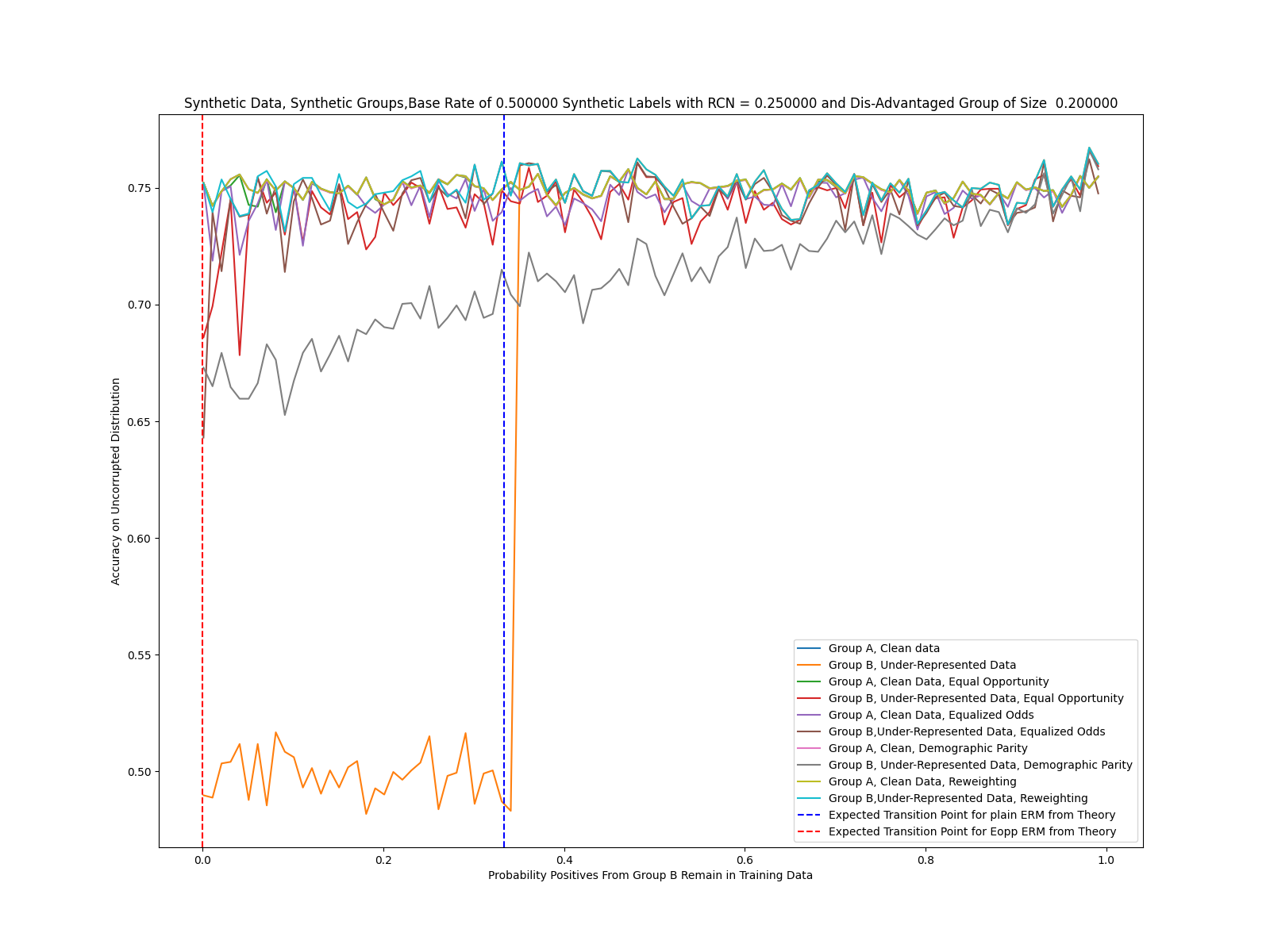}
    \caption{Fully Synthetic Experiment Showing Accuracy Loss. Vertical dashed blue line is when $\eta = (1-\eta)\beta$, e.g. to the left of this line.  }
    \label{fig:fullsyn}
\end{figure}

This is a synthetic experiment that verifies the poor performance of Demographic Parity when the Bayes Predictor is \emph{not} a trivial classifier, meaning all positive or all negative, and shows the effective performance of the other fairness constraints.

We have a one-dimensional data-set with $x \in \{0,1 \}$ and $y(x)= x$ with probability $\eta $ and otherwise.
The $x$-axis corresponds to the amount of Under-Representation Bias. The far left hand side means almost all positive examples from Group B are filtered from the training data, while on the right hand side relatively few or none are filtered out.

The $y$-axis reports the accuracy of the relevant classifier on the true, un-corrupted test distribution when we train on this corrupted distribution. We report both Group A and Group B accuracy, where Group B is the group being impacted by the bias models.

 In order to satisfy Demographic Parity, the classifier will have to steadily classify more of the negative region as positive, which is why we observe the linear decrease in test accuracy on Group B. Note, the red line indicates the recovery region for Equal Opportunity in this bias model, e.g. the Equal Opportunity constraint will recover the optimal classifier for the as long as the probability positives stay in the training data is greater than zero. In other words, we are in the Strong Recovery regime for these parameters.

\begin{figure}[H]
    \centering
    \includegraphics[width=1.0\textwidth]{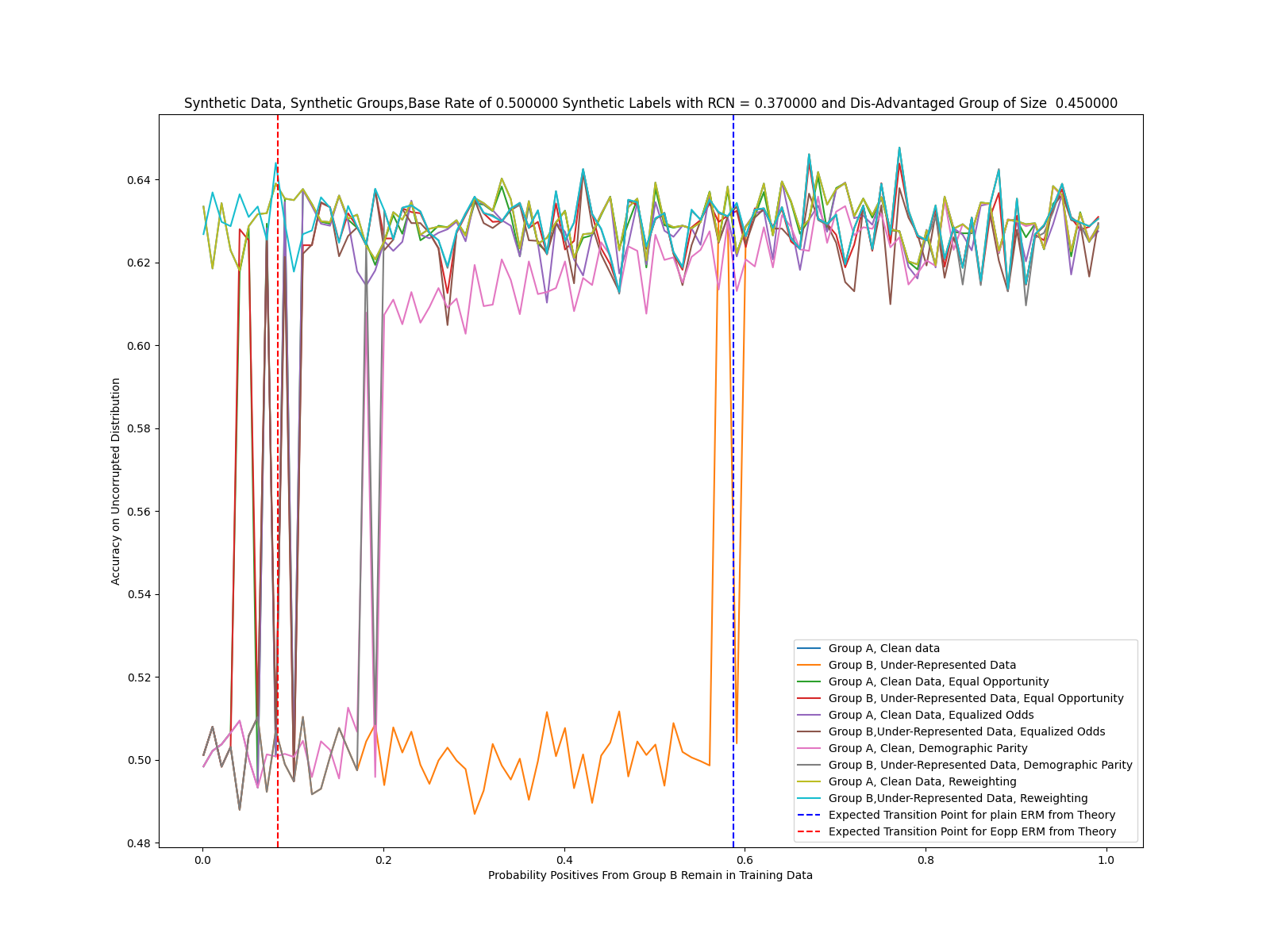}
    \label{fig:underrep synsynsyn}
    \caption{This figure has higher values of $\eta$ and $r$ so we are not in the Strong Recovery Regime, meaning the theoretical bounds for Equal Opportunity recovery in this bias model is greater than zero. We again see close alignment with our theoretical bounds.}
\end{figure}

\subsection{Semi-Synthetic Data, Natural Labels}
In Figure \ref{fig:underrep natural data, true labels} we repeat the above experiments but using the natural labels from ACS-Folktables. 
This corresponds to the Semi-Synthetic experiments described at the start of this section.

\begin{figure}[H]
         \centering
         \includegraphics[width=1.0\textwidth]{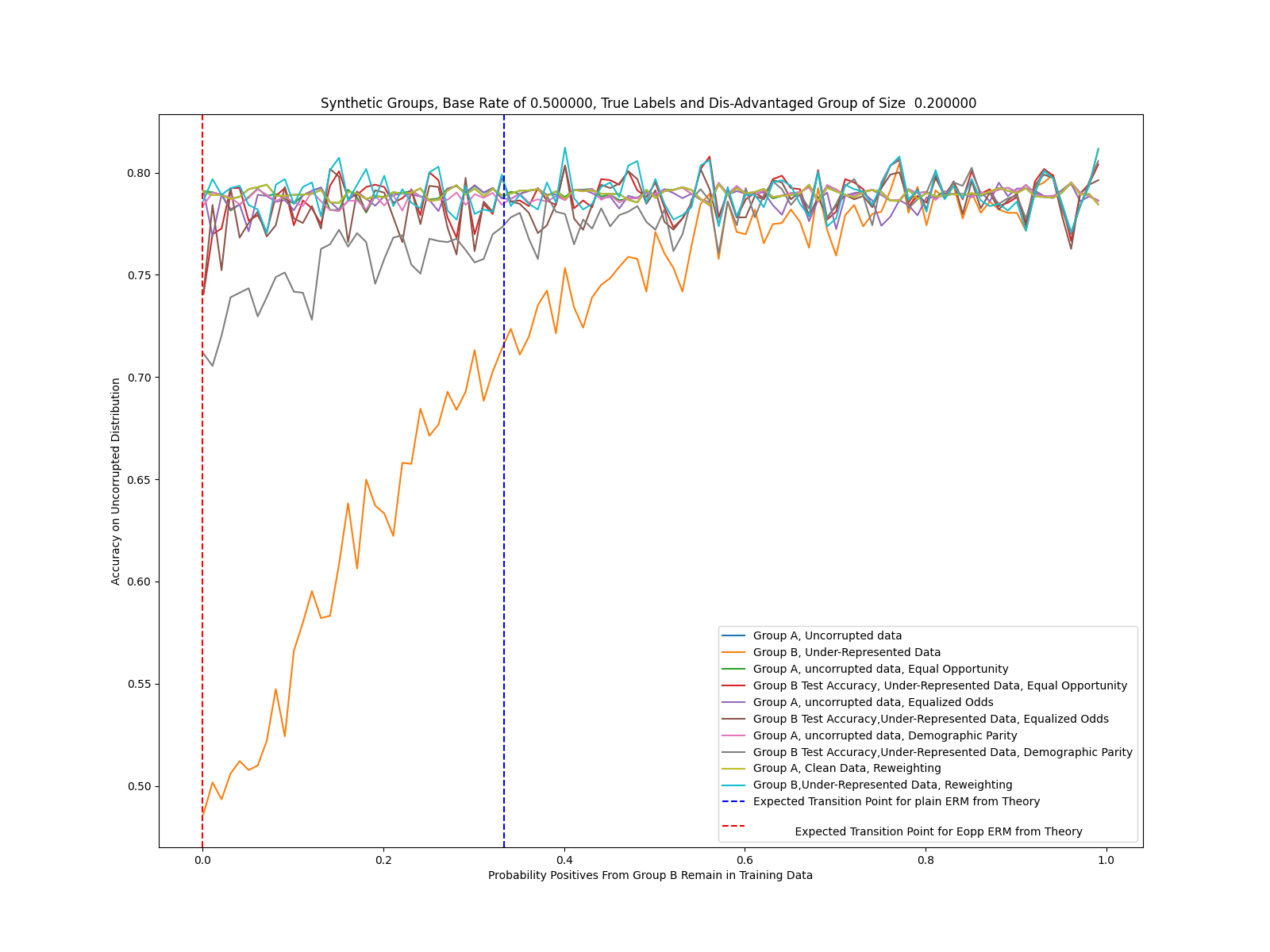}
         \caption{In this case, synthetic groups are still used, but we retain the original data and labels from the folktables ACS dataset. Then we inject Under-Representation Bias. Note the red/blue lines are now vacuous but are included for reference. In contrast to the previous results, which showed a sharp discontinuity, now plain ERM has a more graceful decline in performance. Parity [the grey line] exhibits similar issues to the synthetic data.}
         \label{fig:underrep natural data, true labels}
\end{figure}

\section{Labeling Bias}
Recall Labeling Bias as defined in Section \ref{sec:labelbias}.
When there is only labeling bias, the ERM recovery transition occurs when there are more negative points than positive points in the true positive region of $h^*$. Recall that there are two sources of negative points, 
true negatives and positives flipped to negative \footnote{Plain ERM will recover $h^*$ in the model if the following inequality holds.
\begin{align*}
    & (1-\eta)(1-\nu) > (1-\eta) \nu + \eta \\
    & 1-\eta - (1-\eta) \nu > (1-\eta) \nu + \eta \\
    &  -2(1-\eta) \nu > 2 \eta  - 1 \\
    & \nu < \frac{(1-2 \eta)}{2(1-\eta)}
\end{align*}}
For data corrupted with only Labeling Bias, we expect Equal Opportunity to recover in the correct parameter regime, as our main Theorem shows. Re-weighting and Parity should also be effective. Equalized Odds in contrast should be ineffective. 

\subsection{Fully Synthetic Experiment}

\begin{figure}[H]
         \centering
\includegraphics[width=1.0\textwidth]{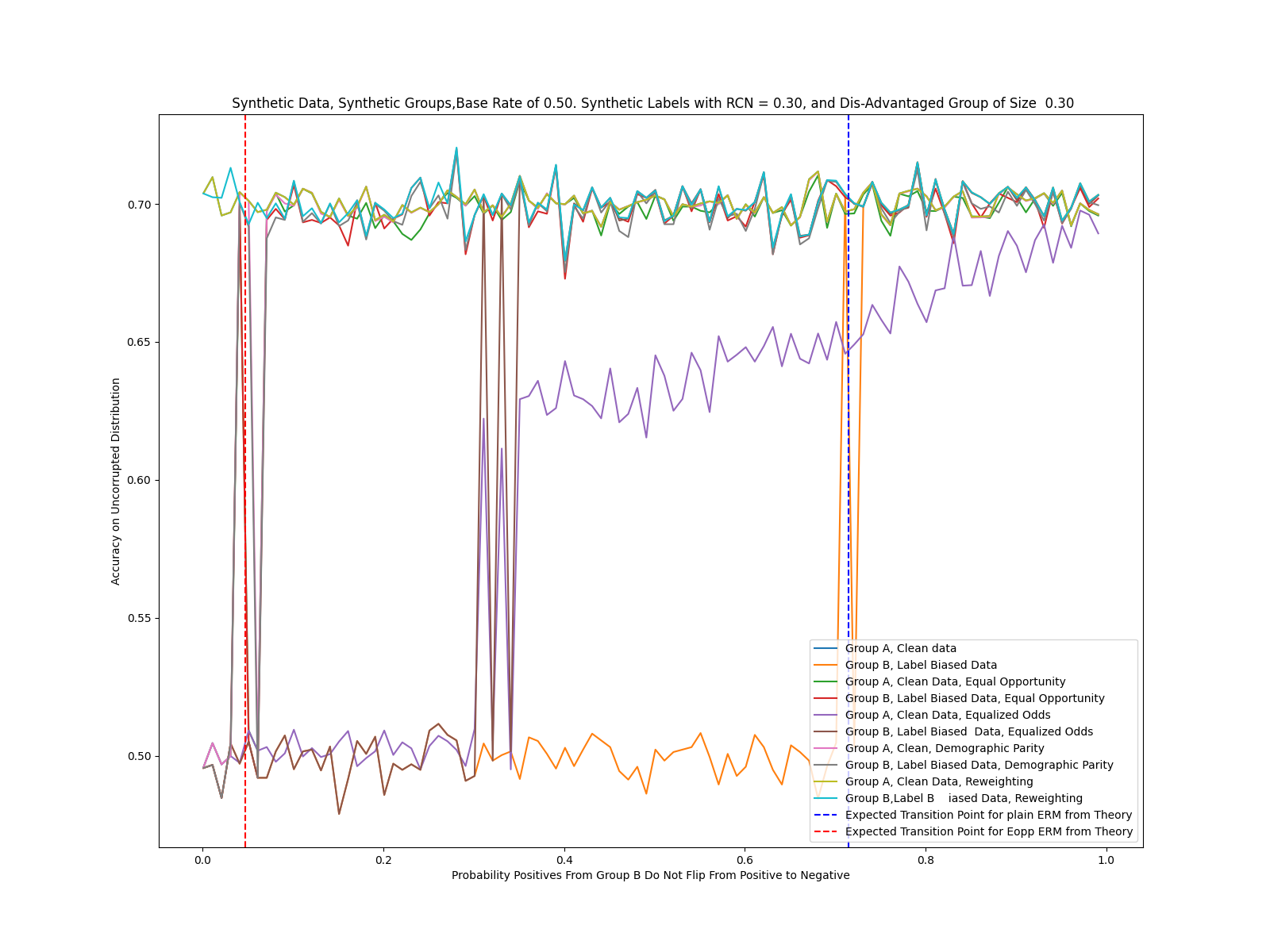}
         \caption{Fully Synthetic Experiment Showing Accuracy Loss. This is the analogous experiment to \ref{fig:fullsyn}, but with Labeling Bias instead of Under-Representation Bias. Note that we see the steady drop in performance of Equalized Odds, even in the regimes where plain ERM is sufficient. Interestingly, in this case both Group A and Group B accuracy is harmed by Equalized Odds, which is consistent with some arguments in \cite{hardt16}. An interpretation of this is that in this case, the Fairness constraint Equalized Odds might be actively harmful to the disadvantaged group, when compared with doing \emph{nothing}, e.g. learning with an un-modified base classifier.}
         \label{fig:labelnoisesyndata }
\end{figure}

\subsection{Semi-Synthetic Experiment}
Now we shift back to the semi-synthetic experiment.

\begin{figure}[H]
\includegraphics[width=1.0\textwidth]{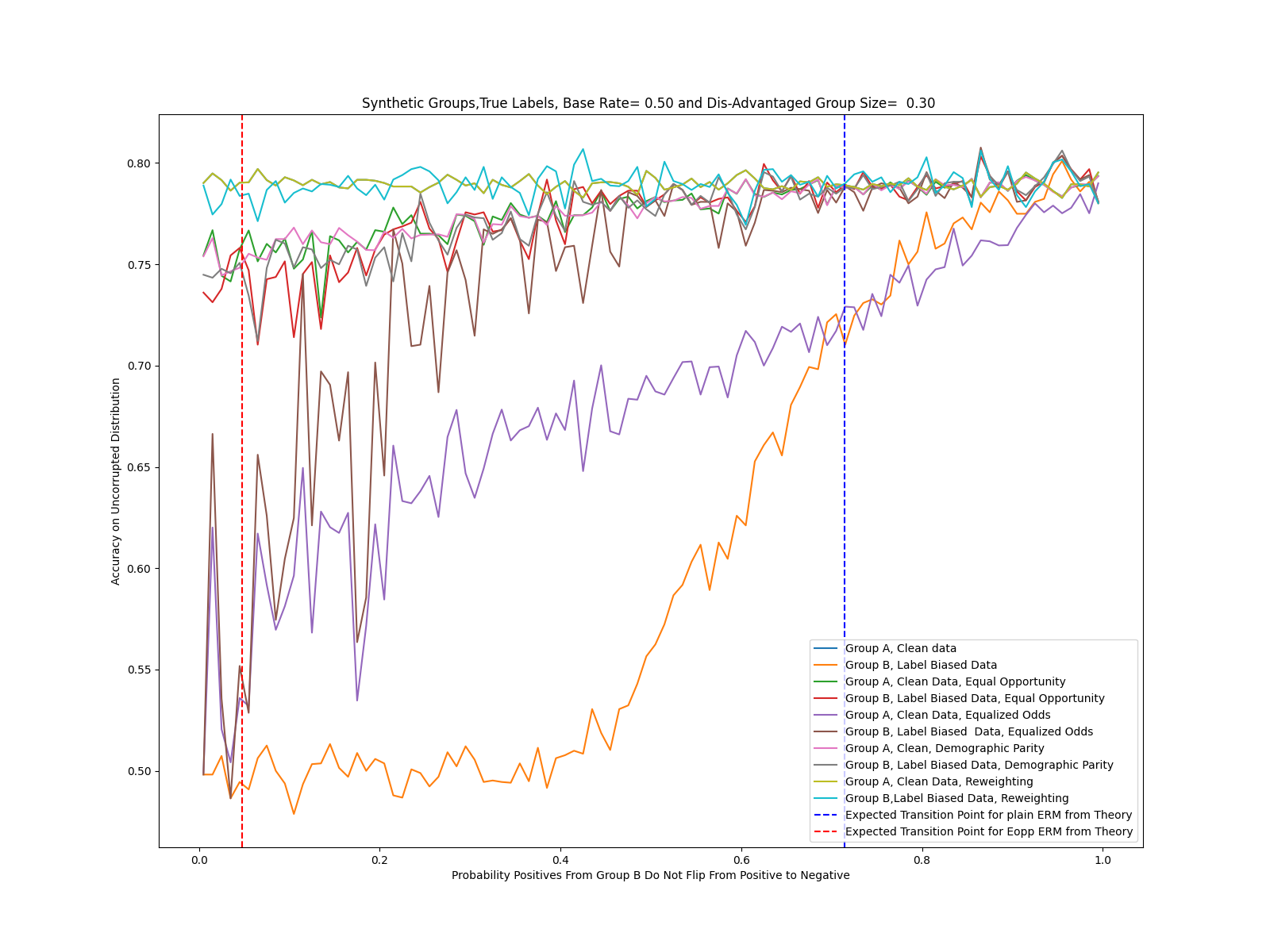}
         \caption{Semi-Synthetic Experiment Showing Accuracy Loss. This is the analogous experiment to \ref{fig:fullsyn}, but with Labeling Bias instead of Under-Representation Bias. Interestingly, in this plot, we see some decay of Equal Opportunity earlier than expected from theory, but other than that results are generally consistent with the theoretically expected behavior, namely the poor performance of Equalized Odds specifically.}
         %Note that the close alignment of this plot with Synthetic Experiment.
         \label{fig:semilabelnoisesyndata }
\end{figure}

%Our results are not meant to halt research on this family of algorithms, but to expose a critical failure mode

\subsection{Under-Representation Bias and Labeling Bias}
% other plots would be NICE but are optional.
We now discuss combining Under-Representation and Labeling Bias. 
Recall that re-weighting is the intervention that failed  when both forms of bias are included. 
Running the parameters of \ref{sec:labelbias} with the logistic regression classifier for synthetic data actually resulted in re-weighting still learning the correct classifier. 
This is because logistic regression is a linear model and the `inductive bias' of logistic regression would result in 
the classifier having weights such it would still make the correct prediction.
This points to the surprisingly durable effectiveness of the re-weighting intervention. Future research should consider exploring re-weighting interventions in broader contexts. 

Shifting to a classifier that definitely learns the Bayes Risk [e.g. k-nearest neighbors] would result in the poor behavior we describe in \ref{sec:labelbias}, where running ERM on the re-weighted data would oscillate between recovering $h^*$ and in this case the all negative classifier, thus swapping between approx $100\%$ accuracy and $75 \%$ [because the base rate is only $25\%$, so the all negative prediction].

\begin{table}[!h]
\begin{centering}
\caption{Average Accuracies over 50 trials of the Synthetic Experiment with parameters 
    $r=0.45, p=0.25, \eta = 0,\beta_{POS}=0.95, \beta_{NEG}$ and $\nu=0.58$.}
\label{table:knn-syn}
\begin{tabular}{||c c c c c c ||} 
 \hline
 Group & ERM & Equal Opp & Equalized Odds & Parity & Reweighting \\ 
 \hline \hline
 A & 1.0 & 0.925 & 0.750 & 0.925 & 1.0 \\ 
 \hline
 B & 0.855 & \textbf{0.925} &  0.750 & 0.842 & 0.875\\
 \hline
\end{tabular}
\end{centering}
\end{table}
Interestingly, due to the `knife's edge' nature of this lower-bound and the randomness of the label noise, the Reweighting Accuracy would oscillate neatly between $75 \%$, [e.g. due to the base rate being $25 \%$] and $100\%$.
Our recorded average for re-weighting is almost exactly the average of these two values.

Now we repeat the same experiment, but with an artificial base rate with true labels. 
\begin{table}[!h]
\begin{centering}
\caption{Average Accuracies over 50 trials of the Semi-Synthetic Experiment with parameters 
    $r=0.45, p=0.25, \eta = 0,\beta_{POS}=0.95, \beta_{NEG}$ and $\nu=0.58$.}
\label{table:knnsemisyn}
\begin{tabular}{||c c c c c c ||} 
 \hline
 Group & ERM & Equal Opp & Equalized Odds &  Parity & Reweighting \\ 
 \hline \hline
 A & 0.801 & 0.791 & 0.759  & 0.790 & 0.806 \\ 
 \hline
 B & 0.764 & 0.792 &  0.793 & 0.790 & 0.759 \\
 \hline
\end{tabular}
\end{centering}
\end{table}

These results are somewhat hard to interpret because for the natural data all the values of the interventions are quite close together. 
\emph{This lower bound does not strongly replicate for the natural data from folk-tables}, in that re-weighting still has tolerable performance.
Re-weighting is still the worst recovery notion on Group $B$, but the margin between the recovery notions is seems to be small.

This raise open questions about whether or not such re-weighting counter-examples will occur for natural distributions.
Despite this lower bound for re-weighting, the surprising durability of re-weighting as an intervention across all bias models points to the need for ongoing research for this notion
as a fairness \emph{intervention} rather than constraint. 

One open question is that all of our bias models are agnostic to the how far examples are from the decision boundary.
Likely bias models that take that into account would complicate and challenge our theoretical/empirical results, but pose challenges to our method of analysis.

\begin{figure}[H]
\label{fig:knnsemisyn}
         \centering
         \includegraphics[width=1.0\textwidth]{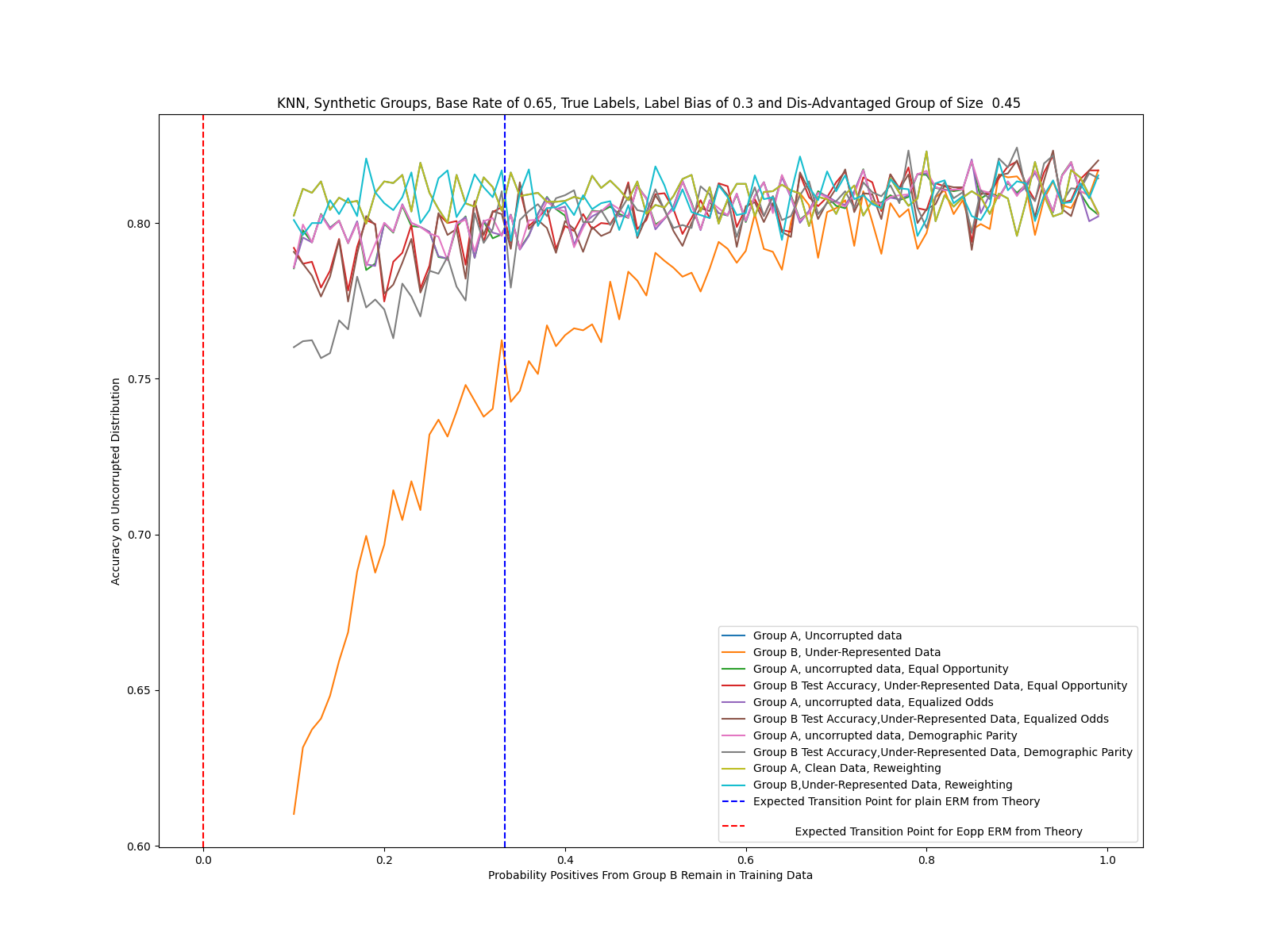}
         \caption{Semi-Synthetic Under-Representation Bias experiment with 3-nearest neighbors classifier, exhibited to vouchsafe effectiveness of k-nn on this dataset.}
         \label{fig:k-nn, underrep natural data, true labels}
\end{figure} 

Summarizing the experimental section, we would note that these experiments broadly support our theoretical claims. Additionally, we argue that re-weighting based interventions deserve further study due to their simplicity to implement and generally effectiveness, in addition to the effective performance of Equal Opportunity.

\section{Discussion}
Ensuring the reliable and beneficial performance of machine learning algorithms in fairness sensitive settings 
requires understanding the impact of noisy labeling and other data reliability issues, 
since, we claim that in some settings the
root cause of disparate model performance is disparate data quality.

While in some settings the optimal ground truth accuracy on different demographic groups really may be very distinct and policy concerns militate in favor of not deploying such disparate models or carefully selecting among models that witness un-avoidable trade-offs, we believe there are meaningful contexts in where better training procedures can overcome biases in the data and enable improved predictive performance across multiple groups.
Our work is an ongoing contribution to that broader vision by complementing existing theoretical work with further empirical work. 
%We emphasize the on-going research questions raised by \emph{our results}, name

In this section we have shown that Equal Opportunity constrained ERM will recover from several forms of training data bias, including Under-Representation Bias (where positive and/or negative examples of the disadvantaged group show up in the training data at a lower rate than their true prevalence in the population) and Labeling Bias (where each positive example from the disadvantaged group is mislabeled as negative with probability $\nu\in (0,1)$), in a clean model where the Bayes optimal classifiers $h_A^*,h_B^*$ satisfy most fairness constraints on the {\em true} distribution and the errors of $h_A^*,h_B^*$ are uniformly distributed. 

 The high-level message of this section is that fairness interventions need not be in competition with accuracy and may improve classification accuracy if training data is unrepresentative or biased; however these results will be connected to the true data distributions and features of the biased data-generation process.  It would be interesting to consider other ways in which training data could be biased, and other assumptions on the optimal classifiers, to determine what kinds of interventions might be most appropriate for different biased-data scenarios.

 We also show that a simpler reweighting approach succeeds in some but not all of our models.  This approach can be viewed as reweighting the training data to satisfy Demographic Parity, and then running an unconstrained ERM on the reweighted data (which is different from placing a Demographic Parity constraint on ERM on the actual training data, which does {\em not} work in our bias models). 
 Troublingly, we observe that enforcing calibration harms the very group we intended to aid 
 and results in substantially lowered accuracy across both groups, 
 even when the bias
 is small enough that normal ERM would work well.
 This points to more general issues with the class of fairness criteria called outcome tests (of which calibration is one variety) as also observed in \cite{infra}.
 
One limitation of our results is that we have used a stylized model for the generation of labels and how the bias enters the data-set.
However, we believe our results provide
useful insight into how fairness interventions can aid in reducing errors caused by bias in training data.
Even in this simple model, we observe separations between the fairness interventions 
and note that even when 
the fair solution is the right hypothesis in terms of both true accuracy and fairness, 
the fairness interventions can be tricked by the bias in the data.

In our bias models, we observe starkly different behavior of Equal Opportunity and Equalized Odds, two closely related fairness notions, when used to constrain ERM.
This sharp separation recommends that we closely align diagnosing a fairness concern with selecting an intervention, rather than looking towards universal solutions. 
In particular, biased data concerns like those we model in this chapter, appear to be both prevalent 
\cite{bertrand2004emily} and difficult to recognize. 

\subsection{Effectiveness of Equal Opportunity} 
A core high-level message in this Chapter \ref{sec:fairnoise} is the general effectiveness of Equal Opportunity in recovering from biased data in contrast to other fairness notions.

Is this effectiveness unreasonable or surprising? What `causes' this effectiveness? 
As noted in \cite{hardt16}, 
because Equal Opportunity is an easier constraint to satisfy, it will generally lead to a lower loss of utility than Equalized Odds.
Additionally, Equal Opportunity is always consistent with perfect prediction, one aspect of it's `inductive bias'.

A common thread\footnote{But not universal in this thesis, e.g. the lower bound in Theorem \ref{thm:lowereopp} } throughout this chapter and we will see again in Chapter \ref{sec:malnoise} is that the data corruption makes a plain learning algorithm `more' pessimistic on Group $B$, meaning that corrupted data from that group looks more and more negative.
This is likely because we think of the positive classification as more desirable,
so our theory of the fairness harms relies on making more of a group be classified as negative.

Equal Opportunity roughly pegs the true positive rate Group $B$ at the correct true positive rate of Group $A$, pushing the decisions in the right direction, especially when Group $B$ is small. Equal Opportunity allows the learner algorithm to take chance on Group B and classify more of them as positive, and the most accurate way to do this, will generally be close to or approximately the original $h^*$, as our theorems show.

Specifically, for Under-representation Bias, the probability of showing up depends on true label, rather than on your position in feature space, which is what makes Equal Opportunity still hold. For Labeling Bias, all positives are flipped to negative at the same rate, regardless of their location with respect to the decision boundary, which is why $h^*$ still satisfies Equal Opportunity in this case.

As we noted earlier in the chapter, $h^*$ fails to satisfy Parity with Under-Rep Bias, and $h^*$ violates Equalized Odds with Labeling Bias.

Equal Opportunity is generally well aligned our objective of recovering from biased data. An interesting research direction is showing when this alignment fails. Our results in the upcoming chapter, specifically Theorem \ref{thm:lowereopp},
are an initial step in this direction, and we show a \emph{partial} breakage.

\subsection{Fairness Diagnostics vs Interventions}
Some \cite{barocas2019fairness} argue in favor of implementing and conceptualizing fairness constraints
as \emph{diagnostic tools} rather than \emph{interventions}.

They even theorize that the research focus on fairness constraints as the primary fairness intervention is primarily motivated by 
minimizing the cost paid by model developers, since post-processing an existing model is presumably cheaper than
other interventions like investing in collecting higher quality data or including more humans in the loop.
Possibly these interventions would be more durable and avoid the possible zero-sum trade-offs that \emph{can} be present in fairness constraints.

Our work in this  chapter, \ref{sec:fairnoise} lies squarely within the \emph{intervention} school of fairness constraints, with our novel contribution being our study on how these interventions recover from biased data.
Even when stakeholders agree that other, more pro-active interventions are required, the speed of action, low cost, and perhaps ability to adapt to unforeseen data quality and provenance issues could all argue in favor of using fairness constraints as interventions. 
An intriguing possibility exists in our case; imagine a system that is continually retrained and augmented with fresh data, where Under-Representation Bias and Labeling Bias are intermittently present. 

Since the Equal Opportunity constraint in our model is consistent with the Bayes Optimal Predictor, our work possibly offers a best of the worlds, in that our recommended intervention [Equal Opportunity], would not foreclose deploying the most accurate classifier when no bias is present, and would robustly recover the accurate classifier when the bias is present.

In contrast, in the next chapter,  Chapter \ref{sec:malnoise}, we focus on characterizing the worst case accuracy loss due \emph{requiring} fairness constraints when an adversary can corrupt data, intentionally using those constraints to amplify their power. 
We argue that our results in these two chapters, which are somewhat in tension, show the need for further research that contrasts these two important fairness concerns.

%% file: mal_intro.tex
\section{Introduction}

The widespread adoption of machine learning algorithms across various domains, including recidivism prediction \cite{flores2016false,dieterich2016compas}, credit lending \cite{Kozodoi_2022}, 
and predictive policing \cite{lum2016predict}, 
has raised significant concerns regarding biases and unfairness in these models. Consequently, substantial efforts have been devoted to developing approaches for learning fair classification models that exhibit effective performance across protected attributes such as race and gender.

One critical aspect of addressing fairness in machine learning is ensuring the robustness of models against small amounts of adversarial corruption present in the training data. 
This data corruption may arise due to flawed data collection or cleaning processes \cite{saunders2013accuracy}, strategic misreporting \cite{hardt2016strategic}, under-representation of certain subgroups \cite{blum2019recovering}, or distribution shift over time \cite{schrouff2022maintaining}.

Empirical studies have demonstrated that such data unreliability is often centered on sensitive groups e.g. \cite{gianfrancesco2018potential}, emphasizing the need to understand the vulnerability of fair learning to adversarial perturbations. 
A concerning possibility is that fairness constraints might allow the adversary to amplify the effect of their corruptions by exploiting how these constraints require the classifier to have comparable performance on every relevant sub-group, even small ones.
%\footnote{Instead of thinking of our results in terms of an explicit adversary, they can also be interpreted as a worst case statement over corrupted data distributions, and that we want our fairness constrained learning pipelines to be `stable}'.

Previous work by \cite{lampert} and \cite{celis2021fair} have explored this topic from a theoretical perspective, considering different adversarial noise models. 
\cite{celis2021fair} focused on the $\eta$-Hamming model, where the adversary selectively perturbs a fraction of the dataset by modifying the protected attribute. 

\cite{lampert} on the other hand, investigated the Malicious Noise model, where an $\alpha$ fraction of the data-set (or distribution) 
is uniformly chosen and those data points are arbitrarily perturbed by the adversary.
We will focus on this Malicious Noise model.
In our study, we extend the framework of fair learning in the presence of Malicious Noise \cite{lampert} by  considering a broader range of fairness constraints and introducing a way to bypass some of their negative results by randomizing the hypothesis class.

\cite{lampert} present a pessimistic outlook, highlighting data distributions in which any proper learner, particularly in scenarios with imbalanced group sizes, exhibits high vulnerability to adversarial corruption when the learner is constrained by Demographic Parity \cite{calders2009building} or Equal Opportunity \cite{hardt16}. 
These results demonstrate novel and concerning challenges to designing fair learning algorithms resilient to adversarial manipulation in the form of 
Malicious Noise. 

The results of \cite{lampert} indicate that fairness constrained learning is much less robust than unconstrained learning.

In this chapter, we  present a more optimistic perspective on the vulnerability of fairness-constrained learning to malicious noise by introducing randomized classifiers. 
By allowing randomized classifiers, we can explore alternative strategies that effectively mitigate the impact of malicious noise and enhance the robustness of fairness-constrained models.
In addition, we extend the analysis beyond the fairness constraints examined in \cite{lampert}, providing a complete characterization of the robustness of each constraint and revealing a diverse range of vulnerabilities to Malicious Noise.

\subsection{Our Contributions}
We bypass the impossibility results in \cite{lampert} by allowing the learner to produce a randomized improper classifier. This classifier is constructed from hypotheses in the base class $\mathcal{H}$ using our post-processing procedure, which we refer to as the $(P,Q)$-Randomized Expansion of a hypothesis class $\mathcal{H}$, or $\PQ$
%\footnote{$\PQ$ is chosen because the base hypotheses are parameterized by $p,q$.}
%in that we are including hypotheses near to $\mathcal{H}$.
\begin{definition}[$\PQ$] \label{defn:pq}
For each classifier $h \in \mathcal{H}$, for $p,q \in[0,1]$ 
% \pcomargincomment{this should actually be $p,q \in[0,1]^{|Z|}$, a vector of biases for each group}
% \kmsmargincomment{I think it is ok to be a bit informal in this section}
\begin{align*}
    h_{p,q}(x) : = 
    \begin{cases}
    h(x)\quad \text{ with probability }1-p \\
    y \sim \Bern(q) \quad \text{otherwise}
    \end{cases}
\end{align*}
We define $\PQ$ as the expanded hypothesis class created by the set of all possible
$h_{p,q}(x)$.
\[ 
%\PQ := \cup_{h \in \calH} \cup_{p,q \in [0,1]} \{ h_{p,q}(x) \} 
\PQ := \{ h_{p,q} \mid h \in \calH, p,q \in [0,1]\} 
\]
\end{definition}
When clear from context we drop the dependence on $p,q$ and simply refer to $\hat{h} \in \PQ$.

Larger $p$ means we ignore more of the information in the base classifier $h$ and rely on the $\Bern(q)$.
The main technical questions we address in this chapter are:
\begin{center}
 \emph{
How susceptible and sensitive are fairness constrained learning algorithms
to Malicious Noise and to what extent does this vulnerability
depend on the specific fairness notion, especially if we allow improper learning?}

%  \emph{To What extent can improper learning bypass prior impossibility results for fairness constrained
% learning?}

%\emph{When a fairness aware ERM-learner is being targeted by an Adversary with malicious noise, by allowing the learner to exhibit a simple improper learning rule ($h^{'} \in \PQ$), can we exhibit hypotheses that are as robust to malicious noise as an ERM learner who is indifferent to fairness?}
\end{center}

%In the rest of the paper, we shall answer this question in the affirmative for some fairness constraints and show constant 
%
%In other words, the classical result \cite{kearns1988learning}
%\subsection{Our Contributions}

We focus on proving the existence of $h^{'} \in \PQ$ that satisfies a given fairness constraint and exhibits minimal accuracy loss on the original data distribution. 
Recall that $\alpha$ is the fraction of the overall distribution that is corrupted by the adversary.
%Our proofs are constructive. 
%We enumerate a hierarchy of fairness constraints and describe our results. 
%Here are the notions within order of robustness to malicious noise. (Again, here more robustness mean more accuracy on the natural data (despite the malicious noise) while still satisfying a given fairness constraint on biased data).
%In other words, train and select a model using the 

\begin{comment}
\begin{align*}
\textit{Demographic Parity} \geq Equal Opportunity \geq  Equalized Odds 
\geq Equal Error Rates
\end{align*}
\end{comment}

\begin{comment}
\begin{theorem} \label{thm: zoo}
Now we combine these fairness notions into one location and discuss their error in the presence of $O(\alpha)$ corruptions. 
Observe that from \cite{malnoise}, unconstrained ERM has an unconditional lower bound of $\Omega(\alpha)$.
This is our baseline and we want Fair ERM in the closure model to compete with this.
\begin{enumerate}
    \item Unconstrained ERM excess error $\Theta(\alpha)$ [exhibit $h^{*}$]
    \item Demographic Parity also has accuracy loss at most $\Theta(\alpha)$
    \item Eopp is at worst $\Omega\sqrt{\alpha})$
    \item Equal Error Rates : $\Omega(1)$
       \item Equalized Odds gets error $\min \{ r_A, r_B \}=\Omega(1)$

    \item Calibration has error in the worst case $\Omega(1)$
    %\item Min-Max Fairness has error in the worst case $\Omega(1)$
\end{enumerate}
\end{theorem}
\end{comment}

Our list of contributions is: 
\begin{enumerate}
\item We propose a way to 
%improperly learn a fair classifier in the 
bypass 
lower bounds \cite{lampert} in Fair-ERM with Malicious Noise by extending the hypothesis class using  the $\PQ$ notion.
%\item We show that this approach returns a classifier that 
%satisfies the fairness guarantee (parity and equal opportunity) with only a small loss in 
%accuracy when compared to the best hypothesis in the class. 
\item For the Demographic Parity \cite{calders2009building} constraint, our approach guarantees no more than $O(\alpha)$ loss in accuracy (which is \emph{optimal}
%the best one can expect 
in the Malicious Noise model \emph{without} fairness constraints \cite{malnoise}). 
In other words, in contrast to the perspective in \cite{lampert} which shows $\Omega(1)$ accuracy loss, we show that Demographic Parity constrained ERM can be made just as robust to Malicious Noise as unconstrained ERM.
\item For the Equal Opportunity \cite{hardt16} constraint, we guarantee no more than $O(\sqrt{\alpha})$ accuracy loss
and show that this is tight, i.e no classifier can do better. 
%We also show that no approach can do better than this.
\item For the fairness constraints Equalized Odds \cite{hardt16}, Minimax Error \cite{minimaxfair}, Predictive Parity, and our novel fairness constraint Parity Calibration, we show strong negative results.
%hat is, 
Namely, for each constraint there exist natural distributions such that an adversary that can force any algorithm to return a fair classifier that has $\Omega(1)$ loss in accuracy.
\item For Calibration \cite{faircalib}, we observe that the excess accuracy loss is at most $O(\alpha)$.
%\item Lastly we present sufficient conditions for a hypothesis class to be robust in the malicious adversarial model.
\end{enumerate}
%These results contradict \pcomargincomment{I think contradict is a strong wording here. Also, I think we're hammering a bit much on this distinction} parts of and extend the landscape introduced in \cite{lampert}. 
%\kmsreplace{This}{Our} work prompts \kmsdelete{high level} questions \pcomargincomment{what high level questions? It might be good to state them directly. Might not be obvious to the reader. I'm not sure what they are.} 
% These results prompts questions \pcomargincomment{"prompt". But also, what questions? This is very vague} about the right sensitivity level of fairness constraints.
%We do not focu

\begin{comment}
\begin{tabular}{ |p{2.7cm}||p{2.7cm}|p{2.7cm}|p{2.7cm}|p{2.7cm}| }
 \hline
 \multicolumn{5}{|c|}{ Results Summary} \\
 \hline
 Size of Group $B$ & Demographic Parity &  Equal Opportunity & Equal Error Rates & Equalized Odds \\
 \hline
 $ |A|=|B|$  & AF    &AFG&   004  & 420 \\
$|A| > O(\alpha) = |B|$ &   AX  & ALA   &248 & 420\\
 \hline
\end{tabular}
\end{comment}

%% file: mal_preliminaries.tex
\section{Preliminaries}
\label{sec:malprelim}
In fairness-constrained learning, the goal is to learn a classifier that achieves good predictive performance while satisfying certain fairness constraints that
connect the performance of the classifier on multiple groups, to ensure effective performance on all groups.

Specifically, we start with a dataset consisting of examples with feature vectors $(x \in \mathcal{X})$, labels $(y \in \mathcal{Y})$, and group attributes $(z \in \mathcal{Z})$. 
We assume that each example is drawn i.i.d from a joint distribution $\mathcal{D}$ of random variables $(X,Y,Z)$. 
There are multiple groups in the dataset, and we aim to ensure that the classifier's predictions do not unfairly favor or disfavor any particular group. 
We will denote $\mathcal{D}_z$ as the conditional distribution of random variables $X$ and $Y$ given $Z = z$. For simplicity, we will assume there are two disjoint groups: $A$ and $B$ in the dataset with B being the smaller and more vulnerable of the two. However, our results apply more broadly to any number of groups.

We aim to use the dataset to learn a classifier $f : \mathcal{X} \rightarrow \mathcal{Y}$ given a hypothesis class $\mathcal{H}$. 
However, in this chapter we suppress sample complexity learning issues and focus on characterizing the
accuracy properties of the best hypothesis in the expanded hypothesis class with a corrupted data distribution $\widetilde{D}$.
The goal is to probe the fundamental sensitivity of Fair-ERM to unreliable data in the large sample limit.
%\footnote{This focus is s}.
% We will typically think of the classifier $f$ as being quite group 
% specific; i.e. the groups are easily identifiable from the data and the distributions $\mathcal{D}_{A}$, $\mathcal{D}_B$
% may be substantially different.\footnote{This is not  necessary for our technical results but is useful for reader clarity.} 
% \pcomargincomment{at this point, we have not yet introduced A and B yet as the groups. Ok, I just added something above} Thus, we will occasionally use the notation $f_{A}$ or $f_{B}$ to refer to the behavior of the classifier only on Group $A$ or $B$ respectively.\pcomargincomment{I actually already mentioned this at the end of the realizability paragraph. It seems we have a couple repetitions.}
% % Additionally, the reader should think of Group $B$ as the smaller and more vulnerable group.

To this end, we consider solving the standard risk minimization problem with fairness constraints, known as Fair-ERM.
\begin{align}
\min_{h\in\mathcal{H}} & ~~\mathbb{E}_{(X,Y,Z)\sim\mathcal{D}} \left[\one(h(X) \neq Y)\right] ~~~~\ \\
\text{subject to} & ~~~~ F_z(h) = F_{z'}(h) \qquad \forall z,z'\in \mathcal{Z}. \label{FairnessConstraint}
\end{align}
where $F_z(h)$ is some fairness statistic of $h$ for group $z$ given the true labels $y$, such as \emph{true positive rate} :
$ \text{(TPR)}: F_z(h) = \mathbb{P}(h(X)=+1|Y=+1,Z=z)$.

% $\delta$ represents the maximum allowable difference between the fairness statistic of the classifier for any two groups in the dataset. We will 
% As we will show, our upper bounds will produce classifiers satisfying the fairness 
% with the same value of $\delta$ as $h^{*}$, the Bayes-optimal fair classifier, but our lower bounds will apply even if we allowed a constant-sized gap in the fairness constraint.
% Equivalently, we will often think of $\delta$ as zero.

We make a mild \emph{realizability assumption} that there exists a solution to this risk minimization problem. That is, there is at least one hypothesis in the class that satisfies the fairness constraint. This optimal solution is denoted as $h^*$.

For the results in this chapter, we only need the assumption that each group has non-trivial fraction of positives.
Formally, we assume that that for each fixed group $A$, where $r_A= P_{(x,y) \sim \mathcal{D} } [x \in A]$,
\begin{equation}
     P_{(x,y) \sim \mathcal{D}} [y = 1 \cap x \in A] := r_{A}^{+} \geq \frac{r_A}{c}
    \label{ass:raplus}
\end{equation}
for some  integer c.
Think $c \leq 20$.
This will allow the adversary to modify each group's true positive rate substantially, but not arbitrarily, because there is some non-trivial fraction of positives in each group.

As noted above, since we allow our hypothesis class to be group-aware, we can reason about $h^*_z$ for all $z \in Z$, where $h^*_z$ is the restriction of the optimal classifier $h^*$ to members of group $z$. In other words, $h_{z}^{*}$ is the optimal group-specific classifier for Group $z$.

\subsection{Fairness Notions}
Different formal notions of group fairness have previously been proposed in literature. These notions include, but are not limited to, Demographic Parity, Equal Opportunity, Equalized Odds, Minimax Fairness, and Calibration\cite{dwork2012fairness, calders2009building, hardt16, klein16, chouldechova2017fair}.

Selecting the “right” fairness measure is, in general, application-dependent.\footnote{We would also note that these fairness constraints are imperfect measures of fairness that likely do not capture all of the normative properties relevant to a specific task or system.  }%\pcomargincomment{need to add a bit more on fairness notions here}
% When we will call a classifier `unfair', it will mean that one of these fairness constraints is violated.
One of our goals in this work is to provide understanding of their implications under adversarial attack, which could aid in the selection process.
For the convenience of the reader, we include a table in Appendix \ref{fairtable} summarizing the fairness notions we consider in this chapter. 
Other than Calibration, these all are notions for binary classifiers. 
%In Appendix \ref{subsec:minimaxfair}, we will consider Minimax Fairness \cite{minimaxfair} but 
%\pcoreplace{elide}{present} 
%omit the definition here for clarity.
In Section \ref{subsec:calib} we will introduce a new variant of Calibration and will defer discussion of that notion until then.

\subsection{Adversary Model}
\label{subsec:adversary}
Throughout this chapter, we focus on the Malicious Noise Model, introduced by \cite{malnoise}.
This model considers a worst-case scenario where an adversary has complete control over a uniformly chosen $\alpha$ proportion of the training data and can manipulate that fraction in order to move the learning algorithm towards their desired outcomes, i.e. increasing test time error [on un-corrupted data].
%\kmsdelete{However, that $\alpha$ fraction is chosen uniformly from natural examples, unlike in \cite{bshouty2002pac}.}

% \kmsreplace{We follow the notion of a Fairness-ware adversary in \cite{lampert}.}{The core fear in \cite{lampert} and our work is that an adversary could use the fairness constraints to amplify their power when they have a small corruption budget $\alpha$.}
%\cite{lampert} introduces the fairness-aware adversary who corrupts a fixed fraction of a data set in order to maximize the test time error [on un-corrupted data] of a learner who uses the corrupted data to train a model.
%\cite{lampert}.
In \cite{malnoise}'s model, the samples are drawn sequentially from a fixed distribution. With probability $\alpha$ and full knowledge of the learning algorithm, data distribution and all the samples that have been drawn so far, the adversary can replace sample $(x,y)$ with an arbitrary sample $(\tilde{x}, \tilde{y})$.

%We reframe the data generating procedure as follows: Each timestep $t$ 
At each time-step $t$,
\begin{enumerate}
    \item The adversary chooses a distribution $\widetilde{\mathcal{D}}_t$ that is $\alpha-$close to the original distribution $\mathcal{D}$ in Total Variation distance.
    \item The algorithm draws a sample $(x_t,y_t)$ from $\widetilde{\mathcal{D}}_t$ instead of $\mathcal{D}$
\end{enumerate}
Note that the adversary's choice at time $t$, $\widetilde{\mathcal{D}}_t$ can depend on the samples $\{x_1,y_1, \ldots, x_{t-1}, y_{t-1}\}$ chosen so far.
%\footnote{}

Reframing the Malicious Noise Model in this manner simplifies analysis and allows us to focus on the fundamental aspect of this model which is how the accuracy guarantees of fairness constrained learning change as a function of $\alpha$.

\subsection{Core Learning Problem}
\label{subsec: learningproblem}
In the fair-ERM problem with Malicious Noise, our goal is to find the optimal classifier $h^*$ subject to a fairness constraint. However, the presence of the Malicious Noise makes this objective challenging. 
Instead of observing samples from the true distribution $\mathcal{D}$, we observe samples from a corrupted distribution $\widetilde{\mathcal{D}}$. 

In the standard ERM setting, \cite{kearns1988learning} show that the optimal classifier that can be learned using this corrupted data is one that is $O(\alpha)$-close to $h^*$ in terms of accuracy [on the original distribution]. 
The fair-ERM problem with a Malicious Noise adversary introduces an additional layer of complexity, as we must also ensure fairness while achieving high accuracy. 

%We say a learning algorithm is robust to  if it returns a classifier $h$ that satisfies two conditions. 
%First, the perceived unfairness, i.e, the difference in the estimated fairness constraint $\widetilde{F}$ between any two %groups $z$ and $z'$ in $\widetilde{\mathcal{D}}$ should be at most $\delta$. Second, the expected error of $h$ and $h^*$ %with respect to $\mathcal{D}$ should differ by at most a function of $\alpha$, where $h^*$ is the optimal classifier for %the fair-ERM problem on the true distribution $\mathcal{D}$. 
\begin{definition}
\label{def:robust}
We say a learning algorithm for the fair-ERM problem is $\beta$-robust with respect to a fairness constraint $F$ in the malicious adversary model with corruption fraction $\alpha$, if it returns a classifier $h$ such that $\widetilde{F}_z(h) = \widetilde{F}_{z'}(h)$ and 
\begin{align*}
| \mathbb{E}_{\mathcal{D}} \ [\one (h(X, Z) \neq Y)] - \mathbb{E}_{\mathcal{D}} \ [\one  (h^{*} (X, Z) \neq Y)] | \leq \beta(\alpha)
\end{align*}
where $h^*$ is the optimal classifier for the fair-ERM problem on the true distribution $\mathcal{D}$ with respect to a hypothesis 
class $\mathcal{H}$ and $\beta$ is a function of $\alpha$.
\end{definition}
This definition captures the desired properties of a learning algorithm that can perform well under the malicious noise model while achieving both accuracy and fairness, as measured by the fairness constraint $F$.

Thus, this is an agnostic learning problem \cite{HAUSSLER199278} with an adversary and fairness constraints.
As referenced in the introduction, we will allow the learner to return $h^{'} \in \PQ$, where $\PQ$ is a way to post-process each
$h \in \mathcal{H}$ using randomness.
In Sections \ref{sec:mainresults} and  \ref{subsec:calib} will characterize the optimal value of $\beta$ given the relevant fairness constraint $F$ and base hypothesis class
$\mathcal{H}$.

%% file: mal_main_results.tex
\section{Main Results: Demographic Parity, Equal Opportunity and Equalized Odds}
\label{sec:mainresults}
%\kmsmargincomment{I want this version in orange and then 4.1, I did kms delete on previous version. 
%I want to strike this paragaph since I think we are being unnecessarily shy about our results. I do think the Lampert results and ours meanignfully clash and I think this extensive dive into Lampert will simply confuse the %reader/make us vulnerable to skeptical reviewers. 
%I could be amenable to a long/detailed comparsion/constrast in appendix}
We now present our technical findings for Demographic Parity, Equal Opportunity, and Equalized Odds, and show how randomization enables better accuracy for Fair-ERM with Malicious Noise.
\cite{lampert} show impossibility results for Demographic Parity and Equal Opportunity  where a \emph{proper} learner is forced to return a classifier with $\Omega(1)$ excess unfairness and accuracy
 compared to $h^{*}$ for a synthetic and finite hypothesis class/distribution. 
 
To overcome this limitation, we propose a novel approach to make the hypothesis class $\mathcal{H}$ more robust, by injecting noise into each hypothesis $h \in \mathcal{H}$. In other words, we allow improper learning, and refer to the resulting expanded set of hypotheses as $\PQ$.
By injecting controlled noise into the hypotheses, we effectively ``smooth out" the hypothesis class $\mathcal{H}$, making it more resilient against adversarial manipulation. 

Since we allow group-aware classifiers, we learn two classifiers $h_{A}, h_{B} \in \PQ$, typically distinct from each other.
% Our approach limits the amount of fairness loss for any hypothesis class and true distribution
% $\mathcal{D}$ that satisfies a mild realizability assumption that at least one classifier in the hypothesis class must satisfy the fairness constraint.
Our method minimizes fairness loss for any hypothesis class and true distribution $\mathcal{D}$, under the assumption that at least one classifier in the original hypothesis class $\mathcal{H}$ satisfies the fairness constraints. 
We aim to find a fair classifier $\hat{h} \in \PQ$ that is as good as the best $h^{*} \in \mathcal{H}$.

\subsection{Demographic Parity}
Demographic Parity \cite{calders2009building} requires that the decisions of the classifier are independent of the group membership;
%o%the samples
 that is, $P_{(x,y) \sim \DA} [h(x)=1] = P_{(x,y) \sim \DB} [h(x)=1]$\footnote{Note there is no reference in the definition to the true labels, so a trivial hypothesis that flips a random coin for all examples would satisfy this notion, albeit at minimal accuracy. }. 

When the original distribution $\mathcal{D}$ is corrupted, a fair hypothesis on $\mathcal{D}$ may seem unfair to the learner. In order to analyze our approach it is important to understand how the fairness violation of a fixed hypothesis changes after the adversary corrupts an $\alpha$ proportion of the distribution.

\begin{restatable}[Parity after corruption]{proposition}{corruptparity}\label{prop:corruptparity}
Let $\widetilde{\mathcal{D}}$ be any corrupted distribution chosen by the adversary, and $h$ be a fixed hypothesis in $\mathcal{H}$. For a fixed group $A$, the following inequality bounds the change in the proportion of positive labels assigned by $h$:
$
\left| P_{(x,y) \sim \DAC} [h(x)=1] - P_{(x,y) \sim \DA} [h(x)=1] \right|  \leq \\
 \frac{\alpha}{(1-\alpha) r_A + \alpha }
$
where $r_A = \RA$, i.e how prevalent the group is in the original distribution. 
\end{restatable}

% \begin{linked}[Parity after corruption]{proposition}{corruptparity}\label{prop:fixedh-parity}
% Let $\mathcal{D}'$ be the corrupted distribution, and $h$ be a fixed hypothesis in $\mathcal{H}$. For a fixed group $A$, the following inequality bounds the change in the proportion of positive labels assigned by $h$:
% \begin{equation}
% \left| P_{(x,y) \sim \DAC} [h(x)=1] - P_{(x,y) \sim \DA} [h(x)=1] \right| \leq \frac{\alpha}{(1-\alpha) r_A + \alpha }
% \end{equation}
% where $r_A = \RA$.
% \end{linked}

This proposition provides an upper bound on the change in the proportion of positive labels assigned by a fixed hypothesis $h$ in $\mathcal{H}$ after the distribution has been corrupted according to the Malicious Noise Model. The full proof can be found in the Appendix \ref{proof:corruptparity}. 
The proof shows that this change is bounded by a function of the corruption rate $\alpha$ and the proportion of the dataset in the fixed group $A$, denoted by $r_A$. 

Intuitively, this means that the smaller a group is, the easier it is for the adversary to make a fair hypothesis seem unfair for members of that group.

% \begin{theorem}
%     For any hypothesis class $\mathcal{H}$ and distribution $ \dist = (\DA, \DB)$, a robust fair-ERM learner for the parity constraint in the Malicious Adversarial Model returns a hypothesis $\hhat$ such that 
%     \begin{equation*}
%         \error{\hhat} \leq O(\alpha)
%     \end{equation*}
% \end{theorem}

\begin{restatable}{theorem}{mainparity}\label{thm:mainparity}
For any hypothesis class $\mathcal{H}$ and distribution $ \dist = (\DA, \DB)$, a robust fair-ERM learner for the parity constraint in the Malicious Adversarial Model returns a hypothesis $\hhat \in \PQ$ such that 

$\error{\hhat} \leq O(\alpha)$ \\
where $h^*$ is the optimal classifier for the fair-ERM problem on the true distribution $\mathcal{D}$ with respect to hypothesis class $\mathcal{H}$.
\end{restatable}

This theorem states that a fair-ERM learner searching over the smoothed hypothesis class 
$\PQ$
returns a classifier that is within $\alpha$ of the accuracy of the best fair classifier in the original class $\mathcal{H}$. The full constructive proof can be found in the appendix \ref{proof:mainparity}.

The proof exhibits classifier $h \in \PQ$ that satisfies the desired guarantee. 
This classifier mostly behaves identically to $h^*$ but deviates with probability $p_A$ on samples from group $A$ (and with probability $p_B$ on samples from group $B$). We give an explicit assignment of these probability values $p_A$, $q_A$, $p_B$, $q_B$ in $[0,1]$ so that $h$ is perceived as fair by the learner. Then, we show that these values are small enough that the proportion of samples where $h(x) \neq h^*(x)$ is small ($O(\alpha)$). \emph{This is the best possible outcome in the malicious adversary model without fairness constraints \cite{kearns1988learning}.}

\subsection{Equal Opportunity}
Equal Opportunity \cite{hardt16} requires that the True Positive Rates of the classifier are equal across all the groups, that is, $P_{(x,y) \sim \DA} [h(x)=1 \mid y = 1] = P_{(x,y) \sim \DB} [h(x)=1 \mid y = 1]$. 
Similarly to Demographic Parity, we first provide bounds on how the fairness violation of a fixed hypothesis changes after the adversary corrupts an $\alpha$ proportion of the dataset. 
This is important because it gives an estimate of how much violation must be offset.

\begin{restatable}[TPR after corruption]{proposition}{corrupttpr}\label{prop:corrupttpr}
    Let $\widetilde{\mathcal{D}}$ be any corrupted distribution chosen by the adversary, and $h$ be a fixed hypothesis in $\mathcal{H}$. For a fixed group $A$, the following inequality bounds the change in True Positive Rate of $h$:
    \begin{equation}
        \left| \text{TPR}_A(h, \widetilde{\mathcal{D}}) - \text{TPR}_A(h, \dist) \right| \leq \frac{\alpha}{(1-\alpha) r_A^+ + \alpha }
    \end{equation}
    where $\text{TPR}_A(h, \dist) = P_{(x,y) \sim \DA} [h(x)=1 | y = 1]$ and $r_A^+ = P_{(x,y) \sim \mathcal{D}} [y = 1 \cap x \in A]$
\end{restatable}

This proposition provides an upper bound on the change to the true positive rate in group $A$ assigned by a fixed hypothesis $h$ in $\mathcal{H}$ after the dataset has been corrupted according to the Malicious Noise Model. 
The full proof can be found in the appendix \ref{proof:corrupttpr}. 

\emph{Since $\alpha \in [0,1]$, $O(\sqrt{\alpha})$ means larger (meaning worse) accuracy loss, compared $O(\alpha)$.}

The function that bounds the change in True Positive rate is similar to that of Demographic Parity with the proportional size of group A $r_A$ replaced with the proportion of the dataset that is positively labeled and in group A, $r_A^+$.
We will see that this slight change in dependence makes the robust learning problem more difficult and leads to a worse dependence on $\alpha$.

\begin{restatable}[Upper Bound]{theorem}{maineopp}\label{thm:maineopp}
For any hypothesis class $\mathcal{H}$ and distribution $ \dist = (\DA, \DB)$, a robust fair-ERM learner for the equal opportunity constraint in the Malicious Adversarial Model returns a hypothesis $\hhat$ such that 
$\error{\hhat} \leq O(\sqrt{\alpha})$
where $h^*$ is the optimal classifier for the fair-ERM problem on the true distribution $\mathcal{D}$ with respect to hypothesis class $\mathcal{H}$.
\end{restatable}

This theorem states that a fair-ERM learner, when applied with the smoothed hypothesis class $\PQ$, returns a classifier that is within $\sqrt{\alpha}$ of the accuracy of the best fair classifier in the original class $\mathcal{H}$. The full proof can be found in Appendix \ref{proof:mainparity}. 
%The proof follows a similar structure to that of demographic parity.

In constructing a classifier $h \in \PQ$, we aim for it to behave mostly identically to $h^*$ but introduce deviations with probability $p_A$ for samples from group $A$ and probability $p_B$ for samples from group $B$. 
However, in the case of the Equal Opportunity fairness constraint, this approach, as used for Demographic Parity, does not work effectively. 
We observe that the amount of correction required for each group depends inversely on the true positive rate, which presents challenges when the true positive rate (TPR) is close to 0 or 1.

For example, suppose the classifier achieves a 95\% TPR for a fixed group. The adversary can manipulate the TPR to reach 100\% by corrupting only a few samples. 
Correcting this change and bringing the TPR back down to 95\% is an incredibly difficult task, similar to finding a \textit{needle in a haystack}, since the learner essentially has to identify the corrupted samples to do so.
In such cases, it might be easier for the learning algorithm to increase the TPR of the other groups from 95\% to 100\% instead. 

The tradeoff lies in equalizing the corrections that only transform the TPR of a fixed group to its original value versus the corrections that transform the TPR of other groups to match the TPR of the group with the most corruptions.

\begin{restatable}[Lower Bound]{theorem}{lowereopp}\label{thm:lowereopp}
There exists a distribution $ \dist = (\DA, \DB)$ and a malicious adversary of power $\alpha$ that guarantees that any hypothesis, $\hhat$, returned by an improper learner for the fair-ERM problem with the equal opportunity constraint satisfies the following:
$\error{\hhat} \geq \Omega(\sqrt{\alpha})$
where $h^*$ is the optimal classifier for the fair-ERM problem on the true distribution $\mathcal{D}$ with respect to a hypothesis class $\mathcal{H}$.
\end{restatable}

In this lower bound, under the given conditions, no proper or improper learner can achieve an error rate lower than a threshold that scales with the square root of the adversary's power. In other words, as the adversary becomes more powerful ($\alpha$ increases), the error rate of the hypothesis returned by an improper learner will unavoidably be at least on the order of $\sqrt{\alpha}$.

The proof of this lower bound result sets up a scenario reflecting the \textit{needle in the haystack} issue described earlier. We present a distribution with two groups, one of size $\sqrt{\alpha}$ and the other of size $1 - \sqrt{\alpha}$. We construct a hypothesis class where the optimal classifier has a high but not perfect true positive rate. 
Then we show that any improper learner must either suffer poor accuracy on the smaller group or lose $\Omega(\sqrt{\alpha})$ accuracy on the larger group. 
The full proof can be found in the Appendix \ref{proof:corrupttpr}.
% \kmsedit{While not the best achievable for no fairness constraints [which \cite{kearns1988learning}$O(\alpha)]$, this $\theta(\sqrt{\alpha})$ accuracy loss is a substantial improvement over the $\theta(1)$ loss in \cite{lampert}.}

\subsection{Equalized Odds}
Equalized Odds \cite{hardt16} is a fairness constraint that requires equalizing
True Positive Rates (TPRs) and False Positive Rates (FPRs) across different groups. 
This notion is very sensitive to the adversary's corrupted data and we exhibit a problematic lower bound, showing the adversary can force terrible performance.

% This is our first \kmsedit{strong} negative result. 
The intuition is as follows; for a small group, the Adversary can set the Bayes Optimal TPR/FPRs rates of that group towards arbitrary values and so the learner must do the same on the larger group, regardless of their hypothesis class, forcing large error.
The full proof is in Appendix \ref{sec:eoddsproof}.
\begin{theorem}[Lower Bound]
\label{thm: Equalized Odds} 
For a learner seeking to maximize accuracy subject to satisfying Equalized Odds, an adversary with corruption fraction $\alpha$ can force an additional $\Omega(1)$ accuracy loss when compared to the performance of the optimal fair classifier on the true distribution.
\end{theorem}

%% file: mal_calibration.tex
\label{sec:calib}
In this section, we explore various notions of calibration \cite{dawid} for our model.
Calibration is a desirable property typically considered for classifiers, where predicted label probabilities should correspond to observed frequencies in the long run. For example, in weather forecasting, a well-calibrated predictor should have approximately 60\% of days with rain when it forecasts a 60\% chance of rain. This calibration requirement should hold for every predicted probability value output by the model.

Calibration has important fairness implications \cite{flores2016false, chouldechova2017fair, faircalib,multicalib} because a mis-calibrated predictor can lead to harmful actions in high-stakes settings, such as over-incarceration \cite{compassgender}. 
We show that varying the exact calibration requirements can substantially impact the model's accuracy loss when malicious noise is present in the training data.
%These results may be of independent interest to the calibration literature.

In this section, we align closely  with \cite{faircalib}, where the learner seeks to maximize accuracy while ensuring the classifier is perfectly calibrated.
Up until now, we have focused on binary classifiers, so in Section \ref{subsec: predparity} we consider a related notion called Predictive Parity \cite{chouldechova2017fair, flores2016false}, before considering calibration notions for hypotheses with output in $[0,1]$.

\begin{comment}
To recall, as before the learning problem is 

\begin{align}
\min_{h\in\mathcal{H}} & ~~\mathbb{E}_{(X,Y,Z)\sim\mathcal{D}} \left[\mathbbm{1}(h(X) \neq Y)\right] ~~~~\ \\
\text{subject to} & |K(z)-K(z')| \leq \delta \qquad \forall z,z'\in \mathcal{Z}. \label{FairnessConstraint}
\end{align}
where $K: h \rightarrow \mathbbm{R}$ is some notion of calibration error for $h$ for group $z$ given the true labels $y$.

Typically calibration requirements are most natural for regression problems where predictor $h$ provides 
\emph{fine-grained} scores that corresponds to the underlying 
probability of some outcome. 
\end{comment}

%Throughout these sections we consider a property titled \emph{shared range}.
%Namely that even though the hypotheses are fine tuned for each group, these calibrated classifiers
%share the same range. 

\subsection{ Predictive Parity Lower Bound}
\label{subsec: predparity}
\begin{definition}[Predictive Parity \cite{chouldechova2017fair}]
A binary classifier $h: \mathcal{X} \rightarrow \{0,1\}$ satisfies predictive parity if for groups A and B, $P_{x \sim \DA}[h(x)=1]>0$, $P_{x \sim \DB}[h(x)=1]>0$ and
% \footnote{This mild technical remark is explained a in the Appendix} 
\[P_{(x,y) \sim \DA} [y=1 | h(x)=1] = P_{(x,y) \sim \DB} [y=1 | h(x)=1] \]
\end{definition}
In later sections we consider other calibration notions.
Here we consider an adversary who is attacking a learner constrained by equal predictive parity when group sizes are \emph{imbalanced}.

\begin{comment}
\begin{theorem}
With probability $1-(1-n)^{\alpha}$, there exists a FAIR ERM learner constrained learner with $O(\alpha)$
excess error. 
\end{theorem}
\end{comment}

\begin{theorem}
\label{thm:predparity}
    For a malicious adversary with corruption fraction $\alpha$, for Fair-ERM constrained to satisfy Predictive Parity, then there is no $h \in \PQ$ with less than $\Omega(1)$ error. 
\end{theorem}

The intuition for this statement is that imbalanced group size will allow the adversary to change the conditional mean substantially.
%In expectation, 
Below, we have an informal proof:
\begin{proof}[Proof Sketch:]
Suppose $P(x \in A)=1-\alpha$ and $P(x \in B)= \alpha$.
Observe that whatever the initial value of $P_{(x,y) \sim \DB} [y=1 | h(x)=1]$, the adversary can drive this value $P_{(x,y) \sim \mathcal\DBC} [y=1 | h(x)=1]$ to $50\%$ or below
by adding a duplicate copy of every natural example in group $B$ with the opposite label.

Since all of these points are information-theoretically indistinguishable, any hypothesis for group $B$ that makes any positive predictions incurs at least $50\%$ error and $1/2=P_{(x,y) \sim \mathcal\DBC} [y=1 | h(x)=1]$ calibration error.
%will have to do the same for
Any classifier for group $A$ satisfying Predictive Parity will have to do the same, yielding our $\Omega(1)$ error.
%The full proof in Section \ref{proof:predparity}.
%\begin{align*}
 %   blehp
%\end{align*}
%Observe that this attack 
\end{proof}

\subsection{Extension to Finer Grained Hypothesis Classes}
\label{subsec:calib}
A criticism of this lower bound might be that these calibration notions are very coarse and calibration is intended for fine-grained predictors, meaning those that have a finer grained discretization of the probabilities in $[0,1]$.
%and inappropriate for a binary classifier that in effect has two bins. 
%While a diversion from the rest of the paper where we tend to focus on binary classifiers, 
We now provide extensions for these lower bounds to real valued $\mathcal{H}$. 
Interestingly, we show if the learner can modify their `binning strategy', the learner can `decouple' the classifiers for the groups in the population and 
thus only suffer $O(\alpha)$ accuracy loss.
%Rather than being an algorithmic trick, this attack is fundamental as it seems to occur in the wild
%organically 
%as a type of red-lining. 
%This is because absent further constraints, calibration is a weak notion of mere self-consistency.
%Attacks of this type motivate more constrained notions of calibration like 
We adopt the version of calibration from \cite{faircalib}.
\begin{definition}[Calibration] \label{def:calib}
A classifier $h: \mathcal{X} \rightarrow [0,1]$ is Calibrated with respect to distribution $\mathcal{D}$ if 
\[\forall r \in [0,1], r= \mathbb{E}_{(x,y) \sim \mathcal{D} }[y=1| h(x)=r]\]
We will primarily focus on the discretized version of this definition where the classifier assigns every data point to one of $R$ bins, each with a corresponding label $r$, that partition $[0,1]$ dis-jointly. 
We will refer to this partition as $[R]$ with $r \in [R]$ corresponding to the prediction of a bin. 
\[ \forall r \in [R], r= \mathbb{E}_{(x,y) \sim \mathcal{D} }[y=1| h(x)=r] \]
\end{definition}
%Observe that nothing in this initial definition references groups. 
%The natural generalization to the above definition with 
Calibration as a fairness requirements with demographic groups requires that the classifier $h$ is calibrated with 
respect to the group distributions $\DA$ and $\DB$ simultaneously. 
In the sections that follow when we say `calibrated' this always refers to calibration with respect to $\DA$ and $\DB$. 

\begin{theorem}
\label{thm:calib}
    The learner wants to maximize accuracy subject to using a calibrated classifier, $h: \mathcal{X} \rightarrow [R]$ where $[R]$ is a partition of $[0,1]$ into bins.%^labelled bins with each label.
    
    The learner may modify the binning strategy after the adversary commits to a corruption strategy.
    Then an adversary with corruption fraction $\alpha$ can force at most $O(\alpha)$ excess accuracy loss over the non-corrupted optimal
    classifier. 
\end{theorem}

\newpage

\subsection{Parity Calibration}
Motivated by Theorem \ref{thm:calib}, we introduce a \emph{novel} fairness notion we call \emph{Parity Calibration}\footnote{We would note that this is initial discussion of a novel fairness constraint that arose naturally from considering Theorem \ref{thm:calib}. The idea is in some cases it might be more desirable to have a more sensitive calibration notion, hence we define Parity Calibration. This notion requires further study and analysis before deployment in sensitive contexts.}
% \footnote{This is a strong fairness constraint and should be thought of as a strong prior that while conditional label distribution $\mathcal{D}_{y|x}$ can be different among groups, how much of each group falls in each risk category is the same.}.
Informally, this notion is a generalization of Statistical/Demographic parity \cite{dwork2012fairness} for the case of classifier with 
$R$ bins partitioning $[0,1]$.
\begin{definition}[Parity Calibration]
\label{def:paritycalib}
Classifier $h: \mathcal{X} \rightarrow [R]$, where $[R]$ is a partition of $[0,1]$ into labelled bins, satisfies
\emph{Parity Calibration} if the classifier is Calibrated (Definition \ref{def:calib}) \emph{and}
\begin{align*}
\forall r \in [R], P_{(x,y) \sim \DA} [h(x)=r] =  P_{(x,y) \sim \DB} [h(x)=r]
\end{align*} 
\end{definition}

%These lower bounds still hold for stronger notions of calibration error, namely $K_1(h, \mathcal{D})$ and $K_2(h, \mathcal{D})$ 
%which are average calibration error for the $l_1$ and $l_2$ norms respectively.
\begin{theorem}
\label{thm:paritycalib} 
Consider a learner maximizing accuracy subject to satisfying Parity Calibration.
%$h: \mathcal{X} \rightarrow [R]$ where $[R]$ is a partition of $[0,1]$ into labelled bins with each label.
    The learner may modify the binning strategy after the adversary commits to a corruption strategy.
    Then an adversary with corruption fraction $\alpha$ can force $\Omega(1)$ excess accuracy loss over the non-corrupted optimal
    classifier. 
\end{theorem}

%We defer the proof of this statement to the appendix, but the intuition is a follows.

If the size of Group $B$ is $O(\alpha)$, then following a similar duplication strategy for Predictive Parity Theorem \ref{thm:predparity},
then the adversary can force Group $B$ to have an expected label of $50\%$, i.e.
$\forall x \in B, \mathbb{E}_{x \sim \DB}[y|x]=50\%$.
Thus, any classifier that is calibrated must assign all of Group $B$ to a $50\%$ bucket.
In order to satisfy \emph{Parity Calibration}, the classifier must do the same to Group $A$, yielding $50\%$ error on Group $A$.

% \kmsdelete{\subsection{Discussion}
% In general these results are consistent with the observed behavior of Calibration in other parts of theoretical computer science.
% If the learner/society really only cares about accuracy, then the insensitivity in Section \ref{subsec:calib} is somewhat of a feature, not a bug, 
% especially if the unreliability of data in Group $B$ optimistically could be transient?
% %However, advocates for stronger notions calibration would instead note that in \ref{thm: calib} 
% In general, when thinking about accuracy loss and malicious in the context of fair ERM; what is the appropriate amount of sensitivity in
% the learning process? We shall discuss this somewhat more in Section \ref{sec: discussion}.}
% %We would observe that are substantial 

\begin{comment}
\begin{definition}[Average Calibration Error]
The avergae calibration error of a predictor $h$ (with $h: \mathcal{X} \rightarrow [0,1]$) on distribution $\mathcal{D}$ is:
\[ K_1(f, \mathcal{D}) = \sum_{v \in R(h) } P_{(x,y) \sim D} [h(x)=v]|v-\mathbbm{E}_{(x,y) \sim \mathcal{D}}[y|h(x)=v] |\]
where $R(h)$ is the range of $h$. 

Similarly, the average squared calibration error is 
    \[ K_1(f, \mathcal{D}) = \sum_{v \in R(h) } P_{(x,y) \sim D} [h(x)=v]|(v-\mathbbm{E}_{(x,y) \sim \mathcal{D}}[y|h(x)=v])^2\]\end{definition}

\begin{theorem}
    
\end{theorem}
\end{comment}

%% file: mal_discussion.tex
\section{Discussion}
\label{sec: discussion}
We study Fair-ERM in the Malicious Noise model, and in some cases allow 
the learner to maintain optimal overall accuracy despite the signal in Group $B$ being almost entirely washed out.
%when we allow learners to use the
%$\PQ$ randomized expansion of the hypothesis class $\mathcal{H}$
In particular, we show that different fairness constraints have fundamentally different behavior in the presence of Malicious Noise, in terms of the amount of accuracy loss that a given level of Malicious Noise could cause a fairness-constrained learner to incur. 
The key to achieving our results, which are more optimistic than those in \cite{lampert}, is allowing for improper learners using the (P,Q)-randomized expansions of the given class $\mathcal{H}$.
%We \kmsreplace{present a picture of the}{prove upper and lower bounds on}
%accuracy loss for a range of fairness notions, given \kmsreplace{this simple randomization step.}{learning over $\PQ$.
%In general our results indicate Fair-ERM (given learning over $\PQ$) is more robust than claimed in \cite{lampert}.
The type of smoothness we create by using $\PQ$ seems to be a natural property that is likely shared by many natural hypothesis classes.

A criticism of our work could be that the pessimistic lower bounds in Equal Opportunity and for Calibration rely on an unrealistically strong adversary. We would note that the corruption strategies in each lower bound in this chapter and in \cite{lampert} hold for any adversary that has the capability to choose a group and add points to that group that look similar to existing points but with opposite labels of the original points.

Fairness notions are motivated as a response to learned disparities when there is systemic error affecting one group. 
Fairness notions are supposed to mitigate this by ruling out classifiers that have worse performance on a sub-group. 
This can peg both classifiers at a lower level of performance in order to \emph{motivate} \cite{hardt16} improving the data collection or labelling process to obtain more reliable performance. 
%So in \kmsreplace{some}{a} sense, sensitivity of the fairness notion to poor sub-group performance caused by malicious noise is the \textit{point} of fairness constraints! 
However, it is also desirable that fairness constraints perform gracefully when subject to Malicious Noise, because fairness constraints will be used in contexts where the data is unreliable and noisy. %without the learner's knowledge.
This tension, exposed by our work, motivates 
%a revisiting of fairness notions from first principles approach and trying to axiomatize the 
%desired properties of a fairness intervention a la cryptography and privacy. \footnote{Work in multi-calibration \cite{multicalib} is a viable direction for this problem but it is unclear how 
%that and related notions behave with unreliable data. }
ongoing work studying the sensitivity level of fairness constraints.

%This work was supported in part by the National Science Foundation under grant CCF-2212968, by the
%Simons Foundation under the Simons Collaboration on the Theory of Algorithmic Fairness, by the Defense
%Advanced Research Projects Agency under cooperative agreement HR00112020003. The views expressed in
%this work do not necessarily reflect the position or the policy of the Government and no official endorsement
%should be inferred. Approved for public release; distribution is unlimited.

%% file: mal_appendix.tex
%\section{Neurips Ethics Review}

%\input{sections/ethics}

\section{Fairness Notions}
\label{fairtable}
\begin{center}
\begin{tabular}{ |p{4cm}||p{9cm}| }
 \hline
 \multicolumn{2}{|c|}{ Fairness Constraints}\\
 \hline
 Demographic Parity \cite{dwork2012fairness} & $ P_{(x,y) \sim \DA} [h(x)=1]= P_{(x,y) \sim \DB} [h(x)=1] $  \\
 \hline 
 Equal Opportunity \cite{hardt16} & $ P_{(x,y) \sim \DA} [h(x)=1| y=1 ]= P_{(x,y) \sim \DB} [h(x)=1| y=1 ] $ \\
 \hline
 Equalized Odds \cite{hardt16} &   $P_{(x,y) \sim \DA} [h(x)=1| y=1 ]= P_{(x,y) \sim \DB} [h(x)=1| y=1 ]$ and \\
  & $P_{(x,y) \sim \DA} [h(x)=1| y=0 ]= P_{(x,y) \sim \DB} [h(x)=1| y=0 ]$ \\
  \hline
  Predictive Parity \cite{chouldechova2017fair} & $P_{(x,y) \sim \DA} [y=1| h(x)=1 ]= P_{(x,y) \sim \DB} [y=1| h(x)=1 ]$ \\
\hline
Calibration\footnote{$h:\mathcal{X} \rightarrow [0,1]$} \cite{klein16,dawid} & $ \forall r \in [0,1],\quad  r = \mathbb{E}_{x,y \sim \mathcal{D}} [y|h(x)=r] $\\
\hline
\end{tabular}
\end{center}

\section{Proofs}

\input{mal_parity}

\input{mal_eopp}

\input{mal_eodds}

\section{Calibration Proofs}

\input{mal_calibproof}

\input{mal_minimax}

\input{mal_multgroup}

%% file: mal_parity.tex
\corruptparity*

\begin{proof}[Proof of Proposition~\ref{prop:corruptparity}]\label{proof:corruptparity}
We want to bound the change in the proportion of positive labels assigned by $h$ when we move from the original distribution $\mathcal{D}$ to the corrupted distribution $\widetilde{\mathcal{D}}$. For a fixed group $A$, we can express the proportion of positive labels assigned by $h$ in $\widetilde{\mathcal{D}}$ in terms of the proportion of positive labels assigned by $h$ in $\mathcal{D}$ as follows:

\begin{equation}
    P_{(x,y) \sim \DAC} [h(x)=1] = \frac{(1-\alpha) P_{(x,y) \sim \DA} [h(x)=1] \cdot \RA + E_A}{(1-\alpha) \RA + \alpha_A}
\end{equation}

where $\alpha_A$ is the proportion of the data set that is corrupted and in group $A$ and $E_A$ is the proportion of the data set that is corrupted, in group $A$ and positively labeled by $h$.

Our goal is to obtain an upper bound on the difference between $P_{(x,y) \sim \DAC} [h(x)=1]$ and $P_{(x,y) \sim \DA} [h(x)=1]$.
We use the fact that $E_A \leq \alpha$ and $\alpha_A \leq \alpha$ to obtain the following upper bound:

\begin{align*}
 &   \left| P_{(x,y) \sim \DAC} [h(x)=1] - P_{(x,y) \sim \DA} [h(x)=1] \right| \\
 &   = \left| \frac{E_A - \alpha_A P_{(x,y) \sim \DA} [h(x)=1] }{(1-\alpha) \RA + \alpha_A} \right| \leq \frac{\alpha}{(1-\alpha) r_A + \alpha  }
\end{align*}
\end{proof}

\mainparity*

\begin{proof}[Proof of Theorem~\ref{thm:mainparity}]\label{proof:mainparity}
For $z \in \{A, B \}$, let $\normalF_z(h)$ and $\corruptF_z(h)$ denote the proportions of positive labels assigned by $h$ in group $z$ in the original and corrupted distributions respectively. That is, for group $A$, $\normalF_A(h) = P_{(x,y) \sim \DA} [ h (x)=1]$ and $\corruptF_A(h) = P_{(x,y) \sim \DAC} [ h (x)=1]$.
    It suffices to show that there exists $h \in \closure$ that satisfies the guarantees above. 
    % \pcocomment{Might need to add a lemma before this where we show that this is sufficient.}
    Consider $\hstar \in \hclass$. By the realizability assumption 
    % \pcocomment{there are two realizability assumptions here, one where $h^*$ satisfies the violation up to $\delta$ and one where it's exact equality. We'll use the one assuming equality and we'll add a lemma showing that things work fine for the $\delta$ violation one.}
    , $\hstar$ satisfies the parity constraint i.e $\normalF_A(h^*) = \normalF_B(h^*)$. 
    After the corruption, the parity violation of $h^*$, $|\corruptF_A(h^*) - \corruptF_B(h^*)|$ may increase. Now we define the following parameters ($p_z$ and $q_z$) for $z \in \{A, B \}$.
    \begin{equation}
        p_z = \begin{cases}
            \frac{\normalF_z(h^*) - \corruptF_z(h^*)}{1 - \corruptF_z(h^*)} & \text{if} \ \normalF_z(h^*) \geq \corruptF_z(h^*)\\
            \frac{\corruptF_z(h^*) - \normalF_z(h^*)}{\corruptF_z(h^*)} & \text{otherwise}\\
        \end{cases} \quad
        q_z = \begin{cases}
            1 & \text{if} \ \normalF_z(h^*) \geq \corruptF_z(h^*)\\
            0 & \text{otherwise}\\
        \end{cases}
    \end{equation}
    Now consider a hypothesis $\hhat$ that behaves as follows: Given a sample $x$:
    \begin{itemize}
        \item  If $x \in A$, with probability $p_A$, return label $q_A$. Otherwise return $h^* (x)$
        \item Similarly, if $x \in B$, with probability $p_B$, return label $q_B$. Otherwise return $h^* (x)$
    \end{itemize}
    $\hhat \in \PQ$ since it follows the definition of our closure model. We will now show that $\hhat$ satisfies the parity constraint in the corrupted distribution (i.e $\corruptF_A(\hhat) = \corruptF_B(\hhat)$). First, observe that for $z \in \{A, B \} $, if $\normalF_z(h^*) \geq \corruptF_z(h^*)$, then $\corruptF_z(\hhat) = \normalF_z(h^*)$. This is because
    \begin{align*}
        \corruptF_z(\hhat) 
        &= (1 - p_z) \corruptF_z(h^*) + p_z q_z \\
        &= \corruptF_z(h^*) + p_z(1 - \corruptF_z(h^*)) \\
        &= \corruptF_z(h^*) + \normalF_z(h^*) - \corruptF_z(h^*) \\
        &= \normalF_z(h^*)
    \end{align*}
    Similarly, if $\normalF_z(h^*) < \corruptF_z(h^*)$, then $\corruptF_z(\hhat) = \normalF_z(h^*)$. This is because
    \begin{align*}
        \corruptF_z(\hhat) 
        &= (1 - p_z) \corruptF_z(h^*) + p_z q_z \\
        &= \corruptF_z(h^*) + p_z(0 - \corruptF_z(h^*)) \\
        &= \corruptF_z(h^*) + \normalF_z(h^*) - \corruptF_z(h^*) \\
        &= \normalF_z(h^*)
    \end{align*}
    Thus, $\corruptF_A(\hhat) = \normalF_A(h^*) = \normalF_B(h^*) = \corruptF_B(\hhat)$. Therefore $\hhat$ satisfies the parity constraint in the corrupted distribution.
    
    We will now show that $\error{\hhat} \leq O(\alpha) $. Since $\hhat$ deviates from $\hstar$ with probability $p_A$ on samples from $A$, and with probability $p_B$ on samples from $B$, we only need to show that the proportion of samples such that $\hhat (x) \neq h^* (x)$ is small. Fix a group $z \in \{A, B\}$. If $\normalF_z (h^*) \geq \corruptF_z (h^*)$, then with probability $p_z = \frac{\normalF_z (h^*) -\corruptF_z (h^*)}{1 - \corruptF_z (h^*)}$, $\hhat$ returns a positive label for samples in group $z$. Thus, the expected proportion of samples in group $z$ such that $\hhat (x) \neq h^* (x)$ is $p_z$ times the proportion of negative labelled samples (by $h^*$) in group $z$ (since those get flipped to positive).
    \begin{align*}
        \EE_{x \in z} [\one (\hhat (x) \neq h^* (x))] &= p_z \cdot P_{(x,y) \sim \dist } [x \in z] (1 - \corruptF_z (h^*) ) \\
        &= \frac{\normalF_z (h^*) -\corruptF_z (h^*)}{1 - \corruptF_z (h^*)} \cdot P_{(x,y) \sim \dist } [x \in z] (1 - \corruptF_z (h^*) ) \\
        &= (\normalF_z (h^*) -\corruptF_z (h^*)) \cdot P_{(x,y) \sim \dist } [x \in z] 
    \end{align*}
    Similarly, if $\corruptF_z (h^*) > \normalF_z (h^*)$, then with probability $p_z = \frac{\corruptF_z (h^*) -\normalF_z (h^*)}{\corruptF_z (h^*)}$, $\hhat$ returns a negative label. Thus, the expected proportion of samples in group $z$ such that $\hhat (x) \neq h^* (x)$ is $p_z$ times the proportion of positively labelled samples (by $h^*$) in group $z$ (since those get flipped to negative).
    \begin{align*}
        \EE_{x \in z} [\one(\hhat (x) \neq h^* (x))] &= p_z \cdot P_{(x,y) \sim \dist } [x \in z] \cdot \corruptF_z (h^*) \\
        &= \frac{\corruptF_z (h^*) -\normalF_z (h^*)}{\corruptF_z (h^*)} \cdot P_{(x,y) \sim \dist } [x \in z] \cdot \corruptF_z (h^*)  \\
        &= (\corruptF_z (h^*) -\normalF_z (h^*)) \cdot P_{(x,y) \sim \dist } [x \in z] 
    \end{align*}
    Therefore, the expected total number of samples such that $\hhat (x) \neq h^* (x)$ across the entire distribution is bounded as follows:
    \begin{align*}
        \mathbb{E}_{(x,y) \sim \mathcal{D}} \ [\one (\hhat (x) \neq h^* (x))] 
        &= \sum_{z \in \{ A, B \}} |\corruptF_z (h^*) -\normalF_z (h^*)| \cdot P_{(x,y) \sim \dist } [x \in z] \\ 
        &\leq \sum_{z \in \{ A, B \}} \frac{\alpha}{(1- \alpha) P_{(x,y) \sim \dist } [x \in z] + \alpha} \cdot P_{(x,y) \sim \dist } [x \in z] \\ \intertext{by proposition \ref{prop:corruptparity}} 
        &\leq \frac{2\alpha}{(1- \alpha)}
    \end{align*}
    Note that even though the adversary can choose a different distribution at each timestep, we can wlog assume the adversary chooses the same distribution $\widetilde{D}$ where the quantity $|\corruptF_z (h^*) -\normalF_z (h^*)|$ is maximized at every timestep, as in Proposition \ref{prop:corruptparity}.
    Although the model in \cite{kearns1988learning} is slightly weaker than \cite{lampert}, this theorem holds in full generality for both models where we replace the difference $|\corruptF_z (h^*) -\normalF_z (h^*)|$ with the bounds from Lemma 2 of \cite{lampert}. The dependence on $\alpha$ remains the same in both cases.
    % \pcocomment{Because of the way I define $z$, the entire proof/construction should work for any number of groups but we would have $n \times \alpha$ in the numerator for accuracy loss. I wonder if that's avoidable} \pcocomment{Update: I think it's avoidable. The bound in proposition 1 could be improved to make the adversary's changes across all groups sum up to $\alpha$}
\end{proof}

%% file: mal_eopp.tex
\subsection{Equal Opportunity}

\corrupttpr*

\begin{proof}[Proof of Proposition~\ref{prop:corrupttpr}]\label{proof:corrupttpr}
For a fixed group $A$, the TPR of $h$ in $\widetilde{\mathcal{D}}$ can be expressed in terms of the TPR of $h$ in the original distribution $\mathcal{D}$ as follows:
\begin{equation}
        \text{TPR}_A(h, \widetilde{\mathcal{D}}) = \frac{(1-\alpha) \text{TPR}_A(h, \mathcal{D}) \cdot \RA + E_A^+}{(1-\alpha) \RA + \alpha_A^+}
\end{equation}
where $\alpha_A$ is the proportion of the data set that is corrupted and in group $A$ and $E_A^+$ is the proportion of the data set that is corrupted, in group $A$, is positive, and is predicted as positive by $h$.
Thus,
\begin{equation}
    \left| \text{TPR}_A(h, \widetilde{\mathcal{D}}) - \text{TPR}_A(h, \mathcal{D}) \right| = \left| \frac{E_A - \alpha_A \text{TPR}_A(h, \mathcal{D}) }{(1-\alpha) \RA + \alpha_A} \right| \leq \frac{\alpha}{(1-\alpha) r_A^+ + \alpha }
\end{equation}
since $E_A \leq \alpha$ and $\alpha_A \leq \alpha$
\end{proof}

% \begin{theorem}
%     For any hypothesis class $\mathcal{H}$ and distribution $ \dist = (\DA, \DB)$, a robust fair-ERM learner for the equal opportunity constraint in the Malicious Adversarial Model returns a hypothesis $\hhat$ such that 
%     \begin{equation*}
%         \error{\hhat} \leq O(\sqrt{\alpha})
%     \end{equation*}
% \end{theorem}

\maineopp*
\begin{proof} [Proof of Theorem~\ref{thm:maineopp}]\label{proof:maineopp}
We will use Proposition \ref{prop:corrupttpr} and the assumption we introduced in Section \ref{sec:malprelim}, Equation \ref{ass:raplus} to show this statement.

To show the proof overall, suffices to show that there exists $h \in \PQ$ that satisfies the guarantees above. 
Consider $\hstar \in \hclass$. By the realizability assumption, $\hstar$ satisfies the equal opportunity constraint i.e $\text{TPR}_A(h^*, \mathcal{D}) = \text{TPR}_B(h^*, \mathcal{D})$. 
After the corruption, the equal opportunity violation of $h^*$, $|\text{TPR}_A(h^*, \widetilde{\mathcal{D}}) - \text{TPR}_B(h^*, \widetilde{\mathcal{D}})|$ may increase. Now we define the following parameters ($p_z^i$ and $q_z^i$) for $i, z \in \{A, B \}$. 
\begin{equation}\label{eq:tpr-prob}
    p_z^i = \begin{cases}
        \frac{\corruptF_i(h^*) - \corruptF_z(h^*)}{1 - \corruptF_z(h^*)} & \text{if} \ \corruptF_i(h^*) \geq \corruptF_z(h^*)\\
        \frac{\corruptF_z(h^*) - \corruptF_i(h^*)}{\corruptF_z(h^*)} & \text{otherwise}\\
    \end{cases} \quad
    q_z^i = \begin{cases}
        1 & \text{if} \ \corruptF_i(h^*) \geq \corruptF_z(h^*)\\
        0 & \text{otherwise}\\
    \end{cases}
\end{equation}
One can think of the parameter $p_z^i$ as the proportion of samples in group $z$ whose outcomes needs to be changed in order to match the true positivity rate of group $i$.
% \pcocomment{The parameters should be clipped so that they are in the $[0,1]$ interval. will fix later} 
    Now consider two hypotheses $\hhat_i$ for $i \in \{ A, B\}$ that behave as follows: Given a sample $x$:
    \begin{itemize}
        \item  If $x \in A$, with probability $p_A^i$, return label $q_A^i$. Otherwise return $h^* (x)$
        \item Similarly, if $x \in B$, with probability $p_B^i$, return label $q_B^i$. Otherwise return $h^* (x)$
    \end{itemize}
One can think of $\hhat_i$ as a hypothesis that deviates from $h^*$ on every other group to make their true positive rate on the corrupted distribution match that of group $i$.
Observe that $\hhat_i \in \PQ$ for $i \in \{A, B\}$ since it follows the definition of our closure model $\PQ$. We will now show that $\hhat_i$ for $i \in \{ A, B\}$ satisfies the True Positive Rate constraint on the corrupted distribution (i.e $\corruptF_A(\hhat_i) = \corruptF_B(\hhat_i)$ for fixed $i \in \{ A, B\}$). First, observe that for $z \in \{A, B \} $, if $\corruptF_i(h^*) \geq \corruptF_z(h^*)$, then $\corruptF_z(\hhat_i) = \corruptF_i(h^*)$. This is because
    \begin{align*}
        \corruptF_z(\hhat_i) 
        &= (1 - p_z) \corruptF_z(h^*) + p_z q_z \\
        &= \corruptF_z(h^*) + p_z(1 - \corruptF_z(h^*)) \\
        &= \corruptF_z(h^*) + \corruptF_i(h^*) - \corruptF_z(h^*) \\
        &= \corruptF_i(h^*)
    \end{align*}
    Similarly, if $\corruptF_i(h^*) < \corruptF_z(h^*)$, then $\corruptF_z(\hhat) = \corruptF_i(h^*)$. This is because
    \begin{align*}
        \corruptF_z(\hhat) 
        &= (1 - p_z) \corruptF_z(h^*) + p_z q_z \\
        &= \corruptF_z(h^*) + p_z(0 - \corruptF_z(h^*)) \\
        &= \corruptF_z(h^*) + \corruptF_i(h^*) - \corruptF_z(h^*) \\
        &= \corruptF_i(h^*)
    \end{align*}
    Thus, $\corruptF_A(\hhat_i) = \corruptF_i(h^*) = \corruptF_B(\hhat_i)$. Therefore $\hhat_i$ for $i \in \{ A, B\}$ satisfies the Equal Opportunity Constraint on the corrupted distribution.
    
    We will now show that the existence of at least one $\hhat_i$ for $i \in \{A, B\}$ satisfies $\error{\hhat} \leq O(\sqrt{\alpha})$. Since $\hhat_i$ deviates from $\hstar$ with probability $p_A^i$ on samples from $A$, and with probability $p_B^i$ on samples from $B$, it suffices to show that $p_A^i \cdot r_A + p_B^i \cdot r_B$ is $O(\sqrt{\alpha})$ for $i \in \{A, B\}$. 
    This is sufficient because of the Assumptionm in Equation \ref{ass:raplus}.
    
    We consider the following cases:
\begin{enumerate}
    \item Suppose wlog $r_B \leq \frac{\sqrt{\alpha}}{1 - \sqrt{\alpha}}$. Then $\hat{h}_{B}$ satisfies the guarantee. This is because $p_A^B = 0$ (by equation~\ref{eq:tpr-prob} ) and $p_B^B \leq 1$. Thus, $p_A^B \cdot r_A + p_B^B \cdot r_B$ is $O(\sqrt{\alpha})$. 
    %In words, this is the case where 

    \item If instead $\min (r_A, r_B) > \frac{\sqrt{\alpha}}{1 - \sqrt{\alpha}}$. wlog let $B$ be a group with the highest true positive rate greater than 0.5 or the smallest true positive rate less than 0.5. At least one group must satisfy this constraint. If $B$ has the highest true positive rate greater than 0.5, then 
    \begin{align*}
    p_B^A &= \frac{\corruptF_B (h^*) - \corruptF_A (h^*)}{\corruptF_B (h^*)} \\
    &\leq \frac{\corruptF_B (h^*) - F_B (h^*) + F_A (h^*) - \corruptF_A (h^*)}{0.5} \intertext{since $\corruptF_B (h^*) \geq 0.5$ and by realizability assumption $\normalF_B (h^*) = \normalF_A (h^*)$}
    &\leq 2 |\corruptF_B (h^*) - F_B (h^*)| + 2 |\normalF_A (h^*) - \corruptF_A (h^*)| \\ \intertext{by proposition~\ref{prop:corrupttpr} and the Assumption in Equation \ref{ass:raplus}}
    &\leq O (\sqrt{\alpha})
    \end{align*} 
    Thus, $p_A^A \cdot r_A + p_B^A \cdot r_B$ is at most $O(\sqrt{\alpha})$
    The case where $B$ has the smallest true positive rate follows similarly.
\end{enumerate}
Similar to the proof of Theorem~\ref{thm:mainparity}, we can assume wlog the adversary chooses the same distribution $\widetilde{D}$ where the quantity $|\corruptF_z (h^*) -\normalF_z (h^*)|$ is maximized at every timestep, as in Proposition \ref{prop:corruptparity}.
Although the model in \cite{kearns1988learning} is slightly weaker than \cite{lampert}, this theorem holds in full generality for both models where we replace the difference $|\corruptF_z (h^*) -\normalF_z (h^*)|$ with the bounds from Lemma 5 of \cite{lampert}. The dependence on $\alpha$ remains the same in both cases.
\end{proof}

%\lowereopp*
%\begin{proof} [Proof of Theorem~\ref{thm:lowereopp}]\label{proof:lowereopp}
%Suppose group A is of size $1 - \sqrt{a}$ and B is of size $\sqrt{a}$. Suppose the best classifier in the hypothesis class can only attain $(1 - \sqrt{\alpha})\%$ percent TPR on both groups. An adversary with $\alpha\%$ can corrupt the distribution so that this classifier has $100 \%$ percent on corrupted distribution for group $B$. Fix a classifier $h$ returned by a learner in this setting. In order to satisfy the perceived fairness constraint of the ERM solver i.e tpr of $h$ must be the same for both groups in the corrupted distribution, then  

%\end{proof}

\lowereopp*

\begin{proof} [Proof of Theorem~\ref{thm:lowereopp}]\label{proof:lowereopp}
We will show a distribution and a malicious adversary of power $\alpha$ such that any hypothesis returned by a learner incurs at least $\sqrt{\alpha}$ expected excess error.
The distribution $\mathcal{D}$ will be such that $P_{x \sim \mathcal{D}}[x \in B]= \Omega(\sqrt{\alpha})$. This distribution will be supported on exactly four points $x_1 \in A, x_2 \in A, x_3 \in B, x_4 \in B$ with labels $y_1 = +, y_2 = -, y_3 = +, y_4 = -$. We also have that 
$$P_{x,y \sim \mathcal{D}}[x = x_1, y = +] = P_{x, y \sim \mathcal{D}}[x = x_2, y = -] = \frac{1 - \sqrt{\alpha}}{2}$$ 
and 
$$P_{x, y \sim \mathcal{D}}[x = x_3, y = +] = P_{x,y \sim \mathcal{D}}[x = x_4, y = -] = \frac{\sqrt{\alpha}}{2}$$
That is, each group has equal proportion of positives and negatives.
%\textcolor{red}{ should this say $\DA$ instead of $\mathcal{D}$}

The adversary commits to a poisoning strategy that places positive examples from Group $B$ into the negative region of the optimal classifier. That is, the adversary changes the original distribution $\mathcal{D}$ so that 
$$P_{x, y \sim \mathcal{D}}[x = x_1, y = +] = P_{x, y \sim \mathcal{D}}[x = x_2, y = -] = \frac{(1-\alpha)(1 - \sqrt{\alpha})}{2}$$
$$P_{x, y \sim \mathcal{D}}[x = x_3, y = +] =
P_{x, y \sim \mathcal{D}}[x = x_4, y = -] = \frac{(1-\alpha)\sqrt{\alpha}}{2}$$ and $P_{x, y \sim \mathcal{D}}[x = x_4, y = +] = \alpha$

We assume the perfect classifier is in the hypothesis class.
Now fix a classifier $h$ returned by a learner. This classifier must satisfy equal opportunity. Let $p_1, p_2, p_3, p_4$ be the probability that 
$h$ classifies $x_1, x_2, x_3, x_4$  as positive, respectively. 
Observe that $\widetilde{\text{TPR}}(h_A) = p_1$ and $\widetilde{\text{TPR}}(h_B) = 1 - (1 - p_4) \alpha' - (1-p_3)(1 - \alpha')$ where $\alpha' = \frac{2\sqrt{\alpha}}{(1 - \alpha) + 2\sqrt{\alpha}}$. The latter is due to the samples $(x_4, +)$ which the adversary added to the distribution. 
The adversary added an $\alpha$ amount which turned out to be an $\alpha'$ proportion of the positives in $B$. 
Since this classifier satisfies equal opportunity on the corrupted distribution, it must be the case that $p_1 = 1 - (1 - p_4) \alpha' - (1-p_3)(1 - \alpha')$. Thus, $(1 - p_1) \geq (1 - p_4) \alpha'$.
The error of $h$ on the original distribution is therefore
\begin{align*}
& (1 - p_1 + p_2) \frac{(1 - \sqrt{\alpha})}{2} + (1 - p_3 + p_4) \frac{\sqrt{\alpha}}{2} \\
\geq & \ (1 - p_1) \frac{(1 - \sqrt{\alpha})}{2} + p_4 \frac{\sqrt{\alpha}}{2} \\ \intertext{by the equal opportunity constraint}
\geq & \ (1 - p_4) \alpha' \frac{(1 - \sqrt{\alpha})}{2} + p_4 \frac{\sqrt{\alpha}}{2} \\
= & \ (1 - p_4) \cdot \frac{2\sqrt{\alpha}}{(1 - \alpha) + 2\sqrt{\alpha}} \cdot \frac{(1 - \sqrt{\alpha})}{2} + p_4 \frac{\sqrt{\alpha}}{2} \\
\geq & \ (1 - p_4) \frac{\sqrt{\alpha}}{2} + p_4 \frac{\sqrt{\alpha}}{2}  \geq \Omega (\sqrt{\alpha}) 
\end{align*}
\end{proof}

%% file: mal_eodds.tex
\section{Equalized Odds}
\label{sec:eoddsproof}
Now we will consider Equalized Odds.
%which will have a similar lower bound to Calibration in Theorem \ref{thm:calib}. 

\begin{proof}[Equalized Odds Proof  of $\Omega(1)$ accuracy loss:] \label{proof:eodds}
%To show that for Equalized Odds requires $\Omega(1)$ error when the adversary has corruption budget $\alpha$, even with our hypothesis class $\PQ$, 
it suffices to exhibit a `bad' distribution and matching corruption strategy; which we exhibit below.

\begin{enumerate}
\item Say Group A has $1-\alpha$ of the probability mass i.e. $P_{(x,y) \sim \calD}[x \in A] \geq 1-\alpha$ and thus $P_{(x,y) \sim \calD}[x \in B] \leq \alpha$.
\item The positive fraction for each group under distribution $\mathcal{D}$ is $P_{(x,y) \sim \DA}[y=1]=P_{(x,y) \sim D_B}[y=1]=\frac{1}{2}$
\item Since $P_{(x,y)\sim \calD}[x \in B] \leq \alpha$, the adversary has sufficient corruption budget such that they can inject a duplicate copy of each example in B but with the opposite label.   
That is, for each example x in Group B in the training set, the adversary adds another identical example but with the opposite label.
\end{enumerate}

This adversarial data ensures that on Group $B$, any hypothesis $h$ (of any form) will now satisfy 
\[ P_{x \sim \hat{\mathcal{D}}_{B}}[h(x)=1 | y=1] = P_{x \sim   \hat{\mathcal{D}}_{B}}[h(x)=1  | y=0] = p  \] for some value $p \in [0,1]$
due to the indistinguishable duplicated examples; i.e. the hypothesis can choose how often to accept examples [e.g. increase or decrease $p$] but it cannot distinguish positive/negative examples in Group $B$.
%\footnote{Since these distributions are evenly balanced by class, $P_{x \sim \DB}[h(x)=1]=p$.}

Note that we can select $p$ using some arbitrary $h$ but that randomness does not help us.
Observe that similarly, the True Negative/False Negative Rates on Groyp $B$ must be $1-p$.
%So, for group A, to satisfy equalized odds, both thes terms must  must also equal $1/2$

Since $A$ is evenly split among positive and negative and we must satisfy Equalized Odds, 
this means that our error rate on group A is
\begin{align*}
& P_{(x,y) \sim \DA}[ h(x) \neq y ] =  P_{(x,y) \sim \DA}[ h(x) \neq y \cap y=1] +  P_{(x,y) \sim \DA}[ h(x) \neq y \cap y=0] \\
& =P_{(x,y) \sim \DA}[ h(x) \neq 1 | y =1  ]P[y=1] + P_{(x,y) \sim \DA}[ h(x) \neq 0 | y = 0  ]P[y=0]  \\
& = P_{(x,y) \sim \DA}[ h(x) \neq 0 | y =1  ]P[y=1] + P_{(x,y) \sim \DA}[ h(x) = 1 | y = 0  ]P[y=0] \\
& = (1-TPR_{A}) \frac{1}{2} + FPR_{A} \frac{1}{2} \\
& =(1-p)(\frac{1}{2}) + p(\frac{1}{2})= \frac{1}{2}
\end{align*}
So, the adversary has forced us to have $50 \%$ error on group A which yeilds the result.
\end{proof}

%% file: mal_calibproof.tex
\begin{proof}[Proof of Theorem ~\ref{thm:predparity}, Predictive Parity Lower Bound]  \label{proof:predparity}

%The proof of this lower bound is similar to that of Proof \ref{proof:eodds} but with the flipped conditioning. 

To show that Predictive Parity requires $\Omega(1)$ error when the adversary has corruption budget $\alpha$, even with our hypothesis class $\PQ$, 
it suffices to exhibit a `bad' distribution and matching corruption strategy; which we exhibit below. 

Recall that we require that $P_{x \sim \DA}[h(x)=1]>0$ and $P_{x \sim \DB}[h(x)=1]>0$. 
This is to avoid the case where the learner rejects all points from Group $B$.

\begin{enumerate}
\item Assume that group A has $1-\alpha$ of the probability mass i.e. $P_{(x,y) \sim \calD}[x \in A] \geq 1-\alpha$ and thus $P_{(x,y) \sim \calD}[x \in B] \leq \alpha$.
\item The positive fraction for each group under distribution $\mathcal{D}$ is $P_{(x,y) \sim \DA}[y=1]=P_{(x,y) \sim D_B}[y=1]=\frac{1}{2}$
\item Since $P_{(x,y) \sim \calD}[x \in B] \leq \alpha$, the adversary has sufficient corruption budget such that they can a duplicate copy of each example in B but with the opposite label.   
That is, for each example x in Group B in the training set, the adversary adds another identical example but with the opposite label.
\end{enumerate}

This adversarial data ensures that on Group $B$, any hypothesis $h$ (of any form) will now satisfy 
\[  P_{(x,y) \sim \hat{\mathcal{D}}_{B}}[y=1 | h(x)=1 ] = P_{(x,y) \sim   \hat{\mathcal{D}}_{B}}[y=0  | h(x)=0]=\frac{1}{2} \]
due to the indistinguishable duplicated examples.  
So, for Group A, to satisfy Predictive Parity, both these terms must also equal $\frac{1}{2}$ and induce $50 \%$ error on Group $A$.
\end{proof}

\begin{proof}[Proof of Theorem~\ref{thm:calib}, Calibration $O(\alpha)$.]\label{proof:calib}

In order to prove this statement, we consider $h^{*}$ which is the Bayes Predictor $h^{*} = \mathbb{E}[y|x]$, but using some finite binning scheme $[R]$.
Clearly $h^{*}$ is calibrated on natural data and $h^{*}: \mathcal{X} \rightarrow [R]$. 

We  will show how to modify $h^{*}$ to still satisfy the fairness constraint on the corrupted data without losing too much accuracy, regardless
of the adversarial strategy.

In the case of Calibration,  we will do this by just separately re-calibrating each group. 

Let $[\hat{R}]:=[R]$.
We will now modify $[\hat{R}]$ from $[R]$ to be calibrated on the malicious data.

That is;
For each group $z$ (i.e $z=A$ or $z=B$), for each bin $r \in [R]$ (i.e., ${x: h^{*}(x)=r}$), we create a new bin if there is no bin in $[R]$ with value  $\hat{r} = E_{(x,y) \sim \mathcal{D}_z}[y | h^{*}(x)=r]$.  

That is, we define $\hat{h}(x) = \hat{r}$ for all $x \in g$ such that $h^{*}(x)=r$.

Observe that by construction, $\hat{h}$ is calibrated separately for each group, so it is calibrated overall.  We just need to analyze the excess error of $\hat{h}$ compared to $h^{*}$. 
We will show this is only
$O(\alpha)$.

%Recall that our notion of accuracy for this problem is given by thresholding the binning scheme at $\frac{1}{2}$ and giving bins above that threshold value the positive prediction and predicting negative otherwise. 
Observe that increase in expected error is how much that bin is shifted from the true probability $h^{*}(x)$.

For each bin $r \in[R]$, the shift in 
$|r - \hat{r}|$ is at most the fraction of points in the bin that are malicious noise. 
Let $x \in MAL$ mean point $x$ is a  corrupted point.

Then 
\begin{align*}
& \mathbb{E}_{x \sim \mathcal{D}}[h^{*}(x)-\hat{h}(x)] \leq \sum_{r \in [R]} P[x \in r] |r-\hat{r}| \\
%\leq \sum_{r \in [R]} P[x \in r] |r-\hat{r}| \\
& = \sum_{r \in [R]} P[x \in r] \frac{P[x \in r \cap x \in MAL]}{P[x \in r]} \\
& \leq \sum_{r \in [R]} P[x \in r \cap x \in MAL] = O(\alpha) \quad \text{ Definition of Malicious Noise Model}
\end{align*}
Note that this is considering $L1$ error, accuracy loss is less than for $L2$ error, immediate for since $\alpha \in [0,1)$.

\end{proof}

%% file: mal_minimax.tex
\section{Minimax Fairness}
\label{subsec:minimaxfair}
In this Section, we will briefly and informally consider Minimax Fairness.
Introduced in \cite{minimaxfair} this notion  optimizes for a different objective. 

Using their notation ($\epsilon_k = \mathbb{E}_{(x,y) \sim \mathcal{D}_k}[h(x) \neq y]$ or group-wise error) with a groupwise max error bound of $1> \gamma > 0$
\begin{align*} 
h^{*} = \argmin_{h \in \Delta{H}} \quad  \mathbb{E}_{(x,y) \sim \mathcal{D}}[h(x)\neq y] \\
\max_{1 \leq k \leq K}  \epsilon_{k} (h) \leq \gamma
\end{align*}
Letting $OPT$ refer to the value of solution of the optimization problem, the learning goal is to find an $h$ that is $\epsilon$-approximately optimal for the mini-max objective, meaning that $h$ satisfies: 
\[ max_{k} \epsilon_{k}(h) \leq OPT + \epsilon\]

Observe that if the goal of the learner is compete with the value of $OPT$ on the unmodified data, in our malicious noise model this objective is
ineffective since if one group is of size $O(\alpha)$, the adversary can always drive the error rate on that group $\Omega(1)$.

This model seems incompatible with malicious noise due to the sensitivity of minimax fairness to small groups. 

Observe that the Minimax Fairness framework includes Equalized Error
rates as a special case.

%\begin{align*}
%& \min_{h \in \simplex{H}} err(h) \\
%& \text{subject to} err_{k}(h) \leq \gamma, k=1, \dots K
%\end{align*}

%Observe that when one sub-group is $O(\alpha)$ size, the adversary can always drive t

%% file: mal_multgroup.tex
%\section{Multiple Groups}
%\label{multgroup}

%In this section will summarize our results for the case of $k$-multiple groups for the convience of the reader ---- extend

%% file: fairscreen_content.tex
\section{Introduction}
Consider what we will term \textit{sequential screening processes}. 
In this setting a decision maker (e.g. a company seeking to hire applicants) makes a decision, 
like hiring, by using a sequence of intermediate decision-making steps that each filter out some candidates, in order to ideally produce a pool of mostly qualified candidates at the final step. 

We assume some people are truly qualified for the position being filled, and we call them positive examples, and others are truly unqualified and we call them negative examples. And then the various intermediate steps have different probabilities of qualified/unqualified applicants passing each step, which could be different for different demographic groups.  We also assume that the final (interview) stage of the process is particularly expensive for the decision-maker, and reveals the true label of the applicant.

%%%%
%Imagine a two-step filtering and selection process to hire employees. In the first stage, the Learner has access only to resumes of individuals from group $A$ and group $B$.  Using the resumes to filter out un-promising applicants, the Learner then selects a subset to receive an interview, which is presumed to be perfectly accurate in selecting candidates to proffer offers.

%However, the stages differ in their cost structures for the firm. Scanning resumes via an automated system is cheap and its cost is negligible, but the interviewing stage is costly. In response, the firm will be very attentive to selecting only the most promising individuals to promote for interviewing, in order to minimize the expected number of interviews required to discover $k$ qualified candidates (verified by passing the final interview). Assume that the final interview can perfectly distinguish true positives from true negatives.

%In contrast to the standard PAC learning set-up,
%The goal of the learner is not to learn an accurate model for the sake 
%of learning an accurate model.
%An accurate model is only helpful if it can identify %enough
%qualified individuals. 

To illustrate a concern that could arise in this setting, suppose there are two demographic groups $A$ and $B$, and just one test $t$ in the screening process prior to the final stage.  Suppose that test $t$ and the underlying base rates of the two groups have the property that
%individuals are interviewed only if $t(x)=1$.
%For example, consider two groups $A$ and $B$ and a test $t$ such that
%with equal base rates and some $x$ and $x'$ identical other than group membership but such that 
$\Pr[y=1|t(x) =1, x \in A]  \geq \Pr[y=1 |t(x)=1, x \in B] + \epsilon$ for some $\epsilon > 0$.  That is, the pool of group-$A$ applicants who pass the test has a higher fraction of positive examples than the pool of group-$B$ applicants who pass the test.
%; the probability that a randomly selected example from group $A$ has `positive' outcome (i.e., will pass the final interview stage') is slightly higher than the one for a randomly selected example from group $B$.
Since the cost of final interviews is assumed to be high, in this case a rational decision maker would be sensitive to even a small $\epsilon$ gap, in order to minimize the expected number of interviews made per hire. 
In particular, small gaps between these groups in the population would be amplified in that
the rational decision-maker 
would then choose not to promote \textit{any} individuals from group $B$ to the final interview round, which clearly violates common sense fairness norms. 
There is empirical evidence that similar phenomenon occurs in real world settings, when employers have limited information~\citep{bertrand2004emily}.

A second concern is that even if the decision-maker interviews all individuals who make it to the final round (and more generally, at each level promotes all individuals who pass the test to the next round), the tests themselves could have the property that qualified individuals from some groups pass them more easily than qualified individuals from others.  So, in the end, a qualified individual from one group might have a much lower chance of making it to the final interview round than a qualified individual from another.

%%%%
Because of fairness violations of this kind, we consider a regulator that requires the screening process to satisfy Equal Opportunity~\citep{hardt16}, that is, qualified individuals of each group have the same chance of receiving an interview. 
This requirement motivates
%brings up
the 
%question
problem
of how to satisfy such a condition in the most efficient way, minimizing the number of interviews needed per successful hire as well as the number of overall applicants needed to enter the screening process per hire. 
This is the question we address in this chapter.

We assume that the tests themselves and their order in the process are fixed beforehand and the action space of the firm (of our algorithm) is solely modifying how individuals move through the pipeline in response to their test outcomes (the promotion policy).
More specifically, for each test, we need to decide the probability that an individual from a given group who passes or fails the test should continue on to the next stage.
One can satisfy the fairness requirement with simple promotion policies (such as promoting all individuals regardless of whether they pass or fail each test), but the tension is how to do so in a way that results in a useful process.

This captures the scenario of performing  modifications to pre-existing screening systems (the test themselves are fixed) in order to respond to fairness issues. 
We assume we are given, for each test, its statistical properties for each group (the probability that a random qualified or unqualified individual will pass the test).\footnote{If we were to design a socio-technical system from first principles using the insights of machine learning research, we might seek to design tests that are ideally more robust to group difference and still predictive, however such a re-design process could be costly and slow. 
In a world of limited resources, re-purposing pre-existing tests to be more fairness aware in a timely manner and still maintaining effectiveness is necessary.} 

\subsection{Our Results}
We study how to implement the fairness requirement of Equal Opportunity in this sequential screening setting and what method of implementing it would achieve a high efficiency.  
One of our core results is that there is a solution that maximizes precision (minimizes the number of interviews needed per successful hire) subject to maintaining Equal Opportunity, that is given by promoting individuals from each group according to what we call the {\em opportunity ratio}. Moreover, it is possible to maximize overall precision subject to satisfying Equal Opportunity by a policy in which each level in the process satisfies Equal Opportunity individually (this property will not hold for the more general objective below). 
%is given by the Opportunity Ratio, in which we do not promote some fraction of group $A$ individuals who pass the first step of the pipeline in order to match the fraction of positives in group $B$ that pass the test.
%After the first step, we promote every sample who passes each test and promote no test-failers.

Then we consider the more general case of satisfying Equal Opportunity while maximizing a linear combination of precision and recall (1/precision is the expected number of interviews needed per successful hire, and 1/recall is proportional to the number of overall applicants needed to enter the screening process per hire). This problem is challenging because, as we show, the space of Equal-Opportunity solutions is non-convex. Moreover, the optimal way to use one test to optimize a linear combination of precision and recall may depend on %other tests in the system.
all other available tests.

Nonetheless, we are able to achieve an FPTAS for maximizing any linear combination of precision and recall, as well as an exact algorithm with running time that is `only' exponential in the number of levels $k$ and the number of the groups. This latter result relies on certain structural properties of optimal solutions that we develop in our analysis. 
Finally, we discuss extensions to our model such as requiring the screening process to be group-blind, and considering the requirement of satisfying Equalized Odds.  Unfortunately, the optimal fair group-blind policy may be much worse than the optimal fair group-aware policy. For example, in some cases it may require a policy that completely bypasses all the tests.
%In particular, in Section \ref{sec:single-policy} we show that when we are required to have a single promotion policy for all groups (group-blind), that requiring Equal Opportunity results in an a vacuous model in the worst case. 
 
%We additionally defer selected proofs to the appendix in order to enhance comprehension.
\subsection{Related Work}
Fairness in pipelines was initiated by \cite{BowerKNSVV17} and follow up work by \cite{dwork2018fairness, dwork2020individual}. This work
%model 
differs from~\citep{dwork2020individual} in several keys ways. 
We both use the word `pipelines' but our work is more focused on the specific case of hiring pipelines in which we are looking at the fairness of the final outcome for a given individual, drawn from the population, rather than considering the individual fairness~\citep{dwork2012fairness} of the cohort context to which one is assigned.
We do not consider cohort based scoring rules.

The structure of our model is very close to that of \cite{downstream18}, but the objective in that work 
is jointly designing college admission and grading schemes that
satisfy Equal Opportunity over the admissions/college process \textit{and in particular} incentivize a rational employer to use a group blind hiring policy.
In contrast, our work considers maximizing precision or a linear combination of recall and precision while satisfying Equal Opportunity.

Another related work by \cite{arunachaleswaran2022pipeline} is the idea of pipeline interventions. 
In that paper there is a wide pipeline with a finite number of states at time $t$ and the goal of the algorithm designer is to modify the transition probabilities from state to state in order to maximize a reward at the final step. 
This corresponds to efficiently allocating a government subsidy to aid dis-advantaged individuals, from the perspective of maximizing social welfare.

Intriguingly, the paper by \cite{khalili2021fair} argues that Equal Opportunity is misaligned with fairness in screening allocation problems with a finite number of available items (think hiring a small number of engineers at a start-up vs accepting applicants for a credit card).
In our work, we do not focus on modeling a finite number of available positions (e.g., we are in the case with a larger number of available items).

Most closely related to our work is \cite{mansourscreen}, in which there is noisy Bernoulli feedback in a hiring setting with sequential tests.
In contrast to our scenario, they assume both underlying candidate skill levels and test results are sampled independently from Bernoulli distributions. 
Furthermore, they allow hiring an applicant before the end of the pipeline (e.g., if you pass the first three of five tests and those tests have high signal, you may skip the next two tests).
In our model, we assume each stage of the process is memoryless (the probability of making it to stage 3 from stage 2 depends only on the result of the stage-2 test and group membership, and not the result of the stage-1 test) and we allow tests to be asymmetric (e.g., it could be that positive examples from a given group pass with probability 0.75 and negative examples pass with probability 0.5).
In our motivation, we model the initial tests as cheap while the ultimate interview is expensive and accurate, while in \cite{mansourscreen}, 
each test is equally accurate and costly and additionally they want to minimize the expected number of tests to hire a candidate. 
Consistent with our perspective, the authors exhibit an impossibility result arguing that satisfying Equal Opportunity requires group dependent thresholds if the tests have different noise rates.  
%In contrast to our work, they exhibit a greedy algorithm based on a random walk, but they are solving for a different objective (i.e. minimizing the expected number of tests) while we focus on the precision at the end of the pipeline and on the linear combination of precision and recall.

Additionally, there are connections between our work and classical economic discussions of statistical discrimination \cite{arrow72, phelps72} in that both perspectives model disparities in outcomes that derive from strategic actors making decisions to allocate
goods differently based on perceived differences in predicted outcomes (termed statistical discrimination). 
Our models do not capture taste based discrimination.

\subsection{Roadmap} %SHRINK shorter
In Section~\ref{model} we formally describe our model and present some examples that show key phenomena. 
In Section~\ref{maxprecsection} we prove and discuss our first main theorem, about how to maximize precision (at the end of the screening process)
subject to Equal Opportunity.

Then we consider the more general case of satisfying Equal Opportunity while maximizing a linear combination of precision and recall. This problem is challenging because, as we show in Section~\ref{sec:problem-statement-and-example}, the space of Equal-Opportunity solutions is non-convex. Moreover,  
how to effectively utilize a test may depend on all other available tests (Section~\ref{sec-test-nonmono}).
On the other hand, as we show in Section~\ref{sec:exact-alg}, the solution space does satisfy certain useful structural properties.
We then use these structural results to to achieve an exact optimal algorithm, and in Section~\ref{sec:fptas} to achieve an FPTAS for maximizing linear combination of precision and recall, as well as other functions of precision and recall.

Finally, in Section~\ref{sec-altmodels} we discuss extensions to our model such as requiring the screening process to be group-blind, and considering the requirement of satisfying Equalized Odds. 
%This work provides a method to implement the Equal Opportunity constraints in this type of screening process, and explains trade-offs of fairness variations in this model. 
%We defer selected proofs to the appendix in order to enhance comprehension.

\section{Preliminaries\label{model}}
Now we formally define our model and introduce some informative examples.
As mentioned above, the scenario to keep in mind is a stylized hiring process, consisting of a sequence of tests or interviews. 
Each candidate takes a test, and depending on their outcome on that test at that stage, is possibly promoted to the next stage of the screening process. 
We focus on modifying this promotion policy in response to satisfying the fairness constraints and achieving a high objective value or a low cost value.
This is a constrained optimization problem, with structure. 

\subsection{Definitions}
We use ${\mathcal X}$ to denote the set of demographic groups, and $X\in {\mathcal X}$ to denote a specific group.
We assume group membership is known to the algorithm, groups are disjoint, and an individual from group $X$ is promoted based on both their test performance and a promotion policy (defined below) for that corresponding group.
We assume individuals are either truly qualified or truly unqualified, and use label $y=1$ to denote a truly-qualified individual and label $y=0$ to denote a truly-unqualified individual. For each group $X$, let $q_X$ denote the base rate for that group, namely $\Pr[y=1 | x \in X]$.

\begin{defn}[{\bf Test Statistics}]
For each test $t$ and each group $X\in {\mathcal X}$, we define $\tau_{X1} := \Pr[t(x, y)=1 | y=1, x\in X]$ to be the probability a qualified candidate from group $X$ passes the test, and $\tau_{X0} := \Pr[t(x, y)=1 | y=0, x\in X]$ to be the probability an unqualified candidate from group $X$ passes the test. 
We assume all tests are {\em minimally effective} for all groups in that positive examples are more likely to pass than negative examples. More precisely, \begin{align}\label{eq:minimally-effective}
    \tau_{X1} > \tau_{X0}\ge 0
     \quad \forall X\in \mathcal X &&\text{(Minimal Effectiveness Property)}
\end{align}
Note that we assume that the probability of an individual passing a given test depends only on their true qualification $y$ and their group membership $X$. We also assume test statistics are given and known to our algorithm.  
\end{defn}
We use 
$\tau_{X1}^{j}$, $\tau_{X0}^{j}$ to denote the test statistics at \emph{stage j} of the interview process.
For convenience, we define $T_{X}^j = (\tau^j_{X1}, \tau^j_{X0})$ as useful shorthand to capture the test statistics at stage $j$ for group $X$. 
Note that the same test may have different effectiveness per group.
%, and we use this effectiveness to decide how to promote or screen out individuals as they move through the screening process. 

\begin{defn} [{\bf Post-Processing Modification}]
We would like to modify the outcomes of the tests in the screening process so 
that some fairness goal (to be specified later) is achieved at the end of the screening (i.e., in the final interview stage).
Further, we assume as part of the problem setting that the only `allowed' correction is to modify how 
candidates are promoted to the next stage.
The promotion probability of each candidate only depends on their group membership and performance at the {\em current} test (whether they passed or failed the test). 
Formally, for each group $X\in {\mathcal X}$, let $\pi_{X1}^j$ denote the probability a candidate $x \in X$ who passes the test at stage $j$ is promoted to stage $j+1$, and $\pi_{X0}^j$ the probability that a candidate who fails the test at stage $j$ is promoted to stage $j+1$.\footnote{Note, in general randomized promotion policies will be necessary to satisfy the fairness criteria.}
We describe a policy for a given stage $j$ as $\{(\pi_{X1}^j, \pi_{X0}^j)\}_{X\in \mathcal X}$.
\end{defn}

For instance, a naive fairness respecting solution is to simply ignore the tests and promote all examples to the end of the pipeline, i.e., $\{(\pi_{X1}^j=1,\pi_{X0}^j=1)\}_{X \in \cX, j \in [k]}$  where $k$ is the number of
tests in this screening process.
However, this would result in a useless process from the perspective of the decision maker. 
The most straightforward use of tests is to promote all who pass and none who fail, i.e., $\{(\pi_{X1}^j=1,\pi_{X0}^j=0)\}_{X \in \cX, j \in [k]}$. However, this might not satisfy 
required fairness properties. 
%\AV{For consistency, better to use $k$ instead of $K$.} 
We now formally describe the fairness properties we consider. 

%If the decision maker only wants to maximize precision, this could be a useful solution, but depending on the $\tau_{Bi}$, could result in no positives from group $B$ reaching the end of the screening process and causing fairness concerns, or the positives in group $B$ have considerably smaller chance to reach the interview stage compared to the positives in group $A$. We emphasize that for each group $X\in \mathcal X$ and each stage $i\in [k]$, $T_{Xi}$ are characteristics of the test $t$ in level $i$ and are given to us and cannot be modified. 

\begin{defn}[\bf Equal Opportunity and Equalized Odds~\citep{hardt16}] 
This chapter primarily discusses two fairness notions, specifically {\em Equal Opportunity} and {\em Equalized Odds}.
The first notion, Equal Opportunity requires that the classifier have equal {\em True Positive Rates} for each group in the population. Equivalently, 
for a classifier $h$ and true labels $y$, $\Pr[h(x) =1 | y(x)=1, x \in A] = \Pr[h(x) =1 | y(x)=1, x \in B]$.
Equalized Odds is similar but it also requires that the {\em False Positive Rates} are equal; formally, $\Pr[h(x) =1 | y(x)=0, x \in A] = \Pr[h(x) =1 | y(x)=0, x \in B]$.
%\footnote{While these notions are somewhat deprecated in the fairness in machine learning literature, their simplicity in auditing and deployment mandates that they continue to be studied from a theoretical perspective. \avnote{not sure we want to keep it.}}
\end{defn}
In our problem, Equal Opportunity is motivated by a desire that qualified individuals should have the same shot at an interview regardless of their group membership.
%, and a concern that without any constraints, a rational decision-maker might only select individuals to interview from whichever group has highest precision at the last level. 
%otherwise would in a sense, over-fit, to small gaps in precision at the final layer between the groups and totally discard the positive examples from a certain group even though there is a relatively small gap between the two groups. 
In our problem, there is additionally a critical distinction between %Equal Opportunity/Equalized Odds
the fairness criteria (e.g. Equal Opportunity or Equalized Odds) being satisfied at the end
%holding over the entire 
pipeline and alternatively that requiring these criteria hold for every transition between stages as individuals move through the pipeline, a stronger notion.

Now that we have described the terms that characterize a problem instance and the action space of the algorithm, we describe the objective value that captures the usefulness of a screening process.
We term these multiple different objective functions `pipeline efficiency'.
 \begin{defn}[{\bf Pipeline Efficiency}]\label{def-pipeline-eff}
% In contrast to the above fairness notions, we also want to capture the utility of the screening process from the perspective of the firm deploying the model. 
 In our work we focus on two core notions of efficacy from the perspective of the firm deploying the screening process.
 {\em Interview efficiency} (equivalently, {\em precision}) is the fraction of candidates in the last round who are qualified, i.e., the fraction of interviews that lead to hires (or at least to job offers). 
 {\em Throughput efficiency} (equivalently, {\em recall}) is fraction of qualified candidates who make it to the final round, and determines the expected number of applicants needed to enter the pipeline to hire one candidate.
% This quantity is proportional to the inverse of the fraction of true positives who make it to the end of the pipeline (recall).
 %
 In this chapter, we study cost functions that are functions of these two quantities only.
 %; interview complexity ($\propto$ 1/precision) and advertisement complexity ($\propto$ 1/recall).
 \end{defn}
 We model the last available test as highly discriminative but extremely expensive per each test utilization  and this is what motivates the interview efficiency. 
 In particular, if we assume that the $k$ stages prior to the interview round have zero or negligible cost per test, and there are many available candidates,
 then we presume that the goal of the firm is to maximize the interview efficiency (precision, at the final round). 

\subsection{Formal Problem Statement and Illustrative Examples}\label{sec:problem-statement-and-example}
Now, we combine the above into a formal statement. 
Given a screening process/pipeline $\mathscr{P}$ with $k$ stages, this pipeline consists of a collection of disjoint groups $\mathcal{X}$ and tests statistics $T_X = (T_{X}^{1}, T_{X}^{2}, \dots T_{X}^{k} )$ for every group $X\in \mathcal{X}$. 

%The goal of the algorithm designer is to exhibit a method to find promotions policies $\{(\pi_{X1}^{1}, \pi_{X0}^{1}), \dots, (\pi_{X1}^{k}, \pi_{X0}^{k})\}_{X\in \mathcal{X}}$ denoted as $\pi$ such that the overall policy satisfies the relevant fairness notion (either at the end of the screening process or at the end of each stage) and maximizes the given pipeline efficiency.
The goal of the algorithm designer is to exhibit a method to find promotion policies $\{(\pi_{X1}^{j}, \pi_{X0}^{j})\}_{X\in \mathcal{X}, j\in [k]}$ denoted as $\pi$ such that the overall policy satisfies the relevant fairness notion (either at the end of the screening process or at the end of each stage) and maximizes the given pipeline efficiency.
Now we move into illustrative examples.

\paragraph{An illustrative one-stage example:} Consider a one-stage pipeline with test parameters $$((\tau_{A1},\tau_{A0}), (\tau_{B1},\tau_{B0})) = ((1,0.5), (0.8,0.5)).$$
Observe that the policy of promoting individuals if and only if they pass the test does not satisfy Equal Opportunity. 
Instead, two policies that satisfy Equal Opportunity are
$P=((\pi_{A1}, \pi_{A0}), (\pi_{B1}, \pi_{B0})) = ((0.8,0), (1,0))$ and policy $Q=((1,0), (1,1))$.
In words, the policy $P$ would promote all individuals who passed the test from group $B$, but would only promote $80\%$ of those from group $A$. 
This down-weighting of group $A$ would suffice to satisfy Equal Opportunity. 
In contrast, policy $Q$ promotes all individuals from group $A$ who pass the test and promotes everyone from group $B$, regardless of their test score.
In this example, $P$ is the optimal Equal Opportunity policy with respect to precision.

\paragraph{The set of policies satisfying Equal Opportunity is not convex:}
Interestingly, for a two stage pipeline with two groups, the set of policies satisfying Equal Opportunity is not convex.
Consider a pipeline with first level $T_{A}^{1} = (3/4, 0)$ and $T_{B}^{1} = (1/2, 1/2)$
and with second level $T_{A}^{2} = (1/2, 1/2)$. and $T_{B}^2 = (3/4, 0)$. 
Consider policy $P$ with ($P^1_{A} = (1, 0)$, $P^1_{B} = (1, 1)$)  and ($P^2_{A} = (1,1)$, $P^2_{B} = (1,0)$).
This policy has recall $3/4$ for each group and therefore satisfies Equal Opportunity.
Consider policy $Q$ with parameters 
($Q^1_{A} = (1, 0)$, $Q^1_{B} = (1, 1/2)$) and
($Q^2_{A} = (1,1)$, $Q^2_{B} = (1,1)$).
This policy also has the recall of $3/4$ for each group and therefore also satisfies Equal Opportunity. 
However, the average of these two policies denoted as $\pi$ is ($\pi^1_{A} = (1,0)$,  $\pi^1_{B} = (1,3/4)$),
while ($\pi^2_{A} = (1,1)$ , $\pi^2_{B} = (1,1/2)$).
The recall for group $A$ is still $\frac{3}{4}$, while the recall for group $B$ is $(\frac{1}{2} + \frac{1}{2} \cdot \frac{3}{4})(\frac{3}{4} + \frac{1}{4} \cdot \frac{1}{2}) = \frac{49}{64} \neq \frac{3}{4}$. 

Thus this convex combination of policies does not satisfy Equal Opportunity and therefore the set of Equal Opportunity promotion policies is not convex.

\paragraph{Requiring Equalized Odds at each level can significantly harm performance:} The above example also shows that requiring Equalized Odds at each level can significantly harm performance.  Notice that policy $P$ above satisfies Equalized Odds overall and has perfect precision and fairly high recall.  However, the only way to satisfy Equalized Odds at each level is to completely bypass both tests, which would be much worse for precision.

Interestingly, as we show below, requiring {\em Equal Opportunity} at each level does {\em not} harm precision relative to requiring it for the pipeline as a whole (though it can hurt recall).

\section{Maximizing Precision Subject to Equal Opportunity \label{maxprecsection}}
% \section{Interview Efficiency: Maximizing Precision and Satisfying Equal Opportunity
In this section, we exhibit a policy $\pi$ that maximizes precision at the end of the screening  process while satisfying Equal Opportunity over the entire process.
To do this, we prove that the optimal method for this objective is given by promoting individuals from
each group according to the {\em Opportunity Ratio} (which we will define shortly).  
%Note, for clarity, we first define and consider these notions for a screening process with a single test before the final interview, and then extend to the general case with $k$ stages.
%\KS{First, consider the notion of True Positive  and False Positive Rates, which we can write out explicitly as a function of the test parameters and our promotion policy.}

\begin{defn}
For a test $\tau$ and associated promotion policy $\{(\pi_{X1}, \pi_{X0})\}_{X\in \mathcal X}$, define $M_{X,\tau, \pi} := (\tau_{X1} \pi_{X1} + (1 - \tau_{X1}) \pi_{X0})$ and $N_{X,\tau, \pi} := (\tau_{X0} \pi_{X1} + (1 - \tau_{X0}) \pi_{X0})$. Note that $M_{X,\tau, \pi}$ and $N_{X,\tau, \pi}$ are the probabilities that a positive and respectively a negative example from group $X$ is promoted to the next level,
%{\em True Positive Rate} and the {\em False Positive Rate} of $\pi$ the probability 
and so will be important quantities for our analysis.

\end{defn}

%\KS{should we say something about k stages here?}
%Now we consider the interview efficiency for a screening process with one single test.
\begin{obsr}\label{obs:EOpp-cond}
For any single-stage policy $\{(\pi_{X1}, \pi_{X0})\}_{X\in \mathcal X}$ satisfying Equal Opportunity for a test $\{(\tau_{X1}, \tau_{X0})\}_{X\in \mathcal X}$, there exists $M$ such that $M_{X,\tau, \pi} = M$ for every $X\in \mathcal X$.

Furthermore, for a $k$-stage screening process $\{\tau^i\}_{i\in [k]}$, a policy $\{(\pi_{X0}, \pi_{X1})\}_{X\in \mathcal X}$ is Equal Opportunity if there exists $M$ such that $\Pi_{i=1}^k M_{X,\tau^i, \pi^i} = M$ for every group $X \in {\mathcal X}$.
\end{obsr}
%These observations are identical to the definition of Equal Opportunity. 

% insert opportunity ratio and an example

\begin{obsr}\label{obsr:interview-eff-formula}
Recall that $q_X$ denotes the base rate for group $X$, and let $u_X = 1-q_X$. For a single-stage pipeline with test $\tau$ and promotion policy $\pi$,
%$\{(\pi_{X1}, \pi_{X0})\}_{X\in \mathcal X}$ and the pre-interview test $t$ with qualified and unqualified fractions $\{(\q_X, \u_X)\}_{X\in \mathcal X}$, 
the interview efficiency (i.e., precision) is equal to
\begin{align}\label{eq:interview-eff}
    \ie(q,u,\tau,\pi):= \frac{\sum_{X\in \mathcal X}\q_X M_{X,\tau, \pi}}{\sum_{X\in \mathcal X}\q_X M_{X,\tau, \pi} + \u_X N_{X,\tau, \pi}}.
\end{align}
Similarly, when we consider the extension to a $k$-stage pipeline, the interview efficiency is equal to
\begin{align}\label{eq:interview-eff-k}
    \ie(q,u,\tau,\pi):= \frac{\sum_{X\in \mathcal X}\q_X \prod_{i=1}^{k} M_{X,\tau^{i}, \pi^{i}}}{\sum_{X\in \mathcal X}\q_X \prod_{i=1}^{k} M_{X,\tau^{i}, \pi^{i}} + \u_X \prod_{i=1}^{k} N_{X,\tau^{i}, \pi^{i}}}.
\end{align}
\end{obsr}

Now, we formally define the policy given by the opportunity ratio as follows.
\begin{defn}[\textbf{Opportunity Ratio Policy}] Consider a screening process with $k$ stages. 

For each $X\in \mathcal{X}$, let $\rho_X := \Pi_{j\in [k]}(\tau^j_{X^*1}/\tau^j_{X1})$, where $X^* = \argmin_{X\in \mathcal X} \Pi_{j\in [k]}\tau^j_{X1}$.
The {\em Opportunity Ratio} policy, at the first stage for each $X\in \mathcal{X}$, promotes $\rho_X$ fraction of those who pass the test and none of those who fail the test. For the remaining stages $(i=2,3,...,k)$, the Opportunity Ratio policy fully trusts the result of the tests; a candidate is promoted to the next stage iff they pass the test at the current stage. Formally, for every $X\in \mathcal{X}, \pi^1_{X1} = \rho_X, \pi^1_{X0}=0$ and $\pi^i_{X1}=1, \pi^i_{X0} =0, \forall i\ge 2$. 
\label{def:OR}
\end{defn}

In the rest of this section, 
%we show that promoting according to this policy is the unique
we study the task of maximizing interview efficiency under different settings and fairness requirements.
\subsection{Maximizing Interview Efficiency subject to Equal Opportunity at the Final Stage} 
As a warm-up, we start with the simplest setting where the screening process has only one test before the interview stage.
%(final stage).
\begin{thm}[{\bf Opportunity Ratio Policy Maximizes Precision for Single-Stage %Screening
Process}]\label{thm:single-pipeline}
Let $t=((\tau_{A1},\tau_{A0}), (\tau_{B1},\tau_{B0}))$ be a test satisfying the Minimal Effectiveness Property.  
The maximum precision policy satisfying Equal Opportunity is the opportunity ratio policy.
Moreover, for any group $X\in \mathcal X$, it is always sub-optimal to promote any candidates who failed the test (i.e., in any optimal policy, $\pi_{X0} = 0, \forall X\in \mathcal{X}$).
\end{thm}
\begin{proof}
First, for any policy $\pi$, we upper-bound the interview efficiency (i.e.,~precision) for a screening process with parameters $\q, \u, \tau$. To bound the interview efficiency, for each $X\in \mathcal X$, we lower-bound the False Positive Rate $N_{X,\tau,\pi}$ in terms of the True Positive Rate $M_{X,\tau, \pi}$. 
\begin{align}
    N_{X,\tau, \pi} 
    &= \tau_{X0} \pi_{X1} + (1-\tau_{X0}) \pi_{X0}\nonumber \\ 
    &= \tau_{X0}(\pi_{X1}-\pi_{X0}) + \pi_{X0} \nonumber\\
    % &\ge \tau_{X0} (\pi_{X1}-\pi_{X0}) + \frac{\tau_{X0}}{\tau_{X1}} \pi_{X0} &&\rhd\text{by Eq.~\eqref{eq:minimally-effective}, $\forall X\in \mathcal X$, $\tau_{X1} > \tau_{X0} \ge 0$} \nonumber\\
    &\ge \frac{\tau_{X0}}{\tau_{X1}} \big(\tau_{X1}(\pi_{X1}-\pi_{X0}) + \pi_{X0}\big) &&\rhd\text{by Eq.~\eqref{eq:minimally-effective}} \nonumber\\
    &= \frac{\tau_{X0}}{\tau_{X1}}\cdot M_{X,\tau, \pi} \label{eq:X-bound}
\end{align}
By Equal Opportunity of $\pi$ and employing Eq.~\eqref{eq:X-bound} in the formula for the interview efficiency, Eq.~\eqref{eq:interview-eff},
\begin{align}
    \ie(\q, \u, \tau, \pi) 
    &= \frac{\sum_{X\in \mathcal X}\q_X M_{X,\tau, \pi}}{\sum_{X\in \mathcal X}\q_X M_{X,\tau, \pi} + \u_X N_{X,\tau, \pi}} \nonumber \\
    &\le \frac{\sum_{X\in \mathcal X}\q_X M_{X,\tau, \pi}}{\sum_{X\in \mathcal X}(\q_X + \u_X\cdot \frac{\tau_{X0}}{\tau_{X1}}) M_{X,\tau, \pi}} \quad\rhd\text{by Eq.~\eqref{eq:X-bound}} \nonumber\\
    &=\frac{\sum_{X\in \mathcal X}\q_X}{\sum_{X\in \mathcal X}\q_X + \u_X\cdot \frac{\tau_{X0}}{\tau_{X1}}} \; \rhd\forall X\in \mathcal{X}, M_{X,\tau, \pi} = M \label{eq:upper-bound-interview-eff}
\end{align}
Note that the inequalities are tight when $\pi_{X0} = 0$ for all $X\in \mathcal X$. 

Next, we show that the {\em opportunity ratio} policy satisfies Equal Opportunity and achieves the bound in Eq.~\eqref{eq:upper-bound-interview-eff}. 
In the opportunity ratio policy $\pi^*$, only a $(\frac{\tau_{X^*1}}{\tau_{X1}})$-fraction of candidates in group $X$ who pass the test $t$ (picked uniformly at random) are promoted to the next stage. In other words, for any group $X\in \mathcal X$, we set $\pi^*_{X1} = \frac{\tau_{X^*1}}{\tau_{X1}}, \pi^*_{X0}=0$. Then,
\begin{align*}
    \ie(\q, \u, \tau, \pi^*) 
    &= \frac{\sum_{X\in \mathcal X}\q_X M_{X,\tau, \pi^*}}{\sum_{X\in \mathcal X}\q_X M_{X,\tau, \pi^*} + \u_X N_{X,\tau, \pi^*}} \nonumber \\
    &= \frac{\sum_{X\in \mathcal X}\q_X \tau_{X1} (\frac{\tau_{X^* 1}}{\tau_{X 1}})}{\sum_{X\in \mathcal X} \q_X \tau_{X1} (\frac{\tau_{X^* 1}}{\tau_{X 1}}) + \u_X \tau_{X0}(\frac{\tau_{X^* 1}}{\tau_{X 1}})} \\
    &= \frac{\sum_{X\in \mathcal X}\q_X}{\sum_{X\in \mathcal X}(\q_X + \u_X\cdot \frac{\tau_{X0}}{\tau_{X1}})} %&&\rhd \text{by~\eqref{eq:minimally-effective}, $\tau_{X^*1}>0$}
\end{align*}
Hence, $\pi^*$ is an equal opportunity policy with the maximum interview efficiency for any screening process with parameters $\q, \u, \tau, \pi$. 
%\AV{Note that our analysis does not use the fact that $q_A + u_A + q_B + u_B =1$. We may need to exploit this property further in the proof for multi-stage pipelines.}
\end{proof}
\begin{rem}
Note that any policy $\pi$ where for each $X\in {\mathcal X}$, $\pi_{X1} = \eta \cdot \pi^*_{X1}, \pi_{X0} = 0$ for a constant $\eta<1$ also satisfies the Equal Opportunity and maximizes the interview efficiency objective (i.e., precision). However, $\pi^*$ has a strictly higher {\em recall}. 
\end{rem}
%\subsection{Maximizing Interview Efficiency in Multi-Stage Processes}
Next, we state our result for the general setting in which there are multiple stages and multiple groups in the screening process. The proof of the theorem is similar to the single test version and is deferred to Appendix~\ref{sec:precision-max-proof}. %of the Appendix. 
\begin{thm}[{\bf Multi-Stage Screening Process}]\label{thm:multi-eq-opp-prec}
Consider a $k$-stage screening process whose all tests are minimally effective. The maximum interview efficiency policy satisfying Equal Opportunity is the Opportunity Ratio policy and has interview efficiency equal to $\frac{\|q\|_1}{\|q\|_1 + \sum_{X\in \mathcal X}\u_X \Pi_{i=1}^k(\tau^i_{X0}/\tau^i_{X1})}$.
\end{thm}

\subsection{Maximizing Interview Efficiency Subject to Equal Opportunity at the End of Each Stage} 
 Here, we consider the setting in which the goal is find a policy that maximizes interview efficiency and satisfy Equal Opportunity at the end of each stage---{\em not only at the interview stage}. 
Following Theorem~\ref{thm:multi-eq-opp-prec}, the maximum interview efficiency in this setting is at most $\|q\|_1/(\|q\|_1 + \sum_{X\in \mathcal X}\u_X \Pi_{i=1}^k\frac{\tau^i_{X1}}{\tau^i_{X0}})$. Next, we show that the following slightly modified opportunity ratio policy $\pi$ that satisfies Equal Opportunity at the end of each stage maximizes the interview efficiency. The policy $\pi$ applies the opportunity ratio at each stage of the pipeline.
\begin{align*}
    \pi^i_{X0} = 0, \pi^i_{X1} = \frac{\tau^i_{X^*_i 1}}{\tau^i_{X 1}} &&\forall i\in [k], X\in {\mathcal X}, \text{ where $X^*_i := \argmin_{X\in \mathcal X} \tau^i_{X1}$}
\end{align*}
Again, it is straightforward to verify that $\pi$ satisfies the Equality of Opportunity. Moreover, %the interview efficiency of $\pi$ is
\begin{align*}
    \ie(q, u, \tau, \pi)
    &= \frac{\sum_{X\in \mathcal X}\q_X M_{X,\tau, \pi}}{\sum_{X\in \mathcal X}\q_X M_{X,\tau, \pi} + \u_X N_{X,\tau, \pi}} \\
    &= \frac{\sum_{X\in \mathcal{X}}q_X \Pi_{i\in [k]} \tau^i_{X^*_i1}}{\sum_{X\in \mathcal{X}}q_X \Pi_{i\in [k]} \tau^i_{X^*_i1} + \sum_{X\in \mathcal X}\u_X \frac{\tau^i_{X^*_i 1} \tau^i_{X 0}}{\tau^i_{X 1}}} \\
    &= \frac{\|q\|_1}{\|q\|_1 + \sum_{X\in \mathcal X}\u_X \Pi_{i=1}^k\frac{\tau^i_{X0}}{\tau^i_{X1}}}
\end{align*}
The only difference compared to the policy of Theorem~\ref{thm:multi-eq-opp-prec} is that in the former policy the recall can be higher.
\begin{rem}
Adding the condition to satisfy the Equality of Opportunity at the end of each stage does not harm interview efficiency. However, this condition may decrease the recall of the optimal policy.  
\end{rem}

\section{Pipeline Efficiency: Maximizing Linear Combinations of Precision and Recall}\label{sec:linear-objective}
%Preliminaries for Linear Combination of Recall/Precision} 
Now we shift our focus to exhibiting a promotion policy that satisfies Equal Opportunity and
maximizes a linear combination of precision and recall given by the positive weight $\alpha \in \mathbb{R}_{\ge 0}$; $f_{\alpha}(\pi):=(1-\alpha)\cdot \mathrm{recall}(\pi) + \alpha\cdot\mathrm{precision}(\pi)$.
As in Definition \ref{def-pipeline-eff}, higher precision corresponds to higher interview efficiency, and higher recall corresponds to higher throughput efficiency. 

We start with a simple $2$-approximation algorithm for maximizing any given linear  of precision and recall.
\begin{thm}[\bf Approximation Algorithm for Linear Combination of Precision and Recall]
There exists a  polynomial time 2-approximation algorithm for maximizing any linear combination of precision and recall. 
\end{thm}
\begin{proof}
Note that the policy that bypasses all tests is an Equal Opportunity policy and maximizes recall---it achieves recall equal to one.
Moreover, by Theorem~\ref{thm:multi-eq-opp-prec}, the Opportunity Ratio is an Equal Opportunity policy maximizing precision. Hence, the better of the ``bypassing all tests'' policy and the Opportunity Ratio policy is a $2$-approximation of any given linear combination of precision and recall. 
\end{proof}

In order to obtain better performance for maximizing linear combinations of precision and recall, we develop structural properties of optimal solutions, and then use them to get an exact algorithm with running time that is exponential only in $k$ and the number of groups.
Additionally, by a dynamic programming approach we exhibit a {\em fully polynomial time approximation scheme (FPTAS)}.
%as well as an FPTAS.

One challenge is that as shown in Section~\ref{sec:problem-statement-and-example}, the space of Equal Opportunity solutions is non-convex. Another is that as shown in Section \ref{sec:optnotopt} below, Opportunity Ratio is no longer optimal, and as shown in Section \ref{sec-test-nonmono} below, there exists no function ranking the efficacy of tests solely based on their statistics.  
%explain the challenges non-convexity/example 5

We begin by presenting the examples mentioned above, and then developing the structural properties we will use.

\subsection{Illustrative Examples}
\subsubsection{Opportunity Ratio not Optimal for Linear Combination of Precision and Recall}\label{sec:optnotopt}
In the previous sections, our 
key algorithmic strategy is to use the Opportunity Ratio to re-weight the promotion  policy. 
Since this policy satisfied Equal Opportunity and maximized precision (among Equal Opportunity policies), if our objective is to only maximize precision, then the Opportunity Ratio is sufficient.
Now we exhibit an example where the Opportunity Ratio solution is not optimal when maximizing any linear combination of precision and recall when there is any nonzero weight on recall.  Specifically, in this example there is an alternative policy with the same precision as the Opportunity Ratio solution but strictly higher recall.
%In a sense, this is obvious, since the opportunity ratio is focused on maximizing precision (recall for the opportunity ratio policy we do not promote any individuals who fail a test).

 Consider a pipeline with $T_{A}^{1} = (3/4, 0)$ and $T_{B}^{1} = (1/2, 1/4)$.
In the second stage, $T_{A}^{2} = (1/2, 1/4)$  and $T_{B}^{2} = (3/4, 0)$.
Consider policy $P$:  ($P_{A}^{1} = (1, 0)$ and $P_{B}^{1} = (1, 1)$, while $P_{A}^{2} = (1,1)$ and  $P_{B}^{2} = (1,0)$.

This policy has recall $3/4$ and precision $1$ for each group and therefore satisfies Equal Opportunity. 
Thus if our objective here is maximize the average of precision and recall, this policy has objective function value $7/8$.
In  contrast, the Opportunity Ratio policy as given in Definition \ref{def:OR} is $P_{A}^{1} = (1, 0)$,$P_{B}^{1} = (1, 0)$ and $P_{A}^{2} = (1,0)$, $P_{B}^{2} = (1,0)$ which reduces our recall to $\frac{3}{4} \cdot \frac{1}{2} = \frac{3}{8}$ while to precision is still $1$, for score of $\frac{11}{16}$.
Clearly this is a lower objective function score than the first policy.
%[ 3/4 0], [1/2, 1/4]

\subsubsection{Optimal Policy Non-Locality for Linear Combination of Precision and Recall \label{sec-test-nonmono}}
% Suppose we have one group in the population and want to optimize a linear combination of recall and precision---this is indeed a much simpler setting compared to the multi-group inputs with Equal Opportunity requirement that we eventually aim to solve). 
% A natural question is whether we can solve this problem with a simple greedy algorithm that makes local decisions in a single pass of the test statistics.
% We answer this question in the negative by exhibiting an example pipeline with test statistics such that when two of three tests are available, using only the first test is strictly optimal, while when considering all three tests the strict optimum is the opposite of the previous case.
% This shows that to find an optimal policy that maximizes a given linear combination of precision and recall, it is not sufficient to follow a ranking function $r : \mathcal{T} \rightarrow \mathbb{R}$ of the tests in the pipeline that determine the ``efficacy'' of each test solely based on its statistics (i.e., for each level $i$, $r$ is a function of $\{(\tau_{X1}, \tau_{X0})\}_{X\in \mathcal{X}}$, where $\mathcal{T}$ is the collection of tests used in various stages of the pipeline. 

%\avnote{Proofread this section.}
Suppose we have one group in the population and want to optimize a linear combination of recall and precision. 
A baseline idea is whether we can solve this problem with a 
%simple
natural
greedy algorithm that makes local decisions in a single pass of the test statistics \footnote{In the related work by \cite{mansourscreen} the answer is in the affirmative, but their model is different and has uniform noise across true positives and true negatives.}.

We answer this question in the negative in
by exhibiting an example pipeline with test statistics such that when two of three tests are available, 
using only the first test is strictly optimal, while when all three tests are available, the optimum is instead to use the other two tests and not the first test.
This shows that an algorithm that maximizes a linear combination of precision and recall cannot simply assign separate scores to each test and then use only the highest-scoring tests.  
Our example is only for one group. 

The counterexample is as follows. 
The base-rate in the population is  $P(y=1) = 1/2$.
Consider test $t_1 = (1/2,0)$ and tests $t_2 =t_3= (1-\delta, 1/2)$ where $\delta=\frac{1}{100}$. The objective function is $f(\pi) = \frac{1}{3} \cdot  \mathrm{recall}(\pi) + \frac{2}{3} \cdot \mathrm{precision}(\pi)$. 
%Further, $\pi^{i}=[ \pi^i_{1}, \pi^i_{0}]$ is the promotion probabilities for those who pass/fail test $t_i$ (there is only one group in this example). 
In the following, 
%as we work with policies that either fully use each test (i.e., $\pi_1 =1, \pi_0 = 0$) or bypass it (i.e., $\pi_1 = \pi_0 = 1$), we 
let $f(t_{1})$ to denote the score of the policy that only promotes those who pass $t_1$ and bypasses all other tests
while $f(t_{2} t_{3})$ denotes bypassing $t_1$ and promoting individuals if and only if they pass tests $t_2$ and $t_3$.
In the Appendix~\ref{appendix-linear-combination-counter} we show while $f(t_1)$ is larger than any policy using $t_1$ and $t_2$ (possibly in fractions), $f(t_2 t_3)$ is strictly larger than any policy using $t_1, t_2$ and $t_3$ (again, possibly in fractions).

% Using test $t_1$ ($\pi^{1}=[1,0]$) and bypassing test $t_2$ has recall $=1/2$ and precision $=1$, so $f(t_1)=5/2$.
% If test $t_2$ is available besides test $t_1$ then using test $t_2$ while keeping $\pi^{1}= [1,0]$ can only harm the recall, so once we are committed to setting $\pi^{1} = [1,0]$ the optimal choice is to bypass $t_2$; set $\pi^{2}=[1,1]$.

% Now, consider if a third test $t_3=t_2$ (e.g. an identical copy of $t_2$) is also available. 
% Then bypassing test $t_1$ and using test $t_2$ followed by test $t_3$ ($\pi^{2}=\pi^{3}=[1,0])$ has recall $(1-\delta)^2$.
% Similarly, this has precision $\frac{(1-\delta)^2}{(1-\delta)^2 + 1/4}$.
% So if $\delta = 1/100$, then 
% \[f(t_2 t_3)= (\frac{99}{100})^2 + 2\frac{(99/100)^2}{(99/100)^2 + 1/4} > 2.57 > \frac{5}{2} = f(t_1) \]
% Observe that $f(t_1 t_2 t_3) < f(t_1)$.
% In the Appendix~\ref{appendix-linear-combination-counter} we will further elaborate why $f(t_2 t_3)$ is the optimal policy for these problem parameters.

\subsection{An Exact Algorithm}\label{sec:exact-alg}
%Maximizing Linear Combination of Precision and Recall}
%\subsubsection{Properties of Pareto Optimal Policies}
%\avnote{starting this section, we switch to the case in which the equal opportunity is required only at the interview stage.}
In this section, we give an exact algorithm for maximizing any given linear combination of precision and recall subject to satisfying Equal Opportunity by the end of the screening process. 
%To recall, Equality of Opportunity in a $k$-stage screening process requires that for every pair of groups $X, Y \in \mathcal{X}$, $\Pi_{i\in [k]} M_{X,\tau^i, \pi^i} = \Pi_{i\in [k]} M_{Y,\tau^i, \pi^i}$.

First we show that for any $k$-stage screening process over a population specified by a collection of groups $\mathcal{X}$, there exists a set of Equal Opportunity policies $\mathcal{P}_{k, \mathcal{X}}$ that {\em weakly Pareto dominate} (w.r.t.~precision and recall) any policy satisfying Equal Opportunity. In particular, we show that each policy $\pi :=(\pi^1, \cdots, \pi^k)\in {\mathcal P}_{k, {\mathcal X}}$ has the following structure, $(1-\pi^i_{X1})\pi^i_{X0} = 0, \forall i\in [k], X\in \mathcal X$.
% \begin{align*}
%     (1-\pi^i_{X1})\pi^i_{X0} = 0, &&\forall i\in [k], X\in \mathcal X
% \end{align*}

\begin{defn}[{\bf Pareto Dominant Policy}]\label{def:pareto-dominant}
For a given screening process, a policy $\pi$ {\em weakly} Pareto dominates a policy $\tilde\pi$ w.r.t.~precision and recall iff, $\mathrm{recall}(\pi) \ge \mathrm{recall}(\tilde{\pi})$ and $\mathrm{precision}(\pi) \ge \mathrm{precision}(\tilde{\pi})$.
% \begin{align*}
%     &\mathrm{recall}(\pi) \ge \mathrm{recall}(\bar{\pi}),  &\mathrm{precision}(\pi) \ge \mathrm{precision}(\bar{\pi})
% \end{align*}
Moreover, $\pi$ {\em strictly} Pareto dominates $\tilde \pi$ if at least one of the above inequalities holds strictly. 

Furthermore, a set of policies $\mathcal P$ weakly Pareto dominates a policy $\tilde\pi$ w.r.t. precision and recall iff there exists a policy $\pi\in \mathcal P$ that $\pi$ weakly Pareto dominates $\tilde \pi$.
\end{defn}
% First, we set up some notations. For any group $X\in \{A, B\}$,
% \begin{align*}
%     &M^X_{\tau, \pi} := \tau_{X1}\pi_{X1} + (1-\tau_{X1})\pi_{X0}
%     &N^X_{\tau, \pi} := \tau_{X0}\pi_{X1} + (1-\tau_{X0})\pi_{X0} 
% \end{align*}
% Then, the Equality of Opportunity requires that 
% \begin{align}\label{eq:eqopp}
% \Pi_{i\in [k]} M^A_{\tau^i, \pi^i} = \Pi_{i\in [k]} M^B_{\tau^i, \pi^i}.
% \end{align}

% Recall that the precision of any policy $\pi$ is defined as follows. 
% \begin{align}\label{eq:precision}
%     \frac{\sum_{X\in \mathcal X} q_X \Pi_{i\in [k]} M^X_{\tau^i, \pi^i}}{\sum_{X\in \mathcal X} q_X \Pi_{i\in [k]} M^X_{\tau^i, \pi^i} + u_X \Pi_{i\in [k]} N^X_{\tau^i, \pi^i}}
% \end{align}
\begin{lem}\label{lem:pi-cond}
For any $k$-stage screening policy that satisfies the Minimal Effectiveness Property \ref{eq:minimally-effective}, the set of Equal Opportunity policies in $\mathcal{P}:=\{\pi \in [0,1]^{2|\mathcal{X}|k} : (1 - \pi^i_{X1}) \pi^i_{X0} = 0, \forall X\in \mathcal{X}, i\in [k]\}$ weakly Pareto dominates all equal opportunity policies w.r.t.~precision and recall.
% \begin{align}\label{eq:pareto-front}
%     (1 - \pi^i_{X1}) \pi^i_{X0} = 0, \quad \forall i\in [k].
% \end{align}

In other words, any equal opportunity policy violating $(1 - \pi^i_{X1}) \pi^i_{X0} = 0$ for a group $X\in \mathcal X$ and a stage $i\in [k]$ is weakly Pareto dominated by ${\mathcal P}$. %(i.e., a policy that satisfies Eq.~\eqref{eq:pareto-front}). 
\end{lem}
\begin{proof}
First, we show that in any policy $\pi$ which is not strictly Pareto dominated (w.r.t.~precision and recall), $\pi^i_{X1}>0$ for every $X\in \mathcal X, i\in [k]$. Hence, we can only consider policies $\pi$ where $\pi_{X1}>0$ for all $X\in \mathcal{X}$. The proof of the following claim is deferred to Appendix~\ref{sec:linear-objective-proofs}. 
\begin{clm}\label{clm:non-zero-pi-one}
Consider a $k$-stage screening process whose tests satisfy the Minimal Effectiveness Property \ref{eq:minimally-effective}.
In any optimal policy of this screening process that satisfies Equal Opportunity, for all $X\in \mathcal X$ and $i\in [k]$, $\pi^i_{X1} >0$. 
% In other words, any Equal Opportunity policy $\pi$ with $\pi^i_{X1} =0$ for a group $X\in \mathcal X$ and a stage $i\in [k]$ is strictly Pareto dominated by an Equal Opportunity policy $\tilde{\pi}$ such that $\tilde{\pi}^i_{X1}>0$ for all $X\in \mathcal X$ and $i\in [k]$. 
\end{clm}
Now, for the sake of contradiction, suppose that there exist a level $i\in [k]$ and a group $X\in \mathcal X$ such that $\pi^i_{X0}> 0$ and $\pi^i_{X1} < 1$. 
Note that w.l.o.g., we can assume that $\tau^i_{X1} <1$; otherwise, by setting $\pi_{X0} =0$, the recall of the policy does not decrease and the precision strictly increases. Hence, there exist $\epsilon_1, \epsilon_0 >0$ such that $\tau^i_{X1} \epsilon_1 - (1-\tau^i_{X1}) \epsilon_0 = 0$ where either $(\epsilon_1 = 1-\pi_{X1}, \epsilon_0 \le \pi_{X0})$ or $(\epsilon_1 \le 1- \pi_{X1}, \epsilon_0 = \pi_{X0})$. 

We define a new policy $\tilde{\pi}$, which differs from $\pi$ only in level $i$ of group $X$, as follows: $\tilde{\pi}^i_{X1} = \pi^i_{X1} +\epsilon_1$ and $\tilde{\pi}^i_{X0} = \pi^i_{X0} - \epsilon_0$.
% \begin{align*}
%     \tilde{\pi}^i_{X1} &= \pi^i_{X1} +\epsilon_1, &&\tilde{\pi}^i_{X0} = \pi^i_{X0} - \epsilon_0 \\
%     \tilde{\pi}^j_{Y1} &= \pi^j_{Y1}, 
%     &&\tilde{\pi}^j_{Y0} = \pi^j_{Y0} \quad \text{if $(Y\neq X) \vee (j\neq i)$}
% \end{align*}
%since for every $i\in [k]$, $N^X_{\tau^i, \pi^i} >0$, the new policy which replaces $\pi^i_{X1}$ with $\pi^i_{X1} +\epsilon_1$ and replaces $\pi^i_{X0}$ with $\pi^i_{X0} - \epsilon_0$, has higher precision which contradicts the optimally of policy $\pi$.
Next, we show that $N_{X,\tau^i, \tilde{\pi}^i} < N_{X,\tau^i, \pi^i}$.
\begin{align*}
    N_{X,\tau^i, \tilde{\pi}^i}
    &= \tau^i_{X0} \tilde{\pi}^i_{X1} +(1-\tau^i_{X0}) \tilde{\pi}^i_{X0} \\
    &= \tau^i_{X0} (\pi^i_{X1} + \epsilon_1) +(1-\tau^i_{X0}) (\pi^i_{X0} - \epsilon_0) \\
    &= \tau^i_{X0}\pi^i_{X1} +(1-\tau^i_{X0})\pi^i_{X0} + (\tau^i_{X0}\epsilon_1 + \tau^i_{X0}\epsilon_0 - \epsilon_0) \\
    &= \tau^i_{X0}\pi^i_{X1} +(1-\tau^i_{X0})\pi^i_{X0} \\
    &\quad + (\tau^i_{X0}\epsilon_1 + \tau^i_{X0}\epsilon_0 - \tau^i_{X1}\epsilon_1 - \tau^i_{X1}\epsilon_0) \;\rhd \epsilon_0 = \tau^i_{X1} (\epsilon_0 + \epsilon_1) \\
    &< \tau^i_{X0}\pi^i_{X1} +(1-\tau^i_{X0})\pi^i_{X0} \;\rhd\tau^i_{X0} < \tau^i_{X1} \\
    &= N_{X,\tau^i, \pi^i}
\end{align*}
%This implies that $\Pi_{j=1}^k N^X_{\tau^j, \pi^j} < \Pi_{j=1}^k N^X_{\tau^j, \tilde{\pi}^j}$. 
Further, since $\tau^i_{X1} \epsilon_1 - (1-\tau^i_{X1}) \epsilon_0 = 0$, $\tilde{\pi}$ satisfies Equal Opportunity and has the same recall as $\pi$. Moreover, since $N_{X, \tau^i, \tilde{\pi}^i} < N_{X, \tau^i, \pi^i}$ and for all $j\in [k]\setminus \{i\}$, $N_{X, \tau^j, \pi^j} \ge0$, $\Pi_{j=1}^k N_{X, \tau^j, \pi^j} \le \Pi_{j=1}^k N_{X, \tau^j, \tilde{\pi}^j}$. Hence the precision of $\tilde{\pi}$ is not less than the one of $\pi$. This contradicts the strict Pareto optimally of policy $\pi$. Thus the statement holds and for any level $i\in [k]$ and any group $X\in \mathcal X$, $(1-\pi^i_{X1})\pi^i_{X0} = 0$.
\end{proof}
% \avnote{need to check if want to keep the following corollary.}
% \begin{cor}\label{cor:opt-single}
% In a $k$-stage screening process with a single group $A$ whose tests satisfy the ``minimally effectiveness'' property and for all $i\in [k], \tau^i_{A1} <1$, the set of policies satisfying the following property strictly Pareto dominates all policies w.r.t.~precision and recall, $\forall i\in [k],\quad (1 - \pi^i_{A1}) \pi^i_{A0} = 0$.
% % \begin{align}%\label{eq:pareto-front-single}
% %     \forall i\in [k],\quad (1 - \pi^i_{A1}) \pi^i_{A0} = 0.
% % \end{align}
% \end{cor}

Next, we show additional structures of the set of Equal Opportunity policies ${\mathcal P}_{k, {\mathcal X}}$ that weakly Pareto dominates {\em all} Equal Opportunity policies.

\begin{lem}\label{lem:pi-zero}
Consider a $k$-stage screening process whose tests satisfy the Minimal Effectiveness Property \ref{eq:minimally-effective}.
The set of Equal Opportunity policies $\mathcal{S} \subseteq \mathcal{P} = \{\pi \in [0,1]^{2|\mathcal{X}|k} : (1 - \pi^i_{X1}) \pi^i_{X0} = 0, \forall X\in \mathcal{X}, i\in [k]\}$ where for each group $X\in \mathcal X$, there exists at most one level $i \in [k]$ such that $0 < \pi^i_{X0} < 1$, weakly Pareto dominates all Equal Opportunity policies.

In other words, any Equal Opportunity policy $\pi$ of the screening process is weakly Pareto dominated by $\tilde{\pi} \in \mathcal{S}$ (in every policy $\tilde{\pi} \in \mathcal{S}$, for each group $X\in \mathcal X$, there exists at most one level $i$ such that $0 < \tilde{\pi}^i_{X0} < 1$). 
\end{lem}
\begin{proof}
Suppose for contradiction that there exist a group $X\in \mathcal X$ and levels $i,j$ such that $0 < \pi^i_{X0}, \pi^j_{X0} < 1$. Next, we show that we can modify $\pi$ in levels $i$ and $j$ and replace $\pi^i_{X0}, \pi^j_{X0}$ with $\tilde{\pi}^i_{X0}, \tilde{\pi}^j_{X0}$ such that 
\begin{align}
    M_{X,\tau^i, \pi^i} M_{X,\tau^j, \pi^j} 
    &= (\tau^i_{X1} + \pi^i_{X0} (1- \tau^i_{X1})) (\tau^j_{X1} + \pi^j_{X0} (1- \tau^j_{X1})) \nonumber\\
    &= (\tau^i_{X1} + \tilde{\pi}^i_{X0} (1- \tau^i_{X1})) (\tau^j_{X1} + \tilde{\pi}^j_{X0} (1- \tau^j_{X1})) \nonumber \\
    &= M_{X,\tau^i, \tilde{\pi}^i} M_{X,\tau^j, \tilde{\pi}^j}, \label{eq:equal-opp-cond} \\
    N_{X,\tau^i, \pi^i} N_{X,\tau^j, \pi^j}
    &= (\tau^i_{X0} + \pi^i_{X0} (1- \tau^i_{X0})) (\tau^j_{X0} + \pi^j_{X0} (1- \tau^j_{X0})) \nonumber \\
    &> (\tau^i_{X0} + \tilde{\pi}^i_{X0} (1- \tau^i_{X0})) (\tau^j_{X0} + \tilde{\pi}^j_{X0} (1- \tau^j_{X0})) \nonumber\\
    &= N_{X,\tau^i, \tilde{\pi}^i} N_{X,\tau^j, \tilde{\pi}^j} \label{eq:prec-opt}
\end{align}
Note that Eq.~\eqref{eq:equal-opp-cond} guarantees that the new policy $\tilde{\pi}$ satisfies Equal Opportunity and has the same recall as the policy $\pi$. Moreover, Eq.~\eqref{eq:prec-opt} shows that precision of the new policy is not less than than the precision of $\pi$. 
Next, we show that in the new policy, either  $\tilde{\pi}^i_{X0} \in \{0,1\}$ or $\tilde{\pi}^j_{X0} \in \{0,1\}$. 

Without loss of generality, we can assume that the feasible range of values for $\tilde{\pi}^i_{X0}$ to satisfy Equal Opportunity is $[\pi^i_{X0} - \epsilon^i, \pi^i_{X0} + \delta^i]$ which corresponds to $[\pi^j_{X0} - \delta^j, \pi^j_{X0} + \epsilon^j]$. Both intervals are sub-intervals of $[0,1]$ and since both $\tilde{\pi}^j_{X0}, \tilde{\pi}^i_{X0}$ belong to $[0,1]$, it is straightforward to verify that 
\[ (\pi^i_{X0} - \epsilon^i) (1 - (\pi^j_{X0} + \epsilon^j)) = (1-(\pi^i_{X0} + \delta^i)) (\pi^j_{X0} - \delta^j) =0 \]

Let $L = \frac{M_X}{\tau^i_{X1}\tau^j_{X1}}$ where $M_X = M_{X, \tau^i, \pi^i} M_{X, \tau^j, \pi^j} = M_{X, \tau^i, \tilde{\pi}^i} M_{X, \tau^j, \tilde{\pi}^j}$.
By the Minimal Effectiveness Property, $1< L < \frac{1}{\tau^i_{X1} \tau^j_{X1}}$. Then, satisfying Equal Opportunity is equivalent to satisfy the following constraint, $(1 + \tilde{\pi}^i_{X0}(\frac{1-\tau^i_{X1}}{\tau^i_{X1}})) (1 + \tilde{\pi}^j_{X0}(\frac{1-\tau^j_{X1}}{\tau^j_{X1}})) = L$.
% \begin{align*}
%     (1 + \tilde{\pi}^i_{X0}(\frac{1-\tau^i_{X1}}{\tau^i_{X1}})) (1 + \tilde{\pi}^j_{X0}(\frac{1-\tau^j_{X1}}{\tau^j_{X1}})) = L
% \end{align*}
Hence, it implies that %$\tilde{\pi}^j_{X0} 
%    = (\frac{L}{1 + \tilde{\pi}^i_{X0}(\frac{1-\tau^i_{X1}}{\tau^i_{X1}})} - 1)/(\frac{1-\tau^j_{X1}}{\tau^j_{X1}})
%    = (\frac{\tau^j_{X1}}{1-\tau^j_{X1}}) (\frac{L - 1 - \tilde{\pi}^i_{X0}(\frac{1-\tau^i_{X1}}{\tau^i_{X1}})}{1 + \tilde{\pi}^i_{X0}(\frac{1-\tau^i_{X1}}{\tau^i_{X1}})})$.
\begin{align*}%\label{eq:pi-j}
    \tilde{\pi}^j_{X0} 
    = (\frac{L}{1 + \tilde{\pi}^i_{X0}(\frac{1-\tau^i_{X1}}{\tau^i_{X1}})} - 1)/(\frac{1-\tau^j_{X1}}{\tau^j_{X1}}) \\
    = (\frac{\tau^j_{X1}}{1-\tau^j_{X1}}) (\frac{L - 1 - \tilde{\pi}^i_{X0}(\frac{1-\tau^i_{X1}}{\tau^i_{X1}})}{1 + \tilde{\pi}^i_{X0}(\frac{1-\tau^i_{X1}}{\tau^i_{X1}})})
\end{align*}

\paragraph{Case 1: $\max(\tau^i_{X1}, \tau^j_{X1}) = 1$.} Without loss of generality, suppose $\tau^i_{X1} = 1$. Then, we can simply set $\tilde{\pi}^i_{X0} = 0$ and the resulting policy $\tilde{\pi}$ will maintain Equal Opportunity. Moreover, since $1-\tau^i_{X0} > 0$, $N_{X, \tau^i,\tilde{\pi}^i} \leq N_{X, \tau^i,\pi^i}$. In the other case, we can similarly set $\tilde{\pi}^j_{X0}=0$.

\paragraph{Case 2: $\tau^i_{X1}, \tau^j_{X1} < 1$.}
The task of finding $\tilde{\pi}^i_{X0}$ is as follows: $\tilde{\pi}^i_{X0} = \argmin_{y\in [\pi^i_{X0} - \epsilon^i, \pi^i_{X0} + \delta^i]} f(y)$ which is equal to
    \begin{align*}
    %  \tilde{\pi}^i_{X0} 
    %  &=\argmin_{y\in [\pi^i_{X0} - \epsilon^i, \pi^i_{X0} + \delta^i]} f(y) \\
    (\tau^i_{X0} + y (1- \tau^i_{X0})) (\tau^j_{X0} + (\frac{\tau^j_{X1}}{1 - \tau^j_{X1}})(\frac{L - 1 - y (\frac{1-\tau^i_{X1}}{\tau^i_{X1}})}{1 + y (\frac{1-\tau^i_{X1}}{\tau^i_{X1}})})(1-\tau^j_{X0})). 
    \end{align*}
Next, we show that for any 
\[ y \in [0,1], f''(y) =- \frac{2L(\tau^j_{X0} -1)(\frac{\tau^j_{X1}}{1-\tau^j_{X1}})(\frac{1-\tau^i_{X1}}{\tau^i_{X1}}) (\frac{\tau^i_{X0}}{\tau^i_{X1}} - 1)}{(1 + (\frac{1-\tau^i_{X1}}{\tau^i_{X1}}) y)^3} < 0 \]
    % \begin{align*}
    %     f''(y) =- \frac{2L(\tau^j_{X0} -1)(\frac{\tau^j_{X1}}{1-\tau^j_{X1}})(\frac{1-\tau^i_{X1}}{\tau^i_{X1}}) (\frac{\tau^i_{X0}}{\tau^i_{X1}} - 1)}{(1 + (\frac{1-\tau^i_{X1}}{\tau^i_{X1}}) y)^3} < 0.
    % \end{align*}
To prove it note that the minimal ``effectiveness property'' of the tests $\{\tau^i\}_{i\in [k]}$ (i.e., $\tau^i_{X1} > \tau^i_{X0} \ge 0, \forall X\in \mathcal X, i\in [k]$) implies that $\frac{\tau^i_{X0}}{\tau^i_{X1}} -1 < 0$. Moreover since by our assumption $\tau^j_{X1}, \tau^i_{X1} <1$, $f''(y)<0$ for all values of $y \in [0,1]$. Since $f$ is a concave function in $[\pi^i_{X0} - \epsilon^i, \pi^i_{X0} + \delta^i]$, the minimum value of $f$ in this interval obtained in one of its endpoints. In other words, the maximum precision corresponds to the case either $\tilde{\pi}^i_{X0} \in \{0,1\}$ or $\tilde{\pi}^j_{X0} \in \{0,1\}$.
\end{proof}

Finally, we show that each group can only have at most one level that partially uses its corresponding test.
\begin{lem}\label{lem:one-partial}
Consider a $k$-stage screening process whose tests satisfy the ``minimal effectiveness'' property.
The set of Equal Opportunity policies $\mathcal{P}_{k,\mathcal X} \subset \mathcal{P} = \{\pi \in [0,1]^{2|\mathcal{X}|k} : (1 - \pi^i_{X1}) \pi^i_{X0} = 0, \forall X\in \mathcal{X}, i\in [k]\}$ where for each group $X\in \mathcal X$, there exists at most one level $i \in [k]$ such that $\pi^i_{X1} < 1$ or $0<\pi^i_{X0} <1$, weakly Pareto dominates all Equal Opportunity policies.
\end{lem}
The proof is similar to the proof of Lemma~\ref{lem:pi-zero} and we defer it to Appendix~\ref{sec:linear-objective-proofs}.
The above lemma enforces a very restricted structure on the set $\mathcal{P}_{k, \mathcal X}$ of Equal Opportunity policies that {\em weakly Pareto dominate} all Equal Opportunity policies. To summarize, in each policy $\pi \in \mathcal{P}_{k, \mathcal X}$, for each group $X\in {\mathcal X}$, the restriction of $\pi$ on $X$ has the following properties
\begin{enumerate}
    \item There is at most one level $i^*\in [k]$ such that $\pi$ {\em partially uses} the test $\tau^{i^*}$; i.e., {\em either}
    $0< \pi^{i^*}_{X1} < 1$ and $\pi^{i^*}_{X0} = 0$, {\em or} $\pi^{i^*}_{X1} = 1$ and $0< \pi^{i^*}_{X0} < 1$.
    
    \item In any remaining level $i$, $\pi^i$ either {\em bypasses} $\tau^i$ (i.e., $\pi^i_{X1} \pi^i_{X0} =1$), or {\em fully exploits} $\tau^i$ (i.e., $\pi^i_{X1} = 1, \pi^i_{X0} =0$).
\end{enumerate}

\begin{thm}[\bf Exact Algorithms for Linear Combination of Precision and Recall]
Given any linear objective function of form $f_\alpha(\pi) := \alpha \cdot \mathrm{precision}(\pi) + (1-\alpha) \cdot \mathrm{recall}(\pi)$, There exists an exact algorithm that runs in time $O(k^{|{\mathcal X}|} \cdot 2^{k|{\mathcal X}|})$ and finds an Equal Opportunity policy of the screening process with parameters $(q,u, \tau, \mathcal X)$ that maximizes $f_\alpha$.
\end{thm}
\begin{proof}
Using the aforementioned set ${\mathcal P}_{k, {\mathcal X}}$ of weakly Pareto optimal policies (w.r.t. precision and recall) that satisfy the Equality of Opportunity, we enumerate over all policies in ${\mathcal P}_{k, {\mathcal X}}$ as follows.
\begin{itemize}
    \item For each group $X\in \mathcal X$, pick a level $i_X \in [k]$ ({\em i.e., $k^{|{\mathcal X}|}$ possible configurations}).
    \item Fix an ``integral'' policy $\pi$ for the rest of levels in each group $X\in \mathcal X$,
    \begin{itemize}
        \item In each group $X\in \mathcal X$, for each level $i \neq i_X$, we decide whether to fully use the test ($\pi^i_{A1} =1, \pi^i_{A0}=0$) or to bypass the test ($\pi^i_{X1} = \pi^i_{X0} = 1$) ({\em i.e., $2^{(k-1)|{\mathcal X}|}$ possible configurations}). 
    \end{itemize}
    \item For each $X\in {\mathcal X}, i_{X}\in [k]$, we fix the policy $\pi^{i_X}$ partially as follows,
    \begin{itemize}
        \item $(1 - \pi^{i_X}_{X1}) \pi^{i_X}_{X0} = 0, \forall X\in {\mathcal X}$ ({\em i.e., $2^{|{\mathcal X}|}$ possible configurations}). 
    \end{itemize}
\end{itemize}
In each of the policies $\pi$ as constructed above, we set the remaining $\pi$ values (i.e., $\pi^{i_X}$) so that Equality of Opportunity is satisfied and the objective function $f_\alpha$ is maximized.  
Finally, we maintain the configuration $\pi$ that maximizes $f_\alpha$. Note that the whole process takes $O(k^{|{\mathcal X}|} \cdot 2^{k|{\mathcal X}|})$ time.  
\end{proof}

Similarly, we can show the following.
\begin{thm}[\bf Exact Algorithms for Linear Combination of reciprocal of Precision and Recall]
Given any objective function $g_\alpha(\pi) := \alpha/\mathrm{precision}(\pi) + (1-\alpha) / \mathrm{recall}(\pi)$, There exists an exact algorithm that runs in time $O(k^{|{\mathcal X}|} \cdot 2^{k|{\mathcal X}|})$ and finds an Equal Opportunity policy of the screening process with parameters $(q,u, \tau, \mathcal X)$ that minimizes $g_\alpha$.
\end{thm}

\begin{rem}[{\bf General Objective Functions}]\label{rem:exact-alg}
Our approach provides an exact algorithm for maximizing (resp., minimizing) a given pipeline efficiency objective $f$ (resp., pipeline complexity cost $g$) over Equal Opportunity policies if $f$ (resp., $g$) satisfies the following natural condition: 
for any pair of policies $\pi_1, \pi_2$ where $\pi_1$ weakly Pareto dominates $\pi_2$ w.r.t. precision and recall, $f(\pi_1) \ge f(\pi_2)$ (resp., $g(\pi_1) \le g(\pi_2)$).   
\end{rem}

\subsection{An FPTAS Algorithm}\label{sec:fptas}
In this section, we present FPTAS algorithms for maximizing a given pipeline efficiency objective (resp., minimizing a given pipeline cost function) while satisfying the Equal Opportunity requirement.
We consider two regimes. 
In this section, as in previous sections, we consider the regime where we are allowed to treat individuals from different groups differently; more precisely, we can set $\pi^j_{Xi} \neq \pi^j_{Yi}$ for $j\in [k], i\in\{0,1\}$. Next, in Section~\ref{sec:single-policy}, we consider a new regime where the goal is to achieve Equal Opportunity while treating individuals from both groups similarly; $\forall i\in [k], X\neq Y\in {\mathcal X}, \pi^i_{X1} = \pi^i_{Y1}, \pi^i_{X0} = \pi^i_{Y0}$.

To exploit our algorithm in different settings, we describe it for the most basic setting of the problem. Given a single group of applicants with parameters $q,u$ and a pipeline $\mathscr{P} = \{\tau^i\}_{i\in [k]}$, the goal is find a policy $\pi$ that maximizes a given {\em pipeline efficiency objective} $f(\mathrm{recall}(\pi, \mathscr{P}), \mathrm{precision}(\pi, q, u ,\mathscr{P}))$. Our approach works for a quite general set of objective functions; more notably, as in the previous section, for two natural settings: maximizing a {\em linear combination of precision and recall} and minimizing a {\em linear combination of reciprocals of precision and recall}. 

\paragraph{High-level Description of Algorithm.}
Now we write a dynamic program (DP) to optimize a given pipeline efficiency objective $f$ up to a given accuracy parameter $\epsilon$.
We create a DP-table $M[i, \p, \n]$ where $i\in [k]$, $\p \in [0,\ell_{\p} := \log_{1-\epsilon} L_{\p}]$ and $\n \in [0,\ell_{\n} := \log_{1-\epsilon} L_{\n}]$ where $L_{\p}, L_{\n}$ are lower bounds on {\em True Positive Rate} and {\em False Positive Rate} respectively. 
For each set of parameters $(i, \p, \n)$, $M[i,\p,\n]$  will be a Boolean value indicating whether there exists a policy such that by the end of level $i$, the true positive rate becomes at least $(1-\epsilon)^\p$ and the False Positive Rate becomes at most $(1-\epsilon)^\n$. 
Without loss of generality and for the simplicity of the exposition, we assume $L_{\p}$ and $L_{\n}$ are powers of $(1-\epsilon)$; otherwise we can simply round the lower bounds to largest powers of $(1-\epsilon)$ smaller than actual bounds. 

\paragraph{Solving the DP} We fill out the DP table starting from $i=1$ as follows. First, for any $j_0 \in [0,\ell_{\n}], j_1 \in [0, \ell_{\p}]$, $M[1, j_1, j_0] = \mathrm{true}$ iff the following system of linear inequalities has a feasible solution.
\begin{align} 
    \tau^1_0 x + (1- \tau^1_0) y &\leq (1-\epsilon)^{j_0}, &\tau^1_1 x + (1- \tau^1_1) y \geq (1-\epsilon)^{j_1}. &&\label{eq:base-rule}
\end{align}

Next, we describe the update rule for $i>1$. For any $\p \in [0, \ell_{\p}], \n \in [0, \ell_{\n}]$, $M[i+1, \p, \n] = \bigvee_{(j_0, j_1) \in {\mathcal F}_{i+1}} M[i, {\p}-{j_1}, {\n}-{j_0}]$,
% \begin{align*}
%     M[i+1, \p, \n] = \bigvee_{(j_0, j_1) \in {\mathcal F}_{i+1}} M[i, {\p}-{j_1}, {\n}-{j_0}], 
% \end{align*} 
where ${\mathcal F}_{i+1}$ is a set of $(j_0 \le \n, j_1 \le \p)$ for which the following linear program has a feasible solution,
\begin{align}
    \tau^{i+1}_0 x+ (1-\tau^{i+1}_0) y\le (1-\epsilon)^{j_0}, \tau^{i+1}_1 x+ (1-\tau^{i+1}_1) y \ge (1-\epsilon)^{j_1}.\label{eq:update-rule}
\end{align}

Note that $x, y$ can be interpreted as $\pi^{i+1}_1, \pi^{i+1}_0$, respectively. Moreover, the system of linear inequalities of the update rule in level $i+1$ (Eq.~\eqref{eq:update-rule}) is similar to the rules for the base case (Eq.~\eqref{eq:base-rule}). 
% Refer to Figure~\ref{fig:system-linear-inequalities}. 
% \begin{figure}[!h]
% \begin{center}
% \includegraphics[scale=.5]{dp_fig}
% \caption{The gray region, which is the set of points $(x,y)$ below the red line, above the blue line and inside $\{0\le x,y \le 1\}$, denotes the feasible region for the system of linear inequalities with $(j_0, j_1, \tau_0, \tau_1, \epsilon)$ as in the left plot. The system of linear inequalities with $(i_0, i_1, \tau_0, \tau_1, \epsilon)$ as in the right plot has no feasible solution.}\label{fig:system-linear-inequalities}
% \end{center}
% \end{figure}
% \avnote{maybe remove the figure?}
\begin{lem}\label{lem:DP}
For any $i\in [k]$, if there exists a policy $\pi$ with True Positive Rate $\t_i \ge L_{\p}/(1-\epsilon)^{i-1}$ and False Positive Rate $\f_i$ by the end of level $i$, then for any $j_1\in [0, \ell_{\p}], j_0 \in [0, \ell_{\n}]$ with $(1-\epsilon)^{j_1} \ge \t_i \cdot (1-\epsilon)^{i-1}$ and $(1-\epsilon)^{j_0} \le \min\{ 1, \max\{ L_{\n}, \f_i\}/ (1-\epsilon)^{i-1}\}$, $M[i, {j_1}, {j_0}] = \mathrm{true}$.

In other words, if the policy $\pi$ exists then the DP approach finds a policy with true positive rate at least $(1-\epsilon)^{j_1}$ and false positive rate at most $(1-\epsilon)^{j_0}$.
\end{lem}
The proof is deferred to Section~\ref{sec:linear-objective-proofs}.

\begin{lem}[\bf DP Main Lemma]\label{lem:DP-table-runtime}
For any group $X\in {\mathcal X}$, an accuracy parameter $\epsilon$ and lower bounds on the false positive rate, $L_{\n}$, and the true positive rate, $L_{\p}$, if there exists a policy $\pi^*$ with true positive rate $\t \ge L_{\p}/(1-\epsilon)^{k-1}$ and false positive rate $\f>0$, then the DP algorithm runs in time $O(\frac{k \log^2 (1/L_{\p}) \log^2 (1/L_{\n})}{\epsilon^4})$ and finds a policy $\pi$ with true positive rate at least $(1-\epsilon)^{k-1} \cdot \t$ and false positive rate at most $\min\{ 1, \max\{ L_{\n}, \f\}/ (1-\epsilon)^{k-1}\}$.
\end{lem}
\begin{proof}
The size of table is $O(k \ell_{\p} \ell_{\n})$ and updating each entry in the table takes $O(\ell_\p \ell_\n)$. Hence, the total runtime to compute all entries in the DP table is $O(k\ell^2_\p \ell^2_\n) = O(\frac{k \log^2(1/L_{\p}) \log^2(1/L_{\n})}{\epsilon^{4}})$.

Now we apply the DP approach and by Lemma~\ref{lem:DP}, the solution returned by the algorithm has the true positive rate and the false positive rate satisfying the guarantee of the statement.
\end{proof}

\paragraph{Implications of DP} Here we present FPTAS algorithms using the described DP approach in different settings. We state the results formally and their proofs are deferred to Appendix~\ref{sec:linear-objective-proofs}.

\begin{thm}[\bf FPTAS for Linear Combination of Precision and Recall]\label{thm:EQ-DP-max}
Consider a $k$-stage screening process with parameters $(u, q, \tau, \mathcal X)$ and for any policy $\pi$, let $f_{\alpha}(\pi) = (1-\alpha) \cdot \mathrm{recall}(\pi) + \alpha \cdot \mathrm{precision}(\pi)$ where $\alpha > 0$. Given an accuracy parameter $\epsilon$, there exists an FPTAS that runs in time $O(\frac{|{\mathcal X}| k^5 \log^4(1/\epsilon)}{\epsilon^4})$ and finds an Equal Opportunity policy $\pi$ such that $f_{\alpha}(\pi) \ge (1-\epsilon)f_{\alpha}(\pi^*)$ where $\pi^*$ maximizes $f_\alpha$ over Equal Opportunity policies. 
\end{thm}

\begin{thm}[\bf FPTAS for Linear Combination of Reciprocals Precision and Recall]\label{thm:EQ-DP-min}
Consider a $k$-stage screening process with parameters $(u, q, \tau, \mathcal X)$ and for any policy $\pi$, let $g_{\alpha}(\pi) = (1-\alpha)/\mathrm{recall}(\pi) + \alpha / \mathrm{precision}(\pi)$ where $\alpha > 0$. Given an accuracy parameter $\epsilon$, there exists an FPTAS that runs in time $O(\frac{|{\mathcal X}| k^7 (\log^2\frac{1}{\epsilon} + k^2)}{\epsilon^4})$ and finds an Equal Opportunity policy $\pi$ such that $g_{\alpha}(\pi) \le (1+\epsilon)g_{\alpha}(\pi^*)$ where $\pi^*$ minimizes $g_\alpha$ over Equal Opportunity policies. 
\end{thm}
% The proof of the above theorem is similar to the proof of Theorem~\ref{thm:EQ-DP-max} and is deferred to Section~\ref{sec:linear-objective-proofs}.
%\paragraph{General Function of Precision and Recall.} 
\begin{rem}[{\bf General Objective Functions}]\label{rem:fptas}
In Theorem~\ref{thm:EQ-DP-max} and~\ref{thm:EQ-DP-min} we presented FPTAS for finding Equal Opportunity policies optimizing two standard pipeline efficiency objective functions. 
Here, we generalize the above theorems when the pipeline efficiency objective function $f: [0,1]^{2} \rightarrow \mathbb{R}$ which maps precision and recall to efficiency scores have certain properties. 
Also, we define $g:[0,1]^{2} \rightarrow \mathbb{R}$ such that for any $\t,\f \in [0,1]^{2}$, $g(\t, \f) := f(\mathrm{recall}(\t),\mathrm{precision}(\t,\f))$.
We describe the properties when the goal is to maximize $f$---{\em the required conditions for the minimization version is similar}.
\begin{itemize}
    \item $f$ is non-decreasing w.r.t.~both precision and recall---{\em equivalently, $g$ is non-decreasing in $\t$ and non-increasing in $\f$}.
    \item There exist $L_{\p}, L_{\n} >0$ such that there exists a $(1-\alpha)$-approximate solution of $f$ with $\t \in (L_{\p},1], \f \in (L_{\n},1]$.
    \item The function $f$ is $\beta$-Lipschitz on $\{(x,y) | x\in (L_{\p}, 1], y \in (L_{\n},1]\}$.
\end{itemize}
In particular, the above properties are sufficient to show that the DP approach finds a $(1-\epsilon)$-approximation of $f$ in time 

$\mathrm{poly}(k, |\mathcal{X}|, \epsilon^{-1}, \log(1/L_{\p}), \log(1/L_{\n}))$.
\end{rem}

%\subsection{Selecting from Available Tests}
\begin{rem}[{\bf Selecting from Available Tests}]\label{rem:select-tests}
Suppose that in contrast to our previous approaches, we do allow for the design of the pipeline in that we allow the firm to select some tests to create a pipeline.
For instance, imagine that there is a budget and the firm is allocating this budget to buy tests.
The goal of the firm is the same, e.g. to exhibit a pipeline satisfying a fairness requirement. 
Our algorithms can be modified to handle to this case by adding a term in the DP table corresponding to the budget remaining, with a decision point of choosing to use a given test or not.  Note that the ordering of tests in the pipeline does not matter for the objectives considered.
\end{rem}

\section{Alternate Models}\label{sec-altmodels} In this section we describe some alternate settings, such as using a single promotion policy for both demographic groups (which might be required by regulation), or requiring Equalized Odds.
%, and requiring the fairness criterion hold independently at every step of the screening process. 
%These alternate scenarios evince some of the relative strengths of our primary scenario, Equal Opportunity with two screening processes.
%However, these alternate scenarios are of practical relevance and are worth discussing independently.

\subsection{Screening Processes with Same Policy for All Groups} \label{sec:single-policy}
One alternate fairness model is to additionally require the same policy be used for all groups.
%there are \textit{not} multiple classifiers fine-tuned for each group. 
While utilizing demographic features can aid in achieving fairness goals (e.g.~\cite{dwork2012fairness, hardt16}), 
%in alternative words, this could mean giving two individuals of different demographic groups  different predictions or decisions even if 
%every other feature in their corresponding data point is identical. 
in some regulatory regimes, this fairness-through-awareness may be illegal or problematic, even when intended to ensure equitable treatment. 

In our setting,  if we are constrained to follow \textit{group-blindness}, there be would only one set of tests and one ordering of the tests that all applicants are tested on. Analogously to the previous setting, the action space of the algorithm remains modifying the promotion probabilities, but we now only have one set of policies to modify. 
We also exhibit a DP algorithm for this setting, which we defer to Section~\ref{sec:single-policy-app}.
However, a simple example shows the inefficiencies in this regime. 
Suppose we have a single test with $T_{A} = (1,0)$ and $T_B =(1/2,0).$
Observe that since we are constrained to use group blindness and satisfy Equal Opportunity, there is no way to use the test without violating Equal Opportunity.  Thus, the only way to satisfy Equal Opportunity is to completely bypass the test.

\subsection{Equalized Odds \label{eodds}}
Next, recall that the requirement of Equalized Odds mandates equal True Positive and False Positive rates for all groups.
%, a stronger requirement than simply requiring Equal Opportunity. 
%Now we consider the analogous results and algorithms for satisfying Equalized Odds in our screening process. 
In the appendix, we show structural properties of an optimal promotion policy that satisfies Equalized Odds. 
However, we also note the interview efficiency cost (precision) of requiring Equalized Odds.
In particular, the gap between the interview efficiency of $\pi_{\mathrm{EOdd}}$ and $\pi_{\mathrm{EOpp}}$ can be as large as $\frac{1}{q}-\epsilon$ for any arbitrary $\epsilon >0$.
See Theorem \ref{thm:Eodd-vs-Eopp} for details. 

\subsection{Discussion  Comparing Equalized Odds and Equal Opportunity} 
From the perspective of a decision maker in the wild, how to interpret and operationalize these results?
A robust take-away is that requiring Equalized Odds and Equal Opportunity have substantially different efficiency consequences.
Based on our examples, it seems unlikely that Equalized Odds is effective in this model, especially when requiring Equalized Odds at each stage.
In contrast, the fact that requiring Equal Opportunity at each stage is equivalent to requiring Equal Opportunity of the overall process with respect to interview efficiency may have benefits in ensuring public confidence in the model. 

\subsection{Intersectionality \label{intersect}}
A natural question is how to think when the demographic groups may have an arbitrarily overlapping structure. 
This suggests several open questions in our model, e.g. if a person is in groups $A$ and $B$, then which test parameter $\tau_{A}$ or $\tau_{B}$ corresponds to that person?
Perhaps a direction is to assign to that person an interpolation between these values.
A naive approach is when there are $k$ groups, to create $2^{k}$ new groups and $2^{k}$ test parameters corresponding to every possible group intersection.
If $k$ is small, this may be computationally feasible, but is not responsive when the relevant sub-groups/intersections may not be known apriori.
Perhaps our model could be merged with multi-calibration notions~\citep{hebert2018multicalibration}.
%Extensions to these models would include more expressive labels e.g. a real valued score for labels with high scores being preferred or the tests returning scores as well, rather than binary outcomes in our model. 

\section{Conclusion}
In contrast to some fairness in machine learning work, we focus on post-processing fairness modifications, rather than thinking about the fairness problem in screening processes where tests can be designed from scratch. 
While we believe that the more a priori design approach will have substantial benefits in practice, our approach of modifying pre-existing tests, combined with a concrete (and simple to evaluate) fairness notion, Equal Opportunity, is closely aligned with real world circumstances and models, especially in short term and iterative improvements to models. In some settings, the firm making hiring decisions will outsource some aspects of its pipeline to third party companies and the tests will be a black box, but possibly that come with statistics that can be used in our algorithms.  
This decoupling allows the effective implementation of fairness aware promotion policies in the short term.

%% file: fairscreen_missing-proofs.tex
\section{Proofs from Section~\ref{maxprecsection}}\label{sec:precision-max-proof}
\begin{proof}[Proof of Theorem~\ref{thm:multi-eq-opp-prec}]
First, we show that for any $M\in (0,1]$, any Equal Opportunity policy $\pi_M$ with recall $M$ has interview efficiency at most
\begin{align}
\ie(q, u, \tau, \pi_M) 
& = \frac{\sum_{X\in \mathcal X}\q_X M_{X, \tau, \pi_M}}{\sum_{X\in \mathcal X}\q_X M_{X, \tau, \pi_M} + \u_X N_{X, \tau, \pi_M}} \nonumber \\
&= %\frac{q M}{q M + \sum_{X\in \mathcal X} \u_X N^X_{\tau, \pi_M}} = 
\frac{\|q\|_1}{\|q\|_1 + \sum_{X\in \mathcal X} \u_X \frac{N_{X, \tau, \pi_M}}{M}} \nonumber \\
&\leq \frac{\|q\|_1}{\|q\|_1 + \sum_{X\in \mathcal X} \u_X \Pi_{i=1}\frac{\tau^i_{X0}}{\tau^i_{X1}}}, \label{eq:upper-bound-interview-eff-mult}  
\end{align}
where the last inequality follows from the minimal effectiveness of tests in the screening process and an argument identical to Eq.~\eqref{eq:X-bound}. Note that the inequality holds no matter what the value of $M$ is.
Next, we show that opportunity ratio policy achieves the maximum possible interview efficiency as shown in Eq.~\eqref{eq:upper-bound-interview-eff-mult}. Let $X^* = \argmin_{X\in \mathcal X} \Pi_{j\in [k]}\tau^j_{X1}$. Recall that the opportunity ratio policy $\pi$ is defined as follow.
\begin{align*}
    \pi^1_{X0} &= 0 \text{ and } \pi^1_{X1} = \Pi_{i\in [k]}(\tau^i_{X^*1}/\tau^i_{X1}) &&\forall X\in \mathcal X \\
    \pi^i_{X0} &= 0 \text{ and } \pi^i_{X1} = 1 &&\forall X\in {\mathcal X}, i\ge 2
\end{align*}
It is straightforward to check that $\pi$ is an Equal Opportunity policy with recall $\Pi_{i\in [k]} \tau^i_{X^*1}$. Moreover, the interview efficiency of $\pi$ is 
\begin{align*}
\ie(q, u, \tau, \pi)
&= \frac{\sum_{X\in \mathcal X}\q_X M_{X, \tau, \pi}}{\sum_{X\in \mathcal X}\q_X M_{X, \tau, \pi} + \u_X N^X_{\tau, \pi}} \\
&= \frac{\sum_{x\in \mathcal{X}}q_X \Pi_{i\in [k]} \tau^i_{X^*1}}{\sum_{x\in \mathcal{X}}q_X \Pi_{i\in [k]} \tau^i_{X^*1} + \sum_{X\in \mathcal X}\u_X \Pi_{i\in [k]} \frac{\tau^i_{X^*1}\tau^i_{X0}}{\tau^i_{X1}}} \\
&= \frac{\|q\|_1}{\|q\|_1 + \sum_{X\in \mathcal X}\u_X \Pi_{i=1}^k\frac{\tau^i_{X0}}{\tau^i_{X1}}}
\end{align*}
Hence, $\pi$ is the Equal Opportunity policy maximizing the interview efficiency.
\end{proof}

\section{Proofs from Section~\ref{sec:linear-objective}}\label{sec:linear-objective-proofs}

\begin{proof}[Proof of Claim~\ref{clm:non-zero-pi-one}]
Suppose for contradiction that there exists a group $X\in \mathcal X$ and a level $i\in [k]$ such that $\pi^i_{X1} = 0$. 
First note that $(1-\tau^i_{X1}) \pi^i_{X0} >0$; otherwise, the policy is useless because it prevents candidates of group $X$, in particular the qualified ones, from reaching the interview stage. Hence, by the Equal Opportunity requirement, no qualified candidate will reach the interview stage. 

Next, we show that there exists a policy $\tilde{\pi}$ (which only differs from $\pi$ in level $i$ of group $X$) that satisfies Equal Opportunity for the given screening process and strictly Pareto dominates $\pi$; $\tilde{\pi}^i_{X1}= (\frac{1-\tau^i_{X1}}{\tau^i_{X1}}) \pi^i_{X0}$ and $\tilde{\pi}^i_{X0} = 0$.
% \begin{align*}
%     \Big(\tilde{\pi}^i_{X1}&= (\frac{1-\tau^i_{X1}}{\tau^i_{X1}}) \pi^i_{X0},\tilde{\pi}^i_{X0} = 0\Big) &&\Big(\tilde{\pi}^j_{Y1} = \pi^j_{Y1}, \tilde{\pi}^j_{Y0} = \pi^j_{Y0}\Big) \quad \text{if $(Y\neq X) \vee (j\neq i)$}
% \end{align*}

Since $ M_{X, \tau^i, \tilde{\pi}^i} = \tau^i_{X1} \tilde{\pi}^i_{X1} + (1-\tau^i_{X1}) \tilde{\pi}^i_{X0} = \tau^i_{X1} \tilde{\pi}^i_{X1} = (1-\tau^i_{X1}) \pi^i_{X0} = \tau^i_{X1}\pi^i_{X1} + (1-\tau^i_{X1}) \pi^i_{X0} = M_{X, \tau^i, \pi^i} $
and $\pi$ satisfies the Equal Opportunity, $\tilde{\pi}$ also satisfies Equal Opportunity and has the same recall as $\pi$. Moreover, since
\[N_{X, \tau^i, \tilde{\pi}^i} = \tilde{\pi}^i_{X1} \tau^i_{X0} = (\frac{1-\tau^i_{X1}}{\tau^i_{X1}})\pi^i_{X0} \tau^i_{X0} < (1-\tau^i_{X0}) \pi^i_{X0} = N_{X, \tau^i, \pi^i},\] $\mathrm{precision}(\tilde{\pi}) > \mathrm{precision}(\pi)$. Note that $\frac{1- \tau^i_{X1}}{\tau^i_{X1}} < \frac{1- \tau^i_{X0}}{\tau^i_{X0}}$ holds by the minimal effectiveness property of tests.
%if for all $i\in [k]$, $\tau^i_{X1} < 1$, by the minimal effectiveness property, $\mathrm{precision}(\tilde{\pi}) > \mathrm{precision}(\pi)$.
\end{proof}

%\begin{align*}
 %   & (\frac{1-\tau_1}{\tau_1} )\pi_0 \tau_0 \leq (1-\tau_0) \pi_0 \\
  %  & (1-\tau_1) \tau_0 \leq (1-\tau_0) \tau_1 \\
  %  & \tau_0 - \tau_1 \tau_0 \leq \tau_1 - \tau_1 \tau_0 \\
   % & \tau_0 \leq \tau_1 \quad \text{(minimal effectiveness)}
%\end{align*}

\begin{lem}\label{lem:pi-one}
Consider a $k$-stage screening process whose tests satisfy the Minimal Effectiveness Property.
The set of Equal Opportunity policies $\mathcal{S} \subseteq \mathcal{P} = \{\pi \in [0,1]^{2|\mathcal{X}|k} | (1 - \pi^i_{X1}) \pi^i_{X0} = 0, \forall X\in \mathcal{X}, i\in [k]\}$, where for each group $X\in \mathcal X$, there exists at most one level $i \in [k]$ such that $\pi^i_{X1} < 1$, weakly Pareto dominates all Equal Opportunity policies.
\end{lem}
\begin{proof}
Suppose for contradiction that there are two levels $i,j$ such that $\pi^i_{X1}, \pi^j_{X1} < 1$. First note that by Claim~\ref{clm:non-zero-pi-one}, $\pi^i_{X1}, \pi^j_{X1} > 0$. Moreover, by Lemma~\ref{lem:pi-cond}, since $\pi^i_{X1}, \pi^j_{X1} < 1$, $\pi^i_{X0} = \pi^j_{X0} = 0$.

Next, we show that we can modify $\pi$ in levels $i$ and $j$ and replace $\pi^i_{X0}, \pi^j_{X0}$ with $\tilde{\pi}^i_{X0}, \tilde{\pi}^j_{X0}$ as follows: $\tilde{\pi}^i_{X1} = \pi^i_{X1} \pi^j_{X1}$ and $\tilde{\pi}^j_{X1} = 1$. Then, $M_{X, \tau^i, \tilde{\pi}^i} M_{X, \tau^j, \tilde{\pi}^j}
    = (\tilde{\pi}^i_{X1} \tau^i_{X1}) (\tilde{\pi}^j_{X1} \tau^j_{X1}) 
    = (\pi^i_{X1} \tau^i_{X1}) (\pi^j_{X1} \tau^j_{X1}) 
    = M_{X, \tau^i, \pi^i} M_{X, \tau^j, \pi^j}$.
% \begin{align*}
%     M^X_{\tau^i, \tilde{\pi}^i} M^X_{\tau^j, \tilde{\pi}^j}
%     = (\tilde{\pi}^i_{X1} \tau^i_{X1}) (\tilde{\pi}^j_{X1} \tau^j_{X1}) 
%     = (\pi^i_{X1} \tau^i_{X1}) (\pi^j_{X1} \tau^j_{X1}) 
%     = M^X_{\tau^i, \pi^i} M^X_{\tau^j, \pi^j}.
% \end{align*}
In other words, the policy $\tilde{\pi}$ satisfies Equal Opportunity and has the same recall as $\pi$. Similarly, this modification does not decrease precision. Formally, $N_{X, \tau^i, \tilde{\pi}^i} N_{X, \tau^j, \tilde{\pi}^j}
    = (\tilde{\pi}^i_{X1} \tau^i_{X0}) (\tilde{\pi}^j_{X1} \tau^j_{X0}) 
    = (\pi^i_{X1} \tau^i_{X0}) (\pi^j_{X1} \tau^j_{X0}) 
    = N_{X, \tau^i, \pi^i} N_{X, \tau^j, \pi^j}$.
% \begin{align*}
%     N^X_{\tau^i, \tilde{\pi}^i} N^X_{\tau^j, \tilde{\pi}^j}
%     = (\tilde{\pi}^i_{X1} \tau^i_{X0}) (\tilde{\pi}^j_{X1} \tau^j_{X0}) 
%     = (\pi^i_{X1} \tau^i_{X0}) (\pi^j_{X1} \tau^j_{X0}) 
%     = N^X_{\tau^i, \pi^i} N^X_{\tau^j, \pi^j}.
% \end{align*}
Hence, for each policy $\pi$, there exists another policy with at most one level $i\in [k]$ such that $\pi^i_{X0} <1$ and weakly Pareto dominates $\pi$.  
\end{proof}

\begin{proof}[Proof of Lemma~\ref{lem:one-partial}]
We follow a similar arguments as in the proof of Lemma~\ref{lem:pi-zero}. Note that by Lemma~\ref{lem:pi-zero} and Lemma~\ref{lem:pi-one} there is at most one level $i_1\in [k]$ such that $0< \pi^{i_1}_{X1} < 1$ and $\pi^{i_1}_{X0} = 0$, and there is at most one level $i_0\in [k]$ such that $\pi^{i_0}_{X1} = 1$ and $0< \pi^{i_0}_{X0} < 1$. Next, we show that we can modify the policy $\pi$ in levels $i_0$ and $i_1$ and replace $\pi^{i_0}_{X0}, \pi^{i_1}_{X1}$ with $\tilde{\pi}^{i_0}_{X0}, \tilde{\pi}^{i_1}_{X1}$ such that
\begin{align*}
    M_{X, \tau^{i_0}, \pi^{i_0}} M_{X, \tau^{i_1}, \pi^{i_1}}
    & = (\tau^{i_0}_{X1} + \pi^{i_0}_{X0} (1- \tau^{i_0}_{X1})) (\pi^{i_1}_{X1}\tau^{i_1}_{X1}) \\
    &= (\tau^{i_0}_{X1} + \tilde{\pi}^{i_0}_{X0} (1- \tau^{i_0}_{X1})) (\tilde{\pi}^{i_1}_{X1}\tau^{i_1}_{X1}) \\
    & = M_{X, \tau^{i_0}, \tilde{\pi}^{i_0}} M_{X, \tau^{i_1}, \tilde{\pi}^{i_1}},  \\
    N_{X, \tau^{i_0}, \pi^{i_0}} N_{X, \tau^{i_1}, \pi^{i_1}}
    &= (\tau^{i_0}_{X0} + \pi^{i_0}_{X0} (1- \tau^{i_0}_{X0})) (\pi^{i_1}_{X1}\tau^{i_1}_{X0}) \\
    &< (\tau^{i_0}_{X0} + \tilde{\pi}^{i_0}_{X0} (1- \tau^{i_0}_{X0})) (\tilde{\pi}^{i_1}_{X1}\tau^{i_1}_{X0}) \\
    &= N_{X, \tau^{i_0}, \tilde{\pi}^{i_0}} N_{X, \tau^{i_1}, \tilde{\pi}^{i_1}} 
\end{align*}
Now, we show that in the new solution, either $\tilde{\pi}^{i_0}_{X0} \in \{0,1\}$ or $\tilde{\pi}^{i_1}_{X1} = 1$. 

Without loss of generality, we can assume that the feasible range of values for $\tilde{\pi}^{i_0}_{X0}$ to satisfy Equal Opportunity is $[\pi^{i_0}_{X0} - \epsilon^{i_0}, \pi^{i_0}_{X0} + \delta^{i_0}]$ which corresponds to $[\pi^{i_1}_{X0} - \delta^{i_1}, \pi^{i_1}_{X0} + \epsilon^{i_1}]$. 
Both intervals are sub-intervals of $[0,1]$ and it is straightforward to verify that $(\pi^{i_0}_{X0} - \epsilon^{i_0}) (1 - (\pi^{i_1}_{X0} + \epsilon^{i_1})) = (1-(\pi^{i_1}_{X0} + \delta^{i_1}))=0$.

Let $L = M_X/ (\tau^{i_0}_{X1}\tau^{i_1}_{X1})$ where 
\[ M_X = M_{X, \tau^{i_0}, \pi^{i_0}} M_{X, \tau^{i_1}, \pi^{i_1}} = M_{X, \tau^{i_0}, \tilde{\pi}^{i_0}} M_{X, \tau^{i_1}, \tilde{\pi}^{i_1}} \]. 
By the Minimal Effectiveness Property, $0< L < 3$. Then, satisfying Equal Opportunity is equivalent to satisfy $(1 + \tilde{\pi}^{i_0}_{X0}(\frac{1-\tau^{i_0}_{X1}}{\tau^{i_0}_{X1}})) \tilde{\pi}^{i_1}_{X1} = L$, which implies that $\tilde{\pi}^{i_1}_{X1} = L/(1 + \tilde{\pi}^{i_0}_{X0}(\frac{1-\tau^{i_0}_{X1}}{\tau^{i_0}_{X1}}))$.
The task of finding $\tilde{\pi}^i_{X0}$ is as follows: 
\begin{align*}
    \tilde{\pi}^i_{X0} 
    &= \argmin_{y\in [\pi^{i_0}_{X0} - \epsilon^{i_0}, \pi^{i_0}_{X0} + \delta^{i_0}]} f(y) \\
    &:= (\tau^{i_0}_{X0} + y (1- \tau^{i_0}_{X0})) (\tau^{i_1}_{X0} \cdot \frac{L}{1 + y(\frac{1-\tau^{i_0}_{X1}}{\tau^{i_0}_{X1}})}).
    \end{align*}
Next, we show that for any $y \in [0,1]$,
\[ f''(y) =\frac{2L\tau^{i_1}_{X0} (\frac{1-\tau^{i_0}_{X1}}{\tau^{i_0}_{X1}})(\tau^{i_0}_{X0} (\frac{1-\tau^{i_0}_{X1}}{\tau^{i_0}_{X1}}) + \tau^{i_0}_{X0} - 1)}{(1 + \tau^{i_0}_{X0} y)^3} < 0 \]
To prove it note that the Minimal Effectiveness Property of the tests $\{\tau^i\}_{i\in [k]}$ (i.e., $\tau^i_{X1} > \tau^i_{X0} \ge 0, \forall X\in {\mathcal X}, i\in [k]$) implies that $\frac{\tau^{i_0}_{X0}}{\tau^{i_0}_{X1}} -1 < 0$. Since $f$ is a concave function in $[\pi^{i_0}_{X0} - \epsilon^{i_0}, \pi^{i_0}_{X0} + \delta^{i_0}]$, the minimum value of $f$ in this interval obtained in one of its endpoints. In other words, the maximum precision corresponds to the case either $\tilde{\pi}^{i_0}_{X0} \in \{0,1\}$ or $\tilde{\pi}^{i_1}_{X1} = 1$.
\end{proof}

\begin{proof}[Proof of Lemma~\ref{lem:DP}]
The proof is by induction. For the base case ($i=1$), let $\t_1$ and $\f_1$ denote the true positive rate and the false positive rate of $\pi$ by the end of level $1$. 
The existence of $\pi$ guarantees that the system of inequalities Eq.~\eqref{eq:base-rule} with $(j_0 = \lfloor \log_{1-\epsilon} \f_1 \rfloor, j_1 = \lceil \log_{1-\epsilon}\t_1 \rceil \leq \ell_{\p})$ has a feasible solution.  
More precisely, by setting $(x = \pi_1, y = \pi_0)$, 
\begin{align*} 
    & \tau^1_0 x + (1- \tau^1_0) y = \f_1 \le (1-\epsilon)^{\lfloor \log_{1-\epsilon} \f_1 \rfloor} \\
    & = (1-\epsilon)^{j_0}, \quad \tau^1_1 x + (1- \tau^1_1) y \\
    &= \t_1 \geq (1-\epsilon)^{\lceil \log_{1-\epsilon} \t_1 \rceil} = (1-\epsilon)^{j_1}
\end{align*}
Next, we consider $i>1$ and we assume that the claim holds for all values $i' < i$. Let $M_i := \tau^i_1\pi^i_1 + (1 - \tau^i_1) \pi^i_0$ and $N_i := \tau^i_0\pi^i_1 + (1-\tau^i_0)\pi^i_0$. Note that $\t_i = \t_{i-1} \cdot M_i$ and $\f_i = \f_{i-1} \cdot N_i$.

By the induction hypothesis and considering the first $i-1$ levels in the pipeline, since $\t_{i-1} \ge \t_i \ge L_{\p}/(1-\epsilon)^{i-1} > L_{\p}/(1-\epsilon)^{i-2}$ and $\f_{i-1} \ge \f_i$, there exist $j'_1\in[0, L_{\p}]$ and $j'_0\in [0, L_{\n}]\cup \{\infty\}$ such that $M[i-1, j'_1, j'_0] = \mathrm{true}$ and $(1-\epsilon)^{j'_1} \ge \t_{i-1} \cdot (1-\epsilon)^{i-2}$ and $(1-\epsilon)^{j'_0} \le \min\{1, \max\{L_{\n}, \f_{i-1}\}/(1-\epsilon)^{i-2}\}$. 
More precisely, the algorithm finds a policy $\bar{\pi}$ with true positive rate at least $(1-\epsilon)^{j'_1}$ and false positive rate at most $(1-\epsilon)^{j'_0}$.

Next, by setting $(\bar{\pi}^i_1 = \pi^i_1, \bar{\pi}^i_0 = \pi^i_0)$ and $(j_1 :=\argmin_{j} \{(1-\epsilon)^j \le \t_i(\bar\pi)\}, j_0 := \argmax_{j} \{(1-\epsilon)^j \ge \f_i(\bar\pi)\})$,  
\begin{align*}
    (1-\epsilon)^{j_1} 
    &> (1-\epsilon)\cdot \t_{i}(\bar \pi) \\
    &= (1-\epsilon)\cdot \t_{i-1}(\bar \pi) \cdot M_i &&\rhd\text{by definition of $j_1$}\\
    &\ge (1-\epsilon)\cdot (1-\epsilon)^{j'_1} \cdot M_i &&\rhd\text{by $\t_{i-1}(\bar\pi) \ge (1-\epsilon)^{j'_1}$}\\ 
    &\ge \t_{i-1} \cdot (1-\epsilon)^{i-1} \cdot M_i &&\rhd\text{by induction hypothesis}\\
    &= \t_i \cdot (1-\epsilon)^{i-1}. 
\end{align*}
Similarly,
\begin{align*}
    (1-\epsilon)^{j_0} 
    &< \min\{1, \frac{\f_{i}(\bar \pi)}{1-\epsilon}\} \\
    &= \min\{1, N_i \cdot \frac{\f_{i-1}(\bar \pi)}{1-\epsilon}\} \;\rhd\text{by definition of $j_0$}\\
    &\le \min\{1, (1-\epsilon)^{j'_0} \cdot \frac{N_i}{1-\epsilon}\} \;\rhd\text{ $\f_{i-1}(\bar\pi) \le (1-\epsilon)^{j'_0}$} \\
    &\le \min\{1, \frac{\max\{L_{\n}, \f_{i-1}\}}{(1-\epsilon)^{i-2}} \cdot \frac{N_i}{1-\epsilon}\} \;\rhd\text{induction hypoth.}\\ 
    &\le \min\{1, \frac{\max\{L_{\n}, \f_i\}}{(1-\epsilon)^{i-1}}\}
\end{align*}
which completes the proof.
\end{proof}

\begin{proof}[Proof of Theorem~\ref{thm:EQ-DP-max}]
First, as we are aiming for a $(1-\epsilon)$-approximation, we only need to consider $\alpha\in (\epsilon, 1-\epsilon)$. Otherwise, either the policy maximizing recall (i.e. bypassing all tests) or the policy maximizing precision (Opportunity Ratio policy) is a $(1-\epsilon)$-approximation for $f_\alpha$.  

Next we show in order to guarantee $(1-\epsilon)$-approximations of recall and precision of the policy maximizing $f_\alpha$, it suffices to run the described DP and consider estimates of $\t$ (true positive rate) and $\f$ (false positive rate) of form $(1-\bar\epsilon)^i$ for $i\in \mathbb{N}$ in intervals $[L_{\p}, 1]$ and $[L_{\n}, 1]$ respectively, where $\bar\epsilon \le \epsilon/(2k)$. We provide tight bounds for $L_{\p}$ and $L_{\n}$. 
Note that since for any policy $\pi$, the true positive rate ($\t_i$) and the false positive rate ($\f_i$) are non-decreasing in $i$, it suffices to provide ``large enough'' lowerbounds $L_{\p}$ and $L_{\n}$ for $\t$ and $\f$ in the final stage respectively. 

\paragraph{Bounding $L_{\p}$.} Consider the policy $\pi_{\bypass}$, which bypasses all the tests in both groups, i.e., $\pi^i_{X0}=\pi^i_{X1} =1$ for all $i\in [k], X\in {\mathcal X}$. Since $\pi_\bypass$ is an Equal Opportunity policy for the pipeline and $f_\alpha(\pi_\bypass) = (1-\alpha) + \alpha \|q\|_1$, any optimal Equal Opportunity policy $\pi^*$ for $f_\alpha$ has recall at least $(1-2\alpha + \alpha \|q\|_1)/(1-\alpha)$. Thus, since $\alpha\in (\epsilon, 1-\epsilon)$, $\t \ge (1-2\alpha + \alpha \|q\|_1)/(1-\alpha) \ge \epsilon / (1-\epsilon)$ which implies that in our DP with accuracy parameter $\bar\epsilon$ it suffices to set $L_{\p} = (\frac{\epsilon}{1-\epsilon}) \cdot (1-\bar\epsilon)^{k-1} \ge (\frac{\epsilon}{1-\epsilon}) \cdot \exp(-\epsilon)$.
%= O((1-\frac{\epsilon}{2k})^k)$. 

\paragraph{Bounding $L_{\n}$.} 
For each $X\in \mathcal X$, let $\f_X$ denote the false positive rate of the optimal Equal Opportunity policy for group $X$. Similarly, let $\t_X$ denote the positive rate of (i.e., recall) the optimal policy $\pi^*$ for group $X\in \mathcal X$. By Equality of Opportunity property of $\pi^*$, $\t_{X} = \t$ for each $X\in {\mathcal X}$.
Next, we consider the following cases.

For any sufficiently small $\epsilon >0$, we need to set $L_{\n}$ so that by running the DP with accuracy parameter $\bar\epsilon$, we can approximate both true positive rate and false positive rate of the optimal Equal Opportunity policy within $(1-\epsilon)$-factor of their values. More precisely, we set $L_{\n}$ so that if for each group $X\in {\mathcal X}$ and any pair $(\t_X, \f_X)$ with $t_X \ge {L_{\p}}/(1-\epsilon/2)$, there exists a pair $(\bar{\t}_X, \bar{f}_X)$ such that $\bar{\t}_X \ge (1-\epsilon/2) \t_X$, $\bar{\f}_X \le \min(1, \max(L_{\n}, \f_X)/(1-\epsilon/2))$.
Finally, once the above property holds for all groups $X\in {\mathcal X}$, then for the corresponding policy $\pi$,  $\mathrm{precision}(\pi) > (1-\epsilon) \cdot \mathrm{precision}(\pi^*)$.

Let ${\mathcal X}_1 := \{X\in {\mathcal X} | \f_{X}/(1-\frac{\epsilon}{2}) \ge L_{\n}\}$ and ${\mathcal X}_2 := \{X\in {\mathcal X} | \f_{X}/(1-\frac{\epsilon}{2}) < L_{\n}\}$. Then,
\begin{align*}
    \frac{\mathrm{precision}(\pi)}{\mathrm{precision}(\pi^*)} 
    &=\frac{\big(\frac{\|q\|_1 \cdot \bar{\t}}{\|q\|_1 \cdot \bar{\t} + \sum_{X\in \mathcal X} u_X \cdot \bar{\f}_X}\big)}{ \big(\frac{\|q\|_1 \cdot \t}{\|q\|_1 \cdot \t + \sum_{X\in {\mathcal X}}u_X \cdot \f_X}\big) }\\
    &\ge \frac{\big(\frac{\|q\|_1 \cdot \bar{\t}}{\|q\|_1 \cdot \bar{\t} + \sum_{X\in {\mathcal X}_1} u_X \cdot \bar{\f}_X + \sum_{X\in {\mathcal X}_2} u_X \cdot \bar{\f}_X}\big) } {\big(\frac{\|q\|_1 \cdot \t}{\|q\|_1 \cdot \t + \sum_{X\in {\mathcal X}_1}u_X \cdot \f_X}\big)} \\
    &\ge \frac{\big(\frac{\|q\|_1 \cdot (1-\epsilon/2) \t}{\|q\|_1 \cdot \t + \sum_{X\in {\mathcal X}_1} \frac{u_X \cdot \f_X}{1-\epsilon/2} + \sum_{X\in {\mathcal X}_2} u_X \cdot L_{\n}}\big) }{ \big(\frac{\|q\|_1 \cdot \t}{\|q\|_1 \cdot \t + \sum_{X\in {\mathcal X}_1}u_X \cdot \f_X}\big)}
\end{align*}
Next, we set $L_{\n}$ so that $\|q\|_1\cdot \t + \sum_{X\in {\mathcal X}_2} u_X L_{\n} \le \frac{\|q\|_1\cdot \t}{1-\frac{\epsilon}{2}}$. Since $\t \ge \epsilon / (1-\epsilon)$, it suffices to set $L_\n = \frac{\epsilon^2 \|q\|_1}{(2-\epsilon) (1-\epsilon) (1 - \|q\|_1)} = \Omega(\epsilon^2)$. Hence,
    \begin{align*}
        \frac{ \big(\frac{\|q\|_1 \cdot (1-\frac{\epsilon}{2}) \t}{\|q\|_1 \cdot \t + \sum_{X\in {\mathcal X}_1} \frac{u_X \cdot \f_X}{1-\frac{\epsilon}{2}} + \sum_{X\in {\mathcal X}_2} u_X \cdot L_{\n}}\big) }{ \big(\frac{\|q\|_1 \cdot \t}{\|q\|_1 \cdot \t + \sum_{X\in {\mathcal X}_1}u_X \cdot \f_X}\big) } \ge (1-\frac{\epsilon}{2})^2 > (1-\epsilon).
    \end{align*}

Finally, for each $X\in \mathcal X$, we run the DP algorithm for each group with accuracy parameter $\bar\epsilon$. 
By Lemma~\ref{lem:DP}, the DP algorithm finds a set $\{\t_X = (1-\bar\epsilon)^{i_X}, \f_x = (1-\bar\epsilon)^{j_X}\}_{X\in \mathcal X}$ (and a policy $\pi$ achieving these rates) where for each $X\in \mathcal X$, $\t_X \in [L_{\p}, 1], \f_X \in [L_{\n}, 1] $ such that
\begin{align*}
    \t_X = \t \ge (1-\frac{\epsilon}{2})\cdot \t(\pi^*),\;
    \f_X \le \min\{1, \frac{\max\{L_{\n}, \f_X(\pi^*)\}}{1-\epsilon/2}\} \;\forall X\in \mathcal X,
\end{align*}
and for each $X\in \mathcal X$, $M_X[k, \t_X, \f_X] = \mathrm{true}$. Thus, by the bounds we just showed for the precision of such a policy, $\mathrm{precision}(\pi) \ge (1-\epsilon) \cdot \mathrm{precision}(\pi^*)$. Thus, $f_\alpha(\pi) \ge (1-\epsilon) \cdot f_\alpha(\pi^*)$.

As we need to run the DP algorithm for any of the $|\mathcal X|$ groups separately with the specified parameters $L_{\p}, L_{\n}$ and $\bar{\epsilon} = O(\epsilon/k)$, by Lemma~\ref{lem:DP-table-runtime}, the total time of the DP approach is 
\begin{align*}
O(\frac{|{\mathcal X}| k \log^2\frac{1}{L_{\p}} \log^2\frac{1}{L_{\n}}}{\bar\epsilon^4})
&= O(\frac{|{\mathcal X}| k^5 (\epsilon^2 + \log^2\frac{1}{\epsilon}) \log^2\frac{1}{\epsilon}}{\epsilon^4}) \\
&= O(\frac{|{\mathcal X}| k^5 \log^4\frac{1}{\epsilon}}{\epsilon^4})
\end{align*}
\end{proof}

\begin{proof}[Proof of Theorem~\ref{thm:EQ-DP-min}]
First we show that in our setting, in order to guarantee $(1+\epsilon)$-approximations of recall and precision, it suffices to run the described DP and consider estimates of $\t$ (true positive rate) and $\f$ (false positive rate) of form $(1-\bar\epsilon)^i$ for $i\in \mathbb{N}$ in intervals $[L_{\p}, 1]$ and $[L_{\n}, 1]$ respectively, where $\bar\epsilon \le \epsilon/(2k)$. We provide tight bounds for $L_{\p}$ and $L_{\n}$. 

Note that since for any policy $\pi$, $\p_{i, \pi}, \n_{i, \pi}$ are non-decreasing in $i$, it suffices to provide ``large enough'' lowerbounds $L_{\p}$ and $L_{\n}$ for true positive rate and false positive rate in the final stage respectively (i.e., for $\t$ and $\f$). 

\paragraph{Bounding $L_{\p}$.} Consider the policy $\pi_{\bypass}$, which bypasses all the tests in both groups, i.e., $\pi^i_{X0} = \pi^i_{X1} = 1$ for all $i\in [k], X\in {\mathcal X}$. 
%Since $\pi_\bypass$ is an Equal Opportunity policy for the pipeline with $g_\alpha(\pi_\bypass) = 1-\alpha + \alpha/\|q\|_1$ and for any policy $\pi$ has $g_{\alpha}(\pi) \ge 1$, $\pi_\bypass$ is a $(1+\epsilon)$-approximation when $\alpha \le \frac{\epsilon \|q\|_1}{1- \|q\|_1}$.
%any optimal Equal Opportunity policy $\pi^*$ for $g_\alpha$ has recall at least $(1-\alpha)/(1-2\alpha + \alpha/\|q\|_1)$. 
Let $\tau_{\min} = \min_{X\in \mathcal{X}, j\in [k]} \tau^j_{X1}$. Then, by Theorem~\ref{thm:multi-eq-opp-prec}, Opportunity Ratio maximizes the precision and has recall at least $(\tau_{\min})^k$, in the optimal policy $\t \ge (\tau_{\min})^k$ which implies that in our DP with accuracy parameter $\bar\epsilon$ it suffices to set $L_{\p} = (\tau_{\min})^k \cdot (1-\bar\epsilon)^{k-1} \ge \exp(-\epsilon - k\ln (1/\tau_{\min}))$. 

\paragraph{Bounding $L_{\n}$.} 
For each $X\in \mathcal X$, let $\f_X$ denote the false positive rate of the optimal Equal Opportunity policy for group $X$. Similarly, let $\t_X$ denote the positive rate of (i.e., recall) the optimal policy $\pi^*$ for group $X\in \mathcal X$. By Equality of Opportunity property of $\pi^*$, $\t_{X} = \t$ for each $X\in {\mathcal X}$.
Next, we consider the following cases. 

For any sufficiently small $\epsilon >0$, we need to set $L_{\n}$ so that by running the DP with accuracy parameter $\bar\epsilon$, we can approximate both true positive rate and false positive rate of the optimal Equal Opportunity policy within $(1-\epsilon)$-factor of their values. More precisely, we set $L_{\n}$ so that if for each group $X\in {\mathcal X}$ and any pair $(\t_X, \f_X)$ with $t_X \ge {L_{\p}}$, there exists a pair $(\bar{\t}_X, \bar{f}_X)$ such that $\bar{\t}_X \ge (1-\epsilon/2) \t_X$, $\bar{\f}_X \le \min(1, \max(L_{\n}, \f_X)/(1-\epsilon/2))$.
Finally, once the above property holds for all groups $X\in {\mathcal X}$, then for the corresponding policy $\pi$,  $\mathrm{precision}(\pi) > (1-\epsilon) \cdot \mathrm{precision}(\pi^*)$.

Let ${\mathcal X}_1 := \{X\in {\mathcal X} | \f_{X}/(1-\epsilon/2) \ge L_{\n}\}$ and let ${\mathcal X}_2 := \{X\in {\mathcal X} | \f_{X}/(1-\epsilon/2) < L_{\n}\}$. Note that ${\mathcal X} = {\mathcal X}_1 \dot\cup {\mathcal X}_2$. Then,
\begin{align*}
    \frac{\mathrm{precision}(\pi)}{\mathrm{precision}(\pi^*)} 
    &=\big(\frac{\|q\|_1 \cdot \bar{\t}}{\|q\|_1 \cdot \bar{\t} + \sum_{X\in \mathcal X} u_X \cdot \bar{\f}_X}\big) \\ 
    &\quad\;/ \big(\frac{\|q\|_1 \cdot \t}{\|q\|_1 \cdot \t + \sum_{X\in {\mathcal X}}u_X \cdot \f_X}\big) \\
    &\ge \big(\frac{\|q\|_1 \cdot \bar{\t}}{\|q\|_1 \cdot \bar{\t} + \sum_{X\in {\mathcal X}_1} u_X \cdot \bar{\f}_X + \sum_{X\in {\mathcal X}_2} u_X \cdot \bar{\f}_X}\big) \\
    &\quad\;/ \big(\frac{\|q\|_1 \cdot \t}{\|q\|_1 \cdot \t + \sum_{X\in {\mathcal X}_1}u_X \cdot \f_X}\big) \\
    &\ge \big(\frac{\|q\|_1 \cdot (1-\epsilon/2) \t}{\|q\|_1 \cdot \t + \sum_{X\in {\mathcal X}_1} \frac{u_X \cdot \f_X}{1-\epsilon/2} + \sum_{X\in {\mathcal X}_2} u_X \cdot L_{\n}}\big) \\ 
    &\quad\;/ \big(\frac{\|q\|_1 \cdot \t}{\|q\|_1 \cdot \t + \sum_{X\in {\mathcal X}_1}u_X \cdot \f_X}\big)
\end{align*}
Next, we set $L_{\n}$ so that $\|q\|_1\cdot \t + \sum_{X\in {\mathcal X}_2} u_X L_{\n} \le \frac{\|q\|_1\cdot \t}{1-\frac{\epsilon}{2}}$. Since $\t \ge (\tau_{\min})^k$, it suffices to set $L_\n = \frac{\epsilon \|q\|_1\cdot (\tau_{\min})^k}{(2-\epsilon) (1-\|q\|_1)} = \Omega(\epsilon \cdot (\tau_{\min})^k)$. Hence,
\begin{align*}
    & \big(\frac{\|q\|_1 \cdot (1-\epsilon/2) \t}{\|q\|_1 \t + \sum_{X\in {\mathcal X}_1} \frac{u_X  \f_X}{1-\epsilon/2} + \sum_{X\in {\mathcal X}_2} u_X L_{\n}}\big) / \big(\frac{\|q\|_1 \t}{\|q\|_1 \t + \sum_{X\in {\mathcal X}_1}u_X \f_X}\big) \\
    & \ge (1-\frac{\epsilon}{2})^2 > (1-\epsilon).
\end{align*}

Finally, for each $X\in \mathcal X$, we run the DP algorithm for each group with accuracy parameter $\bar\epsilon$. By Lemma~\ref{lem:DP}, the DP algorithm finds a set $\{\t_X = (1-\bar\epsilon)^{i_X}, \f_X = (1-\bar\epsilon)^{j_X}\}_{X\in \mathcal X}$ (and a policy $\pi$ corresponding to these values) where for each $X\in \mathcal X$, $\t_X \in [L_{\p}, 1], \f_X \in [L_{\n}, 1] $ such that $\forall X \in \mathcal{X}$
\begin{align*}\label{eq:tpr-fpr-bounds}
   & \t_X = \t \ge (1-\epsilon/2)\cdot \t(\pi^*) \newline
   & \f_X \le \min(1, \frac{\max(L_{\n}, \f_X(\pi^*))}{1-\epsilon/2}) 
\end{align*}
and for each $X\in \mathcal X$, $M_X[k, \t_X, \f_X] = \mathrm{true}$. Thus, by the bounds we just showed for the precision of such a policy, $1/\mathrm{precision}(\pi) \le (1+\epsilon) / \mathrm{precision}(\pi^*)$. Thus, $g_\alpha(\pi) \le (1+\epsilon) \cdot g_\alpha(\pi^*)$.

As we need to run the DP algorithm for any of the $|\mathcal X|$ groups separately with the specified parameters $L_{\p}, L_{\n}$ and $\bar{\epsilon} = O(\epsilon/k)$, by Lemma~\ref{lem:DP-table-runtime}, the total runtime is $O(\frac{|{\mathcal X}| k \log^2 (1/L_{\p}) \log^2 (1/L_{\n})}{\bar\epsilon^4}) = O(\frac{|{\mathcal X}| k^7 (\log^2(1/\epsilon) + k^2)}{\epsilon^4})$.
\end{proof}

%\section{Missing Proofs of Section \ref{sec-altmodels} \label{altsection-proofs}
%}

\section{Missing Proofs of Section \ref{eodds}}\label{altsection-proofs}
%First recall that the requirement of Equalized Odds mandates equal true positive and false positive rates for all groups, a stronger requirement than simply requiring Equal Opportunity. 
Similarly to Observation~\ref{obs:EOpp-cond}, we can show the following observation for the policies that satisfies the Equalized Odds requirement.
\begin{obsr}\label{obs:EOdd-cond}
For any policy $\pi$ that satisfies the Equalized Odds for a $k$-stage screening process with parameters

$(\{\tau^i\}_{i\in [k]}, \{q_X, u_X\}_{X\in \mathcal X})$, there exists $M$ and $N$ such that for each $X\in \mathcal{X}$,
\begin{align*}
    M &:= \Pi_{i=1}^k \tau^i_{X1} \pi^i_{X1} + (1 - \tau^i_{X1}) \pi^i_{X0} , &&N := \Pi_{i=1}^k \tau^i_{X0} \pi^i_{X1} + (1 - \tau^i_{X0}) \pi^i_{X0}
\end{align*}
\end{obsr}
Note that as computed in Observation~\ref{obsr:interview-eff-formula}, for policy any satisfying the Equalized Odds, the interview efficiency of a policy $\pi$ for a $k$-stage process with parameters $(q, u, \{\tau^i\}_{i\in [k]})$ is $\frac{\|q\|_1 M}{\|q\|_1 M + \|u\|_1 N}$.

In the rest of the section and for the simplicity of the exposition, we assume there are exactly two groups in the population; $\mathcal X = \{A, B\}$. The result for the general setting can be derived similarly.

\begin{thm}\label{thm:ie-upbound}
The interview efficiency of any policy satisfying Equalized Odds for a single-stage screening process with parameters $(q, u, \tau)$ is at most $    \frac{1}{1 + \frac{u_A + u_B}{q_A + q_B} \cdot \max(\frac{\tau_{A0}}{\tau_{A1}}, \frac{\tau_{B0}}{\tau_{B1}})}$.
% \begin{align*}
%     \frac{q_A + q_B}{(q_A + q_B) + (u_A + u_B) \cdot \max(\frac{\tau_{A0}}{\tau_{A1}}, \frac{\tau_{B0}}{\tau_{B1}})}
% \end{align*}
\end{thm}
\begin{proof}
Maximizing the interview efficiency, is equivalent to minimizing $N_{\tau, \pi} / M_{\tau, \pi}$; a minimizer of the inverse ratio is a maximizer of the interview efficiency and vice versa. Moreover, note that by the Minimal Effectiveness Property of the given test (i.e., Eq.~\eqref{eq:minimally-effective}), $N_{\tau, \pi} < M_{\tau, \pi}$. 
\begin{align*}
    &\frac{N_{\tau, \pi}}{M_{\tau, \pi}} = \frac{\tau_{A0} (\pi_{A1} - \pi_{A0}) + \pi_{A0}}{\tau_{A1} (\pi_{A1} - \pi_{A0}) + \pi_{A0}} \geq \frac{\tau_{A0} (\pi_{A1} - \pi_{A0})}{\tau_{A1} (\pi_{A1} - \pi_{A0})} = \frac{\tau_{A0}}{\tau_{A1}} 
    \text{ and } \\
    &\frac{N_{\tau, \pi}}{M_{\tau, \pi}} = \frac{\tau_{B0} (\pi_{B1} - \pi_{B0}) + \pi_{B0}}{\tau_{B1} (\pi_{B1} - \pi_{B0}) + \pi_{B0}} \geq \frac{\tau_{B0} (\pi_{B1} - \pi_{B0})}{\tau_{B1} (\pi_{B1} - \pi_{B0})} = \frac{\tau_{B0}}{\tau_{B1}}.
\end{align*}
In other words, $N_{\tau, \pi} \geq \max(\frac{\tau_{A0}}{\tau_{A1}}, \frac{\tau_{B0}}{\tau_{B1}}) \cdot M_{\tau, \pi}$. Hence, 
\begin{align*}
    \frac{(q_A + q_B) M_{\tau, \pi}}{(q_A + q_B) M_{\tau, \pi} + (u_A + u_B) N_{\tau, \pi}}
    \leq \frac{1}{1 + \frac{u_A + u_B}{q_A + q_B} \cdot \max(\frac{\tau_{A0}}{\tau_{A1}}, \frac{\tau_{B0}}{\tau_{B1}})}
\end{align*}
% \begin{align*}
%     \frac{(q_A + q_B) M_{\tau, \pi}}{(q_A + q_B) M_{\tau, \pi} + (u_A + u_B) N_{\tau, \pi}} 
%     &\leq \frac{q_A + q_B}{q_A + q_B + (u_A + u_B) \cdot \max(\frac{\tau_{A0}}{\tau_{A1}}, \frac{\tau_{B0}}{\tau_{B1}})}
% \end{align*}
\end{proof}

\begin{rem}
Note that we can generalize the result of Lemma~\ref{thm:ie-upbound} to a $k$-stage screening process with multiple groups $\mathcal{X}$. For any $j\in [k]$, let $\rho := \max_{X\in \mathcal{X}} \Pi_{j\in [k]}\frac{\tau_{X0}^j}{\tau_{X1}^j}$.
Any policy that satisfies Equalized Odds requirement at the end of the process (i.e., before the interview stage) has interview efficiency at most $\frac{\|q\|_1}{\|q\|_1 + \sum_{X\in \mathcal{X}}\rho u_X}$. 
To see this, note that similarly to the proof of Theorem~\ref{thm:ie-upbound} we can show that for every group $X\in \mathcal{X}$, $N_X \geq \rho \cdot M_X$.
\end{rem}

\begin{thm}\label{thm:Eodd-vs-Eopp}
Consider a $k$-stage screening process $(q, u, \tau)$ with multiple groups $\mathcal{X}$ whose tests are minimally effective. Let $\pi_{\mathrm{EOdd}}, \pi_{\mathrm{EOpp}}$ denote the interview efficiency maximizing policy that satisfies Equalized Odds and Equal Opportunity at the end of the process respectively. If $\max_{X\in \mathcal{X}} \Pi_{i\in [k]} \frac{\tau^i_{X0}}{\tau^i_{X1}} > \min_{X\in \mathcal{X}} \Pi_{i\in [k]} \frac{\tau^i_{X0}}{\tau^i_{X1}}$, then $\ie(q,u, \tau, \pi_{\mathrm{EOdd}}) < \ie(q,u, \tau, \pi_{\mathrm{EOpp}})$.

In particular, the gap between the interview efficiency of $\pi_{\mathrm{EOdd}}$ and $\pi_{\mathrm{EOpp}}$ can be as large as $\frac{1}{\|q\|_1}-\epsilon$ for any arbitrary $\epsilon >0$.\footnote{Note that the interview efficiency is always at most $1$ and the trivial Equalized Odds policy that bypasses all tests has interview efficiency $q$.}
\end{thm}
\begin{proof}
The proof of the first part directly follows from the interview efficiency of opportunity ratio policy (Theorem~\ref{thm:multi-eq-opp-prec}) and the upper bound for the interview efficiency of Equalized Odds policies (Theorem~\ref{thm:ie-upbound})

For the second part, consider a pipeline in which there exists a $X^*\in \mathcal{X}$ such that for every $X\in \mathcal{X}\setminus X^*$, $\Pi_{i\in [k]} \frac{\tau^i_{X0}}{\tau^i_{X1}} =0$ and $\Pi_{i\in [k]} \frac{\tau^i_{X^*0}}{\tau^i_{X^*1}} =(1-\delta)^k$. Further, for every $X\in \mathcal{X}\setminus X^*$, let $q_{X} = \frac{\gamma}{k}, u_{X} = \frac{1-\gamma-\mu}{k-1}$ and $q_{X^*} = \frac{\gamma}{k}, u_{X^*} = \mu$. Then, it is straightforward to check that $\ie(\pi_{\mathrm{EOpp}}) = \frac{\gamma}{\gamma + \mu\cdot(1-\delta)^k}$ and $\ie(\pi_{\mathrm{EOdd}}) = \frac{\gamma}{\gamma + (1-\gamma)\cdot (1-\delta)^k}$. As we set $\delta, \mu$ to sufficiently small values, $\ie(\pi_{\mathrm{EOpp}})/\ie(\pi_{\mathrm{EOpp}}) = 1/\gamma -\epsilon = 1/\|q\|_1 -\epsilon$.
\end{proof}

%\avnote{We can keep the following is the section is moved to the appendix. Otherwise, it is not interesting enough to be in the main body.}

%In particular, the above remark shows that if we want to satisfy the Equalized Odds, the optimal policy cannot achieve a guarantee better than a policy that in each stage matches the accuracy of the worst performance of the given test over groups $A$ and $B$ (i.e., $\max(\tau_{A0}/\tau_{A1}, \tau_{B0}/ \tau_{B1})$.  

Next, we show the following structure on a non-trivial optimal solution (i.e., one maximizing the interview efficiency). Note that $\pi= \boldsymbol{1}$ or $\pi=\boldsymbol{0}$ are the two trivial solutions satisfying the Equalized Odds for any given test.
\begin{obsr}\label{obr:Eodd-opt-structure}
For any pipeline $(\tau, q,u)$, in any {\em non-trivial} optimal policy $\pi$, $\min(\pi_{A1}, \pi_{A0}, \pi_{B1}, \pi_{B0}) =0$. Moreover, there exists an optimal policy such that $\max(\pi_{A1}, \pi_{A0}, \pi_{B1}, \pi_{B0}) =1$.
\end{obsr}
\begin{proof}
First, note that by the Minimal Effectiveness Property of the given test (i.e., Eq.~\eqref{eq:minimally-effective}), $N_{\tau, \pi} < M_{\tau, \pi}$.

Suppose that $\min(\pi_{A1}, \pi_{A0}, \pi_{B1}, \pi_{B0}) =\epsilon$. This implies that $M_{\tau,\pi} > N_{\tau, \pi} \ge \epsilon$ 
Then, by subtracting $\epsilon$ from all $\pi$ values, the new policy still satisfies the Equalized Odds and it only increases the interview efficiency. Formally, for $\epsilon>0$
\begin{align*}
    \frac{\|q\|_1 \cdot M_{\tau, \pi}}{\|q\|_1 \cdot M_{\tau, \pi} + \|u\|_1\cdot N_{\tau, \pi}} < \frac{\|q\|_1 \cdot (M_{\tau, \pi} - \epsilon)}{\|q\|_1 \cdot (M_{\tau, \pi} - \epsilon) + \|u\|_1\cdot (N_{\tau, \pi} - \epsilon)}
\end{align*}
The above inequality holds since
\begin{align*}
    & N_{\tau, \pi} < M_{\tau, \pi} \\
    & \;\Rightarrow -\|u\|_1\epsilon N_{\tau, \pi} > - \|u\|_1\epsilon M_{\tau, \pi} \\
    & \;\Rightarrow (\|q\|_1 M_{\tau, \pi}^2 - \|q\|_1 \epsilon M_{\tau, \pi} + \|u\|_1 M_{\tau, \pi} N_{\tau, \pi}) - \|u\|_1 \epsilon N_{\tau, \pi} \\ 
    &\quad> (\|q\|_1 M_{\tau, \pi}^2 - \|q\|_1\epsilon M_{\tau, \pi} + \|u\|_1 M_{\tau, \pi} N_{\tau, \pi}) - \|u\|_1 \epsilon M_{\tau, \pi} \\
    &\;\Rightarrow  M_{\tau, \pi} (\|q\|_1 M_{\tau, \pi} + \|u\|_1 N_{\tau, \pi}) - \epsilon (\|q\|_1 M_{\tau, \pi} + \|u\|_1 N_{\tau, \pi}) \\
    &\quad> M_{\tau, \pi}(\|q\|_1 (M_{\tau, \pi} - \epsilon) + \|u\|_1 (N_{\tau, \pi} - \epsilon)) \\
    &\;\Rightarrow  \frac{M_{\tau, \pi} - \epsilon}{\|q\|_1(M_{\tau, \pi} - \epsilon) + \|u\|_1 (N_{\tau, \pi} - \epsilon)}  \\
    &\quad> \frac{M_{\tau, \pi}}{\|q\|_1 M_{\tau, \pi} + \|u\|_1 N_{\tau, \pi}} \;\rhd\|q\|_1 (M_{\tau, \pi} - \epsilon) + \|u\|_1  (N_{\tau, \pi} - \epsilon) > 0
\end{align*}
In particular, this implies that in any optimal policy, \[ \min(\pi_{A1}, \pi_{A0}, \pi_{B1}, \pi_{B0}) = 0.\]

The second part of the statement follows simply from the fact that if we multiply all $\pi$ values by a constant $c>1$ so that they remain feasible (i.e., none of $\pi$ values goes above one), the interview efficiency of the policy $c\pi$ and the policy $\pi$ are the same. 
\end{proof}
Note that though it seems counter-intuitive, it might be the case
\[ \pi_{A0} = \argmax(\pi_{A1}, \pi_{A0}, \pi_{B1}, \pi_{B0})\]
and/or 
\[ \pi_{A1} = \argmin(\pi_{A1}, \pi_{A0}, \pi_{B1}, \pi_{B0}).\]

%% file: fairscreen_appendix.tex
\section{An FPTAS Algorithm for Screening Processes with Same Policy for All Groups} \label{sec:single-policy-app}
% The first, and, of practical relevance is ensuring fair behavior when there are \textit{not} multiple classifiers fine-tuned for each group. 
% While directly incorporating the the demographic feature into the data representation has been proposed in the fairness in machine learning literature \cite{dwork2012fairness}, in alternative words, this could mean giving two individuals of different demographic groups  different predictions or decisions even if 
% every other feature in their corresponding data point is identical. In some regulatory regimes, this fairness-through-awareness may be illegal or problematic, even when intended to ensure equitable treatment. 

% In our setting,  if we are constrained to follow \textit{group-blindness}, there be would only one set of tests and one ordering of the tests that all applicants are tested on. Analogously to the previous setting, the action space of algorithm remains modifying the promotion probabilities, but we now only have one set of policies to modify. 

Here, we devise a slightly different DP algorithm. Instead of running the DP algorithm for each group separately (as in Section~\ref{sec:fptas}), we run a single DP algorithm for all groups simultaneously. Hence, all policies $\{\pi_X\}_{X \in {\mathcal X}}$ are the same. In our DP approach, we use the same discretization technique and only consider powers of $(1-\epsilon)$.

\paragraph{Solving the DP} 
Consider the first level, $i=1$.
For any given parameters $\{j_{X,0}, j_{X, 1}\}_{X\in \mathcal X}$, where for each group $X\in \mathcal X$, $j_{X,0}\in [0, L_{\n}]$ and $j_{X,1} \in [0, L_{\p}]$, $M[1,\{j_{X0}, j_{X1}\}_{X\in {\mathcal X}}]=\mathrm{true}$ iff the following has a feasible solution.
\begin{align}\label{eq:base-rule-single} 
    \tau^1_{X0} x + (1- \tau^1_{X0}) y &\leq (1-\epsilon)^{j_{X0}} \text{ and}, \nonumber\\
    \tau^1_{X1} x + (1- \tau^1_{X1}) y &\geq (1-\epsilon)^{j_{X1}} \quad \forall X\in {\mathcal X}
\end{align}

Next, we describe the update rule for $i>1$. For any $X\in \mathcal X$, $\n_X \in [0, \ell_\n]$ and $\p_X\in [0, \ell_\p]$,
\begin{align*}
    &M[i+1, \{\p_X, \n_X\}_{X \in {\mathcal X}}]  \\
    &= \bigvee_{\{j_{X1}, j_{X0}\}_{X\in {\mathcal X}} \in {\mathcal F}_{i+1}} M[i, \{{\p_X}-{j_{X1}},  {\n_X}-{j_{X0}}\}_{X\in {\mathcal X}}]
\end{align*}
where ${\mathcal F}_{i+1}$ is the set of $\{j_{X1} \le \p_X, j_{X0}\le \n_X\}_{X\in {\mathcal X}}$ for which the following system of linear inequalities has a feasible solution
$\forall X \in \mathcal{X}$
\begin{align}\label{eq:update-rule-single}
    \tau^{i+1}_{X1} x+ (1-\tau^{i+1}_{X1}) y\ge (1-\epsilon)^{j_{X1}},
    \tau^{i+1}_{X0} x+ (1-\tau^{i+1}_{X0}) y \le (1-\epsilon)^{j_{X0}}
\end{align}
\begin{lem}\label{lem:single-policy-DP}
For any $i\in [k]$, if there exists an Equal Opportunity policy $\pi$ treating all groups similarly, with true positive rate $\t_{X,i} \ge L_{\p}/(1-\epsilon)^{i-1}$, false positive rate $\f_{X,i}$ for $X\in {\mathcal X}$, then there exist $\{j_{X1}, j_{X0}\}_{X\in {\mathcal X}}$ such that $M[i, \{j_{X1}, j_{X0}\}_{X\in {\mathcal X}}] = \mathrm{true}$, where for each $X\in {\mathcal X}$, $(1-\epsilon)^{j_{X1}} \ge \t_{X,i} \cdot (1-\epsilon)^{i-1}$ and $(1-\epsilon)^{j_{X0}} \le \min\{ 1, \max\{ L_{\n}, \f_{X,i}\}/ (1-\epsilon)^{i-1}\}$.

In other words, if the policy $\pi$ exists then the DP approach finds a policy with true positive rate at least $(1-\epsilon)^{j_{X1}}$ and false positive rate at most $(1-\epsilon)^{j_{X0}}$ for each $X\in \mathcal X$.
\end{lem}

\begin{proof}
The proof is by induction. For the base case ($i=1$), let $\t_{X,1}$ and $\f_{X,1}$ denote the true positive rate and the false positive rate of $\pi$ by the end of level $1$ for each group $X\in \mathcal X$. 
The existence of $\pi$ guarantees that the system of inequalities Eq.~\eqref{eq:base-rule-single} with $(j_{X0} = \lfloor \log_{1-\epsilon} \f_{X,1} \rfloor, j_{X1} = \lceil \log_{1-\epsilon}\t_{X,1} \rceil \leq \ell_{\p})$ has a feasible solution.  
More precisely, by setting $(x_X = \pi_{X1}, y_X = \pi_{X0})$, $\forall X\in \mathcal X$, 
\begin{align*} 
    \tau^1_{X0} x_X + (1- \tau^1_{X0}) y_X &= \f_{X1} \le (1-\epsilon)^{\lfloor \log_{1-\epsilon} \f_{X1} \rfloor} = (1-\epsilon)^{j_{X0}} \\ 
    \tau^1_{X1} x_X + (1- \tau^1_{X1}) y_X &= \t_{X1} \geq (1-\epsilon)^{\lceil \log_{1-\epsilon} \t_{X1} \rceil} = (1-\epsilon)^{j_{X1}}
\end{align*}
Next, we consider $i>1$ and we assume that the claim holds for all values $i' < i$. For each $X\in \mathcal X$, let $M_{X,i} := \tau^i_{X1}\pi^i_{X1} + (1 - \tau^i_{X1}) \pi^i_{X0}$ and $N_{X,i} := \tau^i_{X0}\pi^i_{X1} + (1-\tau^i_{X0})\pi^i_{X0}$. Note that for each $X\in \mathcal X$, $\t_{X,i} = \t_{X,i-1} \cdot M_{X,i}$ and $\f_{X,i} = \f_{X,i-1} \cdot N_{X,i}$.

By the induction hypothesis and considering the first $i-1$ levels in the pipeline, since $\t_{X,i-1} \ge \t_{X, i} \ge L_{\p}/(1-\epsilon)^{i-1} > L_{\p}/(1-\epsilon)^{i-2}$ and $\f_{X,i-1} \ge \f_{X, i}$, there exist $j'_{X1}\in[0, L_{\p}]$ and $j'_{X0}\in [0, L_{\n}]$ such that $M[i-1, \{j'_{X1}, j'_{X0}\}_{X\in \mathcal X}] = \mathrm{true}$ and $(1-\epsilon)^{j'_{X1}} \ge \t_{X,i-1} \cdot (1-\epsilon)^{i-2}$ and $(1-\epsilon)^{j'_{X0}} \le \min\{1, \max\{L_{\n}, \f_{X,i-1}\}/(1-\epsilon)^{i-2}\}$. 
More precisely, the algorithm finds a policy $\bar{\pi}$ with true positive rate at least $(1-\epsilon)^{j'_{X1}}$ and false positive rate at most $(1-\epsilon)^{j'_{X0}}$ for each $X\in \mathcal X$.

Next, for each $X\in \mathcal X$, by setting $(\bar{\pi}^i_{X1} = \pi^i_{X1}, \bar{\pi}^i_{X0} = \pi^i_{X0})$ and $(j_{X1} :=\argmin_{j} \{(1-\epsilon)^j \le \t_{X,i}(\bar\pi)\}, j_{X0} := \argmax_{j} \{(1-\epsilon)^j \ge \f_{X,i}(\bar\pi)\})$,  
\begin{align*}
    (1-\epsilon)^{j_{X1}} 
    &> (1-\epsilon)\cdot \t_{X,i}(\bar \pi) \\
    &= (1-\epsilon)\cdot \t_{X,i-1}(\bar \pi) \cdot M_{X,i} &&\rhd\text{by definition of $j_{X1}$}\\
    &\ge (1-\epsilon)\cdot (1-\epsilon)^{j'_{X1}} \cdot M_{X,i} &&\rhd\text{$\t_{X, i-1}(\bar\pi) \ge (1-\epsilon)^{j'_{X1}}$}\\ 
    &\ge \t_{X, i-1} \cdot (1-\epsilon)^{X, i-1} \cdot M_{X, i} &&\rhd\text{induction hypothesis}\\
    &= \t_{X, i} \cdot (1-\epsilon)^{i-1}. 
\end{align*}
Similarly,
\begin{align*}
   (1-\epsilon)^{j_{X0}} 
   &< \min\{1, \frac{\f_{X,i}(\bar\pi)}{1-\epsilon}\} \\
   &= \min\{1, N_{X,i} \cdot \frac{\f_{X, i-1}(\bar \pi)}{1-\epsilon}\} \;\rhd\text{by definition of $j_{X0}$}\\
   &\le \min\{1, (1-\epsilon)^{j'_{X0}} \cdot \frac{N_{X,i}}{1-\epsilon}\} \;\rhd\text{$\f_{X, i-1}(\bar\pi) \le (1-\epsilon)^{j'_{X0}}$} \\
   &\le \min\{1, \frac{\max\{L_{\n}, \f_{X, i-1}\}}{(1-\epsilon)^{i-2}} \cdot \frac{N_{X,i}}{1-\epsilon}\} \;\rhd\text{ induction hyp.}\\ 
   &\le \min\{1, \frac{\max\{L_{\n}, \f_{X,i}\}}{(1-\epsilon)^{i-1}}\}
\end{align*}
which completes the proof.
\end{proof}

\begin{lem}\label{lem:single-policy-DP-runtime}
For an accuracy parameter $\epsilon$ and lowerbounds on the false positive rate, $L_{\n}$, and the true positive rate, $L_{\p}$, the (single policy) DP algorithm runs in time $O(\frac{k \log^{2|{\mathcal X}|}(1/L_{\p}) \log^{2|{\mathcal X}|}(1/L_{\n})}{\epsilon^{4|{\mathcal X}|}})$ and finds a policy $\pi$ with true positive rate at least $(1-\epsilon)^{k-1} \cdot \t_X$ and false positive rate at most $\min\{ 1, \max\{ L_{\n}, \f_X\}/ (1-\epsilon)^{k-1}\}$ for each $X\in \mathcal X$.
\end{lem}
\begin{proof}
The size of table is $O(k \ell_{\p}^{|{\mathcal X}|} \ell_{\n}^{|{\mathcal X}|})$ and updating each entry in the table takes $O(\ell_\p^{|{\mathcal X}|} \ell_\n^{|{\mathcal X}|})$. Hence, the total runtime to compute all entries in the DP table is 
 \[ O(k\ell^{2|{\mathcal X}|}_\p \ell^{2|{\mathcal X}|}_\n) = O(\frac{k \log^{2|{\mathcal X}|}(1/L_{\p}) \log^{2|{\mathcal X}|}(1/L_{\n})}{\epsilon^{4|{\mathcal X}|}}).\]

Now we apply the DP approach and by Lemma~\ref{lem:single-policy-DP}, the solution returned by the algorithm has the true positive rate and the false positive rate satisfying the guarantee of the statement.
\end{proof}
\paragraph{Implications of DP} Here, similarly to Section~\ref{sec:fptas}, we present FPTAS algorithms for the single policy setting with various pipeline efficiency objective using the modified DP approach described above when the number of different protected groups in the population is a fixed constant; $|\mathcal{X}| = O(1)$. 
\begin{thm}\label{thm:EQ-DP-max-single}
Consider a $k$-stage screening process with parameters $(u, q, \tau, \mathcal X)$ where $|\mathcal{X}| = O(1)$. For any policy $\pi$, let $f_{\alpha}(\pi) = \mathrm{recall}(\pi) + \alpha \cdot \mathrm{precision}(\pi)$ where $\alpha > 0$. Given an accuracy parameter $\epsilon$, there exists an FPTAS that runs in time $O(\frac{k^{4|{\mathcal X}|} \log^{2|{\mathcal X}|}(1/\epsilon)}{\epsilon^{4|{\mathcal X}|}})$ and finds an Equal Opportunity policy $\pi$ treating all groups similarly such that $f_{\alpha}(\pi) \ge (1-\epsilon)f_{\alpha}(\pi^*)$ where $\pi^*$ maximizes $f_\alpha$ over Equal Opportunity policies treating all groups similarly. 
\end{thm}
\begin{thm}\label{thm:EQ-DP-min-single}
Consider a $k$-stage screening process with parameters $(u, q, \tau, \mathcal X)$ where $|\mathcal{X}| = O(1)$. For any policy $\pi$, let $g_{\alpha}(\pi) = 1/\mathrm{recall}(\pi) + \alpha / \mathrm{precision}(\pi)$ where $\alpha > 0$. Given an accuracy parameter $\epsilon$, there exists an FPTAS that runs in time $O(\frac{k^{4|{\mathcal X}|} \log^{2|{\mathcal X}|}(1/\epsilon)}{\epsilon^{4|{\mathcal X}|}})$ and finds an Equal Opportunity policy $\pi$ treating all groups similarly such that $g_{\alpha}(\pi) \le (1+\epsilon)g_{\alpha}(\pi^*)$ where $\pi^*$ minimizes $g_\alpha$ over Equal Opportunity policies treating all groups similarly.
\end{thm}
The proof of above theorems are identical to Theorem~\ref{thm:EQ-DP-max} and Theorem~\ref{thm:EQ-DP-min}.

\section{Additional details in Linear Combination Counter Examples \label{appendix-linear-combination-counter}}
%\avnote{Revise this section. My suggestion: (1) Before F.1., state the goal of this section. (2) Add a theorem statement and prove it formally (if the proof has multiple part, instead of having subsections, use lemmas.)}
In this section, we show that one cannot ``locally score'' tests when determining the optimum policy (the policy that maximizes a linear combination of precision and recall).  Specifically, we give a setting with three levels of tests $t_1, t_2, t_3$ such that if only the first two levels $t_1$ and $t_2$ are available, then the optimal solution is to use $t_1$ and bypass $t_2$, but if $t_3$ is also available then the optimal solution is to bypass $t_1$ and use $t_2$ and $t_3$.  Therefore, the question of how to best use two tests may depend on what tests are available at other levels.
Note that in this example there is only one group and we do not have fairness constraints. 

First, we show the following useful property of optimal policies for a pipeline where the first level has test statistics $(1/2, 0)$ and all other levels have test statistics $(1-\delta, 1/2)$.
\begin{lem}\label{lem:prop-single}
In any $k$-stage pipeline where the first stage has test statistics $(1/2, 0)$ and the rest of the stages have tests with statistics $(1- \delta, 1/2)$, the optimal policy is of the form $(1, \pi^1_0), \cdots, (1, \pi^k_0)$.
% \begin{itemize}
%     \item $(1, \pi^1_0), \cdots, (1, \pi^k_0)$ where $\pi^1_0 >0$, or
%     \item $(1, 0), (1, 1)\cdots, (1,1)$, i.e., fully use the first test and bypass the rest.
% \end{itemize}
\end{lem}
\begin{proof}
By Lemma~\ref{lem:pi-cond}, if the False Positive rate is non-zero, in the optimal policy, for every $i\in [k]$, $(1 - \pi^i_1) \pi^i_0 = 0$. Next, we show that in this setting with only one group, for every $i\in [k]$, $\pi^i_1 = 1$. Suppose that there exists a level $i\in [k]$ such that $\pi^i_0 = 0$. Then, if $\pi^i_1 <1$, by increasing $\pi^i_1$ to $1$, the True Positive rate and False Positive rate increase by the same factor. Therefore, the precision remains unchanged and the recall increases; hence, the pipeline efficiency strictly increases.

Next, we consider the case where the optimal policy has precision one (i.e., its False Positive is zero). In any such policy, $\pi^1_0 = 0$. Note that once the precision is $1$, the optimal policy maximizes recall. Hence, the optimal policy is to fully use $t_1$ ($\pi^1_1 =1, \pi^1_0 =0$) and bypass the rest of tests (for every $1<i\le k$, $\pi^i_1 = \pi^i_0 = 1$). 
\end{proof}

\begin{thm}
\label{thmlinearexample}
When the objective is to maximize a linear combination of precision and recall 
%with combinations weights $\alpha, \beta > 0 $ 
in a multi-stage screening process, there exist test parameters $T$ and base rate $p$ such that the maximal score policy 
%for some of the tests 
switches 
%(e.g. two tests $j$ and $k$ with $\pi_{j} = (1,0)$ and $\pi_{k} = (1,1)$) 
when more tests become available.  Specifically, when only tests $t_1$ and $t_2$ are available, the optimal policy is to use $t_1$ and bypass $t_2$ $((1,0), (1,1))$, but if test $t_3$ is also available, the optimal policy is to bypass $t_1$ and use $t_2$ and $t_3$ $((1,1), (1,0), (1,0))$.
\end{thm}
\begin{proof}
Consider base rate $p= P(x=1) = 1/2$ and test $t_1=(\tau_1, \tau_0) = (1/2,0)$ and test $t_2 =t_3= (1-\delta, 1/2)$. 
Let $\delta=\frac{1}{100}$. The linear objective function is $f(\pi) = \mathrm{recall}(\pi) + 2 \cdot \mathrm{precision}(\pi)$. 
Next, we consider two cases: (1) $k=2$ and (2) $k=3$.
\paragraph{Case 1: Two test ($k=2$).} By Lemma~\ref{lem:prop-single}, the optimal policy is of form $((1, \pi^1_0), (1, \pi^2_0))$. By numerical analysis\footnote{Using WolframAlpha.}, the local optimum policies (w.r.t.~$f$) are $((1,0), (1,1))$ and $((1,1), (1,0))$. Next, we compute the score of these two policies: $f((1,0), (1,1)) = 2.5$ and $f((1,1), (1,0)) < 2.32$. Hence, in this case, the optimal policy is to fully use $t_1$ and bypass $t_2$, i.e., $((1,0), (1,1))$.

% either to fully use $t_1$ and bypass $t_2$ (i.e., $((1, 0), (1, 1))$) or to pick $((1, \pi^1_0), (1, \pi^2_0))$ where $\pi^1_0 >0$. 
%\AB{[AB: Isn't the first option a special case of the second option?  Or should $\pi_0^1 =1$ here?]} \AV{Sure, I added the condition $\pi^1_0 > 0$ to make the cases disjoint.}
% \begin{itemize}
%     \item {\bf $((1, 0), (1, 1))$.} $f((1, 0), (1, 1)) = 1/2 + 2\cdot 1 = 5/2$.
%     \item {\bf $((1, \pi^1_0), (1, \pi^2_0))$ where $\pi^1_0 >0$.} By numerical analysis\footnote{Using WolframAlpha.}, the maximizer of $f$ of this form is obtained by the policy $((1,1), (1,0))$ and $f((1,1), (1,0))<2.32$. 
%     %\AB{[AB: According to my calculations, this policy has recall $=1-\delta$ and precision $=\frac{1-\delta}{(1-\delta)+1/2} < 2/3$, so its overall score is less than $2 + 1/3$, which is smaller than the above $2 + 1/2$.]} \AV{You are right. I fixed it.}
% \end{itemize}
\paragraph{Case 2: Three tests ($k=3$).} Similarly to the previous case, the optimal policy for the given pipeline efficiency objective is of form $((1, \pi^1_0), (1, \pi^2_0), (1, \pi^3_0))$.
By numerical analysis, the local optimum policies (w.r.t.~$f$) are $((1,0), (1,1), (1,1))$ and $((1,1), (1,0), (1,0))$. 
Next, we compute the score of these two policies: \[f((1,0), (1,1), (1,1)) = 2.5\text{ and }((1,1), (1,0), (1,0)) > 2.57.\] This time, the optimal policy is to bypass $t_1$ and fully use $t_2, t_3$, i.e. $((1,1), (1,0), (1,0))$.

% \begin{itemize}
%     \item {\bf $((1, 0), (1, 1), (1,1))$.} $f((1, 0), (1, 1), (1,1)) = 1/2 + 2\cdot 1 = 5/2$.
%     \item {\bf $((1, \pi^1_0), (1, \pi^2_0), (1, \pi^3_0))$ where $\pi^1_0 >0$.} By numerical analysis, the maximizer of $f$ of this form is obtained by the policy $((1,1), (1,0), (1,1))$ and $f((1,1), (1,0), (1,1))>2.57$. 
% \end{itemize}

Therefore, while in the first setting ({\em only $t_1$ and $t_2$ are available}) the optimal policy is to fully use $t_1$ and bypass $t_2$, once $t_3$ becomes available, the optimal policy changes to bypass $t_1$ and fully use $t_2$ and $t_3$.  
\end{proof}

%% file: 0_abstract.tex
We now shift topics somewhat; our last screening work focused on screening and 
algorithmic fairness. Now we will focus on screening and strategic behavior.
Specifically, we initiate the study of strategic behavior in screening processes with \emph{multiple} classifiers. 
We focus on two contrasting settings: a ``conjunctive'' setting in which an individual must satisfy all classifiers simultaneously, and a sequential setting in which an individual to succeed must satisfy classifiers one at a time. In other words, we introduce the combination of \emph{strategic classification} with screening processes. We show that sequential screening pipelines exhibit new and 
surprising behavior where individuals can exploit the sequential ordering of the tests to ``zig-zag'' between classifiers without having to simultaneously satisfy all of them. We demonstrate 
%how this can allow
an individual can 
%successfully 
obtain a positive outcome using a limited manipulation budget even when %very 
far from the intersection of the positive regions of every classifier. %deployed simultaneously. 
Finally, we consider 
%the point of view of 
a learner whose goal is to design a sequential screening process that is robust to such manipulations, and provide a construction for the learner that optimizes a natural objective.
We also briefly discuss some of the fairness implications of this work, but note that there are substantial open fairness research directions based on our model.

%% file: 1_intro.tex
%Many of these high stakes decisions involve a screening process, in which an applicant must generally pass multiple tests sequentially rather than a single classifier. 
%For example, many hiring processes involve multiple rounds of interviews; university admissions can involve a combination of standardized tests, essays, or interviews.
Screening processes~\citep{arunachaleswaran2022pipeline, blum2022multi, cohen2019efficient} involve evaluating and selecting individuals for a specific, pre-defined purpose, such as a job, educational program, or loan application. 
% Screening processes \cite{arunachaleswaran2022pipeline, blum2022multi, cohen2019efficient} are a way of making a decision using multiple classifiers or tests.
These screening processes are generally designed to identify which individuals are qualified for a position or opportunity, often using multiple sequential classifiers or tests. For example, many hiring processes involve multiple rounds of interviews; university admissions can involve a combination of standardized tests, essays, or interviews. They have substantial practical benefits, in that they can allow a complex decision to be broken into a sequence of smaller and cheaper steps; this allows, for example, to split a decision across multiple independent interviewers, or across smaller and easier-to-measure criteria and requirements.%, perhaps due to distributed control.
%\footnote{E.g. responsibility for a hiring decision is split across multiple independent interviewers.}  

%The early stages could use cheap tests (e.g.,  resume screening) to eliminate clearly unqualified individuals, while in the later stages more expensive steps allow finer gradations between individuals.
%Alternatively, perhaps the Screener's utility function depends on a mix of skills and using a sequence of classifiers where each classifier is fine-tuned to assess a specific skill allows for more precision. 
%\juba{commented out the part about how the process is set up/early stages being cheaper/etc. We don't really talk about this anymore so not too relevant to the current version of the paper}

Many of the decisions made by such screening processes are high stakes. For example, university admissions can affect an individual's prospects for their entire life. Loan decisions can have a long-term (sometimes even inter-generational) effect on a family's wealth or socio-economic status. When these decisions are high stakes, i.e. when obtaining a positive outcome is valuable or potentially life-changing or obtaining a negative outcome can be harmful, individuals may want to manipulate their features to trick the classifier into assigning them a positive outcome. 

In machine learning, this idea is known as strategic classification, and was notably introduced and studied by~\cite{bruckner2011stackelberg,hardt2016strategic}.
The current work aims to incorporate strategic classification within screening processes, taking a departure from the classical point of view in the strategic classification literature that focuses on a single classifier (see related work section). %~\cite{kleinberg2020classifiers,dong2018strategic,hu2019disparate,shavit2020causal,harris2021stateful,chen2020learning,zhang2021incentive,bechavod2021gaming}. 
%\ali{To Juba (or Lee or Kevin), Pls add the missing citations.}\juba{we don't need to recite here, we put all the cites upfront in related work. Maybe a forward reference. }

%\juba{moved this part down. I think this was jumping the gun, too early, and coming a bit out of nowhere. This is the novel modeling element so will highlight it by having it last.}
The key novel idea of our model of {\em strategic screening processes (or pipelines)}, compared to the strategic classification literature, comes from the fact that i) an individual has to pass and manipulate her way through \emph{several} classifiers, and ii) that we consider \emph{sequential} screening pipelines.

In a sequential screening pipeline, once an individual (also called \emph{Agent}) has passed a test or stage of this pipeline, she can ``forget'' about the said stage; whether or not she passes the next stage depends \textit{only on her performance in that stage}. For example, a job candidate that has passed the initial human resources interview may not need to worry about convincing that interviewer, and can instead expand her effort solely into preparing for the first technical round of interviews.
Alternatively, imagine a student `cramming' for a sequence of final exams, where one has a finite capacity to study that is used up over a week of tests. One wants to achieve a minimum score on each test, with a minimum of effort, by studying in between each test.

Our goal in this work is to examine how considering a pipeline comprised of a sequence of classifiers affects and modifies the way a strategic agent manipulates her features to obtain a positive classification outcome, and how a learner (which we primarily call the \emph{Firm}) should take this strategic behavior into account to design screening pipelines that are robust to such manipulation. In our model, 1) the firm deploys a sequential pipeline of classifiers, 2) the agent is given full knowledge of the pipeline and computes their optimal manipulation strategy, then 3) the agent goes through the screening pipeline and implements said optimal manipulation strategy in order to pass the tests sequentially, one at a time.
%, manipulating in between each step, with overall cost being the sum of the manipulations.

%\ali{To Kevin, maybe better to include the ``cramming for test'' example Avrim suggested here.}

%Our goal in this work is to first examine how, given a fixed pipeline, a strategic individual would manipulate their features to implement a Zig-Zag Strategy and how to modify a pre-existing pipeline to be more robust to these strategic manipulations. %at a cost to true positives who must now manipulate to pass.

We make a distinction between the following two cases: 1) the firm deploys its classifiers sequentially which we refer to as a {\em sequential screening process}; 2) the firm deploys a single classifier whose positive classification region is the intersection of the positive regions of the classifiers that form the pipeline which we sometimes refer to as \emph{simultaneous (or conjunctive) testing}---this single classifier is basically the {\em conjunction} or intersection of classifiers from the pipeline. The former corresponds to a natural screening process that is often used in practice and for which we give our main results, while the latter is primarily considered as a benchmark for our results for the sequential case. %\juba{eeeeeh or it could be a complex classifier, since nobody really uses just linear rules in real-life? A conjunction of linear classifiers is just a decision-tree/logical formula? Not sure this makes sense}\saeed{removed the conjunction case being "hypothetical". have a look.}

%\juba{math model is too soon, commenting out}
%This is modeled by thinking of the pipeline consisting of $k$ stages with classifiers $f^1, \dots f^k$, with $f^i : \rightarrow \mathcal{X} \rightarrow \{0, 1 \}$, where examples or individuals are represented as vectors in $\mathcal{X}$ with feature values.

%\juba{commented out/moved the part about this sounding like it should be harder to game. I think this goes with the contributions and is not part of the intro}

%In particular, it seems like screening processes should be relatively robust to manipulations because getting the desirable outcome of the screening process requires passing every stage, so if the probability of a true negative somehow passing a stage is $p < \frac{1}{2}$ and there are $k$ stages, $\frac{1}{2^{k}}$ decays quickly!

\paragraph{Our Contributions.}

%One may imagine that a sequential screening process is harder for an individual to manipulate his way through, since this requires passing multiple stages and tests before obtaining a positive classification outcome. However, we show the opposite result. Depending on the geometric structure of the tests, we exhibit a strategy such that individuals can exploit the sequential nature of the screening process and move through the whole pipeline despite being very far from the intersection of all positive regions of each classifier.

%\juba{I don't know if I like the commented out version. We are comparing apples and oranges. The implicit comparison is a single static classifier versus a sequence of many classifiers when we say that the sequential case is harder. I would rather say:}

\begin{figure}[t]
\centering
\includegraphics[scale=.75]{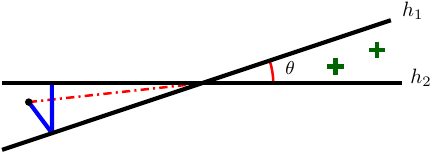}
\caption{Suppose the agent is the disqualified (i.e., placed in the negative region of the conjunctions of $h_1, h_2$) point.
A trivial manipulation strategy is to use the shortest {\em direct} path to the positive region, which is the dashed red path. However, the agent may also first manipulate slightly to pass $h_1$, then manipulate minimally again to pass $h_2$, as depicted with the blue solid path. This is what we call a zig-zag strategy.}
\label{fig:zigzagjumps}
\centering
\end{figure}

We show a perhaps surprising result: an agent can exploit the sequential nature of the screening process and move through the whole pipeline even when she started far from the intersection of the positive classification regions of all classifiers. 
In other words, the sequentiality of screening processes can \emph{improve} an agent's ability to manipulate her way through multiple classifiers compared to the simultaneous screening. We name the resulting set of strategies for such an agent in the sequential case \textit{``zig-zag" strategies}. In other words, whenever the agent does not manipulate straight to a point that is classified as positive by the conjunction of all classifiers, we call it a zig-zag strategy. An example of such a strategy that zig-zags between two classifiers is provided in Figure \ref{fig:zigzagjumps}. 
%\juba{writing below too informal}
% Now we will look at an informal schematic of a successful Zig-Zag strategy, where
%In this example, the Agent evades the classifiers despite being very far from the conjunction of all the classifiers. 
%\juba{Very far from the "positive region of"? It's technically close to all classifiers so we want to be very precise here}

In Figure \ref{fig:zigzagjumps}, since there is a small angle $\theta$ between the two tests, an agent at the bottom of the figure can zag right and then left as shown by the blue lines.  %it This is a Zig-Zag strategy.
In this case, the agent is classified as positive in every single step, and by making $\theta$ arbitrarily small, will have arbitrarily lower total cost (e.g., the cumulative $\ell_2$ distance) compared to going directly to the intersection point of the classifiers. We provide concrete classifiers and an initial feature vector for such a case in Example \ref{exp:zigzag}.
% Label the cost of the Zig-Zag solution as $c_{Zag}$, while define $c_{cjn}$ as in Equation \ref{eq:cost all}.
% \begin{equation}
% \label{eq:cost all}
% \begin{aligned}
% c_{all} = \min_{z} \quad & ||z-x||_2 \\
% \textrm{s.t.} \quad & w_{i}^{T} \geq b_i \quad \forall i \in [k]
% \end{aligned}
% \end{equation}

\begin{figure}[ht]
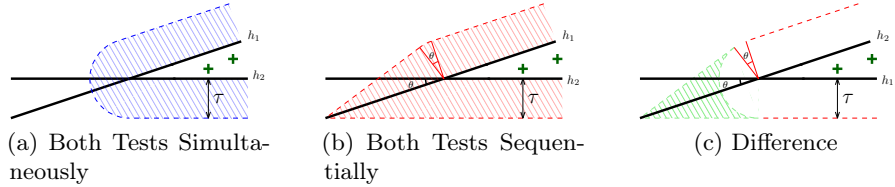

  \centering
  \subfloat[Both Tests Simultaneously]{{\includegraphics[width=0.27\textwidth]{2d-1.pdf}}}
\qquad
\subfloat[Both Tests Sequentially]{{\includegraphics[width=0.27\textwidth]{2d-2.pdf}}}
\qquad
\subfloat[Difference]{{\includegraphics[width=0.27\textwidth]{2d-diff.pdf}}}
\caption{Each agent has a manipulation budget of $\tau$ and the cost function is $\ell_2$ distance. Then,
  ($a$) shows the region of agents who afford to manipulate their feature vectors to pass both tests simultaneously, ($b$) shows the region of agents who afford to manipulate their feature vectors to pass the tests sequentially (i.e, first $h_1$, then $h_2$), and ($c$) shows the difference in these two regions.}
  \label{fig:2d-manipulation}
\end{figure}

In fact, in Section \ref{sec:example}  we show that for a given point, as $\theta$ goes to zero, the ratio between the total cost of the zig-zag strategy and the cost of %$c_{all}/c_{Zag}$
going directly to the intersection can become arbitrarily large.
As we assume that conjunction of the classifiers captures the objective of the firm, using a pipeline can allow more disqualified people to get a positive outcome by manipulating their features. We show this in Figure \ref{fig:2d-manipulation}: This figure shows the region of the agents space that can successfully manipulate to pass two linear tests in the two-dimensional setting, given a budget $\tau$ for manipulation. As shown by the figure, individuals in the green region of Figure \ref{fig:2d-manipulation}.c can pass the tests in the sequential setting but would not be able to do so if they had to pass the tests simultaneously.

%can allow quite unqualified individuals to evade the testing process.
%%

We further show how the optimal zig-zag strategy of an agent can be obtained computationally efficiently via a simple convex optimization framework in Section \ref{sec: algo} and provide a closed-form characterization of this strategy in the special case of $2$-dimensional features and a pipeline of exactly two classifiers in Section  \ref{subsec:closed-form}. 

In Section \ref{subsec-monotone} we consider a ``monotonicity" condition under which, agents prefer to use the simple strategy which passes all classifiers simultaneously in a single move and does not zig-zag between classifiers. 

 Finally, %in Section~\ref{sec:hardness}, we provide an NP-hardness result for the firm's problem of designing a sequential screening pipeline that maximizes accuracy subject to bounded strategic manipulation budget of agents. In response, 
 in Section~\ref{sec:conservative-defense}, we exhibit a defense strategy that maximizes true positives subject to not allowing any false positives. Interestingly, we show that under this strategy, deploying classifiers sequentially allows for a higher utility for the firm than using a conjunction of classifiers. %, despite our results of Section \ref{sec: algo} showing that sequentiality seems to play in favor of agents.
 %\juba{is this really what the result says? Double check once result is finalized}

\begin{figure}
\centering
\subfloat[A photograph of the apparatus used]{\label{fig:capparatus}
\centering
\includegraphics[width=0.29\textwidth]{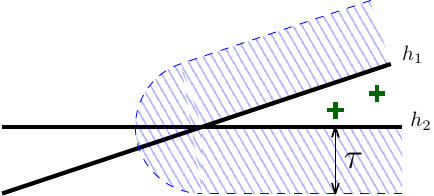}
}
\hfill
\subfloat[A diagram of the apparatus sited inside the evacuated chamber]{\label{fig:cdiagram}
\centering
\includegraphics[width=0.29\textwidth]{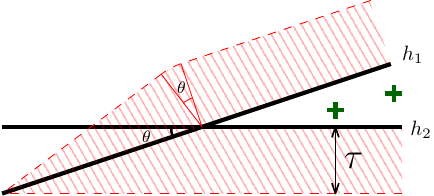}
}
\hfill
\subfloat[A diagram of the apparatus sited inside the evacuated chamber]{\label{fig:cddiagram}
\includegraphics[width=0.29\textwidth]{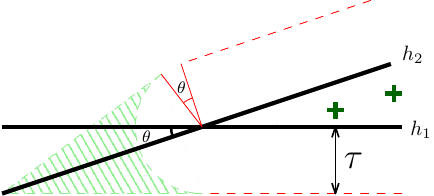}
}
\caption{A graphical depiction of the setup}
\label{fig:fig:2d-manipulation}
\end{figure}

\begin{comment}
\begin{figure}[!t]
\begin{subfigure}
  \includegraphics[width=0.9\textwidth]{stratscreening/figures/2d-1.pdf}
  \end{subfigure}
  \begin{subfigure}
    \includegraphics[width=0.4\textwidth]{stratscreening/figures/2d-2.pdf}
    \end{subfigure}
  \begin{subfigure}
      \includegraphics[width=0.4\textwidth]{stratscreening/figures/2d-diff.pdf}
      \end{subfigure}
  \caption{Each agent has a manipulation budget of $\tau$ and the cost function is $\ell_2$ distance. Then,
  ($a$) shows the region of agents who afford to manipulate their feature vectors to pass both tests simultaneously, ($b$) shows the region of agents who afford to manipulate their feature vectors to pass the tests sequentially (i.e, first $h_1$, then $h_2$), and ($c$) shows the difference in these two regions.}
\label{fig:2d-manipulation}
\end{figure}
\end{comment}

 %\juba{first time we say "Firm". Does not match intro} 

%% file: 1.1_related.tex
\paragraph{Related Work.} Our work inscribes itself at the intersection of two recent lines of work. The first one studies how strategic behavior affects decision-making algorithms (e.g. regression or classification algorithms), and how to design decision rules that take into account or dis-incentivize strategic behavior. This line of work is extensive and comprised of the works of~\citep{bruckner2011stackelberg, hardt2016strategic,kleinberg2020classifiers, randnoisestrat, miller2020strategic,strat1,strat2,strat3,strat4,strat5,strat6,strat8,strat9,strat10,strat11,strat12,strat13,strat14,strat15,strat16,stratperceptron,tang2021linear,hu2019disparate,socialcost18,performative2020,stratindark,randnoisestrat,gameimprove,bechavod2021gaming, bechavod2022information,shavit2020causal,dong2018strategic,chen2020learning,harris2021stateful}.

The second line of work is separate and aims to understand how decisions compose and affect each other in decision-making and screening pipelines~\citep{cohen2019efficient,BowerKNSVV17,blum2022multi,arunachaleswaran2022pipeline, dwork2020individual,faircomp}. These works study settings in which \emph{multiple} decisions are made about an individual or an applicant.
\cite{harris2021stateful} has a similar motivation to ours in studying how multiple rounds of interaction change strategic dynamics, however, the linearity of their model allows them to treat time-steps independently while our agents can benefit from using information on the subsequent steps of the pipeline.
%they observe that
 % under their linear effort conversion functions, the agent’s best response
%problem is linearly separable across time, and the agent’s effort profile at each %time is given by a linear program

However, and to the best of our knowledge, there is little work bringing these two fields together and studying strategic behavior in the context of decision \emph{pipelines} comprised of \emph{multiple} classifiers. This is where the contribution of the current work lies. 

%%%kevin addeded today

Interestingly, there are interesting connections between our model with classical work in learning intersections of half-spaces~\cite{klivans2004learning,klivans2009cryptographic}. 
In our model, we think of the half-spaces as %\emph{fixed} and 
\emph{known} in advance, so our model differs in that agents do not need to \emph{learn} half-spaces. However, future work could instead consider a learner who must learn the intersection of half-spaces while simultaneously considering the effect of strategic behavior, a complex learning problem. Further, there is a subtle distinction that agents in our work that agents may modify their features to pass half-spaces sequentially, but without needing to be in the intersection of all half-spaces; the crux of our contribution is in fact to show that sequentiality often leads to very different agent behavior than modifying features to reach the intersection of the classifiers' positive region. 

The sequentiality of our framework is related to the line of work on convex body chasing~\cite{sellke2020chasing,friedman1993convex,bubeck2019competitively,argue2021chasing, guan2022chasing, bansa2018nested,bubeck2020chasing}, but once again, a distinction between our approach and this line of work is that agents know all classifiers in advance and does not need to plan for an adversary.

Finally, perhaps closest to our work is the line of work on Online Convex Optimization (OCO) with switching costs and known loss functions. These works also assume that (1) the (single) agent observes the loss function before picking a point at each round or even observes the next (fixed size) loss functions sequence, and (2) the cost functions are dependent on the previous point $x_t$, (e.g., $\ell_2$ distance between the current and the previous point). However, our work differs in some of the specific assumptions we make (for example, an agent cannot choose their initial features, while one can choose the starting point in Online Convex Optimization with switching costs and known loss functions~\cite{Shi20OCO,Li21OCO, cesa2013online}), but more importantly, our main focus is different: beyond characterizing the optimal strategy for a strategic agent, we are interested in i) understanding how sequentiality affects and potentially increases agents' ability to strategize and ii) developing screening pipelines that are robust to strategic behavior.

%% file: 2_model.tex
Formally, individuals (or agents) are represented by a set of features $x \in \mathcal{X}$, where $\mathcal{X} \subseteq \mathbb{R}^d$, for $d \ge 1$. 
The firm has a fixed sequence of binary tests or classifiers $h_1, h_2, \dots, h_k:\mathcal{X} \rightarrow \{ 0, 1 \}$ that are deployed to select qualified individuals while screening out unqualified individuals. Here, an outcome of $1$ (positive) corresponds to an acceptance, and an outcome of $0$ (negative) corresponds to a rejection. Once a person is rejected by a test they leave the pipeline. 

In this chapter, we assume that the classifiers are linear and defined by half-spaces; i.e. $h_{i}(x) = 1 \iff w^{\top}_{i} x \geq b_i $ for some vector $w_i \in \reals^d$ and real threshold $b_i \in \reals$. Equivalently, we often write $h_i(x) = \mathbbm{1} \left[w^{\top}_{i} x \ge b_i\right]$.\footnote{While more general classes of classifiers could be considered, linear classifiers are a natural starting point to study strategic classification. This linearity assumption arises in previous work, e.g.~\citep{kleinberg2020classifiers,tang2021linear,gameimprove} to only name a few.} 

In this work we assume that the true qualifications of individuals are determined by the conjunction of the classifiers adopted by the firm in the pipeline, i.e. an agent $x$ is qualified if and only if $h_i (x) = 1$ for all $i$. In other words, the firm has designed a pipeline that makes no error in predicting individuals' qualifications \emph{absent strategic behavior}.

However, in the presence of strategic behavior, individuals try to manipulate their feature vectors to become positively classified by the classifiers simply because they receive a positive utility from  a positive outcome. Similar to prior works, throughout this work, we assume a ``white box" model meaning agents know the parameters for each classifier. More precisely, the firm commits to using a sequential screening process consisting of classifiers $h_1, h_2 \ldots h_k$, and each agent knows the parameters of each hypothesis, the order of the tests, her own feature value $x$, and the cost to manipulate to any other point in the input space.

An agent's cost function is modeled by a function $c: \X \times \X \to \mathbb{R}_{\ge 0}$ that takes two points $x, \hat{x}$ and outputs the cost of moving from $x$ to $\hat{x}$. One can think of $x$ as the initial feature vector of an agent and $\hat{x}$ as the manipulated features. In the sequential setting that we consider, we take the cost of manipulation to be the cumulative cost across every single manipulation. In particular, for a manipulation path $x^{(0)} \to x^{(1)} \to x^{(2)} \to \ldots \to x^{(k)}$ taken by an agent whose true feature values are $x^{(0)}$, the cost of manipulation is given by $\sum_{i=1}^k c(x^{(i-1)}, x^{(i)})$. We assume such manipulations do not change nor improve one's true qualifications\footnote{E.g., in a loan application, such manipulations could be opening a new credit card account: doing so may temporarily increase an agent's credit score, but does not change anything about an agent's intrinsic financial responsibility and ability to repay the loan.} and we discuss how the firm mitigates this effect of manipulation. 

In turn, the firm's goal is to have an accurate screening process whose predictions are as robust to and unaffected by such strategic: the firm modifies its classifiers $h_1, \cdots, h_k$ to $\Tilde{h}_1, \cdots, \Tilde{h}_k$ so that the output of $\Tilde{h}_1, \cdots, \Tilde{h}_k$ on manipulated agents' features can identify the qualified agents optimally with respect to a given ``accuracy measure"; we will consider two such measures in Section~\ref{sec:defense}.   

\subsection{Agent's Manipulation}
We proceed by formally defining the minimal cost of manipulation, which is the minimal cost an agent has to invest to pass all classifiers, and the best response of an agent for both sequential and simultaneous testing. 
\begin{definition}[Manipulation Cost: Sequential]
    Given a sequence of classifiers $h_1, \ldots, h_k$, a global cost function $c$, and an agent $x^{(0)} \in \mathcal{X}$, the manipulation cost of an agent in the sequential setting is defined as the minimum cost incurred by her to pass all the classifiers sequentially, i.e.,
    \begin{align*}
    \manipseq{x^{(0)},\{h_1,\ldots,h_k\}}  \\
    = \min_{x^{(1)}, \ldots, x^{(k)} \in \mathcal{X}}~~~&\sum_{i=0}^{k-1} c(x^{(i)}, x^{(i+1)}) 
    \\\text{s.t.}~~~~~~~~~&h_i(x^{(i)}) = 1~~\forall i \in [k].
    \end{align*}
    The \textit{best response} of $x^{(0)}$ to the sequential testing $h_1,\ldots,h_k$ is the path $x^{(1)},\ldots,x^{(k)}$ that minimizes the objective.
\end{definition}

\begin{definition}[Manipulation Cost: Conjunction or Simultaneous]
    Given a set of classifiers $\{h_1, \ldots, h_k\}$, a global cost function $c$, and an agent $x$, the manipulation cost of an agent in the conjunction setting is defined as the minimum cost incurred by her to pass all the classifiers at the same time, i.e.,
    \begin{align*}
    \manipconj{x,\{h_1,\ldots,h_k\}} = \min_{z \in \mathcal{X}}&~~~c(x, z)
    \\\text{s.t.}&~~~h_i(z)=1 \ \forall i \in [k].
    \end{align*}
    
    The \textit{best response} of $x$ to the conjunction of $h_1\ldots,h_k$ is the $z$ that minimizes the objective.
\end{definition}

%% file: 3_manipulation_charac.tex
In this section, we study the manipulation strategy of an agent. In particular, we present algorithms to compute optimal manipulation strategies efficiently.
For brevity, some of the proofs are relegated to the appendix. We make the following assumption on the cost function in most of the section, unless explicitly noted otherwise:

\begin{assumption}\label{asst: l2_cost}
The cost of moving from $x$ to $\hat{x}$ is given by $c(x,\hat{x}) = \Vert \hat{x} - x \Vert_2$, where $\Vert . \Vert_2$ denotes the standard Euclidean norm.
\end{assumption}
\subsection{Optimal Strategies in the Conjunction Case}
\label{sec: optstragcojn}

As a warm-up to our zig-zag strategy in Section \ref{sec: algo}, we first consider the optimal strategy for our benchmark, which is the case of the simultaneous conjunction of $k$ classifiers.
In the case where agents are supposed to pass a collection of linear classifiers simultaneously, the best response of an agent $x \in \mathbbm{R}^d$ is given by solving the following optimization problem
\begin{align}
\begin{split}\label{eq: convex_optim}
\min_{z}~& c(x,z)\\
\text{s.t.}~~~&w_i^\top z \geq b_i \ \forall i \in [k].
\end{split}
\end{align}
which is a convex program as long as $c$ is convex in $z$. 

In the special case in which $d = 2$ and $k = 2$, i.e. when feature vectors are two-dimensional and an agent must be positively classified by the conjunction of two linear classifiers $h_1 (x) = \mathbbm{1} (w_1^\top x \ge b_1 )$ and $h_2 (x) = \mathbbm{1} (w_2^\top x \ge b_2 )$, we provide a closed form characterization of an agent's strategy. 

We assume that the two classifiers are \emph{not} parallel to each other because if $w_2 = kw_1$ for some $k \in \mathbb{R}$, then one can show that either the acceptance regions of $h_1$ and $h_2$ do not overlap, or the optimal strategy of an agent is simply the orthogonal projection onto the intersection of the acceptance regions of $h_1$ and $h_2$.

We further assume, without loss of generality, that $b_1 = b_2 = 0$ because if either $b_1$ or $b_2$ is nonzero, one can use the change of variables $x' \triangleq x + s$ to write the classifiers as $h_1 (x') = \mathbbm{1} (w_1^\top x' \ge 0)$ and $h_2 (x') = \mathbbm{1} (w_2^\top x' \ge 0)$. Here $s$ is the solution to $\{w_1^\top s = -b_1, w_2^\top s = -b_2\}$.
% \[
% \begin{cases}
%     w_1^\top s = -b_1 \\ w_2^\top s = -b_2
% \end{cases}
% \]

For any $w \in \mathbb{R}^2$ with $\Vert w \Vert_2 = 1$, let $P_w (x)$ and $d_w (x)$ be the orthogonal projection of $x$ onto the region $\{y \in \mathbb{R}^2 : w^\top y \ge 0 \}$, and its orthogonal distance to the same region, respectively. We have
\begin{align*}
& P_w (x) \triangleq
\begin{cases}
    x & \text{if }w^\top x \ge 0 \\ x - (w^\top x) w & \text{if }w^\top x < 0
\end{cases}, \\
&d_w (x) \triangleq
\begin{cases}
    0 & \text{if }w^\top x \ge 0 \\ |w^\top x| & \text{if }w^\top x < 0
\end{cases}.
\end{align*}
Given this setup, the best response characterization of an agent $x$ can be given as follows. If $h_1 (x) = h_2 (x) = 1$ then $z=x$. Otherwise, the best response is either the orthogonal projection onto the acceptance region of $h_1$ or $h_2$, or moving directly to the intersection of the classifiers ($\Vec{0}$):
\begin{enumerate}
    \item If $h_1 (P_{w_2} (x)) = 1$, then $z = P_{w_2} (x)$ and the cost of manipulation is $\manipconj{ x^{(0)} ,\{h_1,h_2\}} = d_{w_2} (x)$.
    \item If $h_2 (P_{w_1} (x)) = 1$, then $z = P_{w_1} (x)$ and the cost of manipulation is $\manipconj{ x^{(0)} ,\{h_1,h_2\}} = d_{w_1} (x)$.
    \item if $h_1 (P_{w_2} (x)) = h_2 (P_{w_1} (x)) = 0$ then $z = \Vec{0}$ and the cost of manipulation is $\manipconj{ x^{(0)} ,\{h_1,h_2\}} = \Vert x\Vert_2$.
\end{enumerate}

Given a budget $\tau$, agents who can manipulate with a cost of at most $\tau$ to pass the two tests simultaneously, i.e. $\{x^{(0)}: \manipconj{ x^{(0)} ,\{h_1,h_2\}} \le \tau \} $ is highlighted in Figure \ref{fig:2d-manipulation}.a.

\subsection{A Zig-Zag Manipulation on Sequential Classification Pipelines}\label{sec:example}
Here, we make the observation that the sequential nature of the problem can change how an agent will modify her features in order to pass a collection of classifiers, compared to the case when said classifiers are deployed simultaneously. We illustrate this potentially counter-intuitive observation via the following simple example: 
\begin{example}\label{exp:zigzag}
Consider a two-dimensional setting. Suppose an agent going up for classification has an initial feature vector $x = (0,0)$. Suppose the cost an agent faces to change her features from $x$ to a new vector $\hat{x}$ is given by $\Vert \hat{x} - x\Vert_2$. Further, imagine an agent must pass two classifiers: $h_1(x) = \mathbbm{1}\left\{4 x_2 - 3 x_1 \geq 1\right\}$, and $h_2(x) = \mathbbm{1} \left\{x_1 \geq 1\right\}$, where $x_i$ is the $i-$th component of $x$.

It is not hard to see, by triangle inequality, that if an agent is facing a conjunction of $h_1$ and $h_2$, an agent's cost is minimized when $\hat{x} = (1,1)$ (this is in fact the intersection of the decision boundaries of $h_1$ and $h_2$), in which case the cost incurred by an agent is $\sqrt{1 + 1} = \sqrt{2}$ (see the red manipulation in Figure \ref{fig:zigexample}). 

\begin{figure}
    \centering
    \includegraphics[scale=0.8]{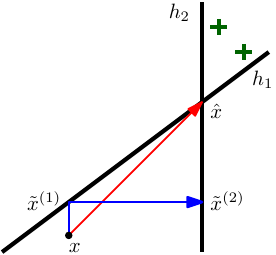}
    \caption{An example for a zig-zag strategy being better for an agent that starts at $x$ in the sequential case than moving in a single step. Here, an agent would prefer to first manipulate to $\tilde{x}^{(1)}$ then to $\tilde{x}^{(2)}$ (the blue arrows) instead of straightforwardly moving from $x$ to $\hat x$ as would be optimal in the conjunction case (the red arrow). 
    }
    \label{fig:zigexample}
\end{figure}

However, if the classifiers are offered sequentially, i.e. $h_1$ then $h_2$, consider the following feature manipulation: first, the agent sets $\tilde{x}^{(1)} = (0,1/4)$, in which case she passes $h_1$ and incurs a cost of $1/4$. Then, the agent sets $\tilde{x}^{(2)} = (1,1/4)$; the cost to go from $\tilde{x}^{(1)}$ to $\tilde{x}^{(2)}$ is $\Vert_2 (1,1/4) - (0,1/4) \Vert = 1$ (see the blue manipulation in Figure \ref{fig:zigexample}). In turn, the total cost of this manipulation to pass (i.e., get a positive classification on) both classifiers is at most $1 + 1/4 = 5/4$, and is always better than the $\sqrt{2}$ cost for the conjunction of classifiers! $\square$
\end{example}

Intuitively, here, the main idea is that in the ``conjunction of classifiers'' case, an agent must manipulate her features a single time in a way that satisfies all classifiers at once. However, when facing a sequence of classifiers $h_1, \ldots, h_k$, once an agent has passed classifier $h_{i-1}$ for any given $i$, it can ``forget'' classifier $h_{i-1}$ and manipulate its features to pass $h_i$ while \emph{not being required to pass $h_{i-1}$} anymore. In turn, the potential manipulations for an agent in the sequential case are less constrained than in the conjunction of classifiers case. This result is formalized below: 

\begin{restatable}{claim}{claimConjUpBoundsSeq}\label{claim:conjBoundSeq}
Let $h_1, \ldots, h_k$ be a sequence of $k$ linear classifiers. For any agent with initial feature vector $x \in \mathbb{R}^d$ ($d\ge 1$), 
%\[
$\manipconj{x,\{h_1,\ldots,h_k\}} \geq \manipseq{x,\{h_1,\ldots,h_k\}}$.
%\]
\end{restatable}
Intuitively, the above claim follows from the observation that any best response solution to the conjunction case in particular still passes all classifiers and has the same cost in the sequential case.

However, there can be a significant gap between how much budget an agent needs to spend in the conjunctive versus in the sequential case to successfully pass all classifiers (for illustration, see Figure~\ref{fig:2d-manipulation}). In fact, we show below that the multiplicative gap between the conjunctive and sequential manipulation cost can be unbounded, even in the two-dimensional setting:
\begin{lemma}
\label{lem:costgap}
Consider $d = 2$. For any constant $M>0$, there exists two linear classifiers $h_1$ and $h_2$ and an initial feature vector $x^{(0)}$ such that 
%\[
$\frac{\manipconj{x^{(0)},\{h_1,h_2\}} }{\manipseq{x^{(0)},\{h_1,h_2\}}} \geq M$.
%\]
\end{lemma}

\begin{proof}
Pick $x^{(0)} = (0,0)$. Let $\gamma > 0$ be a real number. Consider $h_1(x) = \mathbbm{1} \left\{\frac{x_1}{\gamma} + x_2 \geq 1 \right\}$ and $h_2(x) = \mathbbm{1} \left\{\frac{x_1}{\gamma} - x_2 \geq 1 \right\}$. Let $\hat{x}$ be the agent's features after manipulation. To obtain a positive classification outcome, the agent requires both  $\hat{x}_1 \geq \gamma (1 - \hat{x}_2)$ and $\hat{x}_1 \geq \gamma (1 + \hat{x}_2)$. Since one of $1 - \hat{x}_2$ or $1 + \hat{x}_2$ has to be at least $1$, this implies $\hat{x}_1 \geq \gamma$. In turn, $c(x,\{h_1,h_2\}) = \Vert \hat{x}\Vert \geq \gamma$. 

However, in the sequential case, a manipulation that passes $h_1$ is to set $x^{(1)} = (0,1)$. Then a manipulation that passes $h_2$, starting from $x^{(1)}$, is to set $x^{(2)} = (0,-1)$. The total cost is $\Vert (0,1) - (0,0) \Vert + \Vert (0,-1) - (0, 1)\Vert = 1 + 2 =3$. In particular, 
%\[
$\frac{\manipconj{x,\{h_1,\ldots,h_k\}} }{\manipseq{x,\{h_1,\ldots,h_k\}}} \geq \gamma/3$.
%\]
The result is obtained by setting $\gamma = 3M$.
\end{proof}

\subsection{An Algorithmic Characterization of an agent's Optimal Strategy in the Sequential Case}\label{sec: algo}
In this section, we show that in the sequential setting, an agent can compute her optimal sequences of manipulations efficiently. Consider any initial feature vector $x^{(0)} \in \mathbb{R}^d$ for an agent. Further, suppose an agent must pass $k$ linear classifiers $h_1,\ldots,h_k$. For $i \in [k]$, we write once again $h_i (x) = \mathbbm{1}[w_i^\top x\geq b_i]$ the $i$-th classifier that an agent must get a positive classification on. Here and for this subsection only, we relax our assumption on the cost function to be more general, and not limited to $\ell_2$ costs:

\begin{assumption}\label{convex_cost}
The cost $c(x,\hat{x})$ of moving from feature vector $x$ to feature vector $\hat{x}$ is convex in $(x,\hat{x})$.
\end{assumption}

This is a relatively straightforward and mild assumption; absent convexity, computing the best feature modifications for even a single step can be a computationally intractable problem. The assumption covers but is not limited to a large class of cost functions of the form $c(x,\hat{x}) = \Vert \hat{x} - x \Vert$, for \emph{any} norm $\Vert . \Vert$. It can also encode cost functions where different features or directions have different costs of manipulation; an example is $c(x,\hat{x}) = \left(\hat{x} - x \right)^\top A \left(\hat{x} - x \right)$ where $A$ is a positive definite matrix, as used in~\citep{ shavit2020causal,bechavod2022information}.

In this case, an agent's goal, starting from her initial feature vector $x^{(0)}$, is to find a sequence of feature modifications $x^{(1)}$ to $x^{(k)}$ such that: 1) for all $i \in [k]$, $h_i(x^{(i)}) = 1$. I.e., $x^i$ passes the $i$-th classifier; and 2) the total cost $\sum_{i=1}^k c(x^{(i-1)},x^{(i)})$ of going from $x^{(0)} \to x^{(1)} \to x^{(2)} \to \ldots \to x^{(k)}$ is minimized. This can be written as the following optimization problem: 
\begin{align}
\begin{split}\label{eq: convex_optim2}
\min_{x^{(1)},\ldots,x^{(k)}}~&\sum_{i=1}^k c(x^{(i-1)},x^{(i)})\\
\text{s.t.}~~~&w_i^\top x^{(i)} \geq b_i \ \forall i \in [k].
\end{split}
\end{align}

\begin{claim}
Program~\eqref{eq: convex_optim2} is convex in $(x^{(1)},\ldots,x^{(k)})$.
\end{claim}

In turn, we can solve the problem faced by an agent's computationally efficiently, through standard convex optimization techniques. 

\subsection{A Closed-Form Characterization in the 2-Classifier, 2-Dimensional Case}
\label{subsec:closed-form}
We now provide closed-form characterization of an agent's best response in the sequential case, under the two-dimensional two-classifier ($d=k=2$) setting that we considered in Section~\ref{sec: optstragcojn}. Here, we take the cost function to be the standard Euclidean norm, i.e. $c(x,\hat{x}) = \Vert \hat{x} - x \Vert_2$, as per Assumption~\ref{asst: l2_cost}.

\begin{restatable}{theorem}{twoDcharacthm}\label{thm: 2d_charac}
Consider two linear classifiers $h_1 (x) = \mathbbm{1} (w_1^\top x \ge 0)$ and $h_2 (x) = \mathbbm{1} (w_2^\top x \ge 0)$ where $\Vert w_i \Vert_2 = 1$ for $i \in \{1,2\}$ and an agent $x^{(0)} \in \mathbb{R}^2$ such that $h_1 (x^{(0)}) = 0$ and $h_2 (P_{w_1} (x^{(0)})) = 0$. Let $0 < \theta < \pi$ be the angle between (the positive region of) the two linear classifiers; i.e. $\theta$ is the solution to $\cos \theta = - w_1^\top w_2 $. Then:
\begin{enumerate}
\item If $|\tan \theta| > \Vert P_{w_1} (x^{(0)}) \Vert_2 / d_{w_1} (x^{(0)})$, then the best response for an agent is to pick 
%\[
$x^{(2)} = x^{(1)} = \Vec{0}$.
%\]
In this case, the cost of manipulation is 
%\[
$\manipseq{ x^{(0)} ,\{h_1,h_2\}} = \Vert x^{(0)} \Vert_2$. 
%\]
\item If $|\tan \theta| \le \Vert P_{w_1} (x^{(0)}) \Vert_2 / d_{w_1} (x^{(0)})$, then the best response is given by
\[
x^{(1)}  = \left( 1 - \frac{d_{w_1} (x^{(0)})}{\Vert P_{w_1} (x^{(0)}) \Vert_2} |\tan \theta| \right) P_{w_1} (x^{(0)})
\] 
and $x^{(2)} = P_{w_2} (x^{(1)})$, and the cost of manipulation is given by 
\begin{align*}
&\manipseq{ x^{(0)} ,\{h_1,h_2\}} 
\\&= d_{w_1} (x^{(0)}) | \cos \theta | + \Vert P_{w_1} (x^{(0)})\Vert_2  \sin \theta.
\end{align*}
\end{enumerate}
\end{restatable}
The proof of this theorem is provided in the Appendix. First, note that once the first feature modification has happened and an agent has passed classifier $h_1$ and is at $x^{(1)}$, the theorem states that an agent picks $x^{(2)}$ to simply be the orthogonal projection onto the positive region of $h_2$. This is because the cost for going from $x^{(1)}$ to $x^{(2)}$ is simply the $l_2$ distance between them, in which case picking $x^{(2)}$ to be the orthogonal projection of $x^{(1)}$ on $h_2$ minimizes that distance. The main contribution and challenge of Theorem~\ref{thm: 2d_charac} are therefore to understand how to set $x^{(1)}$ and what is the minimum amount of effort that an agent expands to do so.

\begin{figure}[t]
    \centering
    \includegraphics[scale=0.85]{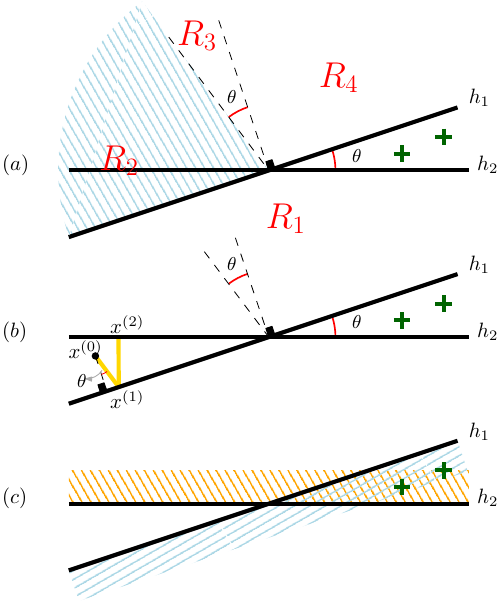}
    \caption{(a) Different cases for how agents best respond: agents in $R_1$ stay at their location to pass the first test and project onto $h_2$ to pass the second. Agents in $R_2$ deploy a zig-zag strategy. Agents in $R_3$ move to the intersection of $h_1$ and $h_2$. Agents in $R_4$ project onto $h_1$. (b) Geometric characterization of the zig-zag strategy: the line passing through $x^{(0)}$ and $x^{(1)}$ has angle $\theta$ with the line perpendicular to $h_1$. (c) This figure highlights the positive regions of $h_1$, $h_2$, and their intersection.}
    \label{fig:zigzag-region}
\end{figure}

Now let's examine different cases in Theorem~\ref{thm: 2d_charac}. Note that we assumed $h_1 (x^{(0)}) = 0$ and $h_2 (P_{w_1} (x^{(0)})) = 0$, i.e. that an agent is not in the positive region for the first test and $P_{w_1} (x^{(0)})$ is not in the positive region for the second test, because otherwise, the solution is trivial. In fact, if $h_1 (x^{(0)}) = 1$, then the solution is simply staying at $x^{(0)}$ for the first test and then projecting orthogonally onto the positive region of $h_2$ to pass the second test:
\begin{align*}
& x^{(1)} = x^{(0)}, \ x^{(2)} = P_{w_2} (x^{(1)}) \\
& \manipseq{ x^{(0)} ,\{h_1,h_2\}} = d_{w_2} (x^{(0)})
\end{align*}
This corresponds to region $R_1$ of agents in Figure~\ref{fig:zigzag-region}. If $h_1 (x^{(0)}) = 0$, but $h_2 (P_{w_1} (x^{(0)})) = 1$, then the best response solution is simply the orthogonal projection onto the positive region of $h_1$:
\begin{align*}
&x^{(2)} = x^{(1)} = P_{w_1} (x^{(0)}) \\
&\manipseq{ x^{(0)} ,\{h_1,h_2\}} = d_{w_1} (x^{(0)})
\end{align*}
This corresponds to region $R_4$ of agents in Figure~\ref{fig:zigzag-region}. Additionally, the first case in the closed-form solutions in Theorem~\ref{thm: 2d_charac} corresponds to the region of the space where agents prefer to travel directly to the intersection of the two classifiers than deploying a zig-zag strategy: this corresponds to region $R_3$ in Figure~\ref{fig:zigzag-region}. The second case corresponds to the region where agents do find that a zig-zag strategy is less costly and gives the algebraic characterization of the optimal zig-zag strategy. This region for an agent is denoted by $R_2$ in Figure~\ref{fig:zigzag-region}. Also, as shown by Figure~\ref{fig:zigzag-region}.b, the zig-zag strategy of agents in $R_2$ has the following geometric characterization: pick $x^{(1)}$ on $h_1$ such that the line passing through $x^{(0)}$ and $x^{(1)}$ has angle $\theta$ with the line perpendicular to $h_1$.

Given a budget $\tau$, agents who can manipulate with a cost of at most $\tau$ to pass the two tests in the sequential setting, i.e. $\{x^{(0)}: \manipseq{ x^{(0)} ,\{h_1,h_2\}} \le \tau \} $ is highlighted in Figure~\ref{fig:2d-manipulation}.b.

We conclude this section by showing that if $\theta \ge \pi / 2$, then agents incur the same cost in the sequential setting as they would under the conjunction setting. In other words, agents can deploy the strategy that they would use if they had to pass the two tests simultaneously. The proof of this theorem is provided in the Appendix.

\begin{restatable}{theorem}{nozigzag}\label{thm: no-zigzag}
    If $\pi/2 \le \theta < \pi$, then for every agent $x^{(0)}$ there exists optimal strategies $x^{(1)}$ and $x^{(2)}$ s.t. $x^{(1)} = x^{(2)}$, i.e.,
    %\[
    $\manipseq{ x^{(0)} ,\{h_1,h_2\}} = \manipconj{ x^{(0)} ,\{h_1,h_2\}}$.
    %\]
\end{restatable}

%% file: 3.1_best_response.tex
\subsection{Monotonicity}
\label{subsec-monotone}
We now consider a monotonicity property that excludes the possibility of a zig-zag strategy arising. 
A similar property
is noted in 
\citep{socialcost18}.

\begin{definition}[Feature Monotone Classifiers]
Classifier $h_i:\mathbb{R}^d \rightarrow \{0,1\}$ is  \textit{monotone} if for every  individual $x$ that is classified as positive by $h_i$, any feature-wise increase in the features of $x$ results in a positive classification by $h_i$. 
Formally, 
\[
\forall x\in \mathbb{R}^d: h_i(x)=1 \Rightarrow h_i(x+\alpha)=1\quad \forall \alpha\in (\mathbb{R}_{\geq 0})^d.
\]
\end{definition}

Note that this monotonicity property may not hold in some classification problems. For example, when applying for a mortgage for $\$100,000$, presumably monotonically increasing income means one is more credit-worthy. However, if an individual reports a $\$3$ million a year income for a loan of $\$100,000$, such a large income could instead indicate fraudulent income reporting or remarkably poor financial planning since presumably such a high net worth individual should not need such a small loan. 

In fact, in case $k=2$, the angle $\theta$ 
measures the ``alignment'' between the classifiers. In the above example, the classifiers may not be aligned. Increases in income are desirable to show financial responsibility; yet, beyond a certain point (for example, when the income becomes much larger than the desired loan), income may become an indicator of poor financial planning or fraudulent transactions. In some hiring settings, having sufficient qualifications is desirable; yet, over-qualification can often be grounds for rejection of a job application.

\begin{restatable}{theorem}{NEthm}\label{thm:ne_thm}
 Let $h_1,\ldots,h_k$ be a sequence of monotone classifiers, and let the initial  feature vector $x^{(0)}$ be such that $h_i(x^{(0)})=0$ for every $i\in[k]$. Assume the cost function can be written as $c(x,\hat{x}) = \Vert \hat{x} - x \Vert$ for some norm $\Vert . \Vert$. Then, we have that
 \[
 \manipseq{ x^{(0)} ,\{h_1,\ldots, h_k\}} = \manipconj{ x^{(0)} ,\{h_1,\ldots, h_k\}}.
 \]
\end{restatable}

Theorem~\ref{thm:ne_thm} in particular implies that under our monotonicity assumption and for a large class of reasonable cost functions, an agent has no incentive to zig-zag in the sequential case and in fact can simply follow the same strategy as in the simultaneous or conjunctive case. This insight immediately extends even when $x^{(0)}$ is positively classified by some but not all of the $h_i$'s as any best response is guaranteed to increase the feature values and thus will maintain the positive classification results of these classifiers. 

%% file: 3_0greedycomparison.tex
\subsection{Myopic or Greedy Strategy}

A natural question that reader might have is how the cost of the zig-zag strategy compares to the cost of a greedy strategy that simply manipulates to the nearest passing point of the current test. One advantage of a greedy strategy is that an agent only needs to know what the next classifier they face is, rather than the entire screening pipeline in advance. 
%That could be a more realistic threat model since the agent needs less information in advance to compute the strategy. 

Given that the agent has full information about the pipeline, the zig-zag manipulation is by definition the optimal strategy and the greedy strategy can be sub-optimal. In the two-classifier two-dimensional case that we consider in our work here, our theorem states that the zig-zag manipulation is the unique optimal manipulation and that this manipulation is different from the greedy manipulation (see Figure~\ref{fig:zigzag-region}(b)). In fact, for $k=2$, the additive gap between the cost of the zig-zag strategy and the greedy strategy can be shown to be $(1 -\cos(\theta))\cdot r$ where $\theta$ is the angle between the two classifiers and $r$ is the distance of the agent from the first classifier. 

One can also show that the gap is unbounded when $k$ grows large: previous work \cite{friedman1993convex} shows an unbounded gap between
the movement cost of being greedy and directly going to the closest point at the intersection of the half-spaces. Because the optimal zig-zag strategy cannot do worse than directly reaching this closest point, the gap between zig-zag and greedy is also unbounded.

%% file: 4_defense.tex
Up to this point in the chapter, we have focused mainly on the existence and feasibility of a zig-zag manipulation strategy from the perspective of an agent. 
We now shift gears and discuss the firm's decision space. We are interested in understanding how the firm can modify its classifiers to maintain a high level of accuracy (if possible), despite the strategic manipulations of an agent. To this end, we assume there is a joint distribution of features and labels $\mathcal{D}$ over $\mathcal{X}\times \{0,1]\}$. 
Interestingly, previous works \citep{bruckner2011stackelberg, hardt2016strategic} show hardness results for finding optimal strategic classifiers, where the objective is finding a single classifier $h$ that attains the strategic maximum accuracy.

Now, we can introduce the defender's game for a typical strategic classification problem.
\begin{align}
\begin{split}
\min_{h \in \mathcal{H}}&~~~P_{(x,y) \sim \mathcal{D}} [h(z^{*} (x)) \neq y ] \\
\text{s.t.}&~~~z^{*}(x) = \arg\max_{z}~h(z) - c(x,z) 
\end{split}
\end{align}
In our work here, $h$ is actually given by the sequential composition of classifiers in the screening process and $c(x,z)$ is the sum of manipulation costs per stage. The objective function in this optimization problem is a direct generalization of $0$-$1$ loss for normal learning problems, only complicated by the strategic behavior of an agent. 

As \cite{bruckner2011stackelberg} observe, this is a bi-level optimization problem and is NP-hard \cite{Jeroslow85} to compute, even when constraints and objectives are linear. 
Interestingly, \cite{hardt2016strategic} also show a hardness of approximation result for general metrics. Because of these past hardness results, we instead focus on a more tractable defense objective. 

\subsection{Conservative Defense}\label{sec:conservative-defense}
Here, we consider a different objective motivated by the hiring process in firms, in which avoiding false positives and not hiring unqualified candidates can  be seen as arguably more important than avoiding false negatives and not missing out on good candidates. This objective, described below, has been previously studied in the context of strategic classification, in particular in~\citep{gameimprove}. 
\begin{definition}[No False Positive Objective]
Given the manipulation budget $\tau$ and the initial linear classifiers $h_1, \cdots, h_k$, the goal of the firm is to design a modified set of linear classifiers $\tilde{h}_1, \cdots, \tilde{h}_k$ that maximize the true positive rate of the pipeline on manipulated feature vectors subject to no false positives. Recall that the ground truth is determined by the conjunction of $h_1, \cdots, h_k$ on unmanipulated feature vectors of agents.   
\end{definition}
Without loss of generality, we assume the pipeline is non-trivial: the intersection of acceptance regions of $h_1, \cdots, h_k$ is non-empty. 

We prove that, under standard assumptions on linear classifiers of the firm, a defense strategy that ``shifts" all classifiers by the manipulation budget, is the optimal strategy for the firm in both pipeline and conjunction settings. We formally define the defense strategy as follows: 
\begin{definition}[Conservative Strategy]\label{def:conservative-defense}
Given the manipulation budget $\tau$, the firm conservatively assumes that each agent has a manipulation budget of $\tau$ per test. For each test $h_i(x) = \mathbbm{1}[w_i ^\top x \geq b_i]$, the firm replaces it by a ``$\tau$-shifted" linear separator $\Tilde{h}_i(x) = \mathbbm{1}[w_i ^\top x \geq b_i + \tau])$. In this section, without loss of generality, we assume that all $w_i$'s have $\ell_2$-norm equal to one. 
\end{definition}

\begin{figure}
    \centering
    \includegraphics[width=0.5\textwidth]{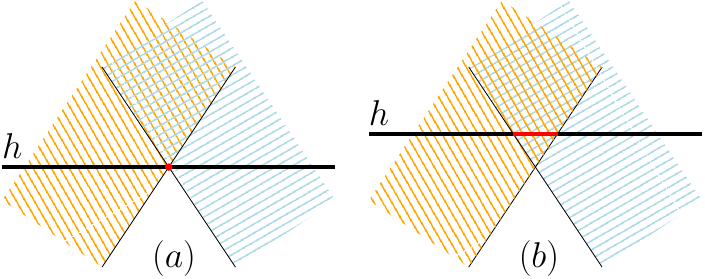}
    \caption{In ($a$), the intersection of $h$ with the positive half plane of the other two classifiers that are in blue and gray shadows is a point which is of zero dimension. This case is not in the general position and $h$ is a redundant classifier. However, in ($b$), the intersection of $h$ with the described positive regions is a line segment, a one-dimensional object. Here, $h$ is not redundant.}
    \label{fig:general-assumption}
\end{figure}

Our statement holds when the linear classifiers satisfy the following ``general position" type condition. \begin{definition}\label{def:gen-position}
We say a collection of linear classifiers $\mathcal{H} = \{h_1(x) = \mathbbm{1}[w_1 ^\top x \geq b_1], \cdots, h_k(x) = \mathbbm{1}[w_k ^\top x \geq b_k]\}$ with $w_1, \cdots, w_k\in \mathbb{R}^d$ are in ``general position" if for any $i\in [k]$, the intersection of $\{x | w_i^\top x = b_i\}$ and $\{x | \bigwedge_{j\in [k], j \neq i} h_j(x)=1\}$ lies in a $(d-1)$-dimensional subspace but in no $(d-2)$-dimensional subspace. In $\mathbb{R}^2$, this condition is equivalent to the standard general position assumption (i.e., no three lines meet at the same point). Moreover, this condition implies that no test in $\mathcal{H}$ is ``redundant", i.e., for every $i\in[k]$, the positive region of $\mathcal{H}$ (i.e., $\bigwedge_{h\in \mathcal{H}} \{x|h(x)=1\}$) is a proper subset of the positive region of $\mathcal{H}\setminus{h_i}$.   See Figure~\ref{fig:general-assumption} for an example in $\mathbb{R}^2$.
\end{definition} 
Now, we are ready to state the main result of this section.
\begin{restatable}{theorem}{thmConservativeDef} \label{thm:conservative-defense-optimality}
Consider a set of linear classifiers $\mathcal{H}= \{h_1, \cdots, h_k\}$ that are in ``general position" (as in Definition~\ref{def:gen-position}). Moreover, suppose that each agent has a manipulation budget of $\tau$. Then, in both the conjunction and sequential settings, the conservative defense is
a strategy that maximizes true positives subject to zero false positives.
\end{restatable}

The proof is provided in Appendix~\ref{app:conservative_proof}. Note that while the conservative defense strategy has the maximum possible true positive subject to zero false positive in both simultaneous and sequential settings, by Claim~\ref{claim:conjBoundSeq}, the conservative defense achieves a higher true positive rate in the sequential setting compared to the simultaneous case. Informally, from the firm's point of view, {\em under manipulation, the sequential setting is a more efficient screening process}.

%% file: 5_discussion.tex
We have initiated the study of \textit{Strategic Screening}, combining screening problems with strategic classification. 
This is a natural and wide-spread problem both in automated and semi-automated decision making. 
We believe these examples and our convex program  can aid in the design and monitoring of these screening processes. Substantial open questions remain regarding fairness implications (Appendix \ref{app:fair}) of the defender's solution and exactly how susceptible real world pipelines are to zig-zagging.

\section{Fairness and Strategic Screening} %As noted in the related work section, there are multiple works that consider fairness and strategic manipulation. 
\label{app:fair}
Some of the works cited in the related work section consider fairness considerations in the space of strategic manipulation, stemming either from unequal abilities to manipulate~\cite{socialcost18,hu2019disparate} or unequal access to information about the classifiers~\cite{bechavod2022information} across different groups. We do not consider these connections in our work, but these considerations are of significant interest and a natural direction for further research, especially due to the importance of making fair decision in high-stake, life altering contexts. We finish with a few interesting examples for this.

Disparities might arise both in the conjunction and in the sequential setting, with or without defense. 
consider the classifiers presented in Example \ref{exp:zigzag} and an instance in which candidates belong to two groups, $G^1$ and $G^2$ with initial feature vector distributed identically and characterized by different total manipulation budgets,  $\sqrt{2}=\tau^2>\tau^1=5/4$. 
The narrative of the fairness disparities in the conjunction case is a simple generalization of the single classifiers case (e.g., \cite{hardt2016strategic}). 

If the distribution is such that a significant fraction of individuals (from both groups) starts at a feature vector that is classified by both classifiers as $0$ and that requires $\sqrt{2}$ manipulation cost to reach their intersection, then only the individuals form $G_2$ will be able to manipulate. 
For the sequential case, consider a distribution with a large enough fraction of individuals starting  at $(0,0)$. Example \ref{exp:zigzag} demonstrates that only individuals from $G_2$ will have sufficient budget to manipulate (using the zig-zag strategy). 
If the firm applies the conservative defense, individuals from $G_1$ that should have been classified as positive might not have sufficient  budget to manipulate their way to acceptance,  which in turn implies  higher false negative rates. 
This indicates, similarly to prior results in strategic classification (e.g.,~\cite{hu2019disparate}), how the  members of the advantaged group are more  easily admitted or hired. 

%ali{The rest is to be removed??}

\section{Proofs of Section \ref{sec:bestRes}}
\label{sup:bestRes}
\input{sup-3_bestRes}

\section{Proofs of Section \ref{sec:defense}}\label{sup:defense}

%% file: sup-3_bestRes.tex
The following is a restatement of Claim~\ref{claim:conjBoundSeq}.

\claimConjUpBoundsSeq*
%\lee{does anybody knows why it says claim b1 instead of 3.2?}\saeed{I don't -- that's weird.}\juba{don't know either. WE could just copy the claim here and give it the right tag}
\begin{proof}[Proof of Claim~\ref{claim:conjBoundSeq}]
Let $\cost$ be the agent's cost function. Let $\hat{x}$ be a vector such that $h_i(\hat{x}) = 1$ for all $i \in [k]$, and such that $\cost(x,\hat{x}) \leq \tau$ where $\tau$ is the manipulation budget available to the agent. Since $\hat{x}$ satisfies $h_i(\hat{x}) = 1$ for all $i \in [k]$, the feature modification $x \to \hat{x}$  gives a positive classification outcome to the agent in the sequential case. Further, the cost of this manipulation is $\cost(x,\hat{x}) + 0 + \ldots + 0 = \cost(x,\hat{x})$. In turn, for any feasible one-shot manipulation that passes all classifiers in the conjunctive case, there exists a feasible sequential manipulation that passes all classifiers in the sequential case which could be of a lower cost; this concludes the proof. 
\end{proof}

\twoDcharacthm*
\begin{proof}[Proof of Theorem~\ref{thm: 2d_charac}]
%For any given $x^{(0)} \in \mathbb{R}^d$, let $P_w (x)$ and $d_w (x)$ be the projection of $x$ onto the line $\{y \in \mathbb{R}^d : w^\top y = 0 \}$ with $\Vert w \Vert_2 = 1$, and its distance to the same line, respectively. We have
%\[
%P_w (x) \triangleq x - (w^\top x) w, \quad d_w (x) \triangleq |w^\top x|
%\]
\begin{figure}
    \centering
    \includegraphics{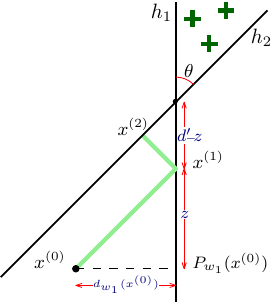}
    \caption{This figure shows how we reduced the optimization problem in Equation~\ref{eq:bestresponse} to the one in Equation~\ref{eq:reduction}. %\ali{tbd. Saeed, pls check the figure.}\juba{can we note the intersection of the classifiers is 0 on the figure to make it slightly easier to read? And add an arrow from $P_{w1}$ to $0$ showing it is $d'$?}\ali{check it now.}
    }
    \label{fig:zigzag-example}
\end{figure}
Given classifiers $h_1$ and $h_2$, the best response of an agent $x^{(0)}$ is a solution to the following optimization problem, as noted in Section~\ref{sec: algo}:
\begin{align*}
&\manipseq{ x^{(0)} ,\{h_1,h_2\}} \\
= &\min_{x^{(1)}, x^{(2)}} \left\{ \Vert x^{(0)} - x^{(1)} \Vert_2 + \Vert x^{(1)} - x^{(2)} \Vert_2 : w_1^\top x^{(1)} \ge 0, w_2^\top x^{(2)} \ge 0 \right\}
\end{align*}
First, we remark that given any $x^{(1)}$, the optimal choice of $x^{(2)}$ is the orthogonal projection of $x^{(1)}$ on classifier $f_2$. Therefore, the best response can be written as:
\begin{equation}
\label{eq:bestresponse}
\manipseq{ x^{(0)} ,\{h_1,h_2\}} = \min_{x^{(1)} \in \mathbb{R}^2} \left\{ \Vert x^{(0)} - x^{(1)} \Vert_2 + d_{w_2} \left(x^{(1)}\right) : w_1^\top z \ge 0 \right\}
\end{equation}
To simplify notations, we will denote $x \triangleq x^{(0)}$. Under the assumptions of the theorem (more specifically, $h_1 (x) = 0$ and $h_2 (P_{w_1} (x)) = 0$), Equation~\eqref{eq:bestresponse} can be rewritten as an optimization over a one-dimensional variable:
\begin{equation}\label{eq:reduction}
\min_{0 \le z \le d'_{w_1} (x)} \left\{ g(z) \triangleq \sqrt{d_{w_1}^2 (x) + z^2} + (d'_{w_1} (x) - z) \sin \theta \right\}
\end{equation}
where $d'_{w_1} (x) \triangleq \Vert P_{w_1} (x)\Vert_2$ -- see Figure~\ref{fig:zigzag-example} for a graphical justification of this rewriting. %\juba{added, check:} this comes from the fact that since $w_1^\top z = 0$, we can write $z-x$ as the sum of two orthogonal components: $P_{w_1} (x)$ the projection of $x$ on the line $w_1^\top x = 0$, and $z - P_{w_1} (x)$ which is orthogonal to said line. 
Note that $g(z)$ achieves its minimum either at the boundaries or at the point where $g'(z) = 0$. Therefore, we have that the minimum is one of the following:
\begin{align*}
%\begin{cases}
& z = 0 \Longrightarrow  g(z) = d_{w_1} (x) + d'_{w_1} (x) \sin \theta \\
&z = d'_{w_1} (x) \Longrightarrow  g(z) = \sqrt{d_{w_1}^2 (x) + d^{'2}_{w_1} (x)} = \Vert x\Vert_2 \\
& z = d_{w_1} (x) | \tan \theta| \Longrightarrow g(z) = d_{w_1} (x) \cos \theta + d'_{w_1} (x) \sin \theta \ (g'(z) = 0)
%\end{cases}
\end{align*}
We can show that if $|\tan \theta| > d'_{w_1} (x) / d_{w_1} (x)$, then the minimizer $z^\star =  d'_{w_1} (x)$, meaning $x^{(2)} = x^{(1)} = \Vec{0}$, and that
\[
\manipseq{ x ,\{h_1,h_2\}} = \Vert x\Vert_2
\]
and if $|\tan \theta| \le d'_{w_1} (x) / d_{w_1} (x)$, then the minimizer $z^\star =  d_{w_1} (x) | \tan \theta|$ which implies
\[
x^{(1)}  = \left( 1 - \frac{d_{w_1} (x^{(0)})}{\Vert P_{w_1} (x^{(0)}) \Vert_2} |\tan \theta| \right) P_{w_1} (x^{(0)})
\] 
and $x^{(2)} = P_{w_2} (x^{(1)})$, and that
\[
\manipseq{ x ,\{h_1,h_2\}} = d_{w_1} (x) | \cos \theta | + d'_{w_1} (x) \sin \theta
\]
Therefore, putting the two cases together,
\begin{align*}
& \manipseq{ x ,\{h_1,h_2\}} = 
\begin{cases}
\Vert x\Vert_2 & \text{if } |\tan \theta| > d'_{w_1} (x) / d_{w_1} (x)\\
d_{w_1} (x) | \cos \theta | + d'_{w_1} (x) \sin \theta & \text{if } |\tan \theta| \le d'_{w_1} (x) / d_{w_1} (x)
\end{cases}
\end{align*}
\end{proof}

\nozigzag*
\begin{proof}
    Let $(x^{(1)}, x^{(2)} = P_{w_2} (x^{(1)}))$ be an optimal strategy of the agent in the sequential setting. Suppose $x^{(1)} \neq x^{(2)}$. We have that
    \begin{align*}
    w_1^\top x^{(2)} &= w_1^\top \left( x^{(1)} - (w_2^\top x^{(1)}) w_2 \right) \\
    &= w_1^\top x^{(1)} - (w_2^\top x^{(1)}) (w_1^\top w_2)
    \end{align*}
    But note that $w_1^\top x^{(1)} \ge 0$ because $x^{(1)}$ passes the first classifier by definition, $w_2^\top x^{(1)} \le 0$ because $x^{(1)} \neq x^{(2)}$, and $w_1^\top w_2 \ge 0$ because $\pi/2 \le \theta < \pi$. Therefore, $w_1^\top x^{(2)} \ge 0$ which implies $h_1 (x^{(2)}) = 1$. However, if $h_1 (x^{(2)}) = 1$, then the following manipulation: $y^{(0)} = x^{(0)}$ and $y^{(1)} = y^{(2)} = x^{(2)}$ passes both tests and that its cost is: $\Vert x^{(2)} - x^{(0)} \Vert_2 \le \Vert x^{(2)} - x^{(1)} \Vert_2 + \Vert x^{(1)} - x^{(0)} \Vert_2$ by the triangle inequality. Given the optimality of $(x^{(1)}, x^{(2)})$, we conclude that $(y^{(1)}, y^{(2)})$ is another optimal strategy that the agent can deploy.
\end{proof}

\NEthm*
\begin{proof}
Let $f_{1,\ldots,k}:\mathbb{R}^d\rightarrow \{0,1\}$ denote the function that returns the conjunction of all the classifiers, i.e., $f_{1,\ldots,k}(x)=h_1(x)\land \ldots \land h_k(x)$. 

Let $z^*_{1,\ldots,k}(x^{0})$ denote the point on $f_{1,\ldots,k}$ that minimizes the cost, i.e., $z^*_{1,\ldots,k}(x^{0})=\text{argmin}_{x^{(1)}}\Vert x^{(0)}-x^{(1)}\Vert_p$. Note that by definition,  points on $f_{1,\ldots,k}$ are classified as positive by all classifiers $h_1,\ldots,h_k$ (i.e., $z^*_{1,\ldots,k}(x^{0})$ this is the best response for the conjunction case). 
% Formally, $z^*_{1,\ldots,k}(x)=\text{argmin}_z\{\Vert x-z \Vert_2: f_{1,\ldots,k}(z)=1\}$ and $cost(- z^*_{1,\ldots,k}(x)\Vert=c_{conj}^*(x,\{h_1,\ldots,h_k\})$. 

It follows from the triangle inequality that any $x^{(1)}$ such that $h_1(x^{(1)})\land \ldots \land h_k(x^{(1)})=1$ has cost $c(x^{(0)},x^{(1)})\geq c(x^{(0)},z^*_{1,\ldots,k}(x^{0}))$.

We proceed by induction on the number of classifiers. 
For the induction base, consider $k=1$. Clearly, in this case moving to $z^*_{1,\ldots,k}(x)$ yields the best response. 

For the induction step, assume that for every initial point $x'$, and every $k-1$ monotone classifiers $h_2,\ldots,h_k$ it holds that
\[
    \Vert x'-z^*_{2,\ldots,k}(x')\Vert_p\leq \Vert x'-z_2\Vert_2+\ldots+\Vert z_{k-1}-z_{k}\Vert_p.
\]
for every $z_2,\ldots,z_k\in \mathbb{R}^d$ such that $h_i(z_i)=1$.

Adding the additional classifier in the beginning, $h_1$ and considering the initial point, $x$.
Assume by contradiction that there exists a path $x=z_0,z_1\ldots, z_k$ such that $h_i(z_i)\geq 0$ for every $i\in [k]$ and that 
\begin{align}\label{eq:monContra}
    c_{seq}^*(x,\{h_1,\ldots,h_k\}) 
    &= \Vert x-z_1\Vert_p+\ldots+\Vert z_{k-1}-z_k\Vert_p \nonumber\\
    &<\Vert x-z^*_{1,\ldots,k}(x)\Vert_p.
\end{align}
Since the path from $z_1$ to $z_k$ is a best response for $h_2,\ldots,h_k$ when the initial feature vector $z_1$, by setting $x'=z_1$ we can apply the induction step we and replace this path by $x,z_1,z^*_{2,\ldots,k}(x')$ without increasing the sum of manipulations. 
If $f_{1,\ldots,k}(z^*_{2,\ldots,k}(z_1))=1$, we have that $\Vert x- z_1\Vert_p+\Vert z_1- z^*_{2,\ldots,k}(z_1)\Vert_p\leq \Vert x-z^*_{1,\ldots,k}(x)\Vert_p$ due to the triangle inequality and the definition of $z^*_{1,\ldots,k}(x)$ and this is a contradiction to Eq.~\ref{eq:monContra}.

So assume $f_{1,\ldots,k}(z^*_{2,\ldots,k}(z_1))=0$.
Since $h_i(z^*_{2,\ldots,k}(z_1))=1$ for every $i\geq 2$ by definition, we have that $h_1(z^*_{2,\ldots,k})=0$. As $h_1(z_1)=1$, we can define $z'\in \mathbb{R}^d$ such that 
\[ 
z'[j]=\max\{{z^*_{2,\ldots,k}(z_1)[j], z_1[j]}\}, 
\]
and from monotonicity it follows that $f_{2,\ldots,k}(z')=1$.

Finally, we have that $\Vert x-z_1\Vert_p+ \Vert z_1-z'\Vert_p < \Vert x-z_1\Vert_p+ \Vert z_1-z^*_{2,\ldots,k}(z_1)\Vert_p$, which is a contradiction to the minimiality of $z^*_{2,\ldots,k}(z_1)$ and thus to the minimality of $z_2,\ldots,z_k$.
\end{proof}

%% file: sup-4_defense.tex
\thmConservativeDef*

\begin{proof}[Proof of Theorem~\ref{thm:conservative-defense-optimality}]
First, we prove that conservative defense achieve zero false positive in both cases. To show this, by Claim~\ref{claim:conjBoundSeq}, it suffices to show it for the sequential setting only. Consider an agent $x$ who initially (i.e., before manipulation) is not in the positive region of conjunctions of $h_1, \cdots h_k$; i.e., $\Pi_{j\in [k]}h_j(x^{(0)})=0$. Hence, there exists a classifier $h_i$ such that $w^\top_i x^{(0)}< b_i$. Now, let $x^{(i)} : x^{(0)} + \epsilon_i$ denote the (manipulated) location of $x$ right before stage $i$. Since the total manipulation budget of $x$ is $\tau$, $w_i^\top x^{(i)} \le w_i^\top x^{(0)} + w_i^\top \epsilon_i < b_i + \tau$ (the choice of $\varepsilon_i$ that maximizes $w_i^\top \epsilon_i$ is $\epsilon_i = \tau w_i$, and $w_i^\top (\tau w_i) = \tau$ since $\Vert w_i \Vert_2 = 1$). Hence, $\tilde{h}(x^{(i)}) = 0$ and agent $x$ cannot pass the modified pipeline $\Tilde{h}_1, \cdots, \Tilde{h}_k$. %\juba{see email. Think we are skipping a step here? Think we are relying on the fact that at best you manipulate $\tau w_i/\Vert w_i \Vert$?}

Next, consider test $i$ and let $\Delta^i$ denote the subspace of points (i.e.,~agents) in the intersection of $\{x | h_i(x)=0\}$ and $ \bigwedge_{j\in [k], j\neq i} \{x| h_j(x)=1\}$. By the general position assumption, $\Delta^i$ is a $(d-1)$-dimensional subspace and is a subset of the $(d-1)$-dimensional hyperplane corresponding to $w^\top_i x = b_i$. Then, there exists only a unique linear separator which is at distance exactly $\tau$ from $\Delta^i$ (and is in the positive side of $h_i$); $\hat{h}_i(x) := \mathbbm{1}[w^\top_i x \ge b_i + \tau]$. Given that any defense strategy with zero false positive has to classify an agents in $\Delta^i$ as negative, it is straightforward to verify that any ``feasible" modified linear separator $h'_i$ (i.e., achieving zero false positive) results in true positive rate less than or equal to the one replaces $h'_i$ with $\hat{h}_i$.
\end{proof}

%% file: rob_introduction.tex
\section{Introduction}

Robustness to adversarial examples is considered a major contemporary challenge in machine learning. Adversarial examples are carefully crafted perturbations or manipulations of natural examples that cause machine learning predictors to miss-classify at test-time \citep{goodfellow2014explaining}. One particularly challenging aspect of this problem is the asymmetry between the learner and the adversary. Specifically, a learner needs to produce a predictor that is \textit{correct} on a randomly drawn natural example and \textit{robust} to potentially \textit{exponentially} many possible perturbations of it; while, the adversary needs to find just a single perturbation that fools the learner. In fact, because of this, adversarially robust learning has proven to require more sophisticated learning algorithms that go beyond standard Empirical Risk Minimization (ERM) in non-robust learning \citep{pmlr-v99-montasser19a}. 

%One of the main challenges arises when the adversary can insert carefully crafted perturbations on a natural example in an exponential number of ways at test time to induce errors. 
%A fundamental challenge in adversarial robustness is the intense asymmetry between the learner and the adversary.
%A classifier needs to be robust on every natural example and for each natural example typically there can be exponentially many possible perturbations. The adversary just needs to find one effective perturbation, while the adversary needs to correctly classify each one. 

%One example is in the robust image classification task where 
%In image classification, for instance, an adversary can put an exponential number of adversarial patches (patch attacks) in different locations, sizes, and shapes on an image to incur errors. 
In patch attacks on images, for instance, an adversary can select one of an exponential number of designs for a patch to be placed in the image in order to cause a classification error. To address this exponential asymmetry between the learner and the adversary, recently \cite{xiang2022patchcleanser} introduced a clever algorithmic scheme, known as Patch-Cleanser, that provably reduces the exponential number of ways that an adversary can attack to a polynomial number of ways through the idea of masking images. 

%\kmsmargincomment{this paragraph is much better-still worry it might distract the reader from our approach?}
Specifically, Patch-Cleanser's \emph{double-masking} approach is based on zero-ing out  two different contiguous blocks of an input image, hopefully to remove the adversarial patch. For each one-masked image, if for all possible locations of the second mask, the prediction model outputs the same classification, it means that the first mask removed the adversarial patch, and the agreed-upon prediction is correct. Any disagreements in these predictions imply that the mask was not covered by the first patch. 

\paragraph{Our Contributions}
When no predictor is perfectly correct on {\em all} perturbations (e.g., all two-mask operations), which we refer to as the 
the \textit{non-realizable} or \emph{agnostic} setting, we exhibit an example where plain $\ERM$ on the augmented dataset fails (See \prettyref{exmpl:ERM-failure}%and \prettyref{fig:example}
). At  a high-level, the main issue is that plain $\ERM$ on the augmented data-set treats all mistakes equally and so this could lead to learning a predictor with very high robust loss, i.e. on {\em many} training examples.
%This type of lower bound exposes some of the fundamental complexities in adversarial robustness in that 
%given an input $x$ a learner needs to be robust against all of a certain perturbation ball $U(x)$ compared to the adversary
%who only needs to find one successful perturbation per input.
%there is a two-mask operation where the predictor makes a mistake, as opposed to learning a predictor with low robust loss, i.e. there are a \emph{few} training examples where the predictor makes mistakes on many of the two-mask operations. 
%Our first contribution is to investigate the following theoretical question:
%\begin{center}
%\textit{Can we use $\ERM$ to learn predictors robust against adversarial \\perturbations in the non-realizable setting?}
%\textit{Can we extend the reduction proposed by~\citet{xiang2022patchcleanser} to the non-realizable setting?}
%\end{center}
%\omcomment{In Q above: patches-->examples. We are posing a general theoretical question.}
Our first contribution is to investigate whether the reduction proposed by~\citet{xiang2022patchcleanser} can be extended to the \emph{non-realizable} setting.
%That is, we are interested in learning a predictor that is adversarially robust to (polynomially many) perturbations only by interacting with an ERM algorithm which itself can be viewed as a non-robust learner. %While\ref{exmpl:ERM-failure} shows that performing ERM on an augmented training dataset (that contains all perturbations) can fail, %even when there is a robust predictor
%To answer this question, 
We answer this question affirmatively in\ref{sec:non-realizable-oracle}, by building upon a prior work by~\citet{DBLP:conf/colt/FeigeMS15}.%by using an algorithm proposed by~\citet{DBLP:conf/colt/FeigeMS15} for learning predictors robust against adversarial perturbations in the non-realizable setting and in the process extend their results. %In\ref{sec:feigecomparison}, we give a detailed comparison of our results in this setting with prior work.

Next, in Section \ref{sec:multi-robustness}, we consider a multi-group setting and investigate the question of \emph{agnostic multi-robust learning} using an $\ERM$ oracle. This question is inspired by the literature on \emph{multi-calibration} and \emph{multi-group learning} ~\citep{multicalib, multi-accuracy-Kim-etal,multiagnostic,multiagstrat, bugbounty}.
Our objective is that given a hypothesis class $\calH$ and a (potentially) rich collection of subgroups $\calG$, learn a predictor $h$ such that for each group $g\in\calG$, $h$ has low robust loss on $g\in \calG$. However, we highlight that the prior work on multi-group learning does not extend to the setting of robust loss since they do not consider adversarial perturbations of natural examples. To our knowledge, our work is the first to consider the notion of multi-group learning for robust loss. That being said we emphasize that there is a trade-off here; our guarantees are for the \emph{more challenging objective of robust loss}, but they are weaker than the ones given for PAC learning in the prior work. A detailed comparison is given in \ref{sec:related-work}. 
%Beyond this, in Section \ref{sec:multi-robustness}, we initiate the study of multi-group robust learning guarantees. We consider a generalized setting with multiple groups where the goal is to learn a predictor using an $\ERM$ oracle that simultaneously has low robust loss on all groups. 

Our motivation for studying multi-robustness is two-fold. First, to prohibit the adversary from targeting a specific demographic group
for adverse treatment. Additionally, it can increase the overall performance of the model by forcing the model to be robust on vulnerable examples. For instance, imagine a self-driving car system with a vision system recording a drive and we consider adversarial examples attacking individual frames of the video.
Ideally, the system would have robust performance over every frame. 
However, average robust error of $1\%$ could be very problematic if those errors instead of occurring uniformly then those errors concentrated on a specific adjacent set of frames.
%These concentrated errors would be more likely to cause the control system of the car to cause a catastrophic error. 
In this %car vision 
example, imagine that the protected groups are nearby frames so that we maintain smooth and reliable performance \emph{locally and globally}.
%This will be expanded on in Section \ref{sec:multi-robust}.

To achieve multi-robustness, using plain $\ERM$ can fail by \emph{concentrating} the overall robust loss on a \emph{few} groups,  instead of \emph{spreading}
the loss across \emph{many} groups. 
However, building on our algorithm in \ref{sec:non-realizable-oracle} 
we propose\ref{alg:boosting} that runs an additional layer of boosting with respect to groups to achieve multi-robustness guarantees across groups. We propose two types of multi-robustness guarantees, 
the first one is a randomized approach that guarantees the expected robust loss on each group is low (\ref{thm:generalization-multi-groups}). Next, we add a de-randomization step to derive deterministic guarantees for the robust loss incurred on each group (\ref{thm:generalization-multi-groups-deterministic}).

\subsection{Related Work}
\label{sec:related-work}
\paragraph{Patch Attacks} Patch attacks \citep{brown2017adversarial,karmon2018lavan,yang2020patchattack} 
are an important threat model in the general field of test-time evasion attacks \citep{goodfellow2014explaining}. 
%In a patch attack, the adversary replaces a contiguous block of pixels with an adversarially crafted pattern. 
Patch attacks realize adversarial test time evasion attacks to computer vision systems in the wild
by printing and attaching a patch to an object.
%\kmsdelete{To secure the performance of computer vision systems against patch-attacks,}
%\kmsedit{
To mitigate this threat, 
%}
there has been an active line of research for providing certifiable robustness guarantees against them \citep[see e.g.,][]{minorityreport, patchguard, patchguard++, bagcert, clippedbag,chiang2020}.

\paragraph{Adversarial Learning using $\ERM$}Recent work by \cite{DBLP:conf/colt/FeigeMS15} gives a reduction algorithm for adversarial learning using an ERM oracle, but their guarantee is only for finite hypothesis classes. We observe in this work that we can apply their reduction algorithm to our problem, and along the way, we extend the guarantees of their algorithm. A more detailed comparison is provided in \ref{sec:feigecomparison}.

\paragraph{Multi-group Learning}
%\kmscomment{I think this section needs a bit more editing}
Interestingly, the notion of multi-robustness has connections with a thriving area of work in algorithmic fairness centered on the notion of multi-calibration ~\cite{multicalib, multi-accuracy-Kim-etal,multiagnostic,multiagstrat, bugbounty, gopalan2022lossminimizationlensoutcome}. %Generally speaking, these notions require a fairness or accuracy constraint hold on a large collection of sub-groups. 
The promise of these multi-guarantees, given a rich set of groups, is to ensure uniformly acceptable performance
on many groups simultaneously. 

Specifically, \cite{multiagnostic} show how to learn a predictor such that the loss experienced by every group is not much larger than the best
possible loss for this group within a given hypothesis class. However, we highlight that the prior work on multi-group learning does not extend to the setting of robust loss since their goal is not to minimize the robust loss by taking into consideration different perturbations of natural examples. In contrast, our approach can achieve multi-robustness guarantees by utilizing two layers of boosting to ensure `emphasis' on both specific groups and the adversarial perturbations.

\cite{multiagstrat,bugbounty} study the problem of minimizing a general loss function over a collection of subgroups. Their approach can capture the robust loss, however, the main distinction between their algorithm and our approach is that unlike them, we do not use group membership during the test time. This is essential when groups correspond to protected features, and therefore in some scenarios, it would be undesirable to incorporate them in decision models. Additionally, if we interpret some of the groups in our setting as objects to be classified like a stop-sign group or fire-hydrant group, %(and we want low robust loss on both),
then an approach that needs to detect group membership is too strong an assumption since the correct classification of those objects is our original goal.

However, we highlight that there is a trade-off here; To our knowledge, our work is the first one to achieve guarantees for the \emph{more challenging objective of robust learning without having access to the group membership of examples} but at the cost of achieving a weaker upper bound on the robust loss incurred on each group compared to the previous work on multi-group PAC learning. A detailed comparison is given in\ref{sec:comparison-prior-work-multi-group-learning}.

%More specifically,~\citet{multiagnostic} study agnostic multi-group \emph{PAC learning} and guarantee that for each group $G_j$ in a collection of groups $\calG$:
%\[\Ex\insquare{\ell(f(x),y)|x\in G_j}\leq \min_{h_G\in\calH}\Ex\insquare{\ell(h_{G_j}(x),y)|x\in G_j}\]
%That is, the hypothesis $f$ must compete against a hypothesis $h_{G_j}\in \calH$ trained specifically to minimize the error over the group $G_j\in\calG$, for every group in the collection. \emph{However, their results do not extend to the case of robust loss.}

%In contrast, we study the following notion for the \emph{more challenging objective of robust learning} over a collection of groups. For all $G\in\calG$:
 %\[\Ex\insquare{\ell(f(x),y)|x\in G}\leq \min_{h\in\calH}\max_{G^*\in\calG}\Ex\insquare{\ell(h_G(x),y)|x\in G^*}\]
 %We leave it as an open question to study whether our upper bounds for the robust loss over a collection of groups can be strengthened.

\section{Setup and Notation}
%\section{Setup and Notation} 
%\sacomment{define robust loss?}
Let $\calX$ denote the instance space and $\calY$ denote the label space. Our main objective is to be robust against adversarial patches $\calA:\calX\to 2^\calX$, where $\calA(x)$ represents the (potentially infinite) set of adversarially patched images that an adversary might attack with at test-time on input $x$. \cite{xiang2022patchcleanser} showed that even though the space of adversarial patches $\calA(x)$ can be exponential or infinite, one can consider a ``covering'' %set 
function $\calU:\calX\to 2^\calX$ of masking operations on images where $\abs{\calU(x)}$ shows the covering set on input image $x$ and is polynomial in the image size. Thus, for the remainder of the paper, we focus on the task of learning a predictor robust to a perturbation set $\calU:\calX \to 2^{\calX}$, where %$\calU(x)\subseteq \calX$
$\calU(x)$is the set of allowed masking operations that can be performed on $x$. We assume that $\calU(x)$ is finite where $\abs{\calU(x)} \leq k$. 

We observe $m$ iid samples $S\sim \calD^m$ from an unknown distribution $\calD$, and our goal is to learn a predictor $\hat{h}$ achieving small robust risk:
\begin{align}
\label{eqn:robrisk}
\mathbb{E}_{(x,y)\sim \calD} [ \max_{z\in \calU(x)} \ind[\hat{h}(z)\neq y]].
\end{align}

Let $\calH\subseteq \calY^\calX$ be a hypothesis class, and denote by $\vc(\calH)$ its VC dimension. Let $\ERM_\calH$ be an $\ERM$ oracle for $\calH$ that returns a hypothesis $h\in \calH$ that minimizes empirical loss. For any set arbitrary set $W$, denote by $\Delta(W)$ the set of distributions over $W$. 

In\ref{sec:non-realizable-oracle}, we focus on a single-group setting where the benchmark $\OPT_{\calH}$ is defined as follows:
\begin{equation}
\label{eqn:opt}
    \OPT_{\calH} \triangleq \min_{h\in \calH} \mathbb{E}_{(x,y)\sim \calD} \max_{z \in \calU(x)} \ind\insquare{h(z)\neq y}. 
\end{equation}

In\ref{sec:multi-robustness}, we consider a multi-group setting, where the instance space $\mathcal{X}$ is partitioned into a set of $g$ groups $\mathcal{G}=\{G_1,\dots,G_g\}$. These groups solely depend on the features $x$ and not the labels. The goal is to learn a predictor that has low robust loss on all the groups simultaneously. In this setup, the benchmark $\OPTDMax$ is as follows:
\begin{align}
\label{optmax}
   \OPTDMax =  \min_{h\in\calH} \max_{j\in[g]} \Ex_{(x,y)\sim D}\insquare{\max_{z\in \calU(x)} \ind[h(z)\neq y] \big| x\in G_j}
\end{align}

%% file: rob_single-population.tex
%\section{MINIMIZING ROBUST LOSS USING AN ERM ORACLE \label{erm}}
\section{Minimizing Robust Loss Using an ERM Oracle \label{erm}}
\label{sec:non-realizable-oracle}

First, we show an example where the approach of \cite{xiang2022patchcleanser} of calling $\ERM_\calH$ on the inflated dataset, i.e., original training points plus all possible perturbations resulting from the allowed masking operations, fails by obtaining %the largest possible gap 
a multiplicative gap of $k-1$ %in true robust loss and apparent 0-1 loss on the inflated dataset.
in the robust loss between the optimal robust classifier and the classifer returned by $\ERM_\calH$, where $k$ is the size of the perturbation sets. This gap exists since $\ERM$ can exhibit a solution that incorrectly classifies at least one perturbation per natural example, while there is a robust classifier that concentrates error on one natural example, thus getting low robust loss.

\begin{example}
\label{exmpl:ERM-failure}
%We exhibit an example that achieves a multiplicative gap of $k-1$ in the robust loss gap between the inflated $\ERM$ solution and the optimal robust classifier, where $k$ is the size of the perturbation sets.
%This gap exists because $\ERM$ can exhibit a solution that incorrectly classifies at least one perturbation per natural example, while there is a robust classifier that concentrates error on one natural example, thus getting low robust loss. %Our only/primary constraint for this problem is that the perturbation balls need to be finite. 
%We can even show this type of example in one dimension with $x \in [-2,2]$ in Figure \ref{fig:example}. 
Consider the following example in $\mathcal{R}$. There is a training set $\{z_1,\cdots,z_{2n}\}$ of original examples, where examples $\{z_1,\cdots,z_n\}$ are positively labeled and are located at $x=1$. $\{z_{n+1},\cdots,z_{2n}\}$ are negatively labeled and are at $x=-1$. Each example $z_i$ has $k=n$ perturbations denoted by $\{z_{i,1},\cdots,z_{i,k}\}$. 

For each of the negative examples $\{z_{n+1},\cdots,z_{2n-1}\}$, all their perturbations are at $x=-0.75$. For the negative example $z_{2n}$, all its perturbations, i.e. $\{z_{2n,1},\cdots,z_{2n,k}\}$, are at $x=0$. For each positive example $z_i$ where $i\in\{1,\cdots,n-1\}$, one of their perturbations $z_{i,1}$ is at $x=0$ and the rest, i.e. $\{z_{i,2},\cdots,z_{i,k}\}$, are at $x=0.75$. For the positive example $z_n$, all its perturbations $z_{n,1},\cdots,z_{n,k}$ are at $x=0.75$.

 The adversarial training procedure considered in the paper by \cite{xiang2022patchcleanser} runs ERM on the augmented dataset (original examples and all their perturbations) to minimize the 0/1 loss. ERM finds a threshold classifier $h_{ERM}$ with threshold $\tau=\eps_1$ for any $0<\eps_1<0.75$ that classifies any points with $x<\tau$ as negative and points with $x\geq \tau$ as positive. As a result, for each positive example $z_i$ for $i\in\{1,\cdots,n-1\}$, the perturbation $z_{i,1}$ is getting classified mistakenly which causes a robust loss on $z_i$. Therefore, $h_{ERM}$ has a robust loss of $(n-1)/2n$  since $n-1$ of the positive examples are not robustly classified. 
However, there exists a threshold classifier $h^*$ with threshold $\tau=\eps_2$ for any $-0.75<\eps_2<0$ that only makes mistakes on perturbations of $z_{2n}$ and thus has a robust loss of $1/2n$. However, its 0/1 loss is higher than $h_{ERM}$ and therefore ERM does not pick it.
Therefore, ERM can be suboptimal up to a multiplicative factor of $n-1$ for any arbitrary value of $n$. An illustration is given in \ref{fig:agnosticexample}.

%In\ref{fig:example}, blacks points indicate adversarial perturbations. There are $2N$ examples, where half of them are true positives and the rest are true negatives. Each example has exactly $k=N$ perturbations. For $N-1$ of the true negatives, all their perturbations  are in $[-1,-1/2]$ and thus are benign.
%However, for one of the true negatives $x_{neg}^{*}$, all $N$ of its perturbations land exactly at $0$. 
%For the positive case, this is flipped. There is one example $x_{pos}^{*}$ where all its perturbations are in $[1/2,1]$. For every other true positive example, however, one perturbation is at $0$ and the rest are in $[1/2,1]$.  $\ERM$ classifies the points at $0$ as negative points and exhibits $h_{ERM}$. This induces a high robust loss since $N-1$ of the true positives are not robustly classified.
%Thus $h_{ERM}$ has robust loss $\frac{N-1}{2N} \sim 50 \%$ while the optimal robust classifier $h_{ROB}^{*}$ correctly classifies all points (natural and adversarial) other than $x_{neg}^{*}$ and so has robust loss $\frac{1}{2N}$, which obtains the multiplicative gap of $k-1$.

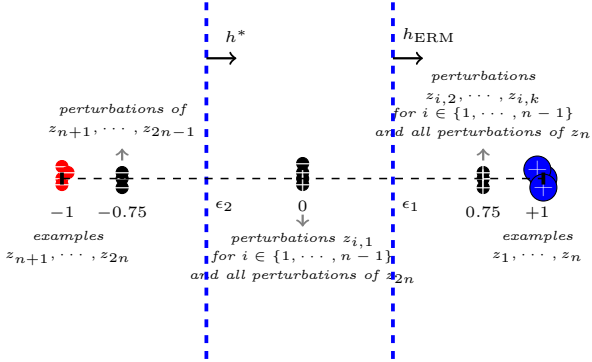
\begin{figure}[ht]

\centering
\begin{tikzpicture}[scale=0.8]

%%%%%-1
 \draw[red, fill=red,text=white] (-4,0) circle (0.1cm);
 %\node[white] at (-4,0) {\small{$-$}};
 \node[white] at (-4,0) {$-$};

 \draw[red, fill=red,text=white] (-4,0.2) circle (0.1cm);
 %\node[white] at (-4,0.2) {\small{$-$}};
 \node[white] at (-4,0.2) {$-$};

 \draw[red, fill=red,text=white] (-3.9,0.1) circle (0.1cm);
 \node[white] at (-3.9,0.1) {$-$};

 \draw[red, fill=red,text=white] (-4,-0.1) circle (0.1cm);
 \node[white] at (-4,-0.1) {$-$};
 %\node[below] at (-4,-0.3) {-1}
 
 \draw[line width=0.5mm] (-4,-0.1) -- (-4,0.1);
 \node[below] at (-4,-0.3) {\tiny{$-1$}};
 \node[below,align=center,font=\fontsize{5}{7}\itshape\selectfont] at (-3.9,-0.7) {examples\\ $z_{n+1},\cdots,z_{2n}$};
 
%\draw[->,line width=0.3mm,gray] (-4,-0.6)--(-4,-0.8);
%%%%-0.75
\draw[black, fill=black,text=white] (-3,0) circle (0.1cm);
\node[white] at (-3,0) {$-$};

 \draw[black, fill=black,text=white] (-3,0.1) circle (0.1cm);
\node[white] at (-3,0.1) {$-$};

 \draw[black, fill=black,text=white] (-3,-0.15) circle (0.1cm);
\node[white] at (-3,-0.15) {$-$};
\node[below] at (-3,-0.3) {\tiny{$-0.75$}};
\node[above,align=center,font=\fontsize{5}{7}\itshape\selectfont] at (-3,0.5) {perturbations of\\ $z_{n+1},\cdots,z_{2n-1}$};
\draw[->,line width=0.3mm,gray] (-3,0.3)--(-3,0.5);
%%%%0.75
\draw[black, fill=black,text=white] (+3,0) circle (0.1cm);
\node[white] at (+3,0) {$+$};

 \draw[black, fill=black,text=white] (+3,0.1) circle (0.1cm);
\node[white] at (+3,0.1) {$+$};

 \draw[black, fill=black,text=white] (+3,-0.15) circle (0.1cm);
\node[white] at (+3,-0.15) {$+$};
\node[below] at (+3,-0.3) {\tiny{$0.75$}};
\node[above,align=center,font=\fontsize{5}{7}\itshape\selectfont] at (+3,0.5) {perturbations \\ $z_{i,2},\cdots,z_{i,k}$\\ for $i\in\{1,\cdots,n-1\}$\\ and all perturbations of $z_n$};
\draw[->,line width=0.3mm,gray] (3,0.3)--(3,0.5);

%%%%0
\draw[black, fill=black,text=white] (0,0) circle (0.1cm);
\node[white] at (+3,0) {$+$};

 \draw[black, fill=black,text=white] (0,0.1) circle (0.1cm);
\node[white] at (0,0.1) {$+$};

 \draw[black, fill=black,text=white] (0,-0.1) circle (0.1cm);
\node[white] at (0,-0.1) {$+$};

 \draw[black, fill=black,text=white] (0,0.25) circle (0.1cm);
\node[white] at (0,0.25) {$-$};

 %\draw[black, fill=black,text=white] (0.1,-0.1) circle (0.1cm);
%\node[white] at (0.1,-0.1) {\small{$-$}};

 %\draw[black, fill=black,text=white] (0,-0.25) circle (0.1cm);
%\node[white] at (0,-0.25) {\small{$+$}};

\node[below] at (0,-0.2) {\tiny{$0$}};
\node[above,align=center,font=\fontsize{5}{7}\itshape\selectfont] at (0,-1.9) {perturbations $z_{i,1}$\\ for $i\in\{1,\cdots,n-1\}$\\ and all perturbations of $z_{2n}$};

\draw[->,line width=0.3mm,gray] (0,-0.6)--(0,-0.8);
%%%%%%+1
\node[circle,draw, minimum size = 0.1cm, inner sep=0pt,fill=blue] (C) at  (+4,0) [text=white] {\small{+}};

\node[circle,draw, minimum size = 0.1cm, inner sep=0pt,fill=blue] (C) at  (+3.9,0.15) [text=white] {\small{+}};

\node[circle,draw, minimum size = 0.1cm, inner sep=0pt,fill=blue] (C) at  (+4,-0.15) [text=white] {\small{+}};
\node[below] at (3.9,-0.3) {\tiny{$+1$}};
\node[below,align=center,font=\fontsize{5}{7}\itshape\selectfont] at (3.9,-0.7) {examples\\ $z_{1},\cdots,z_{n}$};

\draw[line width=0.5mm] (4,-0.1) -- (4,0.1);

%%%%%%%%%1.5
\draw[line width=0.5mm,blue,dashed] (1.5,-3) -- (1.5,3);
%\draw[line width=0.5mm,blue] (1.5,-3) -- (1.5,3);
\node[below] at (1.8,-0.2) {\tiny{$\eps_1$}};

\draw[->,line width=0.3mm] (1.5,2) -- (2,2);
\node[above] at (2.1,2.1) {\tiny{$h_{\text{ERM}}$}};

%%%%%%%%%-1.5
\draw[line width=0.5mm,blue,dashed] (-1.6,-3) -- (-1.6,3);
%\draw[line width=0.5mm,blue] (-1.6,-3) -- (-1.6,3);
\node[below] at (-1.3,-0.2) {\tiny{$\eps_2$}};

\draw[->,line width=0.3mm] (-1.6,2) -- (-1.1,2);
\node[above] at (-1.1,2.1) {\tiny{$h^*$}};

\draw[dashed, line width=0.2mm] (-4,0) -- (4,0);
%\draw[line width=0.2mm] (-4,0) -- (4,0);

\end{tikzpicture}
\caption{
$\ERM$ failure mode in the robustly un-realizable case. Blue, red, and black points show respectively original examples with a positive label, original examples with a negative label, and perturbations of original examples.
}
\label{fig:agnosticexample}
\end{figure}
\end{example}

Next, we present our first contribution: %an algorithm
we show in\ref{thm:generalization-FMS} that \ref{alg:FMS} proposed by 
\cite{DBLP:conf/colt/FeigeMS15} learns a predictor that is simultaneously robust to a set of (polynomially many) masking operations, using an $\ERM_\calH$ oracle. The algorithm is based on prior work% due to \citet{DBLP:conf/colt/FeigeMS15}
, but the analysis and application are novel in this work. A detailed comparison with \cite{DBLP:conf/colt/FeigeMS15} is given in \ref{sec:feigecomparison}. The main interesting feature of this algorithm is that it achieves stronger robustness guarantees in the non-realizable regime when $\OPT_\calH \gg 0$, where the approach of \cite{xiang2022patchcleanser} can fail as mentioned in \ref{exmpl:ERM-failure}. 
%\SetKw{STATE}{}
\begin{algorithm}%[H]
%\begin{algorithmic}[1]
\SetKw{INPUT}{Input}
\SetKw{OUTPUT}{Output}
\caption{\cite{DBLP:conf/colt/FeigeMS15}}
\label{alg:FMS}
  \INPUT weight update parameter $\eta>0$, number of rounds $T$, and training dataset $S=\{(x_1,y_1),\dots, (x_m,y_m)\}$ and corresponding weights $p_1,\cdots,p_m$\;
  
  %and corresponding weights $p_1,\dots,p_m$.}
  %\STATE 
  Set $w_1(z, (x,y)) = 1$, for each $(x,y)\in S, z\in \calU(x)$.\;
  
  %\STATE 
  Set $P^1(z,(x,y)) = \frac{w_1(z,(x,y))}{\sum_{z'\in\calU(x)} w_1(z',(x,y))}$, for each $(x,y)\in S, z\in \calU(x)$.\;
  
\For{each $t\in \{1,\cdots T\}$}{
%\STATE 
Call $\ERM$ on the empirical weighted distribution:\;
%\STATE 
{\small\[h_t = \argmin_{h\in\mathcal{H}} \sum_{(x,y)\in S} \sum_{z\in \calU(x)} {p_{(x,y)} }P^t(z,(x,y)) \ind\insquare{h_{t}(z)\neq y}\]}\;
\For{each $(x,y)\in S$ and $z\in\calU(x)$}{
%\STATE 
{\small $w_{t+1}(z,(x,y)) = (1+\eta \ind\insquare{h_{t}(z)\neq y}) \cdot w_{t}(z, (x,y))$}\;

%\STATE 
$P^{t+1}(z,(x,y))=\frac{w_t(z,(x,y))}{\sum_{z'\in\calU(x)} w_t(z',(x,y))}$\;

}
}
%\ENDFOR
%\ENDFOR
\OUTPUT The majority-vote predictor $\MAJ(h_1,\dots, h_T)$. \\
%\end{algorithmic}
\end{algorithm}

\begin{thm}
\label{thm:generalization-FMS}
Set $T(\eps) = \frac{32 \ln k}{\eps^2}$ and $m(\eps, \delta) = O\inparen{\frac{\vc(\calH)(\ln k)^2}{\eps^4}\ln \inparen{\frac{\ln k}{\eps^2}}+\frac{\ln(1/\delta)}{\eps^2}}$. Then, for any distribution $\calD$ over $\calX\times \calY$, with probability at least $1-\delta$ over $S\sim \calD^{m(\eps,\delta)}$, running \ref{alg:FMS} {where $p_{(x,y)}=1/m$ for all $(x,y)\in S$ }for $T(\eps)$ rounds produces $h_1,\dots,h_{T(\eps)}$ satisfying:

\small\[\Ex_{(x,y)\sim \calD} \insquare{ \max_{z\in \calU(x)} \ind\insquare{\MAJ(h_1,\dots, h_{T(\eps)})(z)\neq y} } 
\leq 2\OPT_{\calH} + \eps\]
where $\MAJ(h_1,\dots, h_{T(\eps)})$ shows the majority-vote of predictors $h_1,\dots, h_{T(\eps)}$.
\end{thm}

\begin{rem}
In the approach proposed by \cite{xiang2022patchcleanser}, the robust loss with respect to the (exponentially many) patches is upper bounded by the robust loss with respect to the (polynomially many) masking operations. Therefore, \ref{thm:generalization-FMS} implies that the robust loss against patches is at most $2\OPT_{\calH} + \eps$.
\end{rem}

\subsection{Comparison with prior related work} 
\label{sec:feigecomparison}
As presented, \cite{DBLP:conf/colt/FeigeMS15} only considered \emph{finite} hypothesis classes $\calH$ and provided generalization guarantees depending on $\log\abs{\calH}$. On the other hand, we consider here infinite classes $\calH$ with bounded VC dimension and provide tighter robust generalization bounds (see \ref{thm:generalization-FMS}). We would also like to highlight another difference. Given an output of $h_1,\dots,h_T$ from \ref{alg:FMS}, the guarantee provided by \cite{DBLP:conf/colt/FeigeMS15} is on average and does not exactly capture the notion of robust loss %for a mixed strategy
%on average,
i.e. the loss on input $x$ is $\sup_{z\in\calU(x)} \frac{1}{T}\sum_{t=1}^{T} \ind[h_t(z)\neq y]$ (\ref{lem:FMS} states their result). We emphasize that this is different from the \emph{robust loss} guarantee that we obtain in \ref{thm:generalization-FMS} for a %pure strategy:
single classifier, i.e. the loss on input 
$x$ is captured as $\sup_{z\in\calU(x)} \ind[\MAJ(h_1,\dots,h_T)(z)\neq y]$. %(\prettyref{thm:generalization-FMS}). 
%This can also be seen as guarantees for two different versions of robust loss. 
In particular, unlike the %``classic'' adversarial setting 
guarantee provided by \cite{DBLP:conf/colt/FeigeMS15} in which the adversary chooses $z\in\calU(x)$ and then we can probabilistically choose a classifier to classify it, to implement the Patch-Cleanser reduction we need a single classifier that is \emph{simultaneously} correct on \emph{all} 
$z\in\calU(x)$. Because of the difference in guarantees derived, we incur a multiplicative factor of 2 compared with their bound. 

The robust learning guarantee \cite[Theorem 2]{attias2022improved} assumes access to a \emph{robust} $\ERM$ oracle, which minimizes the robust loss on the training dataset. On the other hand, at the expense of higher sample complexity, we provide a robust learning guarantee using only an $\ERM$ oracle which is a more common and simpler assumption in the challenging \emph{non-realizable} setting. Prior work due to \cite{DBLP:conf/nips/MontasserHS20} considered using an $\ERM$ oracle for robust learning but only in the simpler realizable setting (when $\OPT_\calH=0$).
%\saedit{Additionally,~\cite{DBLP:conf/colt/FeigeMS15} allow the learner to pick a mixed-strategy. In contrast, our learner's algorithm is a pure strategy, i.e. majority-vote classifier. This results in losing a factor of $2$ compared to their result. }

%\kmsedit{Our guarantees with respect to our baseline $\OPT_{\calH}$ can be interpreted as being more closely
%aligned with the typical adversarial robustness formalization \cite{pmlr-v99-montasser19a}, since the adversary
%must ultimately commit to a single attack and we assume an adversary that always plays second.}

\subsection{Proof of \ref{thm:generalization-FMS}}
Before proceeding with the proof of\ref{thm:generalization-FMS}, we describe at a high-level the proof strategy.
%We exhibit a sketch for the proof of\ref{thm:generalization-FMS} and defer the details to the Appendix. 
The main insight is to solve a finite zero-sum game. In particular, our goal is to find a mixed-strategy over the hypothesis class that is approximately close to the value of the game:
\[\OPT_{S,\calH} \triangleq \min_{h\in \calH} \frac{1}{m}\sum_{i=1}^{m} \max_{z_i\in \calU(x_i)} \ind\insquare{h(z_i)\neq y_i}.\]

We observe that\ref{alg:FMS} due to \cite{DBLP:conf/colt/FeigeMS15} solves a similar finite zero-sum game (see\ref{lem:FMS}), and then we relate it to the value of the game we are interested in (see \ref{lem:opt}). Combined together, this only establishes that we can minimize the robust loss on the empirical dataset using an $\ERM$ oracle. We then appeal to uniform convergence guarantees for the robust loss in\ref{lem:unif-robloss} to show that, with a large enough training data, our output predictor achieves robust risk that is close to the value of the game. 

\begin{lem}
\label{lem:opt}
For any dataset $S %=\SET{(x_1,y_1),\dots,(x_m,y_m)}\in (\calX\times\calY)^m$,
=\{(x_1,y_1),\dots,(x_m,y_m)\}\in (\calX\times\calY)^m$ {with corresponding weights $p_1,\cdots,p_m=1/m$},
\begin{align*}
    &\OPT_{S,\calH} = \min_{h\in \calH} \frac{1}{m}\sum_{i=1}^{m} \max_{z_i\in \calU(x_i)} \ind\insquare{h(z_i)\neq y_i} \\
    &\geq
    \min_{Q\in \Delta(\calH)} \max_{\substack{P_{1}\in \Delta(\calU(x_1)),\\ \dots\\ P_{m} \in \Delta(\calU(x_m))}} \frac{1}{m} \sum_{i=1}^{m} \Ex_{z_i\sim P_i } \Ex_{h\sim Q} \ind\insquare{h(z_i)\neq y_i}
\end{align*}
\end{lem}

\begin{lem} [\cite{DBLP:conf/colt/FeigeMS15}]
\label{lem:FMS}
For any data set $S %=\SET{(x_1,y_1),\dots,(x_m,y_m)}\in (\calX\times\calY)^m$,
=\{(x_1,y_1),\dots,(x_m,y_m)\}\in (\calX\times\calY)^m$ {with corresponding weights $p_1,\cdots,p_m=1/m$}, running\ref{alg:FMS} for $T$ rounds produces a mixed-strategy $\hat{Q} = \frac{1}{T} \sum_{t=1}^{T} h_t \in \Delta(\calH)$ satisfying:
{\small
\begin{align*}
    &\max_{\substack{P_1\in \Delta(\calU(x_1)),\\ \dots,\\ P_m\in \Delta(\calU(x_m))}} \frac{1}{m}\sum_{i=1}^{m} \Ex_{z_i\sim P_i} \frac{1}{T} \sum_{t=1}^{T} \ind\insquare{h_t(z_i)\neq y_i} \\
    &\leq
    \min_{Q\in \Delta(\calH)} \max_{\substack{P_{1}\in \Delta(\calU(x_1)),\\ \dots,\\ P_{m} \in \Delta(\calU(x_m))}} \frac{1}{m} \sum_{i=1}^{m} \Ex_{z_i\sim P_i } \Ex_{h\sim Q} \ind\insquare{h(z_i)\neq y_i} + \\&
    2\sqrt{\frac{\ln k}{T}}
\end{align*}
}%
\end{lem}

\removed{
\begin{lem} [Extension to weighted samples]
For any data set $S %=\SET{(x_1,y_1),\dots,(x_m,y_m)}\in (\calX\times\calY)^m$
=\{(x_1,y_1),\dots,(x_m,y_m)\}\in (\calX\times\calY)^m$ and any corresponding weights $p_1,\dots, p_m > 0$ such that $\sum_{i=1}^{m} p_i = 1$, running\ref{alg:FMS} for $T$ rounds produces a mixed-strategy $\hat{Q} = \frac{1}{T} \sum_{t=1}^{T} h_t \in \Delta(\calH)$ satisfying:
\begin{align*}
    \max_{P_1\in \Delta(\calU(x_1)),\dots,P_m\in \Delta(\calU(x_m))} &\sum_{i=1}^{m} p_i\cdot \Ex_{z_i\sim P_i} \frac{1}{T} \sum_{t=1}^{T} \ind\insquare{h_t(z_i)\neq y_i} \leq\\
    &\min_{Q\in \Delta(\calH)} \max_{P_{1}\in \Delta(\calU(x_1)),\dots, P_{m} \in \Delta(\calU(x_m))} \sum_{i=1}^{m} p_i\cdot \Ex_{z_i\sim P_i } \Ex_{h\sim Q} \ind\insquare{h(z_i)\neq y_i} + 2\sqrt{\frac{\ln k}{T}}.
\end{align*}
\label{lem:extension-FMS-weights}
\end{lem}
}

\begin{lem} [VC Dimension for the Robust Loss \citep{attias2022improved}]
\label{lem:unif-robloss}
For any class $\calH$ and any $\calU$ such that $\sup_{x\in\calX}\abs{\calU(x)}\leq k$, denote the robust loss class of $\calH$ with respect to $\calU$ by
\[\calL^{\calU}_{\calH} = \{(x,y)\mapsto \max_{z\in\calU(x)} \ind\insquare{h(z)\neq y}: h\in\calH\}.\]
Then, it holds that $\vc(\calL^{\calU}_{\calH})\leq O(\vc(\calH) \log(k))$. 
\end{lem}

We are now ready to proceed with the proof of\ref{thm:generalization-FMS}.
%\begin{proof}[Proof of\ref{thm:generalization-FMS}]
%\label{sec:proof-thm-generalization-FMS}
\begin{proof}[Proof of\ref{thm:generalization-FMS}]
Let $S\sim \calD^m$ be an iid sample from $\calD$, where the size of the sample $m$ will be determined later. By invoking\ref{lem:FMS} and\ref{lem:opt}, we observe that running\ref{alg:FMS} on $S$ {with corresponding weights $p_1,\cdots,p_m=1/m$} for $T$ rounds, produces $h_1,\dots, h_{T}$ satisfying

{\small
\[\max_{\substack{P_1\in \Delta(\calU(x_1)),\\ \dots,\\ P_m\in \Delta(\calU(x_m))}} \frac{1}{m}\sum_{i=1}^{m} \Ex_{z_i\sim P_i} \frac{1}{T} \sum_{t=1}^{T} \ind\insquare{h_t(z_i)\neq y_i} \leq \OPT_{S,\calH} + \frac{\varepsilon}{4}
\]
}
Next, the average robust loss for the majority-vote predictor $\MAJ(h_1,\dots, h_T)$ can be bounded from above as follows:
\begin{align*}
    &\frac{1}{m} \sum_{i=1}^{m} \max_{z_i\in \calU(x_i)} \ind\insquare{\MAJ(h_1,\dots,h_T)(z_i)\neq y_i}\\
    &\leq \frac{1}{m} \sum_{i=1}^{m} \max_{z_i\in \calU(x_i)}  2 \Ex_{t\sim [T]}\ind\insquare{h_t(z_i)\neq y_i}\\
    &= 2 \frac{1}{m} \sum_{i=1}^{m} \max_{z_i\in \calU(x_i)} \frac{1}{T} \sum_{t=1}^{T} \ind\insquare{h_t(z_i)\neq y_i}\\
    &\leq 2 \max_{\substack{P_1\in \Delta(\calU(x_1)),\\ \dots,\\ P_m\in \Delta(\calU(x_m))}} \frac{1}{m}\sum_{i=1}^{m} \Ex_{z_i\sim P_i} \frac{1}{T} \sum_{t=1}^{T} \ind\insquare{h_t(z_i)\neq y_i}\\
    &\leq 2 \OPT_{S,\calH} + \frac\varepsilon2.
\end{align*}
In the second line above, the factor $2$ shows up since for any arbitrary example $(z,y)$, if at least half the predictors make a mistake then the majority-vote is wrong, and otherwise it is correct. The factor $2$ is used as a correction so that RHS is bigger than LHS, where the edge case is exactly when half the predictors make a mistake.

Next, we invoke \ref{lem:unif-robloss} to obtain a uniform convergence guarantee on the robust loss. In particular, we apply\ref{lem:unif-robloss} on the \emph{convex-hull} of $\calH$: $\calH^{T} = \{\MAJ(h_1,\dots, h_T): h_1,\dots, h_T\in \calH\}$. By a classic result due to \cite{blumer:89}, it holds that $\vc(\calH^T)=O(\vc(\calH)T\ln T)$. Combining this with\ref{lem:unif-robloss} and plugging-in the value of $T= \frac{32 \ln k}{\varepsilon^2}$, we get that the VC dimension of the robust loss class of $\calH^T$ is bounded from above by
\[\vc(\calL_{\calH^T}^\calU) \leq O\inparen{\frac{\vc(\calH)(\ln k)^2}{\varepsilon^2}\ln\inparen{\frac{\ln k}{\varepsilon^2}}}.\]
Finally, using Vapnik's ``General Learning'' uniform convergence \citep{vapnik:82}, with probability at least $1-\delta$ over $S\sim \calD^m$ where $m =  O\inparen{\frac{\vc(\calH)(\ln k)^2}{\varepsilon^4}\ln \inparen{\frac{\ln k}{\varepsilon^2}}+\frac{\ln(1/\delta)}{\varepsilon^2}}$, it holds that
\begin{align*}
&\forall f\in \calH^T: \Ex_{(x,y)\sim \calD} \insquare{\max_{z\in\calU(x)}\ind\insquare{f(z)\neq y}}\\
&\leq \frac{1}{m}\sum_{i=1}^{m} \max_{z_i\in\calU(x_i)}\ind\insquare{f(z_i)\neq y_i} + \frac\varepsilon4
\end{align*}
This also applies to the particular output $\MAJ(h_1,\dots, h_T)$ of\ref{alg:FMS}, and thus
\begin{align*}
    &\Ex_{(x,y)\sim \calD} \insquare{ \max_{z\in \calU(x)} \ind\insquare{\MAJ(h_1,\dots, h_{T(\varepsilon)})(z)\neq y} }\\ 
    &\leq \frac{1}{m} \sum_{i=1}^{m} \max_{z_i\in \calU(x_i)} \ind\insquare{\MAJ(h_1,\dots,h_T)(z_i)\neq y_i} + \frac{\varepsilon}{4}\\
    &\leq 2\OPT_{S,\calH} + \frac\varepsilon2 + \frac\varepsilon4.
\end{align*}

Finally, by applying a standard Chernoff-Hoeffding concentration inequality, we get that $\OPT_{S,\calH} \leq \OPT_\calH + \frac\varepsilon8$. Combining this with the above inequality concludes the proof.
\end{proof}

%% file: rob_unified-boosting.tex
\section{Multi-Robustness Guarantees On a Set of Groups}
\label{sec:multi-robustness}
%In this section, we consider the problem of learning a predictor that has low robust loss across multiple groups. %This objective can be justified from a fairness perspective. For instance, when the groups correspond to sensitive social or demographic groups like gender and race, ensuring reliable performance on each group is crucial. However, note that even absent fairness requirements, our multi group scheme has important benefits since these groups can be arbitrary. A learner could thus intentionally select groups that are known to be especially vulnerable to adversarial attacks or important to defend.
%For instance, imagine a self driving car system with a vision system recording a drive and we consider adversarial examples attacking individual frames of the video.Ideally, the system would have robust performance over every frame. However, average robust error of $1\%$ could be very problematic if those errors instead of occurring uniformly then those errors concentrated on a specific adjacent set of frames.These concentrated errors would be more likely to cause the control system of the car to cause a catastrophic error. In this car vision example, imagine that the protected groups are nearby frames so that we maintain smooth and reliable performance \emph{locally and globally}.
%In this setting, if we use \ref{alg:FMS} to minimize the overall robust loss, it may result in concentrating the overall robust loss on a few groups, instead of spreading the loss across many groups. 
In this section, we propose a boosting algorithm that learns a predictor with a low robust loss on a collection of subgroups simultaneously. 
%First, in\ref{sec:multi-robust}, we investigate the case of disjoint groups and propose a two-layer boosting algorithm (\ref{alg:boosting}) that achieves multi-robustness on the training dataset $S$. We show\ref{alg:boosting} provides two multi-robustness guarantees: average multi-robustness(\ref{def:avg-multi-robustness}) and $\beta$-multi-robustness (\ref{defn:beta-multi-robustness}) for $\beta<12$.
First, we consider the case of disjoint groups and present our training-time algorithm for this case in\ref{sec:multi-robust}. 
\ref{sec:generalization-guarantees} provides generalization guarantees. In\ref{sec:overlapping-groups-reduction}, we show a reduction from overlapping groups to disjoint groups. In the following, first we formalize the notions of robust loss on a specific group and multi-robustness. 
%Hi saba!!

%\paragraph{Summary of Results.} \ref{sec:multi-robust} investigates the case of disjoint groups and proposes a two-layer boosting algorithm (\ref{alg:boosting}) that achieves multi-robustness on the training dataset $S$. First, we show that $\calH'=\{h_1,\dots,h_T\}$ returned by\ref{alg:boosting} is multi-robust on average (\ref{thm:randomized-multi-robustness}).\ref{thm:deterministic-multi-robustness} exhibits that the majority-vote classifier over $\calH'$, i.e. $\MAJ(h_1,\dots,h_T)$, obtains $\beta$-multi-robustness for $\beta<12$. We remark that although\ref{thm:randomized-multi-robustness} achieves a tighter upper bound on the multi-robustness guarantee,\ref{thm:deterministic-multi-robustness} gives a guarantee for the stronger notion of multi-robustness. In\ref{sec:overlapping-groups-reduction}, we show a reduction from overlapping groups to disjoint groups.\ref{sec:generalization-guarantees} provides generalization guarantees.

%First, we present our formal definitions for multi-robustness.
When the training dataset $S$ is partitioned into $g$ groups $\calG=\{G_1,\dots,G_g\}$,
%\kmsedit{These groups solely depend on the features $x$ and not the labels.}
the empirical robust loss of a predictor $h$ on group $G_j$ is defined as follows:
\begin{align}
&\RLoss_j(h)=\frac{1}{|G_j|}\sum_{(x,y)\in G_j}\max_{z\in\mathcal{U}(x)}\ind[h(z)\neq y]\label{eqn:unweighted-robust-loss}
\end{align}

The learning benchmark that we compete with on a dataset $S$ for the robust loss on each group is $\OPT^{S}_{\max}$ that is defined as follows:
%\begin{defn}
%\label{defn: optmax}
    %\[ 
\begin{align}
&\OPT^{S}_{\max}=\min_{h\in \mathcal{H}}\max_{j\in[g]}\frac{1}{|G_j|}\sum_{(x,y)\in G_j}\max_{z\in\mathcal{U}(x)}\ind[h(z)\neq y]
\label{defn: optmax}
\end{align}
    %\]
%\end{defn}

%In the following, we formalize the notion of multi-robustness over a set of groups $\mathcal{G}=\{G_1,\dots,G_g\}$:

\begin{defn}[Multi-Robustness]
\label{defn:multirob}
A hypothesis $h$ is multi-robust on a dataset $S$ if it achieves the following guarantee:
\begin{align*}
&\max_{j\in[g]}\frac{1}{|G_j|}\sum_{(x,y)\in G_j}\max_{z\in\mathcal{U}(x)}\ind[h(z)\neq y]\leq \OPT^S_{\max}+\eps 
%\label{eq:multi-robustness}
\end{align*}
\label{def:multi-robustness}
\end{defn}
%A hypothesis $h$ satisfies Definition \ref{defn:multirob} if it is within $\varepsilon$ robust loss of the min-max optimal classifier where the adversary gets to have two maximization options, e.g. maximizing over the worst-off group and for each example $x$ in that group, picking a worst-case perturbation. 

\begin{defn}[$\beta$-Multi-Robustness]
A hypothesis $h$ is $\beta$-multi-robust on a dataset $S$ if it achieves the following guarantee:
\begin{align*}
&\max_{j\in[g]}\frac{1}{|G_j|}\sum_{(x,y)\in G_j}\max_{z\in\mathcal{U}(x)}\ind[h(z)\neq y]\leq \beta(\OPT^{S}_{\max}+\eps) \label{eq:beta-multi-robustness}
\end{align*}
\label{defn:beta-multi-robustness}
\end{defn}

\begin{defn}[Multi-Robustness on Average]
A set of hypotheses $\mathcal{H'}=\{h_1,\dots,h_T\}$ is multi-robust on a dataset $S$ on average if the the following property holds:
\[\frac{1}{T}\max_{j\in [g]}\sum_{t=1}^T\RLoss_j(h_t)\leq \OPT^{S}_{\max}+\eps\]
\label{def:avg-multi-robustness}
\end{defn}

\begin{rem}
\ref{def:multi-robustness} is a stronger notion of multi-robustness compared to\ref{def:avg-multi-robustness}.
\end{rem}

\paragraph{Summary of Results.}
\ref{sec:multi-robust} investigates the case of disjoint groups and proposes a two-layer boosting algorithm (\ref{alg:boosting}) that achieves multi-robustness on the training dataset $S$. First, we show that $\calH'=\{h_1,\dots,h_T\}$ returned by\ref{alg:boosting} is multi-robust on average (\ref{thm:randomized-multi-robustness}).\ref{thm:deterministic-multi-robustness} exhibits that the majority-vote classifier over $\calH'$, i.e. $\MAJ(h_1,\dots,h_T)$, obtains $\beta$-multi-robustness for $\beta=2$. We remark that although\ref{thm:randomized-multi-robustness} achieves a tighter upper bound on the multi-robustness guarantee,\ref{thm:deterministic-multi-robustness} gives a guarantee for the stronger notion of multi-robustness. In\ref{sec:overlapping-groups-reduction}, we show a reduction from overlapping groups to disjoint groups.\ref{sec:generalization-guarantees} provides generalization guarantees for both notions of average multi-robustness and $\beta$-multi-robustness. %In\ref{sec:overlapping-groups-reduction}, we show a reduction from overlapping groups to disjoint groups.

%\begin{comment}
\subsection{Comparison to Prior Work on Multi-group Learning}
\label{sec:comparison-prior-work-multi-group-learning}
\cite{multiagnostic} study agnostic multi-group \emph{PAC learning} and their algorithm returns a hypothesis $h$ such that for each group $G_j$ in a collection of groups $\calG$:
%$\forall G_j\in \calG$:
\[\Ex\insquare{\ell(h(x),y)|x\in G_j}\leq \min_{h_{G_j}\in\calH}\Ex\insquare{\ell(h_{G_j}(x),y)|x\in G_j}\]
That is, the hypothesis $h$ must compete against a hypothesis $h_{G_j}\in \calH$
 trained specifically to minimize the error over the group $G_j\in\calG$, for every group in the collection. \emph{However, their results do not extend to the case of robust loss.} In contrast, in our notion of multi-robustness loss that holds for the \emph{more challenging objective of robust learning}, our benchmark is weaker (\ref{def:multi-robustness}). 
We leave it as an open question to study whether our upper bounds for the robust loss over a collection of groups can be strengthened.
%In contrast, we study the following notion for the \emph{more challenging objective of robust learning} over a collection of groups. For all $G\in\calG$:
% \[\Ex\insquare{\ell(f(x),y)|x\in G}\leq \min_{h\in\calH}\max_{G^*\in\calG}\Ex\insquare{\ell(h_G(x),y)|x\in G^*}\]
% We leave it as an open question to study whether our upper bounds for the robust loss over a collection of groups can be strengthened.
%\end{comment}

\subsection{Boosting algorithm achieving multi-robustness guarantees:}
\label{sec:multi-robust}

In this section, we present\ref{alg:boosting} that obtains %both multi-robustness on average and $\beta$-multi-robustness for $\beta<12$
multi-robustness guarantees on a set of \emph{disjoint} groups. The algorithm follows the idea proposed by \cite{freund1996game} that obtains boosting by playing a repeated game. Initially a sample set $S=\{(x_1,y_1),\dots,(x_m,y_m)\}$ partitioned into a set of disjoint groups $\mathcal{G}=\{G_1,\dots,G_{g}\}$ is received as input. %First, the weight of each group $G_j$ is set to $1/g$. At each round of boosting, initially, group weights are normalized. 
$P_j^t$ shows the normalized weight of group $G_j$ in step $t$. Initially, for each group $G_j$, $P^t_j=1/g$.
In each round $t$, the weight of each group gets split between its examples equally: $p_i = P^t_j/|G_j|$ where $(x_i,y_i)\in G_j$. Subsequently, an oracle call is made to{\ref{alg:FMS}} with sample weights $p_1,\dots,p_m$. %\sadelete{\ref{alg:weighted-FMS} is an extension of\ref{alg:FMS} to weighted samples.}
\ref{lem:avg-robust-loss-upper-bound} shows at each iteration $t$,{\ref{alg:FMS}} returns a hypothesis $h_t$ such that its average robust loss across the groups is at most $\OPT^{S}_{\max}+\eps$. In the next iteration $t+1$, for each group $G_j$, the weights of examples in $G_j$ get decreased by a multiplicative factor of 
$1-\delta m_j^{\text{rob}}(h_t)$ where $m_j^{\text{rob}}(h_t)=1-\RLoss_j(h_t)$ and $\delta=\sqrt{{\ln g}/{T}}$.\ref{thm:randomized-multi-robustness} exhibits that after %the average multi-robustness guarantee: for 
$T=\calO({\ln g}/{\eps^2})$ rounds,\ref{alg:boosting} outputs
a set of hypotheses $\mathcal{H'}=\{h_1,\dots,h_T\}$ such that for each group $G_j$ the average multi-robustness guarantee is obtained, i.e., $\frac{1}{T}\sum_{t=1}^T \RLoss_j(h_t)\leq \OPT^S_{\max}+\eps$.\ref{thm:deterministic-multi-robustness} provides that $\MAJ(h_1,\dots,h_t)$ achieves
$\beta$-multi-robustness guarantee for $\beta=2$.

\begin{algorithm}[H]
\SetKw{INPUT}{Input}
\SetKw{OUTPUT}{Output}
\caption{Boosting Algorithm Achieving Multi-Robustness}
\label{alg:boosting}
    \INPUT training dataset $S=\{(x_1,y_1),\dots,(x_m,y_m)\}$ partitioned into a set of groups $\{G_1,\cdots,G_g\}$\;
    
    %\STATE 
    Initially, $\forall 1\leq j\leq g: P_j^t = 1/g$\;
    
    \For{$t=1,\dots,T$}{
    %\STATE 
    $p_i = P^t_j/|G_j|$ where $(x_i,y_i)\in G_j$\;
    
    %\STATE 
    Call{\ref{alg:FMS}} on $S$ with weights $(p_1,\dots,p_m)$ for $T'=\frac{36\ln k}{\eps^2}$ rounds.\;
    
    %\STATE 
    Update $P^t_j,  \text{ for all }j\in[g]$:\;
    
    %\STATE 
    \[ P^{t+1}_j=\frac{P^t_j\cdot\left(1-\delta m_j^{\text{rob}}(h_t)\right)}{Z_t}\]
    where $m_j^{\text{rob}}(h_t)=1-\RLoss_j(h_t)$, $Z_t$ is a normalization factor, and     $\delta=\sqrt{\frac{\ln g}{T}}$.\;
    }
    %\ENDFOR
    \OUTPUT 
    $\mathcal{H'}=\{h_1,\cdots,h_T\}$
\end{algorithm}

\begin{rem}
We remark that the output of\ref{alg:boosting} is a set of majority-vote classifiers over $\calH$:
%\[
\begin{align*}
&\calH'=\Big\{\MAJ(h_{1,1},\dots, h_{1,T'}),\dots,\MAJ(h_{T,1},\dots, h_{T,T'})\\
&:\forall i\in[T], \forall j\in[T'], h_{i,j}\in \calH\Big\}
\end{align*}
%\]
\end{rem}

Before proving the multi-robustness guarantees, we show that\ref{lem:avg-robust-loss-upper-bound} holds. {In order to prove that\ref{lem:avg-robust-loss-upper-bound} holds, first we show in \ref{lem:extension-FMS-weights} that an extension of\ref{lem:FMS} holds when $p_1,\cdots,p_m$ are arbitrary weights such that $\sum_{i=1}^m p_i = 1$.} Next, we restate the guarantee of the Multiplicative Weights algorithm that is a generalization of \emph{Weighted Majority} algorithm \cite{littlestone1994weighted} and is equivalent to \emph{Hedge} developed by \cite{freund1997decision}.

\begin{lem} [Extension to general weights]
For any dataset $S =\{(x_1,y_1),\dots,(x_m,y_m)\}\in (\calX\times\calY)^m$ and any corresponding weights $p_1,\dots, p_m > 0$ such that $\sum_{i=1}^{m} p_i = 1$, running
{\ref{alg:FMS}}
for $T$ rounds produces a mixed-strategy $\hat{Q} = \frac{1}{T} \sum_{t=1}^{T} h_t \in \Delta(\calH)$ satisfying:
%{\small
\begin{align*}
   &\max_{\substack{P_1\in \Delta(\calU(x_1)),\\ \dots,\\P_m\in \Delta(\calU(x_m))}} \sum_{i=1}^{m} p_i\cdot \Ex_{z_i\sim P_i} \frac{1}{T} \sum_{t=1}^{T} \ind\insquare{h_t(z_i)\neq y_i} \\
   &\leq
    \min_{Q\in \Delta(\calH)} \max_{\substack{P_{1}\in \Delta(\calU(x_1)),\\ \dots, \\P_{m} \in \Delta(\calU(x_m))}} \sum_{i=1}^{m} p_i\cdot \Ex_{z_i\sim P_i } \Ex_{h\sim Q} \ind\insquare{h(z_i)\neq y_i}\\ 
    &+ 2\sqrt{\frac{\ln k}{T}}
\end{align*}
%}%
\label{lem:extension-FMS-weights}
\end{lem}
\begin{lem}
In each round $t$ of\ref{alg:boosting}, by making an oracle-call to{\ref{alg:FMS}} after $T'=\frac{4\ln k}{\eps^2}$ rounds, we can find a hypothesis $h_t$ is outputted such that $\E_{j\sim P^t}[\ell^{rob}_j(h_t)]=\sum_{j\in[g]}P_j^{t}\ell_j^{rob}(h_t)\leq \OPT^S_{\max}+\eps$.
\label{lem:avg-robust-loss-upper-bound}
\end{lem}

\begin{thm}[Mutiplicative Weights Algorithm \citep{kale2007efficient}]
\label{thm:MW_alg}
For any sequence of costs of experts $\vec{m}_1,\cdots,\vec{m}_T$ revealed by nature where all the costs are in $[0,1]$, the sequence of mixed strategies $\vec{p}_1,\cdots,\vec{p}_T$ produced by the Multiplicative Weights algorithm satisfies:
\[\sum_{t=1}^T \vec{m}_t\cdot \vec{p}_t\leq (1+\delta)\min_{\vec{p}}\sum_{t=1}^T\vec{m}_t\cdot \vec{p}+\frac{\ln n}{\delta}\]
where $n$ is the number of experts.
\end{thm}

\begin{thm}
\label{thm:avg-empirical-boosting}
When $T=\calO(\frac{\ln g}{\eps^2})$,\ref{alg:boosting} computes a set of hypotheses $\mathcal{H}'=\{h_1,\cdots,h_T\}$, such that for each group $G_j$, 
$\frac{1}{T}\sum_{t=1}^T\RLoss_j(h_t)\leq \OPT^S_{\max}+\eps$.
\label{thm:randomized-multi-robustness}
\end{thm}
\begin{proof}

In each iteration $t$, we define average loss and reward terms as follows:

\[L(P^t,h_t)=\E_{j\sim P_t}\Big[\RLoss_j(h_t)\Big]=\sum_{j\in[g]}P^t_j\RLoss_j(h_t),\] 
\[M(P^t,h_t)=\E_{j\sim P_t}\Big[m_j^{\text{rob}}(h_t)\Big]\]
Substituting $\RLoss_j(h_t)=1-m_j^{\text{rob}}(h_t)$ provides:
\begin{align*}
&M(P^t,h_t)=\sum_{j\in[g]}P^t_j(1-\RLoss_j(h_t))=1-\sum_{j\in[g]}P^t_j\RLoss_j(h_t)\\
&=1-L(P^t,h_t)
\end{align*}
Now by setting $T=\frac{9\ln g}{\eps^2}$ which implies that $\delta=\sqrt{\frac{\ln g}{T}}=\frac{\eps}{3}$, and by using the guarantee of\ref{thm:MW_alg}, the following bound is obtained.
\begin{align*}
&\frac{1}{T}\sum_{t=1}^T M(P^t,h_t)\leq \frac{(1+\delta)}{T}\min_{j\in[g]}\sum_{t=1}^T M(j,h_t)+\frac{\ln g}{\delta T}\\
&\rightarrow \frac{1}{T}\sum_{t=1}^T M(P^t,h_t)\leq \frac{1}{T}\min_{j\in[g]}\sum_{t=1}^T M(j,h_t)+\delta+\frac{\ln g}{\delta T}\\
%&\rightarrow \frac{1}{T}\sum_{t=1}^T M(P^t,h_t)\leq \frac{1}{T}\min_{j\in[g]}\sum_{t=1}^T M(j,h_t)+2\sqrt{\frac{\ln g}{T}}
&\rightarrow \frac{1}{T}\sum_{t=1}^T M(P^t,h_t)\leq \frac{1}{T}\min_{j\in[g]}\sum_{t=1}^T M(j,h_t)+\frac{2\eps}{3}
\end{align*}
where $M(j,h_t)$ is the reward term when the whole probability mass is concentrated on group $G_j$. Therefore for each group $j\in[g]$:
\begin{align}
&\frac{1}{T}\sum_{t=1}^T M(j,h_t)\geq \frac{1}{T}\sum_{t=1}^T M(P^t,h_t)-\frac{2\eps}{3}
\label{eqn:randomized-boosting-eq1}
\end{align}

\ref{lem:avg-robust-loss-upper-bound} provides that in each iteration $t$, $L(P^t,h_t)\leq \OPTSMax+\eps/3$ given that \sareplace{\ref{alg:weighted-FMS}}{\ref{alg:FMS}} is 
executed for $T'=\frac{36\ln k}{\eps^2}$ rounds.
Thus, at each iteration $t$, $M(P^t,h_t)\geq 1-(\OPTSMax+\eps/3)$. Therefore, $\frac{1}{T}\sum_{t=1}^T M(P^t,h_t)\geq 1-(\OPTSMax+\eps/3)$; combining with\ref{eqn:randomized-boosting-eq1} implies that:
\begin{align*}
&\frac{1}{T}\sum_{t=1}^T M(j,h_t)\geq \frac{1}{T}\sum_{t=1}^T M(P^t,h_t)-\frac{2\eps}{3}\\
&\geq 1-(\OPT^S_{\max}+\frac{\eps}{3})-\frac{2\eps}{3}=1-(\OPT^S_{\max}+\eps)
\end{align*}
Plugging in the definition of $L(P^t,h_t)$ implies that:
\[\frac{1}{T}\sum_{t=1}^T L(j,h_t)\leq \OPT^S_{\max}+\eps\]
Which concludes the proof.
\end{proof}

\begin{cor}
\ref{thm:randomized-multi-robustness} implies that if for each example a predictor is picked uniformly at random from $\calH'$ to predict its label, then for each group $G_j\in \calG$, the expected robust loss is at most $\OPT^S_{\max}+\eps$.
\label{cor:interpret-avg-loss}
\end{cor}

\begin{thm}
When $T=\calO(\frac{\ln g}{\eps^2})$,\ref{alg:boosting} computes a set of hypotheses $\calH'=\{h_1,\dots,h_T\}$ such that for each group $G_j$, $\RLoss_j(\MAJ(h_1,\cdots,h_T))\leq 2(\OPT^S_{\max}+\eps)$.
\label{thm:deterministic-multi-robustness}
\end{thm}

\begin{proof}
By\ref{thm:randomized-multi-robustness}, after $T=\calO(\frac{\ln g}{\eps^2})$ rounds, for each group $G_j$, $\frac{1}{T}\sum_{t=1}^T \RLoss_j(h_t)\leq \OPT^S_{\max}+\eps$. Therefore, %at most $1/c$ of the classifiers $\mathcal{H'}=\{h_1,\cdots,h_T\}$ have robust loss greater than $c(\OPT^S_{\max}+\eps)$ on $G_j$. Let's call the set of these classifiers $\mathcal{H}''$. 
the total number of robustness mistakes on $G_j$ across all the classifiers %in $\mathcal{H}'\setminus\mathcal{H}''$ 
$h_1,\cdots,h_T$ is at most $T(\OPT^S_{\max}+\eps)|G_j|$ which is equal to $T/2\cdot 2(\OPT^S_{\max}+\eps)|G_j|$.
%$T(1-1/c)c(\OPT^S_{\max}+\eps)|G_j|$ which is equal to:

%\begin{align*}
%&T(1-1/c)c(\OPT^S_{\max}+\eps)|G_j|\\
%=&(T/2-T/c).\frac{2c(c-1)}{c-2}(\OPT^S_{\max}+\eps)|G_j|
%\end{align*}
Therefore, the fraction of examples in $G_j$ that at least $T/2$ of the classifiers in $h_1,\cdots h_T$ make a robustness mistake on is at most $2(\OPT^S_{\max}+\eps)$. %Let $T_j$ denote the set of these examples. Thus, for each example in $G_j\setminus T_j$, at least $T-T/c-(T/2-T/c-1)=T/2+1$ of the classifiers in $\calH'\setminus \calH''$ are making no robustness mistakes on them, i.e., classifying all their perturbations correctly.
Hence, the fraction of examples in $G_j$ that are not robustly classified by the majority-vote classifier is at most $2(\OPT^S_{\max}+\eps)$.%To find the best value of $c$, we solve the following optimization problem:
%\begin{align*}
%&\min \frac{2c(c-1)}{c-2}\\
%&\text{s.t. } c>2
%\end{align*}
%Which gives $c\approx 3.41421$, and $\frac{2c(c-1)}{c-2}\approx 11.6569$.
\end{proof}
\subsection{Reduction from overlapping groups to disjoint groups}
\label{sec:overlapping-groups-reduction}
When the groups are overlapping, we reduce it to the case of disjoint groups. The reduction is as follows: for an input instance $\calI(\calG=\{G_1,\dots,G_g\}, S)$ of overlapping groups, create a new instance $\calI'(\calG'=\{G'_1,\dots,G'_g\},S')$ as follows. Initially, for all $G'_j\in \calG'$, $G'_j$ is an empty set. For each example $(x_i,y_i)\in S$ that belongs to a set of groups $\calG_i=\{G_{i,1},\cdots, G_{i,|\calG_i|}\}\subseteq \calG$ in $\calI$, create identical copies of $(x_i,y_i)$ and assign each copy including the original example to exactly one of the groups in $\calG'_i=\{G'_{i,1},\cdots, G'_{i,|\calG'_i|}\}$. Now we have an instance $\calI'$ with disjoint groups. By executing\ref{alg:boosting} on $\calI'$, it returns a predictor $h$ that achieves %$\beta$-multi-robustness guarantee of at most $\beta\cdot \OPT^{\calI'}_{\max}$ on it. 
a $\beta$-multi-robustness guarantee. First, we argue that if $h$ is used on $\calI$, it achieves a multi-robustness guarantee of $\beta\cdot (\OPT^{\calI'}_{\max}+\eps)$. This is the case since either $h$ makes a robustness mistake on all copies of an example or does not make any robustness mistakes on any of them. Next, we show that $\OPT^{\calI'}_{\max}\leq \OPT^{\calI}_{\max}$. Consider a predictor $h^*\in \calH$ that achieves multi-robustness of $\OPT^{\calI}_{\max}$ on $\calI$. If $h^*$ is used on $\calI'$, for each example $(x,y)\in S$ that $h^*$ has zero robust loss on, it does not make any mistakes on any of its copies in $\calI'$. Additionally, if $h^*$ makes a robustness mistake on $(x,y)$, then it makes a robustness mistake on all its copies in $\calI'$. Thus, $h^*$ achieves a multi-robustness guarantee of $\OPT^{\calI}_{\max}$ on $\calI'$. Therefore, $\OPT^{\calI'}_{\max}\leq \OPT^{\calI}_{\max}$, and a $\beta (\OPT^{\calI'}_{\max}+\eps)$ multi-robustness guarantee on $\calI$ implies $\beta( \OPT^{\calI}_{\max}+\eps)$ multi-robustness. A similar argument holds for the average multi-robustness guarantee.

\begin{rem}
When $|\calG|$ is large, this reduction becomes computationally inefficient, since in the worst case, the number of samples gets increased by a multiplicative factor of $|\calG|$. However, this reduction is equivalent to keeping only one copy of each sample $(x_i,y_i)\in S$ and when executing\ref{alg:boosting}, in each iteration $t$, assigning it a weight of $p_i=\sum_{j\in[g]:(x_i,y_i)\in G_j}P^t_j/|G_j|$.
\end{rem}

\subsection{Generalization Guarantees}
\label{sec:generalization-guarantees}
In this section, we derive generalization guarantees for multi-robustness. First,\ref{lem:vc-robustloss-groups} shows how to bound the VC-Dimension of the intersection of robust loss and groups. We can then invoke this Lemma to get uniform convergence guarantees that will allow us to get concentration for the conditional robust loss across groups (see \ref{def:multi-robustness}).

\begin{lem} [VC Dimension of Intersection of Robust Loss and Groups]
\label{lem:vc-robustloss-groups}
For any class $\calH$, any perturbation set $\calU$, and any group class $\calG$, denote the intersection function class by
%\small{\[
\begin{align*}
&\calF^\calU_{\calH,\calG} \triangleq \{ (x,y)\mapsto \max_{z\in\calU(x)} \ind\insquare{h(z)\neq y} \wedge \ind[x\in G_j]:\\ 
&h\in\calH, G_j\in\calG \}
\end{align*}
%.\]}
Then, it holds that $\vc(\calF^\calU_{\calH,\calG}) \leq \Tilde{O}\inparen{\vc(\calL^{\calU}_{\calH}) + \vc(\calG)}$.
\end{lem}

\begin{thm}[Generalization guarantees for average multi-robustness]
\label{thm:generalization-multi-groups}
With $T=\calO(\ln g/\varepsilon^2)$ and $m= \Tilde{O}\inparen{\frac{\vc(\calH)\ln^2(k)}{\eps^4}+\frac{\vc(\calG) + \ln(1/\delta)}{\varepsilon^2}}$,
\ref{alg:boosting} computes a set of hypotheses $\calH'=\{h_1,\dots, h_T\}$, such that $\forall G_j\in \calG$, 
\small{\begin{align*}
&\frac{1}{T}\sum_{t=1}^T\Prob_{(x,y)\in \calD}\Big[\exists z\in \calU(x): h_t(z)\neq y \mid x\in G_j \Big]\\
&\leq\inparen{1 + \frac{\varepsilon}{\Prob_{\calD}(x\in G_j)}}\inparen{\OPTSMax + \varepsilon}+\frac{\varepsilon}{\Prob_{\calD}(x\in G_j)}
\end{align*}}
\end{thm}

\begin{thm}[Generalization guarantees for $\beta$-multi-robustness]
\label{thm:generalization-multi-groups-deterministic}
With $T=\calO(\ln g/\varepsilon^2)$, $m=\Tilde{O}\inparen{\frac{\vc(\calH)\ln(g)\ln^2(k)}{\varepsilon^6}+\frac{\vc(\calG) + \ln(1/\delta)}{\varepsilon^2}}$, and $\beta=2$,
\ref{alg:boosting} computes a set of hypotheses $\calH'=\{h_1,\dots, h_T\}$, such that $\forall G_j\in \calG$, 
\small{\begin{align*}
&\Prob_{(x,y)\in \calD}\Big[\exists z\in \calU(x): \MAJ(h_1,\dots,h_T)(z)\neq y \mid x\in G_j \Big]\\
&\leq\inparen{1 + \frac{\varepsilon}{\Prob_{\calD}(x\in G_j)}}\inparen{\beta(\OPTSMax + \varepsilon)}+\frac{\varepsilon}{\Prob_{\calD}(x\in G_j)}
\end{align*}}
\end{thm}

\begin{rem}
In\ref{sec:proof-generalization-guarantees-deterministic}, we show how to achieve generalization guarantees in terms of $\OPTDMax$ instead of $\OPTSMax$.
\end{rem}

\section{Discussion}
We exhibited an example showing how %inflated 
using $\ERM$ on an augmented dataset to learn a robust classifier can fail when the examples are robustly un-realizable. Next, we provided a ``boosting-style'' algorithm that uses $\ERM$ and obtains strong robust learning guarantees in the non-realizable regime. 
%In this work, we introduced a novel counter-example exhibited an example showing how %inflated using $\ERM$ on an augmented dataset to learn a robust classifier in the non-realizable setting can fail. % in robust learning.
%Then, we provided a ``boosting-style'' algorithm that uses $\ERM$ and obtains strong robust learning guarantees in the challenging non-realizable regime. 
This work provides theoretical evidence that our existing methods of learning accurate classifiers i.e. \ERM, can be modified effectively to learn robust classifiers even in the agnostic robust regime. Next, we introduced a new multi-robustness objective to obtain robustness guarantees simultaneously across a collection of subgroups. We showed this objective can be achieved by adding a second layer of boosting to the first algorithm. %We leave it as an open question to improve our bounds for the $\beta$-multi-robustness guarantee. 

Adversarial examples exist for many types of classifiers but are especially salient with modern neural-based vision methods. 
%Interestingly, as noted by Adam Kalai in\footnote{\url{https://simons.berkeley.edu/talks/adam-kalai-microsoft-2023-04-25}} neural networks tend not to benefit from boosting.  Kalai argues that due to the large capacity of networks, they likely have an `implicit' type of boosting with different sub-networks that can do a weighted majority vote, if such a majority vote is useful. 
However, due to the large capacity of these networks, it is not clear that they would benefit from boosting. Therefore, the fact that our algorithms rely on boosting should not be interpreted as a firm recommendation 
to use boosting with neural networks, but instead as a theoretical proof-of-concept that plain ERM can be used to learn robust models, given the right algorithmic scheme, especially if such a scheme can reduce the effective number of perturbations available to the adversary.

%We also introduced a new multi-robustness objective and provided algorithms that obtain robustness guarantees simultaneously across groups. We leave it as an open question to improve our bounds for the $\beta$-multi-robustness guarantee.%Future work can continue to explore the multi-robustness objective, connections with multi-calibration (and related `multi' notions), and other ways to use $\ERM$ oracles to learn robust classifiers. 

%% file: rob_appendix_2.tex
\onecolumn

\section{Supplementary Materials}
\subsection{Proof of Lemma~\ref{lem:opt}}
\begin{proof}
By definition of $\OPT_{S,\calH}$, it follows that 
\begin{align*}
    &\OPT_{S,\calH} = \min_{h\in \calH} \frac{1}{m}\sum_{i=1}^{m} \max_{z_i\in \calU(x_i)} \ind\insquare{h(z_i)\neq y_i}\\
    &\geq \min_{h\in \calH} \max_{z_1\in \calU(x_1),\dots,z_m\in\calU(x_m)}\frac{1}{m}\sum_{i=1}^{m} \ind\insquare{h(z_i)\neq y_i}\\
    &\geq \min_{Q\in \Delta(H)} \max_{z_1\in \calU(x_1),\dots,z_m\in\calU(x_m)}\frac{1}{m}\sum_{i=1}^{m} \Ex_{h\sim Q}\ind\insquare{h(z_i)\neq y_i}\\
    &\geq  \min_{Q\in \Delta(\calH)} \max_{\substack{P_{1}\in \Delta(\calU(x_1)),\\ \dots, \\P_{m} \in \Delta(\calU(x_m))}} \frac{1}{m} \sum_{i=1}^{m} \Ex_{z_i\sim P_i } \Ex_{h\sim Q} \ind\insquare{h(z_i)\neq y_i}.
\end{align*}
\end{proof}

\subsection{Proof of \ref{lem:FMS}}
\begin{proof}
By the minimax theorem and \citep*[][Equation 3 and 9 in proof of Theorem 1]{DBLP:conf/colt/FeigeMS15}, we have that 
\begin{align*}
&\max_{\substack{P_1\in \Delta(\calU(x_1)),\\ \dots,\\ P_m\in \Delta(\calU(x_m))}} \sum_{i=1}^{m} \Ex_{z_i\sim P_i} \frac{1}{T} \sum_{t=1}^{T} \ind\insquare{h_t(z_i)\neq y_i} \leq\\
&\min_{Q\in \Delta(\calH)} \max_{\substack{P_{1}\in \Delta(\calU(x_1)),\\ \dots, \\P_{m} \in \Delta(\calU(x_m))}} \Ex_{z_i\sim P_i } \Ex_{h\sim Q} \ind\insquare{h(z_i)\neq y_i}\\
&+ 2\frac{\sqrt{\mathcal{L}^*m\ln k}}{T},
\end{align*}
where $\calL^*=\sum_{i=1}^{m} \max_{z\in\calU(x_i)} \sum_{t=1}^{T}\ind\insquare{h_t(z)\neq y}$. By observing that $\calL^*\leq m T$ and dividing both sides of the inequality above by $m$, we arrive at the inequality stated in the lemma.
\end{proof}

\subsection{Proof of Lemma~\ref{lem:extension-FMS-weights}}
\begin{proof}
We generalize the argument in \cite{DBLP:conf/colt/FeigeMS15} to accommodate the weights on the samples $p_1,\dots,p_m$. %First, we present a generalization of the algorithm presented by %\citet*{DBLP:conf/colt/FeigeMS15} in the following:
Specifically, let
\[L^{ON}_T = \sum_{t=1}^{T}\sum_{i=1}^{m}\sum_{z \in \calU(x_i)} p_iP^{t}(z,(x_i,y_i)) \ind\insquare{h_{t}(z)\neq y_i}\]
be the loss of \ref{alg:FMS} after $T$ rounds, and let 
\[L^*=\max_{P} \sum_{t=1}^{T}\sum_{i=1}^{m}\sum_{z \in \calU(x_i)} p_iP(z,(x_i,y_i)) \ind\insquare{h_{t}(z)\neq y_i}\]
be the benchmark loss. We show that $L^*(1-\eta)-\frac{\ln k}{\eta}\leq L^{ON}_T$. %The remainder of the analysis follows similarly to \citet*[][Equation 3-10 in proof of Theorem 1]{DBLP:conf/colt/FeigeMS15}. 

To this end, define $W^t_i=\inparen{\sum_{z\in\calU(x_i)}w_t(z,(x_i,y_i))}^{p_i}$ and $W^t=\prod_{i=1}^{m}W^t_i$. Let
\begin{align*}
&F^t_{i} = p_i \cdot \frac{\sum_{z\in\calU(x)} w_t(z,(x,y))\ind\insquare{h_t(z)\neq y}}{\sum_{z\in\calU(x)} w_t(z,(x,y))} \\
&= p_i \sum_{z\in \calU(x_i)} P^{t}(z,(x_i,y_i))\ind\insquare{h_t(z)\neq y}
\end{align*}
be the loss of \ref{alg:FMS} on example $(x_i,y_i)$ at round $t$. Observe that by the Step \sareplace{6}{7} in \sareplace{\ref{alg:weighted-FMS}}{\ref{alg:FMS}}, it holds that $W^T_i\geq (1+\eta)^{p_i \max_{z\in \calU(x_i)}\sum_{t=1}^{T} \insquare{h_t(z)\neq y}}$, and therefore $W^T\geq (1+\eta)^{L^*}$.

Observe also
{\small
\begin{align*}
&W^{t+1}_{i} = \\
&\inparen{\sum_{z:\insquare{h_t(z)\neq y}=0} w_t(z,(x,y)) + \sum_{z:\insquare{h_t(z)\neq y}=1} (1+\eta)w_{t}(z,(x,y)) }^{p_i} \\
&= W^t_i\inparen{1+\eta \frac{F^t_{i}}{p_i}}^{p_i}
\end{align*}
}

This implies that
\begin{align*}
&W^T=\prod_{i=1}^{m} W^T_i = \prod_{i=1}^{m} \insquare{k \prod_{t=1}^{T} \inparen{1+\eta \frac{F^t_{i}}{p_i}}}^{p_i} = \\
&k^{\sum_{i=1}^{m}p_i} \prod_{i=1}^{m}\prod_{t=1}^{T} \inparen{1+\eta \frac{F^t_{i}}{p_i}}^{p_i}
\end{align*}

Combining the above we have,
\[(1+\eta)^{L^*} \leq k\prod_{i=1}^{m}\prod_{t=1}^{T} \inparen{1+\eta \frac{F^t_{i}}{p_i}}^{p_i}.\]
We then apply a logarithmic transformation on both sides
\[L^{*}\ln(1+\eta) \leq \ln k + \sum_{i=1}^{m}\sum_{t=1}^{T} p_i\ln\inparen{1+\eta \frac{F^t_{i}}{p_i}}.\]
Since $a-a^2\leq \ln(1 + a) \leq a$ for $a\geq 0$, we have
\[L^*(\eta-\eta^2)\leq \ln k + \sum_{i=1}^{m}\sum_{t=1}^{T} \eta F^t_i = \ln k+ \eta L^{ON}_{T}.\]
By dividing by $\eta$ and rearranging terms we get $L^*(1-\eta)-\frac{\ln k}{\eta}\leq L^{ON}_T$. 

By setting $\eta=\sqrt{\frac{\ln k}{L^*}}$ and observing that $L^*\leq T$, the remainder of the analysis follows similar to \cite[][Equation 3-10 in proof of Theorem 1]{DBLP:conf/colt/FeigeMS15}.
\end{proof}

\subsection{Proof of Lemma~\ref{lem:unif-robloss}}
\begin{proof}
By finiteness of $\calU$, observe that 
for any dataset $S\in (\calX\times \calY)^m$, each robust loss vector in the set of robust loss behaviors:
$$ \Pi_{\calL^{\calU}_{\calH}}(S) = \{(f(x_1,y_1),\dots, f(x_m,y_m)): f \in \calL^{\calU}_{\calH}\}$$ maps to a 0-1 loss vector on the \emph{inflated set} 
$S_\calU=\{(z^1_1,y_1),\dots, (z^k_1,y_1), \dots, (z_m^1,y_m),\dots, (z_m^k,y_m)\}$, 
{\small
\[\Pi_{{\calH}}(S_\calU)= \{(h(z^1_1),\dots, h(z^k_1),\dots, h(z_m^1),\dots, h(z_m^k)): h\in \calH\}\]
}
Therefore, it follows that $\abs{\Pi_{\calL^{\calU}_{\calH}}(S)}\leq \abs{\Pi_{{\calH}}(S_\calU)}$. Then, by applying the Sauer-Shelah lemma, it follows that $\abs{\Pi_{{\calH}}(S_\calU)} \leq O((mk)^{\vc(\calH)})$. Then, by solving for $m$ such that $O((mk)^{\vc(\calH)}) \leq 2^m$, we get that $\vc(\calL^{\calU}_{\calH})\leq O(\vc(\calH) \log(k))$.
\end{proof}

\subsection{Proof of \ref{lem:avg-robust-loss-upper-bound}}

\begin{proof}
{\small
\begin{align}
&\E_{j\in [g]}[\ell^{rob}_j(h_t)]= \sum_j P_j^t(1/|G_j|)\sum_{(x,y)\in G_j} \max_{z\in \calU(x)}\ind\insquare{h_t(z)\neq y}\label{eqn:avg-robust-loss-def}\\
&=\sum_{i=1}^m p_i \cdot \max_{z\in\calU(x)}\ind\insquare{h_t(z)\neq y} \label{eqn:distribution-samples}\\
&\leq  \max_{\substack{P'_1\in \Delta(\calU(x_1)),\\ \dots,\\P'_m\in \Delta(\calU(x_m))}} \sum_{i=1}^{m} p_i\cdot \Ex_{z_i\sim P'_i} \frac{1}{T} \sum_{\tau=1}^{T} \ind\insquare{h^{\text{FMS}}_{\tau}(z_i)\neq y_i}\label{ref:replace-FMS}\\
&\leq \min_{Q\in \Delta(\calH)} \max_{\substack{P'_{1}\in \Delta(\calU(x_1)),\\ \dots, \\P'_{m} \in \Delta(\calU(x_m))}} \sum_{i=1}^{m} p_i \Ex_{z_i\sim P'_i } \Ex_{h\sim Q} \ind\insquare{h(z_i)\neq y_i} + 2\sqrt{\frac{\ln k}{T}}\label{ref:lem-weigthed-FMS}\\
&\leq \min_{h\in \calH} \max_{\substack{P'_{1}\in \Delta(\calU(x_1)),\\ \dots, \\P'_{m} \in \Delta(\calU(x_m))}} \sum_{i=1}^{m} p_i\cdot \Ex_{z_i\sim P'_i } \ind\insquare{h(z_i)\neq y_i}+ 2\sqrt{\frac{\ln k}{T}}\\
&= \min_{h\in\calH} \max_{\substack{z_1\in \calU(x_1),\\ \dots, \\z_{m} \in \calU(x_m)}} \sum_{i=1}^{m} p_i\cdot   \ind\insquare{h(z_i)\neq y_i} + 2\sqrt{\frac{\ln k}{T}}\label{ineq:pure-strategy-suffices}\\
&\leq \min_{h\in\calH} \max_{j\in[g]} (1/|G_j|)\sum_{(x,y)\in G_j}\max_{z\in\calU(x)}\ind\insquare{h(z)\neq y}+2\sqrt{\frac{\ln k}{T}}\label{ineq:relate-to-opt-max}\\
&= OPT_{\max}+2\sqrt{\frac{\ln k}{T}}
\end{align}
}

\ref{eqn:avg-robust-loss-def} holds by plugging in the definition of $\RLoss_j(h_t)$(\ref{eqn:unweighted-robust-loss}).~\ref{eqn:distribution-samples} holds for a distribution $p_1,\dots,p_m$ on the samples. In \ref{ref:replace-FMS}, $h_t$ is replaced with the hypothesis selected by\sareplace{~\ref{alg:weighted-FMS}}{~\ref{alg:FMS}} in each round $t$. \ref{ref:lem-weigthed-FMS} holds by \ref{lem:extension-FMS-weights}. \ref{ineq:pure-strategy-suffices} holds since it suffices for the max-player to pick a pure strategy. \ref{ineq:relate-to-opt-max} holds since the whole probability mass is put as a uniform distribution on the worst-off group. Note that when defining $p_1,\cdots,p_m$, all individuals that belong to the same group have equal weights. 
\end{proof}

\subsection{Proof of~\ref{cor:interpret-avg-loss}}
\begin{proof}
Expected robust loss on each group $G_j\in \calG$ is:
\begin{align}
&\frac{1}{|G_j|}\sum_{(x,y)\in G_j} \max_{z\in\calU(x)}\frac{1}{T}\sum_{t=1}^T \ind[h_t(z)\neq y]\\
=&\frac{1}{|G_j|}\sum_{(x,y)\in G_j} \max_{z\in\calU(x)}\Ex_{h_t\sim U(\calH')} \ind[h_t(z)\neq y]\\
\leq&\frac{1}{|G_j|}\sum_{(x,y)\in G_j} \Ex_{h_t\sim U(\calH')} \max_{z\in \calU(x)}\ind[h_t(z)\neq y]\label{eqn:Jensen-avg-robustness}\\
=&\frac{1}{T}\sum_{t=1}^T \frac{1}{|G_j|}\sum_{(x,y)\in G_j} \max_{z\in\calU(x)}\ind[h_t(z)\neq y]\\
=&\frac{1}{T}\sum_{t=1}^T \RLoss_j(h_t)\leq \OPTSMax+\eps \label{eqn:thm-avg-robust-loss}
\end{align}
Where~\ref{eqn:Jensen-avg-robustness} holds by Jensen's inequality and~\ref{eqn:thm-avg-robust-loss} holds by~\ref{thm:randomized-multi-robustness}.
\end{proof}

\subsection{Proof of~\ref{lem:vc-robustloss-groups}}
\begin{proof}
The proof is inspired by the proof of \citep[claim B.1 in ][]{DBLP:conf/icml/KearnsNRW18} which proved a similar result for the standard $0$-$1$ loss, and here we extend the result to the robust loss using essentially the same proof. 

Let $S\subseteq \calX \times \calY$ be a dataset of size $m$ that is shattered by $\calF^\calU_{\calH,\calG}$. Then, observe that, by definition of $\calF^\calU_{\calH,\calG}$, the number of possible behaviors $\abs{\Pi_{\calF^\calU_{\calH,\calG}}(S)}$ is at most $\abs{\Pi_{\calL^{\calU}_{\calH}}(S)}\cdot \abs{\Pi_\calG(S)}$. By Sauer-Shelah Lemma, $\abs{\Pi_{\calL^{\calU}_{\calH}}(S)} \leq O(m^{\vc(\calL^{\calU}_{\calH})})$ and $ \abs{\Pi_\calG(S)}\leq O(m^{\vc(\calG)})$. Thus, $\abs{\Pi_{\calF^\calU_{\calH,\calG}}(S)} = 2^m \leq O(m^{\vc(\calL^{\calU}_{\calH})+\vc(\calG)})$, and solving for $m$ yields that $m=\Tilde{O}(\vc(\calL^{\calU}_{\calH})+\vc(\calG))$. Hence, $\vc(\calF^\calU_{\calH,\calG}) \leq \Tilde{O}\inparen{\vc(\calL^{\calU}_{\calH}) + \vc(\calG)}$.
\end{proof}

\subsection{Proof of~\ref{thm:generalization-multi-groups}}
\label{sec:proof-generalization-guarantees}

\begin{proof}
The output of~\ref{alg:boosting} is $\calH'=\{h_1,\dots,h_T\}$ where each of the predictors $h_1,\dots,h_T$ is a majority-vote predictor over $\calH$. Due to \cite{blumer:89}, the VC-dimension of the output space is $\vc(\calH^{T'})=\Big(\vc(\calH)T'\ln T'\Big)$ where $T'$ is the number of rounds of\sareplace{~\ref{alg:weighted-FMS}}{~\ref{alg:FMS}} in each oracle call.

Set $m= \Tilde{O}\inparen{\frac{\vc(\calH^{T'})\ln (k)+\vc(\calG) + \ln (1/\delta)}{\varepsilon^2}}$.  
By setting $T'=\calO(\frac{\ln k}{\eps^2})$ and by invoking \ref{lem:unif-robloss} and \ref{lem:vc-robustloss-groups} on the hypothesis class $\calH$ and group class $\calG$, we get the following uniform convergence guarantee. With probability at least $1-\delta$ over $S\sim \calD^m$,

\begin{align*}
&\inparen{\forall h\in \calH^{T'}}\inparen{\forall G_j\in\calG}:\\
&\Bigg\lvert\Ex_{(x,y)\sim \calD} \insquare{\ind[x\in G_j] \wedge \max_{z\in \calU(x)}\ind[h(z)\neq y] } - \frac{1}{m}\sum_{(x,y)\in S} \ind[x\in G_j] \wedge \max_{z\in \calU(x)}\ind[h(z)\neq y]\Bigg\rvert \leq \varepsilon
\end{align*}

We can rewrite the above guarantee in a conditional form which will be useful for us shortly in the proof. Namely, $\forall h\in \calH^{T'}, \forall G_j\in\calG$:
\begin{align}
    & \Prob_{(x,y)\sim \calD}\insquare{\exists z\in \calU(x): h(z)\neq y | x\in G_j } 
    \leq  \\
    & \frac{\Prob_{S}(x\in G_j)}{\Prob_{\calD}(x\in G_j)} \frac{1}{|G_j|} \sum_{(x,y)\in S\wedge x\in G_j}\max_{z\in \calU(x)}\ind[h(z)\neq y]
    + \frac{\varepsilon}{\Prob_{D}(x\in G_j)}\label{eqn:conditional-uniform-convergence}
\end{align}
where $|G_j|=\sum_{(x,y)\in S} \ind[x\in G_j]$. 

\ref{thm:avg-empirical-boosting} shows that running \ref{alg:boosting} produces hypotheses $h_1,\dots, h_T$ such that, $\forall G_j\in \calG$:
\begin{align}
   &\frac{1}{T}\sum_{t=1}^{T} \frac{1}{|G_j|}\sum_{(x,y)\in S\wedge x\in G_j} \max_{z\in \calU(x)}\ind[h_t(z)\neq y] \leq \OPTSMax + \varepsilon\label{eqn:thm-alg-boosting-avg}
\end{align}

\ref{eqn:conditional-uniform-convergence} implies that $\forall G_j\in\calG$,
\begin{align}
    %&\inparen{\forall G_j\in\calG}:\\
    &\frac{1}{T}\sum_{t=1}^T\Prob_{(x,y)\sim \calD}\insquare{\exists z\in \calU(x): h_t(z)\neq y | x\in G_j } \leq 
    \frac{1}{T}\sum_{t=1}^T\frac{\Prob_{S}(x\in G_j)}{\Prob_{\calD}(x\in G_j)} \frac{1}{|G_j|} \sum_{(x,y)\in S\wedge x\in G_j}\max_{z\in \calU(x)}\ind[h_t(z)\neq y]\\
    &+ \frac{\varepsilon}{\Prob_{\calD}(x\in G_j)},\label{eqn:avg-uniform-convergence-conditional}
\end{align}

Combining~\ref{eqn:thm-alg-boosting-avg} and~\ref{eqn:avg-uniform-convergence-conditional} implies:

\begin{align}
&\frac{1}{T}\sum_{t=1}^T\Prob_{(x,y)\in \calD}\insquare{\exists z\in \calU(x): h_t(z)\neq y | x\in G_j } 
\leq\frac{\Prob_{S}(x\in G_j)}{\Prob_{\calD}(x\in G_j)} \inparen{\OPTSMax + \varepsilon}+\frac{\varepsilon}{\Prob_{\calD}(x\in G_j)}
\label{eqn:avg-generalization-prob-sample}
\end{align}

Now, given additional samples $\tilde{m}= O\inparen{\frac{\vc(\calG)+\log(2/\delta)}{\varepsilon^2}}$, in addition to the above, we can guarantee that:
\begin{align}
\forall G_j\in \calG: \frac{\Prob_{S}(x\in G_j)}{\Prob_{\calD}(x\in G_j)}
 \leq \frac{\Prob_{\calD}(x\in G_j) + \varepsilon}{\Prob_{\calD}(x\in G_j)} = 1 + \frac{\varepsilon}{\Prob_{\calD}(x\in G_j)}.
\label{eqn:avg-generalization-prob-distribution}
\end{align}

Combining~\ref{eqn:avg-generalization-prob-sample} and~\ref{eqn:avg-generalization-prob-distribution} implies that:

\begin{align*}
\frac{1}{T}\sum_{t=1}^T\Prob_{(x,y)\sim \calD}\insquare{\exists z\in \calU(x): h_t(z)\neq y | x\in G_j } 
\leq \inparen{1 + \frac{\varepsilon}{\Prob_{\calD}(x\in G_j)}}\inparen{\OPTSMax + \varepsilon}+\frac{\varepsilon}{\Prob_{\calD}(x\in G_j)}
\end{align*}

which completes the proof. We can also obtain a bound in terms of $\OPTDMax$ instead of $\OPTSMax$ using a similar approach used in~\ref{sec:proof-generalization-guarantees-deterministic}.

\end{proof}

\subsection{Proof of~\ref{thm:generalization-multi-groups-deterministic}}
\label{sec:proof-generalization-guarantees-deterministic}

\begin{proof}
The output of~\ref{alg:boosting} is $\calH'=\{h_1,\dots,h_T\}$. Taking majority-vote over the predictors in $\calH'$ is equivalent to taking the majority-vote of majority-vote predictors over $\calH$. Therefore, due to \cite{blumer:89}, the VC-dimension of the output space is $\vc(\calH^{T'})^T=\Big(\vc(\calH)T'\ln T'\Big)T\ln T$, where $T'$ is the number of rounds of\sareplace{~\ref{alg:weighted-FMS}}{~\ref{alg:FMS}} in each oracle call and $T$ is the number of rounds of~\ref{alg:boosting}.

Let the sample size $m= \Tilde{O}\inparen{\frac{\vc(\calH^{T'})^T\log(k)+\vc(\calG) + \log(1/\delta)}{\varepsilon^2}}$. By setting %$T=c\ln(g)/\ln(\frac{c^c\alpha}{(\alpha+c-1)^c})$
$T=\calO(\ln g/\eps^2)$ and $T'=\calO(\frac{\ln k}{\eps^2})$ and by invoking \ref{lem:unif-robloss} and \ref{lem:vc-robustloss-groups} on the hypothesis class $\calH$ and group class $\calG$, we get the following uniform convergence guarantee. With probability at least $1-\delta$ over the sample set $S\sim \calD^m$, $\forall h\in (\calH^{T'})^T$ and $\forall G_j\in\calG$:
\begin{align}
    &\Bigg\lvert\Ex_{(x,y)\sim \calD} \insquare{\ind[x\in G_j] \wedge \max_{z\in \calU(x)}\ind[h(z)\neq y] } 
    -\frac{1}{m}\sum_{(x,y)\in S} \ind[x\in G_j] \wedge \max_{z\in \calU(x)}\ind[h(z)\neq y]\Bigg\rvert
    \leq \varepsilon\label{eqn:uniform-convergence-deterministic-eq1}
\end{align}

We can rewrite the above guarantee in a conditional form which will be useful for us shortly in the proof. Namely, $\forall h\in (\calH^{T'})^T$ and $\forall G_j\in\calG$:
\begin{align}
\Prob_{(x,y)\sim \calD}\insquare{\exists z\in \calU(x): h(z)\neq y | x\in G_j }
\leq &\\
\frac{\Prob_{S}(x\in G_j)}{\Prob_{\calD}(x\in G_j)} \frac{1}{|G_j|} & \sum_{(x,y)\in S\wedge x\in G_j}\max_{z\in \calU(x)}\ind[h(z)\neq y] 
+ \frac{\varepsilon}{\Prob_{\calD}(x\in G_j)}\label{eqn:uniform-convergence-deterministic-proof}
\end{align}
where $|G_j|=\sum_{(x,y)\in S} \ind[x\in G_j]$.
\ref{thm:deterministic-multi-robustness} provides that %\ref{alg:deterministic-boosting} produces hypotheses $h_1,\dots, h_T$ such that 
$h^{\text{maj}}=\MAJ(h_1,\dots,h_T)$ satisfies that $\forall G_j\in \calG$:
\begin{align}
   %\forall G_j\in \calG: 
   &\frac{1}{|G_j|}\sum_{(x,y)\in S\wedge x\in G_j} \max_{z\in \calU(x)}\ind[h^{\text{maj}}(z)\neq y] \leq \beta (\OPTSMax + \varepsilon)\label{eqn:guarantee-deterministic-eq1}
\end{align}
Combining~\ref{eqn:uniform-convergence-deterministic-proof} and~\ref{eqn:guarantee-deterministic-eq1} implies that $\forall G_j\in \calG$:
\begin{align}
&\Prob_{(x,y)\sim \calD}\insquare{\exists z\in \calU(x): h^{\text{maj}}(z)\neq y | x\in G_j } \\
& \leq \frac{\Prob_{S}(x\in G_j)}{\Prob_{\calD}(x\in G_j)} \inparen{\beta(\OPTSMax + \varepsilon)}+\frac{\varepsilon}{\Prob_{\calD}(x\in G_j)}
\label{eqn:deterministic-robustness-eq2}
\end{align}

Now, given additional samples $\tilde{m}= O\inparen{\frac{\vc(\calG)+\log(2/\delta)}{\varepsilon^2}}$, guarantees that:
\begin{align} 
\forall G_j\in \calG: \frac{\Prob_{S}(x\in G_j)}{\Prob_{\calD}(x\in G_j)} \leq \frac{\Prob_{\calD}(x\in G_j) + \varepsilon}{\Prob_{\calD}(x\in G_j)} = 1 + \frac{\varepsilon}{\Prob_{\calD}(x\in G_j)}
\label{eqn:deterministic-robustness-eq3}
\end{align}

Combining~\ref{eqn:deterministic-robustness-eq2} and~\ref{eqn:deterministic-robustness-eq3} gives a refined bound on the average conditional robust loss that holds uniformly across groups. Namely, $\forall G_j\in \calG$,
\begin{align*}
& \Prob_{(x,y)\sim \calD}\insquare{\exists z\in \calU(x): h^{\text{maj}}(z)\neq y | x\in G_j } \\
& \leq\inparen{1 + \frac{\varepsilon}{\Prob_{\calD}(x\in G_j)}}\inparen{\beta(\OPTSMax + \varepsilon)}+\frac{\varepsilon}{\Prob_{\calD}(x\in G_j)}
\end{align*}
We can also obtain a guarantee in terms of $\OPTDMax$ instead of $\OPTSMax$, as follows. Let $h^*\in \calH$ be a predictor which attains $\OPTDMax$ defined as 
{\small
\[\OPTDMax = \min_{h\in\calH} \max_{G_j\in\calG} \Ex_{(x,y)\sim \mathcal{D}}\insquare{\max_{z\in \calU(x)} \ind[h(z)\neq y] \bigg| x\in G_j}.\]
}
%By the first uniform convergence guarantee in the proof, 
Dividing both sides of \ref{eqn:uniform-convergence-deterministic-eq1} by $\Prob_{S}(x\in G_j)$ provides that $\forall G_j\in\calG, \forall h\in\calH$:
\tiny
$
%\abs{
 \Bigg\lvert\frac{\Prob_{\calD}(x\in G_j)}{\Prob_{S}(x\in G_j)} \Prob_{(x,y)\in \calD}\insquare{\exists z\in \calU(x): h(z)\neq y | x\in G_j } -  \Prob_{(x,y)\in S}  \insquare{\exists z\in \calU(x):  h(z)\neq y | x\in G_j }\Bigg\rvert%} 
 \leq  
 \frac{\varepsilon}{\Prob_{S}(x\in G_j)} $
 \normalsize

and thus it implies that %$\forall G_j\in\calG$:
\begin{align*} 
 & \Prob_{(x,y)\in S}\insquare{\exists z\in \calU(x): h(z)\neq y | x\in G_j }\leq  \\
 \inparen{1+\frac{\varepsilon}{\Prob_{S}(x\in G_j)}}\Prob_{(x,y)\sim \calD} & \insquare{\exists z\in \calU(x): h(z)\neq y | x\in G_j } 
+ \frac{\varepsilon}{\Prob_{S}(x\in G_j)}
\end{align*}

Supposing that $\forall G_j\in\calG$, $\Prob_{S}(x\in G_j)\geq \gamma$. By taking a max over groups $G_j \in \calG$, we get
\[\OPTSMax \leq (1+\frac{\varepsilon}{\gamma}) \OPTDMax +\frac{\varepsilon}{\gamma}.\]

\end{proof}